\documentclass[10pt]{article}
\usepackage[preprint]{tmlr}

\usepackage{amsmath,amssymb,amsfonts,amsthm}
\usepackage{mathtools}
\usepackage{bm}
\usepackage{bbm}

\usepackage{booktabs}
\usepackage{placeins}
\usepackage{pdflscape}
\usepackage{longtable}
\usepackage{multirow}
\usepackage{makecell}
\usepackage{array}
\usepackage{tabularx}
\usepackage[most]{tcolorbox}

\usepackage{listings}
\usepackage{graphicx}
\graphicspath{{figures/}{../figures/}}
\usepackage{subcaption}

\usepackage[x11names,dvipsnames,table]{xcolor}
\usepackage{xurl}
\usepackage{hyperref}
\hypersetup{
    colorlinks=true,
    linkcolor=NavyBlue,
    citecolor=NavyBlue,
    filecolor=Mulberry,
    urlcolor=BrickRed
}
\usepackage{pdflscape}
\usepackage[capitalize,noabbrev]{cleveref}
\AddToHook{env/proposition/begin}{\crefalias{theorem}{proposition}}
\AddToHook{env/lemma/begin}{\crefalias{theorem}{lemma}}
\AddToHook{env/corollary/begin}{\crefalias{theorem}{corollary}}
\AddToHook{env/diagnostic/begin}{\crefalias{theorem}{diagnostic}}
\crefname{diagnostic}{Diagnostic identity}{Diagnostic identities}
\Crefname{diagnostic}{Diagnostic identity}{Diagnostic identities}

\newenvironment{tightenum}{\setlength{\topsep}{2pt}\begin{enumerate}\setlength{\itemsep}{1pt}\setlength{\parskip}{0pt}}{\end{enumerate}}
\newtheorem{theorem}{Theorem}[section]
\newtheorem{proposition}[theorem]{Proposition}\newtheorem{diagnostic}[theorem]{Diagnostic identity}

\newtheorem{corollary}[theorem]{Corollary}
\theoremstyle{definition}
\newtheorem{definition}[theorem]{Definition}

\theoremstyle{remark}

\title{PACE: More Agents, More Samples, or More Thinking?\\Untangling Coordination, Ensembling, and Test-Time \mbox{Compute} in Multi-Agent LLMs}

\author{\name Seyed Mahmoud Sajjadi Mohammadabadi \email mahmoud.sajjadi@unr.edu \\
\addr University of Nevada, Reno, NV 89557, USA \quad Code: \url{https://github.com/mahmoudsajjadi/pace}}
\def\month{11}
\def\year{2026}
\def\openreview{\url{https://openreview.net/}}

\newcommand{\numStudiesTotal}{9}
\newcommand{\numContrastsTotal}{516}

\newcommand{\numRawRows}{531}
\newcommand{\numContrastsZhang}{180}
\newcommand{\numContrastsCatA}{358}

\newcommand{\meanCatA}{\ensuremath{-0.032}}
\newcommand{\meanCatANoZhang}{\ensuremath{-0.005}}
\newcommand{\meanCatAStudy}{\ensuremath{-0.010}}
\newcommand{\meanCatANoZhangStudy}{\ensuremath{+0.003}}
\newcommand{\winRateCatA}{0.37}
\newcommand{\winRateCatANoZhang}{0.48}
\newcommand{\numContrastsCatB}{360}

\newcommand{\meanCatB}{\ensuremath{-0.051}}
\newcommand{\meanCatBNoZhang}{\ensuremath{-0.038}}
\newcommand{\meanCatBStudy}{\ensuremath{-0.043}}

\newcommand{\winRateCatB}{0.24}

\newcommand{\numContrastsCatC}{114}
\newcommand{\meanCatC}{\ensuremath{-0.012}}
\newcommand{\meanCatCStudy}{\ensuremath{-0.024}}

\newcommand{\numContrastsCatBC}{474}

\newcommand{\meanCatBC}{\ensuremath{-0.042}}
\newcommand{\meanCatBCNoZhang}{\ensuremath{-0.029}}
\newcommand{\meanCatBCStudy}{\ensuremath{-0.038}}
\newcommand{\meanCatBCNoZhangStudy}{\ensuremath{-0.033}}
\newcommand{\winRateCatBC}{0.26}

\newcommand{\numNotAheadBC}{352}
\newcommand{\pctNotAheadBC}{74}
\newcommand{\numParetoCostObserved}{79}
\newcommand{\numParetoDominated}{51}
\newcommand{\numParetoCheaperMAS}{28}
\newcommand{\numParetoUnobserved}{273}
\newcommand{\pctParetoDominatedBC}{11}
\newcommand{\numNotAheadB}{275}
\newcommand{\pctNotAheadB}{76}
\newcommand{\meanWunderlichSC}{\ensuremath{+0.020}}
\newcommand{\numStudiesDual}{5}
\newcommand{\numContrastsDual}{318}
\newcommand{\numHeroBelow}{279}
\newcommand{\numHeroApparentWins}{53}
\newcommand{\numHeroPositiveA}{123}
\newcommand{\numCellsDual}{72}
\newcommand{\numNegCellsDual}{64}
\newcommand{\meanSamplingGapCells}{\ensuremath{-0.036}}
\newcommand{\meanSamplingGapCellPooled}{\ensuremath{-0.034}}
\newcommand{\meanSamplingGap}{\ensuremath{-0.038}}
\newcommand{\meanSamplingGapNoZhang}{\ensuremath{-0.043}}
\newcommand{\meanSamplingGapContrast}{\ensuremath{-0.028}}

\newcommand{\pctFeasibleBaselineWins}{36}

\newcommand{\meanJwalapuramCoT}{\ensuremath{-0.007}}
\newcommand{\meanJwalapuramSC}{\ensuremath{-0.041}}
\newcommand{\medianCostRatioCoT}{8.1}

\newcommand{\medianCostRatioSC}{1.6}

\newcommand{\kimThresholdPct}{45}
\newcommand{\numRowsExcluded}{15}
\newcommand{\numTranSeq}{48}
\newcommand{\numContrastsKim}{18}
\newcommand{\numBCNoTranSeq}{426}
\newcommand{\meanBCNoTranSeq}{\ensuremath{-0.045}}
\newcommand{\meanTranStudy}{\ensuremath{-0.007}}
\newcommand{\meanTranNoSeq}{\ensuremath{+0.002}}
\newcommand{\numContrastRowsAC}{832}
\newcommand{\numConfigFeasible}{356}
\newcommand{\pctConfigFeasible}{69}
\newcommand{\numConfigAhead}{51}
\newcommand{\pctConfigAhead}{10}
\newcommand{\numConfigUndet}{109}
\newcommand{\pctConfigUndet}{21}
\newcommand{\numSharedCeil}{156}
\newcommand{\numSharedCeilAhead}{44}
\newcommand{\pctSharedCeilAhead}{28}
\newcommand{\numSharedCeilNoHundred}{140}
\newcommand{\numSharedCeilNoHundredAhead}{28}
\newcommand{\pctSharedCeilNoHundredAhead}{20}
\newcommand{\numNoiseBelowBC}{171}
\newcommand{\numNoiseWithinBC}{268}
\newcommand{\numNoiseAboveBC}{35}
\newcommand{\numFeasibleNoiseBelow}{56}
\newcommand{\numSharedCeilNoiseAbove}{12}
\newcommand{\remlANineK}{5}
\newcommand{\remlANineM}{358}

\newcommand{\remlBOnlyK}{6}
\newcommand{\remlBOnlyM}{360}
\newcommand{\remlBOnlyBeta}{\ensuremath{-0.235}}

\newcommand{\remlBOnlyCrLo}{\ensuremath{-0.388}}
\newcommand{\remlBOnlyCrHi}{\ensuremath{-0.082}}

\newcommand{\remlBNineK}{8}
\newcommand{\remlBNineM}{474}
\newcommand{\remlBNineBeta}{\ensuremath{-0.195}}
\newcommand{\remlBNineSE}{0.052}

\newcommand{\remlBNineCrLo}{\ensuremath{-0.323}}
\newcommand{\remlBNineCrHi}{\ensuremath{-0.067}}
\newcommand{\remlBNineNu}{6.4}
\newcommand{\remlBNineITwo}{93}

\newcommand{\remlBNinePiLo}{\ensuremath{-0.515}}
\newcommand{\remlBNinePiHi}{\ensuremath{+0.125}}
\newcommand{\remlBNineContrastPiLo}{\ensuremath{-1.411}}
\newcommand{\remlBNineContrastPiHi}{\ensuremath{+1.021}}
\newcommand{\remlBNineNoZK}{7}
\newcommand{\remlBNineNoZM}{334}
\newcommand{\remlBNineNoZBeta}{\ensuremath{-0.145}}
\newcommand{\remlBNineNoZCrLo}{\ensuremath{-0.246}}
\newcommand{\remlBNineNoZCrHi}{\ensuremath{-0.045}}

\newcommand{\remlBNineNoJK}{7}
\newcommand{\remlBNineNoJM}{367}
\newcommand{\remlBNineNoJBeta}{\ensuremath{-0.190}}
\newcommand{\remlBNineNoJCrLo}{\ensuremath{-0.347}}
\newcommand{\remlBNineNoJCrHi}{\ensuremath{-0.034}}

\newcommand{\remlGapK}{5}
\newcommand{\remlGapM}{318}

\newcommand{\remlGapCellsK}{5}
\newcommand{\remlGapCellsM}{72}
\newcommand{\remlGapCellsBeta}{\ensuremath{-0.178}}

\newcommand{\remlGapCellsCrLo}{\ensuremath{-0.347}}
\newcommand{\remlGapCellsCrHi}{\ensuremath{-0.009}}
\newcommand{\remlGapCellsNu}{3.7}
\newcommand{\remlBNinePhiMinBeta}{\ensuremath{-0.195}}
\newcommand{\remlBNinePhiMaxBeta}{\ensuremath{-0.194}}
\newcommand{\remlBNinePhiWorstCrLo}{\ensuremath{-0.329}}
\newcommand{\remlBNinePhiWorstCrHi}{\ensuremath{-0.059}}
\newcommand{\remlBNinePhiNuMin}{6.2}
\newcommand{\remlBNinePhiNuMax}{6.5}
\newcommand{\remlBNineLooMinBeta}{\ensuremath{-0.233}}
\newcommand{\remlBNineLooMaxBeta}{\ensuremath{-0.145}}
\newcommand{\remlBNineLooWorstCrHi}{\ensuremath{-0.034}}

\newcommand{\remlCheBaseEightCrLo}{\ensuremath{-0.324}}
\newcommand{\remlCheBaseEightCrHi}{\ensuremath{-0.050}}

\newcommand{\remlCheBenchEightCrHi}{\ensuremath{-0.040}}
\newcommand{\remlCheStudyHalfCrHi}{\ensuremath{+0.007}}
\newcommand{\remlCheStudyEightCrHi}{\ensuremath{+0.059}}
\newcommand{\remlBNineFrechetBeta}{\ensuremath{-0.194}}
\newcommand{\remlBNineFrechetCrLo}{\ensuremath{-0.322}}
\newcommand{\remlBNineFrechetCrHi}{\ensuremath{-0.066}}
\newcommand{\remlBNineMergedJKBeta}{\ensuremath{-0.187}}
\newcommand{\remlBNineMergedJKCrLo}{\ensuremath{-0.333}}
\newcommand{\remlBNineMergedJKCrHi}{\ensuremath{-0.040}}
\newcommand{\remlBNineMergedJKNu}{5.47}
\newcommand{\remlBNineMergedJKP}{0.021}
\newcommand{\remlBNineNoAgiBeta}{\ensuremath{-0.203}}
\newcommand{\remlBNineNoAgiCrLo}{\ensuremath{-0.345}}
\newcommand{\remlBNineNoAgiCrHi}{\ensuremath{-0.062}}
\newcommand{\remlBNineRawDeltaBeta}{\ensuremath{-0.037}}
\newcommand{\remlBNineRawDeltaCrLo}{\ensuremath{-0.061}}
\newcommand{\remlBNineRawDeltaCrHi}{\ensuremath{-0.013}}
\newcommand{\remlBNineCohenHBeta}{\ensuremath{-0.086}}
\newcommand{\remlBNineCohenHCrLo}{\ensuremath{-0.141}}
\newcommand{\remlBNineCohenHCrHi}{\ensuremath{-0.031}}
\newcommand{\remlBNineFwClustBeta}{\ensuremath{-0.241}}
\newcommand{\remlBNineFwClustCrLo}{\ensuremath{-0.397}}
\newcommand{\remlBNineFwClustCrHi}{\ensuremath{-0.085}}

\providecommand{\confCommCharsHp}{}
\renewcommand{\confCommCharsHp}{6{,}998}
\providecommand{\confCommCharsMmlu}{}
\renewcommand{\confCommCharsMmlu}{11{,}070}
\providecommand{\confDsCallsHp}{}
\renewcommand{\confDsCallsHp}{800}
\providecommand{\confDsCallsMmlu}{}
\renewcommand{\confDsCallsMmlu}{800}
\providecommand{\confDsCapHp}{}
\renewcommand{\confDsCapHp}{1}
\providecommand{\confDsCapMmlu}{}
\renewcommand{\confDsCapMmlu}{57}
\providecommand{\confDsParseHp}{}
\renewcommand{\confDsParseHp}{1}
\providecommand{\confDsParseMmlu}{}
\renewcommand{\confDsParseMmlu}{59}
\providecommand{\confDsParseOfHp}{}
\renewcommand{\confDsParseOfHp}{800}
\providecommand{\confDsParseOfMmlu}{}
\renewcommand{\confDsParseOfMmlu}{800}
\providecommand{\confFourGap}{}
\renewcommand{\confFourGap}{\ensuremath{-0.072}}
\providecommand{\confFourHi}{}
\renewcommand{\confFourHi}{\ensuremath{-0.028}}
\providecommand{\confFourLo}{}
\renewcommand{\confFourLo}{\ensuremath{-0.118}}
\providecommand{\confFourN}{}
\renewcommand{\confFourN}{250}
\providecommand{\confFourP}{}
\renewcommand{\confFourP}{0.999}
\providecommand{\confFourPholm}{}
\renewcommand{\confFourPholm}{1.000}
\providecommand{\confFourReject}{}
\renewcommand{\confFourReject}{not rejected}
\providecommand{\confFourUpper}{}
\renewcommand{\confFourUpper}{0.0}
\providecommand{\confOneGap}{}
\renewcommand{\confOneGap}{\ensuremath{-0.015}}
\providecommand{\confOneHi}{}
\renewcommand{\confOneHi}{\ensuremath{+0.035}}
\providecommand{\confOneKTwentyFour}{}
\renewcommand{\confOneKTwentyFour}{\ensuremath{-0.026}}
\providecommand{\confOneLo}{}
\renewcommand{\confOneLo}{\ensuremath{-0.065}}
\providecommand{\confOneN}{}
\renewcommand{\confOneN}{250}
\providecommand{\confOneP}{}
\renewcommand{\confOneP}{0.728}
\providecommand{\confOnePholm}{}
\renewcommand{\confOnePholm}{1.000}
\providecommand{\confOneReject}{}
\renewcommand{\confOneReject}{not rejected}
\providecommand{\confOneUpper}{}
\renewcommand{\confOneUpper}{3.5}
\providecommand{\confOneUpperMax}{}
\renewcommand{\confOneUpperMax}{\ensuremath{+0.061}}
\providecommand{\confOneUpperMaxPP}{}
\renewcommand{\confOneUpperMaxPP}{6.1}
\providecommand{\confOneUpperMin}{}
\renewcommand{\confOneUpperMin}{\ensuremath{+0.026}}
\providecommand{\confOneUpperMinPP}{}
\renewcommand{\confOneUpperMinPP}{2.6}
\providecommand{\confPrimaryComplete}{}
\renewcommand{\confPrimaryComplete}{1}
\providecommand{\confReasonCharsHp}{}
\renewcommand{\confReasonCharsHp}{109{,}038}
\providecommand{\confReasonCharsMmlu}{}
\renewcommand{\confReasonCharsMmlu}{80{,}984}
\providecommand{\confReasonShareHp}{}
\renewcommand{\confReasonShareHp}{0.94}
\providecommand{\confReasonShareMmlu}{}
\renewcommand{\confReasonShareMmlu}{0.88}
\providecommand{\confRegStatus}{}
\renewcommand{\confRegStatus}{the plan, with its addendum, was committed to version control and hashed before any outcome was analyzed; it had not been registered on a public registry when data collection began}
\providecommand{\confThreeGap}{}
\renewcommand{\confThreeGap}{\ensuremath{-0.015}}
\providecommand{\confThreeHi}{}
\renewcommand{\confThreeHi}{\ensuremath{+0.048}}
\providecommand{\confThreeLo}{}
\renewcommand{\confThreeLo}{\ensuremath{-0.079}}
\providecommand{\confThreeN}{}
\renewcommand{\confThreeN}{250}
\providecommand{\confThreeP}{}
\renewcommand{\confThreeP}{0.623}
\providecommand{\confThreePholm}{}
\renewcommand{\confThreePholm}{1.000}
\providecommand{\confThreeReject}{}
\renewcommand{\confThreeReject}{not rejected}
\providecommand{\confThreeUpper}{}
\renewcommand{\confThreeUpper}{4.8}
\providecommand{\confTwoGap}{}
\renewcommand{\confTwoGap}{\ensuremath{+0.073}}
\providecommand{\confTwoHi}{}
\renewcommand{\confTwoHi}{\ensuremath{+0.122}}
\providecommand{\confTwoLo}{}
\renewcommand{\confTwoLo}{\ensuremath{+0.024}}
\providecommand{\confTwoN}{}
\renewcommand{\confTwoN}{200}
\providecommand{\confTwoP}{}
\renewcommand{\confTwoP}{0.002}
\providecommand{\confTwoPholm}{}
\renewcommand{\confTwoPholm}{0.007}
\providecommand{\confTwoReject}{}
\renewcommand{\confTwoReject}{rejected}
\providecommand{\confTwoUpper}{}
\renewcommand{\confTwoUpper}{12.2}
\providecommand{\decoyFinalShareDebate}{}
\renewcommand{\decoyFinalShareDebate}{0.761}
\providecommand{\decoyFinalSharePlacebo}{}
\renewcommand{\decoyFinalSharePlacebo}{0.328}
\providecommand{\decoyLeaveFourDebate}{}
\renewcommand{\decoyLeaveFourDebate}{0.044}
\providecommand{\decoyLeaveFourPlacebo}{}
\renewcommand{\decoyLeaveFourPlacebo}{0.630}
\providecommand{\decoyRoneShare}{}
\renewcommand{\decoyRoneShare}{0.787}
\providecommand{\decoyUnanimousRone}{}
\renewcommand{\decoyUnanimousRone}{0.365}
\providecommand{\hpDebateAcc}{}
\renewcommand{\hpDebateAcc}{0.182}
\providecommand{\hpDebateTok}{}
\renewcommand{\hpDebateTok}{23{,}604}
\providecommand{\hpDiscAOne}{}
\renewcommand{\hpDiscAOne}{0.333}
\providecommand{\hpDiscAZero}{}
\renewcommand{\hpDiscAZero}{0.173}
\providecommand{\hpDiscD}{}
\renewcommand{\hpDiscD}{0.060}
\providecommand{\hpDiscNOne}{}
\renewcommand{\hpDiscNOne}{12}
\providecommand{\hpDiscPlAOne}{}
\renewcommand{\hpDiscPlAOne}{0.375}
\providecommand{\hpDiscPlAZero}{}
\renewcommand{\hpDiscPlAZero}{0.240}
\providecommand{\hpDiscQPowSix}{}
\renewcommand{\hpDiscQPowSix}{0.060}
\providecommand{\hpDiscQPowTen}{}
\renewcommand{\hpDiscQPowTen}{0.009}
\providecommand{\hpDiscRate}{}
\renewcommand{\hpDiscRate}{0.626}
\providecommand{\hpFrontier}{}
\renewcommand{\hpFrontier}{0.109}
\providecommand{\hpGapEst}{}
\renewcommand{\hpGapEst}{\ensuremath{+0.073}}
\providecommand{\hpGapHi}{}
\renewcommand{\hpGapHi}{\ensuremath{+0.122}}
\providecommand{\hpGapLo}{}
\renewcommand{\hpGapLo}{\ensuremath{+0.024}}
\providecommand{\hpKtok}{}
\renewcommand{\hpKtok}{30}
\providecommand{\hpKtokRatio}{}
\renewcommand{\hpKtokRatio}{6}
\providecommand{\hpN}{}
\renewcommand{\hpN}{200}
\providecommand{\hpPlaceboAcc}{}
\renewcommand{\hpPlaceboAcc}{0.248}
\providecommand{\hpPlaceboTokFull}{}
\renewcommand{\hpPlaceboTokFull}{34{,}819}
\providecommand{\hpPlaceboTokPct}{}
\renewcommand{\hpPlaceboTokPct}{48}
\providecommand{\hpSampleTok}{}
\renewcommand{\hpSampleTok}{793}
\providecommand{\hpVoteAcc}{}
\renewcommand{\hpVoteAcc}{0.077}
\providecommand{\hpVoteTok}{}
\renewcommand{\hpVoteTok}{3{,}965}
\providecommand{\lamHurdleHalf}{}
\renewcommand{\lamHurdleHalf}{3.67}
\providecommand{\lamHurdleQuarter}{}
\renewcommand{\lamHurdleQuarter}{2.99}
\providecommand{\lamHurdleZero}{}
\renewcommand{\lamHurdleZero}{2.12}
\providecommand{\meanControlledHpDebateGap}{}
\renewcommand{\meanControlledHpDebateGap}{\ensuremath{+0.073}}
\providecommand{\meanControlledMmluDebateGap}{}
\renewcommand{\meanControlledMmluDebateGap}{\ensuremath{-0.015}}
\providecommand{\meanControlledMmluMoaGap}{}
\renewcommand{\meanControlledMmluMoaGap}{\ensuremath{-0.072}}
\providecommand{\mmluDebateAcc}{}
\renewcommand{\mmluDebateAcc}{0.567}
\providecommand{\mmluDebateFrontier}{}
\renewcommand{\mmluDebateFrontier}{0.581}
\providecommand{\mmluDebateTok}{}
\renewcommand{\mmluDebateTok}{19{,}571}
\providecommand{\mmluGreedyAcc}{}
\renewcommand{\mmluGreedyAcc}{0.408}
\providecommand{\mmluGreedyTok}{}
\renewcommand{\mmluGreedyTok}{1{,}172}
\providecommand{\mmluKtok}{}
\renewcommand{\mmluKtok}{22}
\providecommand{\mmluKtokExact}{}
\renewcommand{\mmluKtokExact}{22.4}
\providecommand{\mmluKtokRatio}{}
\renewcommand{\mmluKtokRatio}{1.4}
\providecommand{\mmluMoaAcc}{}
\renewcommand{\mmluMoaAcc}{0.508}
\providecommand{\mmluMoaTok}{}
\renewcommand{\mmluMoaTok}{13{,}409}
\providecommand{\mmluSFourTok}{}
\renewcommand{\mmluSFourTok}{19{,}571}
\providecommand{\mmluSampleTok}{}
\renewcommand{\mmluSampleTok}{874}
\providecommand{\mmluThinkTok}{}
\renewcommand{\mmluThinkTok}{1{,}120}
\providecommand{\numControlledHpHolmP}{}
\renewcommand{\numControlledHpHolmP}{0.007}
\providecommand{\numControlledHpItems}{}
\renewcommand{\numControlledHpItems}{200}
\providecommand{\numControlledMmluItems}{}
\renewcommand{\numControlledMmluItems}{250}
\providecommand{\preregShaAddOne}{}
\renewcommand{\preregShaAddOne}{4e47179fb69f5664cbe6e993b9022c6f5b3e2d08a357277a346c88003769bbbd}
\providecommand{\preregShaMain}{}
\renewcommand{\preregShaMain}{826cbe97c92dca20d152d505dc644c29f78a5956e3f9b4a2124785511f045c90}

\providecommand{\hpThirtyCaveat}{}\renewcommand{\hpThirtyCaveat}{split-view debate exceeds the post hoc 30-call split-view voting comparator (HP30, $0.010$ at $23{,}002$ tokens) yet still trails placebo and chance}

\providecommand{\hpThirtyProvisional}{}\renewcommand{\hpThirtyProvisional}{the HP30 comparison is reported in \Cref{sec:controlled_primary}}
\providecommand{\hpThirtySentence}{}\renewcommand{\hpThirtySentence}{In a post hoc near-budget comparison, a 30-call split-view voting pool (HP30, $n=200$, $23{,}002$ vs.\ $\hpDebateTok$ tokens) scores $0.010$ [$0.000$, $0.025$] against $0.182$ for split-view debate (difference $+0.172$ [$+0.120$, $+0.228$], two-sided paired $p=0.0004$, unadjusted); because HP30 uses $2.5\%$ fewer tokens than debate, it is a near-budget comparator rather than an exact token match, and debate still trails placebo ($\hpPlaceboAcc$) and four-choice chance ($0.250$).}

\providecommand{\twoStageHtwoLo}{}\renewcommand{\twoStageHtwoLo}{-0.001}
\providecommand{\twoStageHtwoHi}{}\renewcommand{\twoStageHtwoHi}{+0.110}
\providecommand{\twoStageHtwoP}{}\renewcommand{\twoStageHtwoP}{0.028}

\providecommand{\allegVerified}{}\renewcommand{\allegVerified}{1}
\providecommand{\allegListed}{}\renewcommand{\allegListed}{15}
\providecommand{\allegUnverified}{}\renewcommand{\allegUnverified}{14}
\providecommand{\allegIntroAuditB}{}\renewcommand{\allegIntroAuditB}{+0.149}
\providecommand{\allegAllegianceTerm}{}\renewcommand{\allegAllegianceTerm}{+0.113}
\providecommand{\allegRungTerm}{}\renewcommand{\allegRungTerm}{+0.036}

\providecommand{\auditBoundPct}{}\renewcommand{\auditBoundPct}{7.22}

\providecommand{\dualBelow}{}\renewcommand{\dualBelow}{279}
\providecommand{\dualPairs}{}\renewcommand{\dualPairs}{318}
\providecommand{\dualStudiesNegShift}{}\renewcommand{\dualStudiesNegShift}{5}
\providecommand{\failsafeKstar}{}\renewcommand{\failsafeKstar}{2}
\providecommand{\failsafeKzero}{}\renewcommand{\failsafeKzero}{12}

\providecommand{\frontCmaxhp}{}\renewcommand{\frontCmaxhp}{3,965}
\providecommand{\frontCmaxmmlu}{}\renewcommand{\frontCmaxmmlu}{13,982}

\providecommand{\frontLhp}{}\renewcommand{\frontLhp}{0.148}
\providecommand{\frontLmmlu}{}\renewcommand{\frontLmmlu}{0.579}
\providecommand{\frontUhp}{}\renewcommand{\frontUhp}{0.148}
\providecommand{\frontUmmlu}{}\renewcommand{\frontUmmlu}{0.581}
\providecommand{\rhoHp}{}\renewcommand{\rhoHp}{0.07}
\providecommand{\rhoMmlu}{}\renewcommand{\rhoMmlu}{0.49}
\providecommand{\specCrNegative}{}\renewcommand{\specCrNegative}{286}
\providecommand{\specN}{}\renewcommand{\specN}{288}
\providecommand{\specNegative}{}\renewcommand{\specNegative}{288}
\providecommand{\traceItems}{}\renewcommand{\traceItems}{408}

\begin{document}
\raggedbottom

\maketitle

\begin{abstract}
Multi-agent language model evaluations often compare a communicating system with a single greedy pass, which confounds coordination with extra samples and test-time compute. We introduce Parity-Aware Compute Evaluation (PACE): a six-dimensional resource vector, four baseline categories, and a directional rule that credits a gain only against a baseline using at least as much of the declared resources and rules a gain out against any baseline using no more. We audit \numStudiesTotal\ published studies. Across the \numCellsDual\ deduplicated baseline cells that report both a greedy and a sampled baseline, the sampling gap, by which the sampled baseline lowers the multi-agent difference, averages \meanSamplingGapCells\ with studies weighted equally (negative in \numNegCellsDual\ cells), and \numHeroApparentWins\ of the \numHeroPositiveA\ paired contrasts that beat a greedy pass become nonpositive against the sampled baseline. Against sampled or budget-matched baselines the mean difference is \meanCatBC; all eight Category~B/C study means are negative and the one Category~D study is positive. In a controlled experiment with an open-weight model, debate and mixture-of-agents do not clear the equal-token self-consistency frontier on MMLU-Pro (debate \meanControlledMmluDebateGap); on Hidden Profile, split-view debate exceeds a lower-budget pool yet trails content-mismatched placebo and four-choice chance. We release a seven-item checklist, an evaluation card, and command-line utilities.
\end{abstract}

\vspace{1ex}

\section{Introduction}
\label{sec:intro}

A multi-agent system running $K$ agents for $T$ rounds makes roughly $KT$ model calls where a single agent makes one. Comparing that topology with one greedy pass therefore entangles inter-agent communication with independent sampling and expanded test-time compute. Although multi-agent architectures are used for reasoning and coding \citep{xi2023rise,wang2024surveyagent,wu2023autogen,hong2023metagpt,chen2023agentverse}, repeated sampling and longer reasoning traces already improve single-agent accuracy without communication \citep{wang2022selfconsistency,snell2024scaling,brown2024large}.

Studies that hold different resources constant diverge. Parity in thinking tokens lets a single agent match or beat debate across nonvacuous budgets \citep{tran2026thinking}; tool tasks with matched iteration ceilings saturate near $\kimThresholdPct\%$ baseline accuracy \citep{kim2026scaling}; auditing against greedy CoT rather than CoT-SC with five samples shifts the mean difference from $\meanJwalapuramCoT$ to $\meanJwalapuramSC$ \citep{jwalapuram2026illusion}; and majority voting over five samples matches or beats debate with five agents \citep{choi2025debate}.

\begin{figure}[t]
\centering
\includegraphics[width=\textwidth]{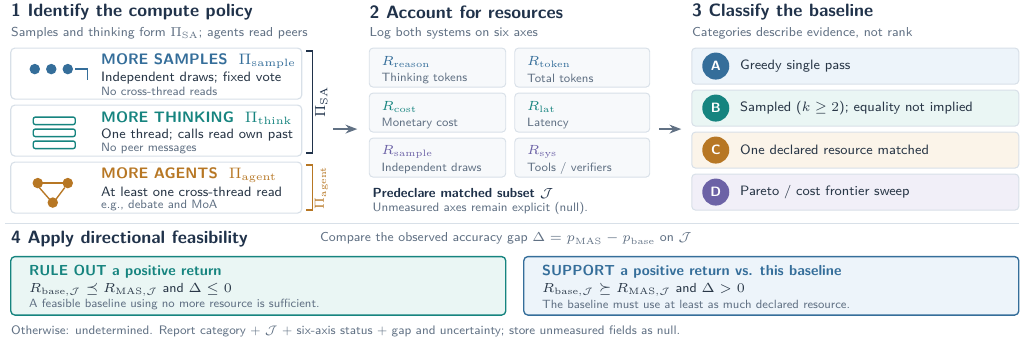}
\caption{PACE overview. Policies are classified by information flow (samples and thinking form the noncommunicating class $\Pi_{\mathrm{SA}}$), both systems are logged on six resource axes with the matched subset $\mathcal{J}$ declared before outcomes are examined, and the baseline is classified as Category A to D, which describe evidence, not rank. On $\mathcal{J}$, a feasible baseline using no more resource with $\Delta\leq 0$ rules out a positive return; a positive gap supports one only against a baseline using at least as much of every declared resource, and otherwise the result is undetermined.}
\label{fig:policy_families}\label{fig:pace_overview}
\end{figure}

Parity-Aware Compute Evaluation (PACE) structures every comparison around four elements: four baseline categories, A to D (\Cref{fig:policy_families}), a six-dimensional resource vector $\mathbf{R}$ (\Cref{eq:resource_vector}), directional resource feasibility (\Cref{sec:decomposition}), and the frontier gap against noncommunicating scaling (\Cref{tab:pace_card}). PACE is an evaluation methodology rather than a survey of multi-agent architectures: its contributions are a formal resource-accounting framework, a directional rule for baseline feasibility, a bounded empirical audit of published comparisons, and a controlled experiment illustrating the resulting evaluation procedure. Extracting \numContrastsTotal\ unique multi-agent configurations (from \numRawRows\ raw table rows) across \numStudiesTotal\ studies, we make five contributions:
\begin{tightenum}
    \item Resource accounting across the six dimensions of $\mathbf{R}$ (\Cref{sec:resource_vector,sec:cme}).
    \item Formal results on information, voting and conformity, with token-equated pool sizing (\Cref{prop:redundancy,prop:vote_amp,prop:keff,thm:conformity,eq:ktok}).
    \item Bounded evidence audit of nine studies: descriptive mean accuracy differences of \meanCatA{} against greedy baselines and \meanCatBC{} against sampled or resource-matched baselines, alongside an exploratory multilevel REML estimate of \remlBNineBeta{} log odds for Category~B/C contrasts (\Cref{sec:meta}).
    \item Controlled evaluation on shared and split evidence, alongside exploratory capability-threshold and transcript diagnostics (\Cref{sec:controlled,app:moved_s5_threshold}).
    \item A reviewer checklist, an evaluation card and software, combined in one table (\Cref{sec:discussion}).
\end{tightenum}

\section{Related Work}
\label{sec:related}

Multi-agent frameworks report gains over single models on reasoning, coding and deduction \citep{du2023improving,wu2023autogen,hong2023metagpt,chen2023agentverse}; \citet{chen2024are} analyze compound inference scaling, and \citet{li2024more} show that independent sampling with majority voting alone (``Agent Forest'') scales with the number of agents, so we test whether communication adds gains beyond this sample multiplicity. \citet{zhang2025stop} observe that new debate methods are often compared only with direct prompting of a single model, which confounds communication with the gains of repeated sampling and longer test-time reasoning \citep{wang2022selfconsistency,snell2024scaling,brown2024large}; \Cref{tab:original_vs_audit} lists the named frameworks audited in our corpus, whose introducing papers we did not re-extract.

Closest prior empirical work compares debate-style systems with single-agent and self-consistency baselines: agent evaluations should be cost-controlled \citep{kapoor2024ai}, and a single agent with strong prompts almost matches the best discussion approach \citep{wang2024rethinking}. \citet{smit2024understanding}, \citet{zhang2025stop} and \citet{yang2025revisiting} differ in what they hold fixed (Yang et al.\ cap self-consistency, refinement and debate at 16 samples, Smit et al.\ report the best and median score over hyperparameter configurations, and Zhang et al.\ report neither matched tokens nor cost); in our extraction, debate-based systems are ahead of the sampled baseline in a minority of contrasts (19\%, 17\% and 21\% for Zhang, Yang and Smit et al.; \Cref{sec:meta}). None places its results beside studies with other matching conditions in one resource-accounting framework or compares the same systems against greedy and sampled baselines; that is what we add. Controlled audits, matching definitions, statistical background and task structure are in \Cref{app:moved_rw_audits,app:moved_rw_more,app:moved_rw_extra}.

\section{The PACE Framework: Resources, Baselines and Directional Feasibility}
\label{sec:theory}

The three channels of a multi-agent difference, context crowding and token dilution, the interpretation of directional feasibility, worked examples of pool sizing and the remaining formal results are in \Cref{app:moved_s3_channels,app:moved_s3_crowd,app:moved_s3_ident,app:moved_s3_pool,app:moved_s3_conformity,app:moved_s3_prop2}.

\subsection{Resource Vector and Verification States}
\label{sec:resource_vector}
Comparing a multi-agent system (MAS) with a single-agent system (SAS) requires the resources of both; rather than one scalar, we use a resource vector with six dimensions:
\begin{equation}
  \mathbf{R} \;=\; \langle R_{\text{reason}},\, R_{\text{token}},\, R_{\text{cost}},\, R_{\text{lat}},\, R_{\text{sample}},\, R_{\text{sys}} \rangle,
  \label{eq:resource_vector}
\end{equation}
where (1)~$R_{\text{reason}}$ (thinking tokens): Tokens generated inside reasoning or scratchpad blocks across all calls. (2)~$R_{\text{token}}$ (total tokens): The sum of input and output tokens across all calls and rounds, with system prompts, role headers, echoed histories and scaffolding counted once. (3)~$R_{\text{cost}}$ (monetary cost): API expenditure in USD when pricing is available; FLOPs are reported separately as a compute proxy, never pooled with dollars. (4)~$R_{\text{lat}}$ (execution latency): Execution latency, stated either as summed serialized call time or as critical-path elapsed time, including synchronization delays across rounds. (5)~$R_{\text{sample}}$ (independent sample count): Number of independent generation trajectories ($k$) drawn and aggregated (total model calls $N_{\text{calls}}$, including sequential rounds and judge calls, are logged separately). (6)~$R_{\text{sys}}$ (external tool and verifier invocations): Invocations of external compilers, unit-test verifiers, retrieval engines, and execution environments (prompt scaffolding tokens $C_{\text{scaffold}}$ are accounted within $R_{\text{token}}$). Coordinates of $\mathbf{R}$ are compared one at a time on the matched subset $\mathcal{J}$, never summed across coordinates.

For any empirical comparison, each dimension of $\mathbf{R}$ is assigned one of four verification states:
(i)~\emph{controlled}: Identical budget or shared ceiling enforced between MAS and SAS (e.g., Tran \& Kiela thinking tokens). (ii)~\emph{approximately matched}: Controlled within bounded thresholds (e.g., Kim et al.\ maximum iterations; Wunderlich et al.\ compute budget sweeps). (iii)~\emph{unmatched (measured)}: Audited and reported, but allowed to diverge (e.g., Jwalapuram et al.\ dollar costs). (iv)~\emph{unreported (omitted)}: Omitted from the evaluation, creating an uncontrolled confound. \Cref{fig:resource_matrix} classifies each dimension of $\mathbf{R}$ for the nine studies; these states reflect our coding of what each paper controls, not a quantity the papers report directly.

\subsection{Additive Decomposition: Policy Classes, Feasibility, and Baseline Strengthening}
\label{sec:decomposition}
Because $\mathbf{R}$ lacks a natural total order, an unconstrained function over it is ill-posed. We call a sequence of model calls a \emph{thread} if each call reads only the task input and that thread's earlier outputs, and partition test-time compute policies for a model set $\mathcal{M}_{\mathrm{mod}}$, tool/verifier set $\mathcal{T}_{\mathrm{tool}}$, and task $t$ by information flow across threads (\Cref{fig:policy_families}): (i)~more samples ($\Pi_{\text{sample}}$), $k$ independent one-call threads combined by a fixed non-model aggregator (such as majority voting or self-consistency); (ii)~more thinking ($\Pi_{\text{think}}$), a single long thread whose later calls read only its own preceding outputs (such as extended chain-of-thought or single-context iterative refinement); and (iii)~more agents ($\Pi_{\text{agent}}$), multi-call schedules in which at least one call reads another thread's output, and may ignore what it reads (including debate, Mixture-of-Agents, orchestrator topologies, or a learned model aggregator over samples). We define the noncommunicating reference class $\Pi_0(\mathcal{M}_{\mathrm{mod}}, \mathcal{T}_{\mathrm{tool}}) \triangleq \Pi_{\text{SA}}$ as the set of policies in which no call reads another thread's output and the final answer is a fixed non-model function of the thread outputs. $\Pi_{\text{sample}}$ and $\Pi_{\text{think}}$ are its two extreme families, and mixtures such as a vote over several multi-call threads also belong to it. Indexing the class by the available models and tools keeps mixed-model voting and external verifiers inside $\Pi_{\text{SA}}$ while excluding cross-thread communication, which prevents the frontier from collapsing onto $\Pi_{\text{agent}}$. A policy in $\Pi_{\text{agent}}$ may choose not to communicate, so the best communicating policy is never below $f$; the coordination effect is therefore defined for a specific system and not for the class. Within $\Pi_{\text{agent}}$, let $\Pi_{\text{agg}}$ be the policies with $k$ independent one-call threads followed by one call that reads all $k$ outputs (a learned selector or synthesizer over samples); it is the smallest communicating policy, and for free-form outputs a multi-round system should be compared with $\Pi_{\text{agg}}$ at equal tokens. Membership in $\Pi_{\text{agent}}$ records architectural capability (the protocol lets a call read another thread's output); realized cross-thread flow is a property of a trace (a message carried evidence that the reader lacked), and the information claims below concern realized flow. For open-ended outputs see \texttt{pace aggregate-open} (\Cref{sec:appendix_cme_cli}). Relative to a declared resource ceiling $\mathbf{B} \in \mathbb{R}_+^6$, the single-agent frontier is
\begin{equation}
f(\mathbf{B}; t, \mathcal{M}_{\mathrm{mod}}, \mathcal{T}_{\mathrm{tool}}) \;\triangleq\; \sup_{\pi \in \Pi_{\text{SA}}: \mathbb{E}[\mathbf{R}(\pi)] \preceq \mathbf{B}} \mathbb{E}\big[\operatorname{Accuracy}(\pi; t)\big].
\label{eq:constrained_frontier}
\end{equation}
When only a subset $\mathcal{J}$ of the axes is matched (below), the relevant frontier is the projected frontier
\begin{equation}
f_{\mathcal{J}}(\mathbf{B}_{\mathcal{J}}) \;\triangleq\; \sup_{\pi \in \Pi_{\text{SA}}: \mathbb{E}[\mathbf{R}(\pi)_{\mathcal{J}}] \preceq \mathbf{B}_{\mathcal{J}}} \mathbb{E}\big[\operatorname{Accuracy}(\pi; t)\big] \;\ge\; f(\mathbf{B}),
\label{eq:projected_frontier}
\end{equation}
which constrains only the matched axes and is never below $f$. The \emph{measured self-consistency frontier} $\hat f_{\mathrm{SC}}(C)$ of \Cref{sec:controlled} (an isotonic fit of the accuracy of $k$ voted samples against measured tokens $C$) is one empirical lower bound on $f_{\mathcal{J}}$ for $\mathcal{J}$ equal to the token axis, not the frontier over all of $\Pi_{\text{SA}}$.
Any empirical baseline $\pi_{\text{baseline}} \in \Pi_{\text{SA}}$ evaluates a lower bound on $f(\mathbf{B})$, so negative gaps remain conservative upper bounds on coordination return; to avoid selection bias from maximizing over multiple baseline settings on one test set, frontiers should be estimated by isotonic regression or split-sample validation. Cached and uncached prompt tokens are logged separately, concurrency assumptions are disclosed for $R_{\text{lat}}$, and $R_{\text{cost}}$ is compared when model prices differ.

Because architectures cannot hold all six dimensions of $\mathbf{R}$ simultaneously equal, an audit declares a matching subset $\mathcal{J} \subseteq \{1,\dots,6\}$ prior to inspecting outcomes and measures the gap against a noncommunicating baseline $\pi_{\text{indep}} \in \Pi_{\text{SA}}$ alongside gaps on every measured axis:
\begin{equation}
\Delta_{\text{comm}}(\pi_{\text{comm}}, \pi_{\text{indep}}; \mathcal{J}, t) \;\triangleq\; \mathbb{E}\big[\operatorname{Accuracy}(\pi_{\text{comm}}; t)\big] \;-\; \mathbb{E}\big[\operatorname{Accuracy}(\pi_{\text{indep}}; t)\big].
\label{eq:pairwise_delta}
\end{equation}
Evaluating sign claims requires only directional resource feasibility on $\mathcal{J}$, not exact equality across all six axes: ruling out a positive coordination return ($\Delta_{\text{comm}} \le 0$) is established by any feasible baseline using no more of the matched resources ($\mathbb{E}[\mathbf{R}(\pi_{\text{indep}})_{\mathcal{J}}] \preceq \mathbb{E}[\mathbf{R}(\pi_{\text{comm}})_{\mathcal{J}}]$), whereas demonstrating a positive coordination return against a specified baseline requires one using at least as much ($\mathbb{E}[\mathbf{R}(\pi_{\text{indep}})_{\mathcal{J}}] \succeq \mathbb{E}[\mathbf{R}(\pi_{\text{comm}})_{\mathcal{J}}]$), and beating it establishes a gain over that specified comparator, not over the frontier. In \citet{choi2025debate}, majority voting over five samples aggregates the round-1 answers of the five debate agents before any debate round occurs, so the baseline is nested inside the debate run ($\mathbf{R}_{\text{vote}} \prec \mathbf{R}_{\text{debate}}$ by construction) without requiring a tolerance band (the interpretation of directional feasibility and its partial identification are in \Cref{app:moved_s3_ident}).

\setcounter{theorem}{-1}\begin{proposition}[Sharp partial-identification bound]\label{prop:manski}
Let $\mathcal{F}_{\mathcal{J}}$ be the observed noncommunicating baselines $\pi_0\in\Pi_{\mathrm{SA}}$ with $\mathbb{E}[\mathbf{R}(\pi_0)_{\mathcal{J}}]\preceq\mathbb{E}[\mathbf{R}(\pi_{\mathrm{comm}})_{\mathcal{J}}]$ and $\Delta_{\mathrm{frontier}}(\pi_{\mathrm{comm}};\mathcal{J})=p_{\mathrm{comm}}-f_{\mathcal{J}}$ (\Cref{eq:projected_frontier}). If $\mathcal{F}_{\mathcal{J}}\neq\emptyset$, then $\Delta_{\mathrm{frontier}}\le\overline{\Delta}_{\mathcal{J}}(\pi_{\mathrm{comm}})\triangleq p_{\mathrm{comm}}-\max_{\pi_0\in\mathcal{F}_{\mathcal{J}}}p(\pi_0)$, and $[\,p_{\mathrm{comm}}-1,\,\overline{\Delta}_{\mathcal{J}}\,]$ is the sharp identified set: without further assumptions the data are compatible with every frontier value between $\max p(\pi_0)$ and $1$. If $\mathcal{F}_{\mathcal{J}}=\emptyset$ the bound is vacuous.
\end{proposition}
\noindent The three adjudication outcomes of \Cref{tab:adjudication} are its three cases (proof in \Cref{sec:appendix_proofs}).

When a multi-agent system $A$ with ceiling $\mathbf{B}_A$ is compared with a greedy single-agent baseline $G$ under ceiling $\mathbf{B}_G \prec \mathbf{B}_A$, the observed raw accuracy delta decomposes through an accounting identity:
\begin{equation}
p_A - p_{G} \;=\; \underbrace{\big[p_A - f(\mathbf{B}_A)\big]}_{\text{Coordination Effect}} \;+\; \underbrace{\big[f(\mathbf{B}_A) - p_{G}\big]}_{\text{Compute Scaling Effect}}.
\label{eq:decomposition}
\end{equation}
This is an accounting identity, not causal identification: the left-hand side is the Category~A contrast, the first term on the right, $p_A - f(\mathbf{B}_A)$, is performance relative to the budget-matched frontier, and in practice an observed sampled baseline $p_{\text{sampled}}$ (Category~B) stands in for $f$. Splitting the compute-scaling term at the vote of the same $K$ samples, with $p_{\mathrm{SC}}(K)$ the accuracy of a $K$-sample vote and $f_{\mathrm{SC}}(k_{\mathrm{tok}})$ the self-consistency frontier at the token-equated pool size of \Cref{eq:ktok}, gives the three-stage telescope identity
\begin{equation}
p_{\mathrm{MAS}}-p_{G}\;=\;\big[p_{\mathrm{SC}}(K)-p_{G}\big]+\big[f_{\mathrm{SC}}(k_{\mathrm{tok}})-p_{\mathrm{SC}}(K)\big]+\big[p_{\mathrm{MAS}}-f_{\mathrm{SC}}(k_{\mathrm{tok}})\big],
\label{eq:telescope}
\end{equation}
the sampling effect at equal agent count, the extra pool at equal tokens, and the coordination effect, which is the bracket tested against the frontier.

The sampling gap is defined as the Category~B delta minus the Category~A delta:
\begin{equation}
\Delta_{\text{gap}} \;\triangleq\; (p_A - p_{\text{sampled}}) - (p_A - p_{G}) \;=\; p_{G} - p_{\text{sampled}}.
\label{eq:sampling_gap_cancel}
\end{equation}
Because $p_A$ cancels, $\Delta_{\text{gap}}$ contains no multi-agent information: it is a sanity check that the accuracy difference shifts by what repeated sampling adds to the single-agent baseline ($p_{\text{sampled}} - p_{G}$), so the headline result is the accuracy difference against the sampled baseline itself, reported alongside inference cost (\Cref{tab:baseline_tiers}, \Cref{fig:empirical_ladder}). Whenever $\mathbf{R}_{\text{baseline}, \mathcal{J}} \preceq \mathbf{R}_{A, \mathcal{J}}$ on the declared matching set, $p_A - p_{\text{baseline}}$ remains a valid upper bound on $p_A - f(\mathbf{B}_A)$ because $f(\mathbf{B}_A) \ge p_{\text{baseline}}$. If $f(\mathbf{B}_A) \ge p_A$, the coordination effect is nonpositive. On the 14 model and benchmark pairs of \citet{choi2025debate} (\Cref{fig:waterfall}, Eqs.~\eqref{eq:decomposition}--\eqref{eq:sampling_gap_cancel}; differences before rounding), single-run accuracy averages $0.670$, majority voting over five samples $0.747$ (a $+0.076$ sampling effect) and decentralized debate ($T=2$, $K=5$) $0.715$, a residual of $-0.031$ against voting that bounds the coordination effect from above (against an apparent $+0.045$ gain over the single-run baseline).

\subsection{Information Partition and Contextual Redundancy}
\label{sec:redundancy}
The decomposition above is silent on when communication could add information. Let $Y$ be the target, $X_{\mathrm{shared}}$ the context every agent receives, and $X_1,\dots,X_K$ the parts of the evidence that only agent $1,\dots,K$ receives. A message is a randomized function of its sender's evidence, $X_{\mathrm{shared}}$ and earlier messages, $\mathcal{H}^{(T)}$ denotes the cumulative transcript after $T$ rounds, and the final answer is a function of $\mathcal{H}^{(T)}$ and independent randomness.

\begin{proposition}[No information gain under redundancy]\label{prop:redundancy}
If $\operatorname{I}(Y;X_1,\dots,X_K\mid X_{\mathrm{shared}})=0$, then for every communication protocol the transcript $\mathcal{H}^{(T)}$ satisfies $\operatorname{I}(Y;\mathcal{H}^{(T)}\mid X_{\mathrm{shared}})=0$, so the Bayes-optimal accuracy of a joint readout of $(X_{\mathrm{shared}}, \mathcal{H}^{(T)})$ equals that of a readout of $X_{\mathrm{shared}}$ alone.
\end{proposition}
\noindent\emph{Proof.} The transcript $\mathcal{H}^{(T)}$ is a function of $(X_{\mathrm{shared}},X_1,\dots,X_K)$ and randomness independent of $Y$. Conditioning on $X_{\mathrm{shared}}$, the data processing inequality gives $\operatorname{I}(Y;\mathcal{H}^{(T)}\mid X_{\mathrm{shared}})\le \operatorname{I}(Y;X_1,\dots,X_K\mid X_{\mathrm{shared}})=0$. \hfill$\square$

The proposition concerns statistical information under Bayes-optimal readout in a closed world without external oracle feedback ($R_{\mathrm{sys}}=0$), not finite-temperature autoregressive computation. A bounded model improves with extra computation on $X_{\mathrm{shared}}$ alone (more samples, longer chains, deliberation), and communication is one way to spend that computation, so the proposition implies no inequality between debate and self-consistency at equal tokens. That inequality is empirical; Principle~1 states it in testable form.

\begin{tcolorbox}[breakable,colback=white,colframe=black!60,boxrule=0.6pt,left=4pt,right=4pt,top=3pt,bottom=3pt]
\textbf{Principle 1 (Contextual Redundancy; empirical form, not a theorem).} When the endowments are redundant, $\operatorname{I}(Y;X_1,\dots,X_K\mid X_{\mathrm{shared}})=0$ (for example, every agent receives the same prompt), communication cannot add information, and we test whether
\begin{equation}
\mathbb{E}\big[\operatorname{Acc}_{\mathrm{Debate}}(C)\big]\;\le\;\mathbb{E}\big[\operatorname{Acc}_{\mathrm{SC}}(C)\big]
\label{eq:redundancy_bound}
\end{equation}
at measured token cost $C$, where $\operatorname{Acc}_{\mathrm{SC}}(C)$ is the self-consistency frontier. A positive premium $\Delta_{\mathrm{comm}}>0$ (\Cref{eq:pairwise_delta}) through pooled evidence requires the joint condition $\operatorname{I}(Y;X_1,\dots,X_K\mid X_{\mathrm{shared}})>0$. The condition is joint because evidence can be synergistic: for independent fair bits $X_1,X_2,X_3$ and $Y=X_1\oplus X_2\oplus X_3$, $\operatorname{I}(Y;X_{1:3})=1$ bit while every pairwise $\operatorname{I}(Y;X_j\mid X_i)=0$. It is necessary for an information gain and not sufficient for a measured premium, because agents must also disclose, transmit, integrate and correctly interpret the partitioned evidence.
\end{tcolorbox}

\paragraph{Sizing the sample pool.}
A corresponding accounting gives the number of single-agent samples whose tokens equal those of one run of a debate. Let one sample cost $c_p+\ell$ tokens (prompt $c_p$, completion $\ell$), and let each agent in rounds $t\ge 2$ read its prompt and only the $K-1$ peer messages of the previous round (rather than the full history of Eq.~\eqref{eq:scaffold_total} in \Cref{app:moved_s3_prop2}). The run then costs $KT(c_p+\ell)+(T-1)K(K-1)\ell$ tokens, which equals
\begin{equation}
k_{\text{tok}} \;=\; KT+(T-1)K(K-1)\,r,\qquad r=\frac{\ell}{c_p+\ell}\in(0,1),
\label{eq:ktok}
\end{equation}
samples (linear in $T$), whereas echoing all $K$ messages from every earlier round as in Eq.~\eqref{eq:scaffold_total} in \Cref{app:moved_s3_prop2} yields $k_{\mathrm{tok}}^{\mathrm{full}} = KT + \tfrac{1}{2}K^2 T(T-1)\,r$ (quadratic in $T$).
\begin{proposition}[Conformity-to-truth threshold]\label{thm:conformity}
Consider $K$ agents whose answers to an item are correct or incorrect, and suppose each round of communication acts on every agent through two constant transition probabilities, independent of how many peers are correct: an incorrect agent becomes correct with probability $\tau_{i\to c}$ and a correct agent becomes incorrect with probability $\tau_{c\to i}$, under the positive-persistence condition $0<\tau_{i\to c}+\tau_{c\to i}\le 1$. Let $p_t$ be the expected per-agent accuracy before round $t$. Then
\begin{equation}
p_{t+1}-p_t=\Delta_{\mathrm{round}}(p_t),\qquad \Delta_{\mathrm{round}}(p)=(1-p)\,\tau_{i\to c}-p\,\tau_{c\to i},
\label{eq:delta_round}
\end{equation}
and $\Delta_{\mathrm{round}}(p_1)<0$ if and only if
\begin{equation}
p_1>p^{*}\;\triangleq\;\frac{\tau_{i\to c}}{\tau_{i\to c}+\tau_{c\to i}},
\label{eq:p_star}
\end{equation}
which equals $\kappa/(1+\kappa)$ with $\kappa\triangleq \tau_{i\to c}/\tau_{c\to i}$ for $\tau_{c\to i}>0$, and $p^{*}=1$ when $\tau_{c\to i}=0$. In addition, $p_t-p^{*}=(1-\tau_{i\to c}-\tau_{c\to i})^{t-1}(p_1-p^{*})$ with $0\le 1-\tau_{i\to c}-\tau_{c\to i}<1$, so convergence is monotone: if $p_1>p^{*}$ then $p^{*}\le p_{t+1}\le p_t\le p_1$ for every $t$, and if $p_1<p^{*}$ then $p_1\le p_t\le p_{t+1}\le p^{*}$.
\end{proposition}

A debate beats the token-equated sample frontier only if its per-agent accuracy clears a hurdle that grows with the correlation of its answers (\Cref{thm:hurdle}); 
under peer imitation a vote over $K$ agents collapses toward the accuracy of one draw (\Cref{prop:coalescent}); and a gain from split evidence needs each agent's private criteria to be disclosed, which happened in only $\hpDiscD$ of items against $\hpDiscQPowTen$ expected if the ten agent-criterion disclosures were independent (\Cref{prop:disclosure}), so debate stays on the decoy when $d<1$. The statements are in \Cref{app:moved_s3_formal} and the proofs in \Cref{sec:appendix_proofs}.

\paragraph{Three routes to positive returns.} Communication can raise accuracy when one of three conditions holds (\Cref{app:moved_s3_routes}): (I)~information partitioning with $\operatorname{I}(Y;X_{1:K}\mid X_{\mathrm{shared}})>0$; (II)~asymmetric external verification (counted in $R_{\mathrm{sys}}$) accepting correct edits at rate $1-\epsilon_{\mathrm{FN}}$ and wrong edits at $\epsilon_{\mathrm{FP}}<\kappa\frac{1-p_1}{p_1}(1-\epsilon_{\mathrm{FN}})$, which pushes $p^{*}\to 1$ as $\epsilon_{\mathrm{FP}}\to 0$ (\Cref{cor:verifier}); or (III)~cross-model error decorrelation ($\rho_{\mathrm{het}}<\rho_{\mathrm{hom}}$), which raises $K_{\mathrm{eff}}$ (\Cref{prop:keff}).

\subsection{Four-Category Baseline Taxonomy}
\label{sec:regimes}
To separate sample multiplicity from compute control, empirical comparisons fall into four baseline Categories, A to D:
(1)~Category~A (greedy single-pass baselines): Multi-agent architectures are compared against a single forward pass ($K=1, T=1$), which confounds coordination with sample count and test-time token expenditure. (2)~Category~B (sampled baselines): The baseline aggregates $k \ge 2$ independent samples without communication (for example self-consistency under majority voting), with $k$ equal to the number of agents, as in \citet{choi2025debate}, or not; no resource is verified equal, so tokens, latency and cost are unconstrained, and the category tests whether communication adds anything beyond sample multiplicity. (3)~Category~C (resource-matched or shared-ceiling baselines): A declared resource constraint or shared ceiling is enforced across conditions, such as parity in thinking tokens ($R_{\text{reason}}^{\text{MAS}} = R_{\text{reason}}^{\text{SAS}}$), shared iteration or generation caps (bounds on the number of calls $N_{\text{calls}}$, which is logged separately from $R_{\text{sample}}$), or matched monetary budgets ($R_{\text{cost}}^{\text{MAS}} = R_{\text{cost}}^{\text{SAS}}$). (4)~Category~D (Pareto frontier sweeps) traces accuracy against inference expenditure over a parametric sweep and has two forms: \emph{D-measured}, an item-level pool evaluated at the measured cost of the system (the frontier of \Cref{sec:controlled}), and \emph{D-reported}, a published Pareto plot read at comparable compute without a per-item match (for example \citet{wunderlich2026mixture}). Categories A to D classify comparison designs by what was matched; they are not a strict ordinal ranking of baseline strength.
The comparator also depends on the evidence structure: with shared evidence it is a vote over $k$ samples, with split evidence a vote over single-view samples, and an external verifier, when present, is given to both arms ($R_{\mathrm{sys}}$).

\begin{table}[!htbp]
\centering
\scriptsize
\setlength{\tabcolsep}{4pt}
\renewcommand{\arraystretch}{1.14}
\caption{What PACE can establish: the admissible claim and the excluded inference for each kind of available baseline and resource evidence.}
\label{tab:what_pace_establishes}
\begin{tabularx}{\textwidth}{@{} >{\raggedright\arraybackslash}p{0.21\textwidth} >{\raggedright\arraybackslash}p{0.37\textwidth} >{\raggedright\arraybackslash}X @{}}
\toprule
\textbf{Available Evidence} & \textbf{Appropriate Claim} & \textbf{Excluded Inference} \\
\midrule
Greedy baseline only (Category~A) & Reports overall accuracy difference over one greedy forward pass ($\Delta_A$). & Cannot separate communication from repeated sampling or compute expansion; cannot rule out parity at matched budgets. \\
Sampled baseline (Category~B) & Tests whether communication improves over $k$ independent draws; rules out a positive frontier gap when $\mathbf{R}_{\mathrm{base},\mathcal{J}} \preceq \mathbf{R}_{\mathrm{MAS},\mathcal{J}}$ and $\Delta \le 0$. & Cannot establish token, latency, or monetary parity when unmeasured; cannot certify coordination gains if the baseline used fewer tokens. \\
Named resource control (Category~C) & Establishes performance under the declared ceiling or iteration cap $\mathbf{B}_{\mathcal{J}}$ relative to the projected frontier $f_{\mathcal{J}}(\mathbf{B}_{\mathcal{J}})$. & Cannot generalize to uncontrolled axes outside $\mathcal{J}$ (such as latency or dollar cost) or extrapolate beyond the tested cap. \\
Measured frontier (Category~D and directional $\hat{f}_{\mathrm{SC}}^-, \hat{f}_{\mathrm{SC}}^+$) & Estimates the frontier gap at matched token cost with paired bootstrap uncertainty, using $\hat{f}_{\mathrm{SC}}^+$ to test positive gains and $\hat{f}_{\mathrm{SC}}^-$ to rule gains out. & Cannot identify the full frontier across all six dimensions if other axes differ, nor attribute gains to message content without placebo checks; points past $C_{\max}$ are flagged as extrapolated. \\
Incomplete resource report & Documents measured dimensions with \texttt{null} for unmeasured axes and states partial identification bounds where feasible baselines exist. & Cannot claim Pareto dominance, net cost savings, or resource parity on unreported axes; unobserved dimensions remain unidentified. \\
Execution traces per item & Measures transition rates ($\tau_{i\to c}, \tau_{c\to i}$), excess correlation ($\Delta\rho_{\mathrm{debate}}$), disclosure rate ($d$), and semantic difference ($\Delta_{\mathrm{arch}}$). & Cannot claim causal semantic persuasion without a character-length-matched placebo control, nor treat plug-in $\hat{p}^*$ as independent without a control log with no communication. \\
\bottomrule
\end{tabularx}
\end{table}

\subsection{The PACE Protocol}
\label{sec:cme}

\def\paceShow{box}\providecommand{\paceShow}{both}\def\paceShowBox{box}\def\paceShowTable{table}
\ifx\paceShow\paceShowTable\else
\ifcsname c@pacealg\endcsname\else\newcounter{pacealg}\fi
\begin{tcolorbox}[colback=white,colframe=black!65,boxrule=0.6pt,left=6pt,right=6pt,top=4pt,bottom=4pt]
\refstepcounter{pacealg}
\ifcsname Cref\endcsname\label[algorithm]{alg:pace_protocol}\else\label{alg:pace_protocol}\fi
\small
\textbf{Algorithm~\thepacealg: The PACE Evaluation Procedure.}
\begin{enumerate}\setlength\itemsep{2pt}\setlength\parskip{0pt}
    \item \textbf{Specify.} Fix the task $t$ (items $N$, shared or split evidence), the models $\mathcal{M}_{\mathrm{mod}}$ and the tools $\mathcal{T}_{\mathrm{tool}}$, identical for both arms.
    \item \textbf{Declare and log.} Predeclare the matched subset $\mathcal{J}\subseteq\{1,\dots,6\}$ before outcomes are seen; log all six dimensions of $\mathbf{R}$, with independent trajectories ($R_{\mathrm{sample}}$) apart from total calls ($N_{\mathrm{calls}}$) and search tokens ($B_{\mathrm{search}}$).
    \item \textbf{Choose a feasible baseline $\pi_0\in\Pi_{\mathrm{SA}}$.} To claim a gain, $\mathbb{E}[\mathbf{R}(\pi_0)_{\mathcal{J}}]\succeq\mathbb{E}[\mathbf{R}(\pi_{\mathrm{MAS}})_{\mathcal{J}}]$ (for example $k\ge k_{\mathrm{tok}}$); to rule one out, $\preceq$ suffices.
    \item \textbf{Estimate the frontier.} Pool accuracy over $k=1\dots K$ draws: the isotonic curve $\hat f_{\mathrm{SC}}$ is primary; $\hat f^{+}_{\mathrm{SC}}$ (monotone envelope) is the estimator used for positive claims and $\hat f^{-}_{\mathrm{SC}}$ (two-fold cross-fit) for nonpositive ones.
    \item \textbf{Bootstrap and control.} Paired bootstrap over items, refitting the frontier in each draw, gives $\Delta_{\mathrm{frontier}}$ and its interval; content claims add $\Delta_{\mathrm{arch}}=p_{\mathrm{comm}}-p_{\mathrm{placebo}}$ and route controls (\Cref{tab:route_tests}).
    \item \textbf{Report} the PACE Card (\Cref{tab:pace_card}), with \texttt{null} for unmeasured axes.
\end{enumerate}
\end{tcolorbox}
\fi
\ifx\paceShow\paceShowBox\else
\begin{table}[!htbp]
\centering
\scriptsize
\setlength{\tabcolsep}{4pt}
\renewcommand{\arraystretch}{1.14}
\caption{\textbf{Hypothetical worked example (illustrative only; not empirical data).} A protocol with $K=3$ agents, $T=2$ rounds, shared task prompt ($c_p = 400$ tokens), completion length $\ell = 100$ tokens ($r = 0.20$), peer message exchange after round~1 ($K-1 = 2$ peer messages read per agent in round~2), and one fixed majority vote aggregation step, mapped across architecture flow, resource accounting across six dimensions, and comparator tiers.}
\label{tab:pace_worked_example}
\begin{tabularx}{\textwidth}{@{} >{\raggedright\arraybackslash}p{0.25\textwidth} >{\raggedright\arraybackslash}p{0.31\textwidth} >{\raggedright\arraybackslash}X @{}}
\toprule
\multicolumn{3}{@{}l}{\textit{Panel A: Architecture and cross-thread flow (hypothetical $K=3$, $T=2$ protocol)}} \\
\midrule
\textbf{Stage} & \textbf{Information read per call} & \textbf{Calls and token accounting} \\
Round~1 (independent draws) & Shared task prompt ($400$ tok) only & $3$ independent threads ($3 \times 500 = 1{,}500$ tok); $R_{\mathrm{sample}} = 3$. \\
Round~2 (peer message exchange) & Task prompt ($400$ tok) plus $2$ peer messages ($200$ tok) & $3$ cross-thread calls ($3 \times 700 = 2{,}100$ tok); class $\Pi_{\mathrm{agent}}$. \\
Aggregation step (fixed vote) & Final round outputs of all $3$ threads & $0$ model tokens ($N_{\mathrm{calls}} = 6$, $C_{\mathrm{MAS}} = 3{,}600$ tok, $k_{\mathrm{tok}} = 7.20$); hypothetical accuracy $0.700$. \\
\midrule
\multicolumn{3}{@{}l}{\textit{Panel B: Resource accounting vector $\mathbf{R}$ across six dimensions (hypothetical values; unmeasured axes stored as \texttt{null})}} \\
\midrule
\textbf{Dimension} & \textbf{Hypothetical logged value} & \textbf{Verification status and convention} \\
$R_{\mathrm{reason}}$ (thinking tokens) & \texttt{null} (unmeasured) & Unreported on standard nonreasoning endpoint (never imputed as $0$). \\
$R_{\mathrm{token}}$ (total tokens) & $3{,}600$ tokens per item & Unmatched (measured) across prompt plus completion tokens ($k_{\mathrm{tok}} = 7.20$). \\
$R_{\mathrm{cost}}$ (monetary cost) & \texttt{null} (unmeasured) & Unreported when local inference pricing is omitted. \\
$R_{\mathrm{lat}}$ (execution latency) & $4.2$ s per item (two sequential rounds) & Unmatched (measured) under concurrent round execution. \\
$R_{\mathrm{sample}}$ (independent draws) & $3$ trajectories ($N_{\mathrm{calls}} = 6$ total calls) & Independent round 1 draws recorded separately from total calls. \\
$R_{\mathrm{sys}}$ (external tool calls) & $0$ tool or verifier calls & Controlled ($0$ tool access on both communicating and baseline arms). \\
\midrule
\multicolumn{3}{@{}l}{\textit{Panel C: Greedy versus sampled versus frontier comparison (hypothetical accuracies; not empirical results)}} \\
\midrule
\textbf{Baseline comparator (rung)} & \textbf{Tokens and hypothetical accuracy} & \textbf{Directional PACE adjudication} \\
Greedy single pass (Category~A) & $500$ tokens; accuracy $0.600$ & $\Delta_A = +0.100$: confounded with sample multiplicity and $7.2\times$ tokens. \\
Sampled vote, $k=3$ (Category~B) & $1{,}500$ tokens; accuracy $0.680$ & $\Delta_B = +0.020$: infeasible to credit gain ($1{,}500 < 3{,}600$ tokens). \\
Sampled frontier, $k=8 \ge k_{\mathrm{tok}}$ (Category~D) & $4{,}000$ tokens ($k=7$ uses $3{,}500$); $\hat{f}_{\mathrm{SC}}^-(3{,}600) = 0.720$ & $\Delta_{\mathrm{frontier}} = 0.700 - 0.720 = -0.020$: system not ahead of the lower frontier. \\
\bottomrule
\end{tabularx}
\end{table}
\fi

\section{Re-Auditing Nine Published Studies}
\label{sec:meta}

This section re-reads the tables of nine published studies and asks what they show once the baseline of each is made explicit. The studies and their extraction, the selection and screening procedure, the evidence flow, the pairing rules and the per-study results are in \Cref{app:moved_s5_nine,app:moved_s5_bounded,app:moved_s5_flow,app:moved_s5_pairing,app:moved_s5_two_baselines}. 
\phantomsection\label{sec:multitier}Because contrasts are unevenly distributed (2 to 180 unique configurations per study; \Cref{tab:appB_ledger}), we weight outcomes by contrast and equally by study (\Cref{tab:baseline_tiers}), omit one study at a time (\Cref{app:sensitivity}), and use three analytical units: (i) the 474 Category~B/C comparison records (\Cref{tab:baseline_tiers} and \Cref{fig:empirical_ladder}), (ii) the 516 unique multi-agent configurations adjudicated by directional resource feasibility (\Cref{tab:adjudication}), and (iii) the 318 linked Category~A and Category~B pairs spanning 72 unique $(\mathrm{study},\mathrm{benchmark},\mathrm{model})$ baseline cells.

\begin{figure}[t]
\centering
\includegraphics[width=\textwidth]{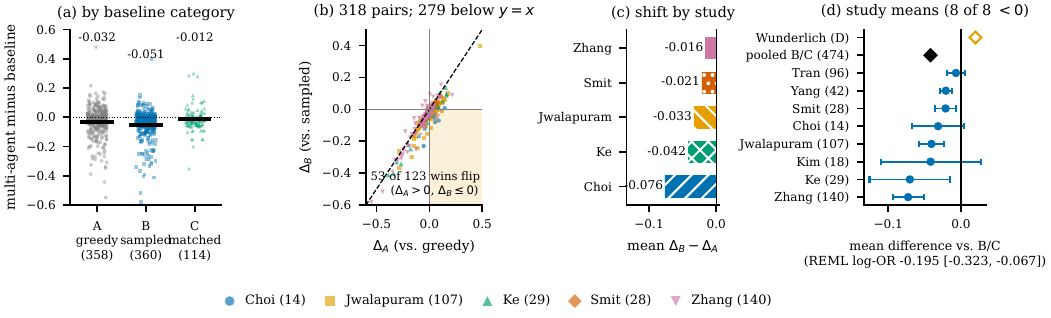}
\caption{The baseline shift. (a) Accuracy difference of each configuration against greedy (A), sampled (B) and resource-matched (C) baselines, bars at means; (b) the \dualPairs{} paired contrasts with both baselines, where \numHeroApparentWins{} of the \numHeroPositiveA{} with $\Delta_A>0$ become nonpositive against the sampled baseline and \dualBelow{} lie below $\Delta_B=\Delta_A$; (c) pair-level mean shift $\Delta_B-\Delta_A$ by study, negative in \dualStudiesNegShift{} of 5; (d) Category B/C study means with 95\% intervals (contrasts treated as independent), the pooled B/C mean and the Category D study. The headline sampling gap (\meanSamplingGapCells) averages the \numCellsDual{} deduplicated baseline cells with studies weighted equally.}
\label{fig:baseline_shift}
\end{figure}

\paragraph{Sampling gap.} For the \numStudiesDual\ studies that report both baselines (\numContrastsDual\ paired contrasts), subtracting $\Delta$ vs.\ Category~A from $\Delta$ vs.\ Category~B gives study means ranging from $-0.016$ (Zhang et al.) to $-0.076$ (Choi et al.; \Cref{fig:baseline_shift}c). The sampling gap is below zero in all \numStudiesDual\ studies. Because several multi-agent arms share one baseline, the canonical estimate deduplicates them into the \numCellsDual\ unique baseline cells: \meanSamplingGapCells\ with studies weighted equally, with \numNegCellsDual\ cells negative. As sensitivity checks, the mean over pairs is \meanSamplingGap\ weighted equally by study, \meanSamplingGapNoZhang\ without Zhang et al.\ and \meanSamplingGapContrast\ weighted by contrast, and the cell-weighted mean is \meanSamplingGapCellPooled. Yang et al., Tran and Kiela, and Kim et al.\ report only a sampled or matched baseline, so no sampling gap can be computed for them. The Jwalapuram et al.\ CoT-SC row carries the upper-bound baseline caveat noted in \Cref{sec:conclusion} and \Cref{app:moved_s5_denoms}. Of the \numHeroPositiveA\ pairs with a positive $\Delta_A$, \numHeroApparentWins\ lose it against the sampled baseline, and 15 of those \numHeroApparentWins\ have $\Delta_A$ above $1.96$ paired binomial standard errors at $\varphi=0.50$ (11 to 25 for $\varphi$ from $0$ to $0.9$), so most of the apparent wins were within noise before the baseline changed.

Against a single greedy pass (Category~A) the contrast mean is \meanCatA; against sampled baselines (Category~B) \meanCatB; against resource-matched baselines (Category~C) \meanCatC; and across the combined Category~B/C pool \meanCatBC, with the multi-agent system not ahead in \pctNotAheadBC\% of contrasts and a feasible baseline matching or beating it in \pctFeasibleBaselineWins\% (per-study verdicts and pooled rows in \Cref{tab:pace_verdict}; means by baseline category in \Cref{app:moved_s5_headline,app:moved_s5_tiers_table}). All eight Category~B/C study means are negative, and the one Category~D study, Wunderlich et al., which reads Pareto fronts without a per-item compute match, has a mean of $+0.020$. At $\varphi=0.50$, \numNoiseAboveBC{} of \numContrastsCatBC{} Category~B/C contrasts ($7.4\%$) exceed $+1.96\,\mathrm{SE}$ versus \numNoiseBelowBC{} below $-1.96\,\mathrm{SE}$. Omitting Zhang et al.\ gives \meanCatBCNoZhang{}, and baselines dominate on accuracy and cost in \numParetoDominated{} of 105 Jwalapuram et al.\ contrasts. \begin{table}[!htbp]
\centering
\scriptsize
\setlength{\tabcolsep}{3pt}
\caption{PACE verdict table: per study, the matching condition, most informative available baseline category (categories classify designs, not a strict ranking), $m$ configurations (Zhang et al.: 180 against greedy, 140 also against self-consistency), mean difference $\bar\Delta$ against Category~A and B/C baselines (D for Wunderlich et al.) and the adjudication counts (not ahead of a feasible baseline / ahead / undetermined). Pooled rows give the REML log odds (B9) with CR2 interval, $I^2$ and 95\% prediction interval; the exact wild cluster bootstrap over all $2^8$ sign patterns \citep{joshi2022cluster} gives $p=2/256=0.0078$ (\Cref{app:certification}).}
\label{tab:pace_verdict}
\begin{tabular*}{\textwidth}{@{\extracolsep{\fill}}llcrrrr@{}}
\toprule
Study & Matching condition & Category & $m$ & $\bar\Delta$ vs.\ A & $\bar\Delta$ vs.\ B/C/D & Not ahead / ahead / undet. \\
\midrule
\multicolumn{7}{@{}l}{\textit{Panel A. Study level}} \\
Tran and Kiela & Thinking tokens & C & 96 & --- & $-0.007$ & 64 / 32 / 0 \\
Kim et al. & Iterations & C & 18 & --- & $-0.042$ & 13 / 5 / 0 \\
Jwalapuram et al. & Cost audited & B & 107 & $-0.007$ & $-0.041$ & 75 / 7 / 25 \\
Choi et al. & Agent count & B & 14 & $+0.045$ & $-0.031$ & 8 / 0 / 6 \\
Wunderlich et al. & Comparable (Pareto) & D & 2 & --- & $+0.020$ & 0 / 0 / 2 \\
Ke et al. & Unreported & B & 29 & $-0.028$ & $-0.070$ & 14 / 0 / 15 \\
Yang et al. & Approx. (16 samples) & B & 42 & --- & $-0.021$ & 35 / 7 / 0 \\
Zhang et al. & Unmatched / unreported & B & 180 (140) & $-0.058$ & $-0.073$ & 131 / 0 / 49 \\
Smit et al. & Unreported & B & 28 & $+0.000$ & $-0.021$ & 16 / 0 / 12 \\
\midrule
\multicolumn{7}{@{}l}{\textit{Panel B. Configuration-level adjudication and pooled estimates}} \\
\multicolumn{2}{l}{All nine studies (configurations)} & & \numContrastsTotal & --- & --- & \numConfigFeasible{} / \numConfigAhead{} / \numConfigUndet \\
\midrule
\multicolumn{2}{l}{Category~A, greedy single pass} & A & \numContrastsCatA & \meanCatA & --- & \\
\multicolumn{2}{l}{Category~B, sampled baselines} & B & \numContrastsCatB & --- & \meanCatB & \\
\multicolumn{2}{l}{Category~C, resource matched} & C & \numContrastsCatC & --- & \meanCatC & \\
\multicolumn{2}{l}{Category~B/C pool, contrast mean} & B/C & \numContrastsCatBC & --- & \meanCatBC & \pctNotAheadBC\% not ahead \\
\multicolumn{2}{l}{Pooled B/C, REML (B9), log odds} & B/C & \numContrastsCatBC & --- & \remlBNineBeta & CR2 [\remlBNineCrLo, \remlBNineCrHi] \\
\multicolumn{2}{l}{Heterogeneity of B9 ($I^2$)} & & & & $\remlBNineITwo\%$ & PI [\remlBNinePiLo, \remlBNinePiHi] \\
\bottomrule
\end{tabular*}
\end{table}

Adjudication details, noise bands, the dependence on Zhang et al., moderator results, denominators, the ladder of study means, the explanation of the disagreement among studies and the capability-threshold analysis are in \Cref{app:moved_s5_headline,app:moved_s5_sens,app:moved_s5_zhang_mod,app:moved_s5_denoms,app:moved_s5_ladder_fig,app:moved_s5_disagree,app:moved_s5_threshold}.

\paragraph{Scope of the corpus.} Tran and Kiela's Sequential arm (\numTranSeq{} contrasts) is single-context iterative refinement, that is $\Pi_{\mathrm{think}}$ in \Cref{sec:decomposition}; it stays in Category~C because its thinking-token budget is matched, and omitting it moves the Category~B/C contrast mean from \meanCatBC{} to \meanBCNoTranSeq{} (\numBCNoTranSeq{} contrasts) and Tran and Kiela's study mean from \meanTranStudy{} to \meanTranNoSeq{}, so the statement that all eight Category~B/C study means are negative holds for the declared slice. Kim et al.\ enter only through Table~16 of the arXiv version (\numContrastsKim{} contrasts on SWE-bench Verified and Terminal-Bench), the one place that reports single-agent and centralized accuracy for each cell at matched iterations; the abstract of the arXiv preprint gives 180 configurations in versions 1 and 2 and 260 in version 3, so statements about the corpus concern what we extracted, not those papers in full. Six of the nine studies are skeptical audits, Kim et al.\ is a scaling study, and the two sources with positive results are Wunderlich et al.\ ($+0.020$) and the introducing rows of Ke et al.\ ($+0.089$ against their own baselines, against $-0.024$ for the same method in a later audit, \Cref{tab:original_vs_audit}), so the corpus is not a random sample of the literature. Accordingly, the pooled estimates describe this purposively assembled corpus and do not estimate the prevalence of coordination gains across the literature.

\paragraph{Multilevel estimate.} A three-level REML model (contrasts within studies) on the log odds of the Category~B/C pool gives \remlBNineBeta{} (CR2 interval [\remlBNineCrLo, \remlBNineCrHi]) at the working correlation $\varphi=0.50$, with $\nu=\remlBNineNu$ degrees of freedom, $I^2_{(2)}=\remlBNineITwo\%$, and a new-study prediction interval of [\remlBNinePiLo, \remlBNinePiHi]; on the \numCellsDual{} unique dual-baseline cells the sampling-gap model gives \remlGapCellsBeta. The model is exploratory because contrasts share baselines and matching conditions (\Cref{sec:reml_main}). The negative Category~B/C direction persists across all five topologies ($-0.015$ to $-0.059$), four domains ($-0.031$ to $-0.075$), reasoning vs.\ instruction models ($-0.028$ vs.\ $-0.053$), and all scales (\Cref{sec:appendix_moderators}), with all 288 specification-curve variants negative (286 CR2 upper bounds $<0$, \Cref{fig:spec_curve}) and leave-one-study-out estimates in [\remlBNineLooMinBeta{}, \remlBNineLooMaxBeta{}] (wild bootstrap $p=0.0078$, \Cref{tab:wild_bootstrap}).

\providecommand{\hpThirtySentence}{}\providecommand{\hpThirtyProvisional}{provisional until cell HP30 is run}\providecommand{\hpThirtyCaveat}{the post hoc 30-call split-view voting comparator (HP30) is not part of these results}

\providecommand{\confNA}{n/a}
\providecommand{\confRegStatus}{[NEEDS\_VERIFICATION: registration status of the design]}
\providecommand{\confPrimaryComplete}{0}
\providecommand{\confOneGap}{\confNA}\providecommand{\confOneLo}{\confNA}\providecommand{\confOneHi}{\confNA}\providecommand{\confOneUpper}{\confNA}
\providecommand{\confOneN}{\confNA}\providecommand{\confOneP}{\confNA}\providecommand{\confOnePholm}{\confNA}\providecommand{\confOneReject}{\confNA}
\providecommand{\confTwoGap}{\confNA}\providecommand{\confTwoLo}{\confNA}\providecommand{\confTwoHi}{\confNA}\providecommand{\confTwoUpper}{\confNA}
\providecommand{\confTwoN}{\confNA}\providecommand{\confTwoP}{\confNA}\providecommand{\confTwoPholm}{\confNA}\providecommand{\confTwoReject}{\confNA}
\providecommand{\confThreeGap}{\confNA}\providecommand{\confThreeLo}{\confNA}\providecommand{\confThreeHi}{\confNA}\providecommand{\confThreeUpper}{\confNA}
\providecommand{\confThreeN}{\confNA}\providecommand{\confThreeP}{\confNA}\providecommand{\confThreePholm}{\confNA}\providecommand{\confThreeReject}{\confNA}
\providecommand{\confFourGap}{\confNA}\providecommand{\confFourLo}{\confNA}\providecommand{\confFourHi}{\confNA}\providecommand{\confFourUpper}{\confNA}
\providecommand{\confFourN}{\confNA}\providecommand{\confFourP}{\confNA}\providecommand{\confFourPholm}{\confNA}\providecommand{\confFourReject}{\confNA}
\providecommand{\confReasonShareMmlu}{\confNA}\providecommand{\confReasonCharsMmlu}{\confNA}\providecommand{\confCommCharsMmlu}{\confNA}
\providecommand{\confReasonShareHp}{\confNA}\providecommand{\confReasonCharsHp}{\confNA}\providecommand{\confCommCharsHp}{\confNA}
\providecommand{\hpVoteAcc}{\confNA}\providecommand{\hpVoteTok}{\confNA}\providecommand{\hpDebateAcc}{\confNA}\providecommand{\hpDebateTok}{\confNA}
\providecommand{\hpFrontier}{\confNA}\providecommand{\hpGapEst}{\confNA}\providecommand{\hpGapLo}{\confNA}\providecommand{\hpGapHi}{\confNA}\providecommand{\hpN}{\confNA}

\providecommand{\mmluGreedyAcc}{\confNA}\providecommand{\mmluGreedyTok}{\confNA}\providecommand{\mmluDebateAcc}{\confNA}\providecommand{\mmluDebateTok}{\confNA}\providecommand{\mmluDebateFrontier}{\confNA}\providecommand{\mmluMoaAcc}{\confNA}\providecommand{\mmluMoaTok}{\confNA}

\providecommand{\hpDiscRate}{\confNA}\providecommand{\hpDiscD}{\confNA}\providecommand{\hpDiscNOne}{\confNA}\providecommand{\hpDiscAOne}{\confNA}\providecommand{\hpDiscAZero}{\confNA}\providecommand{\hpDiscPlAOne}{\confNA}\providecommand{\hpDiscPlAZero}{\confNA}
\providecommand{\hpPlaceboAcc}{\confNA}

\providecommand{\confOneUpperMin}{\confNA}\providecommand{\confOneUpperMax}{\confNA}\providecommand{\mmluSampleTok}{\confNA}\providecommand{\mmluKtok}{\confNA}\providecommand{\mmluSFourTok}{\confNA}\providecommand{\mmluKtokRatio}{\confNA}\providecommand{\mmluThinkTok}{\confNA}\providecommand{\hpSampleTok}{\confNA}\providecommand{\hpKtok}{\confNA}\providecommand{\hpKtokRatio}{\confNA}\providecommand{\hpPlaceboTokFull}{\confNA}\providecommand{\hpPlaceboTokPct}{\confNA}\providecommand{\confDsCapMmlu}{\confNA}\providecommand{\confDsCallsMmlu}{\confNA}\providecommand{\confDsCapHp}{\confNA}\providecommand{\confDsCallsHp}{\confNA}

\providecommand{\confDsParseMmlu}{\confNA}\providecommand{\confDsParseOfMmlu}{\confNA}\providecommand{\confDsParseHp}{\confNA}\providecommand{\confDsParseOfHp}{\confNA}

\providecommand{\confOneUpperMinPP}{\confNA}\providecommand{\confOneUpperMaxPP}{\confNA}\providecommand{\confOneKTwentyFour}{\confNA}
\providecommand{\mmluKtokExact}{\confNA}\providecommand{\lamHurdleZero}{\confNA}\providecommand{\lamHurdleQuarter}{\confNA}\providecommand{\lamHurdleHalf}{\confNA}\providecommand{\decoyRoneShare}{\confNA}\providecommand{\decoyUnanimousRone}{\confNA}\providecommand{\decoyLeaveFourDebate}{\confNA}\providecommand{\decoyLeaveFourPlacebo}{\confNA}\providecommand{\decoyFinalShareDebate}{\confNA}\providecommand{\decoyFinalSharePlacebo}{\confNA}\providecommand{\hpDiscQPowSix}{\confNA}\providecommand{\hpDiscQPowTen}{\confNA}
\providecommand{\preregShaMain}{\confNA}\providecommand{\preregShaAddOne}{\confNA}

\section{A Controlled Test at Measured Token Cost}
\label{sec:controlled}

\subsection{Design}
\label{sec:controlled_design}
Strategies are greedy decoding (S1), a pool of 16 independent samples with plurality voting (S2), one call with more thinking (S3), vanilla debate among five copies of one model for three rounds (S4), debate with confidence-weighted aggregation (S5), debate in which each peer message is replaced by a sample from a different item, truncated or repeated to match the character length of the message it replaces, not token-matched (S7), and a two-layer mixture-of-agents aggregator (S8). Tokens of a strategy are the prompt and completion tokens of every call it used, including the reused first-round calls. The estimand is the frontier gap: a strategy's accuracy minus the measured self-consistency frontier $\hat f_{\mathrm{SC}}(C)$ at its mean measured tokens, where $\hat f_{\mathrm{SC}}$ is the nondecreasing fit of the accuracy of $k$ voted samples ($k=1$ to $16$) against $k$ times the mean tokens per sample, interpolated linearly in tokens. Intervals are 95\% paired bootstrap intervals over items, and the four confirmatory hypotheses use a Holm correction. The confirmatory experiment uses one open-weight model, \texttt{llama3.1:8b}, run locally; exploratory cells add \texttt{deepseek-r1:14b}; the positive control is a synthetic task in which the evidence is split across agents (\Cref{app:controlled}; protocol deviations and post hoc extensions in \Cref{app:moved_s6_deviations}). The frontier is read at each strategy's mean tokens, which can lie beyond the pool: vanilla debate on MMLU-Pro uses \mmluSFourTok{} tokens, equal to about \mmluKtok{} pool samples of \mmluSampleTok{} tokens and so \mmluKtokRatio{} times beyond the 16-sample pool (the post hoc pool of 24 covers it, gap \confOneGap{} becoming \confOneKTwentyFour{}; mixture-of-agents at \mmluMoaTok{} tokens lies inside the pool), and split-view debate on Hidden Profile uses \hpDebateTok{} tokens, about \hpKtok{} samples of \hpSampleTok{} tokens and so about \hpKtokRatio{} times beyond the five-sample split-view pool. H3 is weak for this model: \texttt{llama3.1:8b} has no thinking mode, and more thinking (S3) uses fewer tokens than greedy decoding (\mmluThinkTok{} against \mmluGreedyTok), so evidence on thinking comes from Tran and Kiela.

\subsection{Confirmatory results}
\label{sec:controlled_primary}
\begin{table}[t]
\centering
\scriptsize
\setlength{\tabcolsep}{2.5pt}
\caption{Prespecified contrasts on \texttt{llama3.1:8b} (H1, H2, H4 confirmatory; H3 exploratory but in the four-test Holm family): gap to the self-consistency frontier at equal measured tokens with its 95\% paired bootstrap interval, the directional gap ($\hat f^{-}$ for nonpositive claims, $\hat f^{+}$ for H2) and the one-sided Holm-adjusted $p$ for $H_a$: gap $>0$; H2 is an upper bound against a lower-budget pool.}
\label{tab:confirmatory_primary}
\begin{tabular}{@{}llrrrrr@{}}
\toprule
Hyp.\ & Cell and contrast & Estimate & 95\% interval & Directional gap & $n$ & $p_{\text{Holm}}$ \\
\midrule
H1 & MMLU-Pro: debate (S4) vs.\ frontier & \ensuremath{-0.015} & [\ensuremath{-0.065}, \ensuremath{+0.035}] & $-0.012$ [$-0.065,+0.036$] ($\hat f^{-}$) & 250 & 1.000 \\
H2 & Hidden Profile: split debate (S4v) vs.\ 5-view frontier & \ensuremath{+0.073} & [\ensuremath{+0.024}, \ensuremath{+0.122}] & $+0.035$ [$-0.013,+0.082$] ($\hat f^{+}$) & 200 & 0.007 \\
H3 & MMLU-Pro: gap(S4) minus gap(S3) (exploratory) & \ensuremath{-0.015} & [\ensuremath{-0.079}, \ensuremath{+0.048}] & $-0.018$ [$-0.082,+0.045$] ($\hat f^{-}$) & 250 & 1.000 \\
H4 & MMLU-Pro: MoA (S8) vs.\ frontier (intersection test) & \ensuremath{-0.072} & [\ensuremath{-0.118}, \ensuremath{-0.028}] & $-0.070$ [$-0.113,-0.028$] ($\hat f^{-}$) & 250 & 1.000 \\
\bottomrule
\end{tabular}
\end{table}

\Cref{tab:confirmatory_primary} lists the four prespecified contrasts with their estimates and Holm-adjusted $p$-values; $p$ is one-sided for $H_a$: gap $>0$. For H1, the frontier gap of vanilla debate on MMLU-Pro is \confOneGap{} (95\% interval [\confOneLo, \confOneHi], $n=\confOneN$), which excludes advantages larger than \confOneUpper{} percentage points; across the three frontier estimators and the pools of 16 and 24 samples the gap is between $-0.027$ and $-0.012$ and every interval includes zero. For H2, on the hidden-profile control, split-view debate exceeds the frontier of independent agents who see the same views by \confTwoGap{} ([\confTwoLo, \confTwoHi], $n=\confTwoN$, Holm-adjusted $p=\confTwoPholm$). The isotonic fit pools the drop of the five-view curve from $k=2$ to $k=5$, so it lies above the observed five-view vote and the contrast is an upper bound against a lower-budget pool; against the monotone envelope $\hat f^{+}$ ($0.148$, which the cross-fitted estimator $\hat f^{-}$ also gives on this cell, above the isotonic value $\hpFrontier$; both are estimators, not strict bounds, \Cref{app:certification}) the gap is $+0.035$ ($[-0.013,+0.082]$). H3, the gap of debate minus that of the more-thinking call (\confThreeGap{}, [\confThreeLo, \confThreeHi]), is exploratory but counted in the four-test Holm family. For H4, the gap of mixture-of-agents on MMLU-Pro is \confFourGap{} ([\confFourLo, \confFourHi]), and the intersection test with debate has Holm-adjusted $p=\confFourPholm$.

\begin{figure}[!htbp]
\centering
\includegraphics[width=\textwidth]{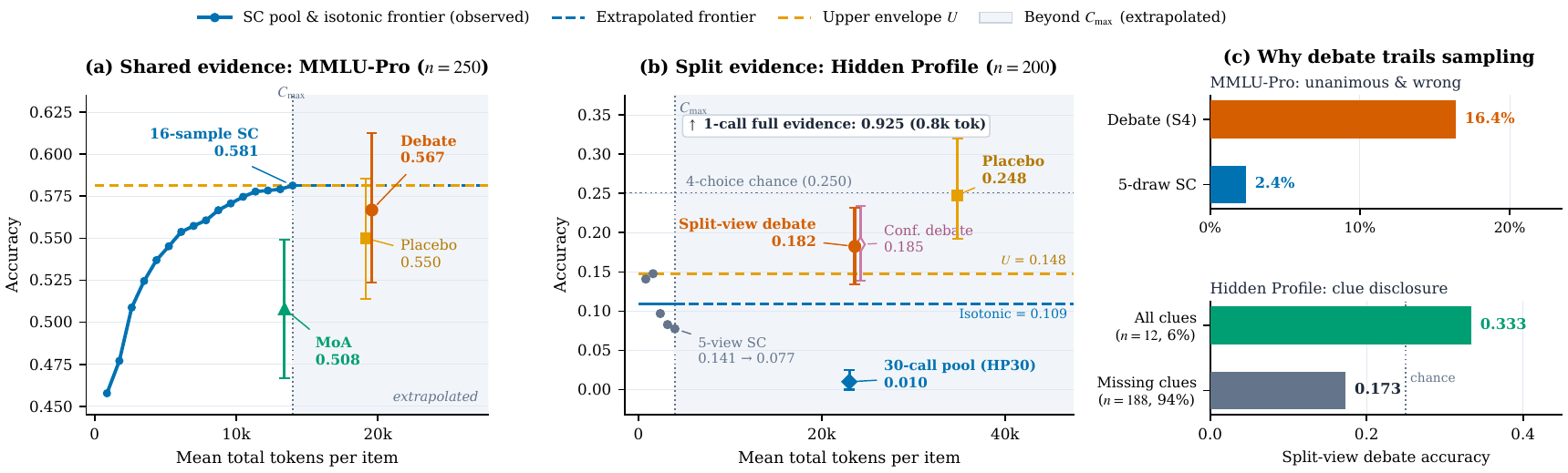}
\caption{Controlled experiment on \texttt{llama3.1:8b}. (a, b) Self-consistency accuracy (points) and its isotonic frontier up to the largest pool cost $C_{\max}$ (\frontCmaxmmlu{} and \frontCmaxhp{} tokens; shaded beyond it, extrapolated), the envelope $U$ and strategies with 95\% paired bootstrap intervals about the frontier at their own cost; (c) on the same logs, the share of MMLU-Pro items on which debate and a five-draw pool end unanimous and wrong, and split-view debate accuracy by whether every private clue was disclosed. All strategies, the cross-fitted estimator and the disclosure diagnostic are in \Cref{fig:controlled_frontier_full,fig:disclosure_diag}.}
\label{fig:controlled_frontier}
\end{figure}

On shared-evidence MMLU-Pro, vanilla debate (S4, accuracy $\mmluDebateAcc$ at $\mmluDebateTok$ tokens) is statistically indistinguishable from the matched self-consistency frontier ($\mmluDebateFrontier$), which at this cost is the standalone 16-sample pool at $13{,}982$ tokens, so the pool scores higher than S4 while using fewer tokens; this agrees with \Cref{prop:redundancy}, which implies no ordering of debate and self-consistency at equal tokens; on the same MMLU-Pro logs, debate ends unanimous and wrong on 16.4\% of items against 2.4\% for five independent draws (ECE $0.271$ vs.\ $0.079$, \Cref{fig:controlled_frontier}c). On split evidence, split-view debate reaches $0.182$ ($23{,}604$ tokens) against a split-view voting frontier of $\hpFrontier$ (the five-view vote scores $\hpVoteAcc$ at $\hpVoteTok$ tokens), whereas greedy full-evidence inference scores $0.925$ in $777$ tokens (\Cref{fig:controlled_frontier}). Split-view debate trails content-mismatched placebo debate (S7, $0.248$ at $34{,}819$ tokens, $\Delta_{\mathrm{arch}}=-0.065$) and four-choice chance ($0.250$); on MMLU-Pro the same contrast is $+0.017$ (S4 $0.567$, S7 $0.550$). The placebo shows messages from other items, uses \hpPlaceboTokPct\% more tokens than split-view debate (\hpPlaceboTokFull{} against \hpDebateTok; prompt $1.39$ and completion $1.67$ times those of S4v, and $0.98$ times the tokens of S4 on MMLU-Pro), and the four candidate names recur in every item, so $\Delta_{\mathrm{arch}}$ is a descriptive contrast and H2 does not establish a semantic benefit of communication. \hpThirtySentence{}

\paragraph{Split-evidence stress test.}\label{sec:poscontrol} The hidden-profile task is a positive control built so that pooling is possible: each of five agents sees the four shared criteria, which favor a decoy, and two of the six split criteria, which favor the target (design in \Cref{app:moved_s6_hpdesign}; \Cref{tab:poscontrol}). No strategy that reads only split views (\hpVoteAcc{} to \hpPlaceboAcc{} above) approaches the $0.925$ of a greedy pass on the full evidence, so this model does not integrate the partitioned evidence. Every agent mentions all of its private criteria in only \hpDiscD{} of items (\Cref{fig:disclosure_diag}); in exploratory logs \decoyRoneShare{} of round-1 answers name the decoy, and an agent whose four peers also name it leaves it with probability \decoyLeaveFourDebate{} in debate but \decoyLeaveFourPlacebo{} under placebo messages, so debate stays on the decoy when $d<1$ (\Cref{prop:disclosure}). Exploratory extensions (\Cref{app:certification,app:moved_s6_deepseek}) show \texttt{deepseek-r1:14b} ($n=50$) devoting $94\%$ of characters to private thinking ($0.000$ split-view accuracy vs.\ $1.000$ full-evidence greedy), \texttt{llama3.1:8b}+\texttt{qwen2.5:7b} ($n=250$) lowering correctness correlation ($\Delta\rho=-0.175$) without lifting voting ($+0.003$), and verified split Python debate ($n=80$) at $+0.087$ [$-0.013$, $+0.188$] vs.\ placebo. 

\section{The PACE Standard for Reporting a Comparison}
\label{sec:discussion}

Positive corpus returns cluster in four cases (\Cref{app:moved_s7_synth}): Pareto fronts without per-item matching ($+0.020$), 100-token chains ($+0.077$, flipping to $-0.012$ to $-0.035$ at $\ge 500$ tokens), unmatched tooling in $R_{\mathrm{sys}}$ ($96.51\%$ vs.\ $57.0\%$), and parallelizable retrieval vs.\ sequential planning. Three rules follow (\Cref{app:moved_s7_decision}): match the single-agent frontier at $k \ge k_{\mathrm{tok}}$ first, use multiple agents only for structural constraints, and give any verifier to both arms while running the matching control in \Cref{tab:route_tests} (\Cref{app:checklist_templates,app:moved_s7_orblock,app:moved_s7_util}). For limited compute, \texttt{PACE-Minimum} votes round-1 draws at zero extra calls (\Cref{app:checklist_templates}). In a null simulation ($K=20$ topologies), ignoring search cost yields $24.6\%$ naive false certification ($5.83\times$ compute) vs.\ $2.0\%$ nested (\Cref{app:certification}). The \texttt{pace} CLI adds open-ended medoid baselines and Rollout Card blocks (\Cref{sec:appendix_cme_cli}). For content claims, a placebo control ($\Delta_{\mathrm{arch}} = p_{\mathrm{comm}} - p_{\mathrm{placebo}} > 0$) separates message semantics from context expansion; on our MMLU-Pro logs, voting round-1 answers instead of debating would counterfactually save $78.4\%$ of tokens at $-0.008$ [$-0.051$, $+0.034$] accuracy (\Cref{app:certification}). Adaptive-stopping and cached pipelines log itemwise budgets ($k_i$) and raw alongside billed tokens (\Cref{app:checklist_templates}).

\begin{table}[!htbp]
\centering
\scriptsize
\setlength{\tabcolsep}{3pt}
\caption{PACE Card v1.0, the seven-field reporting standard: field, reviewer checklist question, required level (Min = \texttt{PACE-Minimum}, Rec = \texttt{PACE-Recommended}, Ext = \texttt{PACE-Extended}; ``Min, Rec'': core field at Minimum, expanded breakdown at Recommended) and companion-package command. Unmeasured dimensions are null, never imputed; \texttt{pace card -{}-v2} is the machine-readable JSON schema of the same fields.}
\label{tab:pace_card}\label{tab:cme_card}
\begin{tabular}{@{}p{0.025\textwidth}p{0.33\textwidth}p{0.40\textwidth}p{0.07\textwidth}p{0.10\textwidth}@{}}
\toprule
 & What to report & Checklist question & Level & Command \\
\midrule
1 & Baseline category: the most informative available Category (A to D) of single-agent baseline, its policy and $k$, and the matched resource if C & Which baseline did the system face, and what was matched? & Min & \texttt{classify} \\
2 & Resource vector: all six dimensions of $\mathbf{R}$ for both systems with their verification states & Does the submission report the verification status of $R_{\mathrm{reason}}$, $R_{\mathrm{token}}$, $R_{\mathrm{cost}}$, $R_{\mathrm{lat}}$, $R_{\mathrm{sample}}$ and $R_{\mathrm{sys}}$? & Min & \texttt{audit} \\
3 & Token accounting: prompt and completion tokens of every call (reused calls included), cached tokens, scaffold tokens & Are system prompts, role headers, echoed histories and routing calls included in $R_{\mathrm{token}}$ and counted once alongside prompt and completion tokens? & Min, Rec & \texttt{card -{}-v2} \\
4 & Resource ratios and feasibility direction on the declared subset $\mathcal{J}$ & Did the baseline receive at least as much of the declared resource (to claim a gain) or no more (to rule one out)? & Rec & \texttt{audit} \\
5 & Frontier gap with its 95\% paired bootstrap interval, pool size and the estimator for the claim ($\hat f^{+}$ positive, $\hat f^{-}$ nonpositive) & Are the frontier gap, 95\% paired bootstrap interval, pool size, and estimator ($\hat f^{+}$ / $\hat f^{-}$) disclosed? & Min, Rec & \texttt{report} \\
6 & Evidence and routes: shared or split evidence, verifier, model heterogeneity, items, seeds, release of logs & Are evidence partition, equal tool or verifier access, model mix, seeds, and item logs disclosed? & Rec & \texttt{card -{}-v2}, \texttt{dynamics} \\
7 & Per-item budgets ($k_i$), cached and uncached input, stopping rules, cost of architecture search and of negative configurations & Are cumulative compute costs of architecture search and negative configurations disclosed? & Ext & \texttt{card -{}-v2} \\
\bottomrule
\end{tabular}
\end{table}

\begin{tcolorbox}[colback=white,colframe=black!60,boxrule=0.5pt,left=4pt,right=4pt,top=2pt,bottom=2pt]
\small
\textbf{Directional feasibility.} Outperforming a baseline that uses at least as much of every declared resource establishes a gain over that comparator; any feasible baseline that uses no more and matches or beats the system refutes a positive frontier gap. \textbf{Pool size.} A debate with $K$ agents, $T$ rounds and token ratio $r$ is compared with at least $k_{\mathrm{tok}}=KT+(T-1)K(K-1)r$ independent samples (\Cref{eq:ktok}). \textbf{How to report.} ``We compare the system with a Category~[A to D] baseline matched on [$\mathcal{J}$] and report the frontier gap [value, 95\% interval] at [mean tokens] using [the envelope $\hat f^{+}$ or cross-fitted $\hat f^{-}$ estimator]. The PACE Card v1.0 is in the supplement.''
\end{tcolorbox}

\section{Limitations and Conclusion}
\label{sec:conclusion}

\paragraph{Limitations.} (i)~The corpus is a bounded synthesis of nine studies, not a systematic review; study, matching condition and baseline category are partly confounded, Category~C rests on two studies, one paper contributes \numContrastsZhang\ of the \numContrastsTotal\ configurations, and the five initial studies were chosen without a search protocol. Extraction used scripted \texttt{pdftotext} parsing and visual inspection; a programmatic parse agrees on every Tran and Kiela table cell and the stratified audit sample found no discrepancy (\Cref{fig:audit_bound}), but there is no complete hand verification. (ii)~The multilevel estimate is exploratory: whole-study error correlation (the adversarial settings of the appendix) moves the CR2 upper bound to $\remlCheStudyHalfCrHi$ and $\remlCheStudyEightCrHi$, the range when one study is omitted at a time is \remlBNineLooMinBeta{} to \remlBNineLooMaxBeta, and omitting Jwalapuram et al., whose baselines are maximum rows that favor the baseline, gives \remlBNineNoJBeta. (iii)~Some contrasts rest on small test sets (such as the coding items of Kim et al.) or on item counts taken from benchmark definitions. (iv)~The confirmatory experiment uses one small open-weight model; exploratory runs use a second model and, for the traces of the appendix, a third (\Cref{sec:controlled_design,app:moved_s4_timeline}), so results need not hold for frontier models. The positive control is synthetic, items within a cell share a model and prompts, token counts can omit cached prefixes, and calls that hit the token cap or return no parsable answer count as incorrect (rates in the appendix). (v)~The plan of the controlled experiment was committed and hashed before any outcome was analyzed but not publicly registered, and the extended sampling pool is post hoc. H1 is a bound, not an equivalence: it excludes advantages of debate over the sampling frontier larger than \confOneUpperMinPP--\confOneUpperMaxPP{} percentage points depending on the frontier definition (primary interval $[\confOneLo, \confOneHi]$). (vi)~On Hidden Profile the H2 contrast is an upper bound against a lower-budget pool and the placebo is content-mismatched, not content-free; \hpThirtyCaveat{}; two-stage resampling over both items and the five-view pool draws widens the H2 interval to $[\twoStageHtwoLo, \twoStageHtwoHi]$ ($p=\twoStageHtwoP$), leaving MMLU-Pro S8 and HP30 unaffected.

\paragraph{Conclusion.} Across \numStudiesTotal\ studies comprising \numContrastsTotal\ unique multi-agent configurations, multi-agent systems show no consistent advantage once single-agent baselines receive repeated samples or a shared budget ceiling. Against sampled or budget-matched baselines the mean difference is \meanCatBC\ (REML $\hat\beta = \remlBNineBeta$, CR2 95\% CI [\remlBNineCrLo, \remlBNineCrHi]); all eight Category~B/C study means are negative, the Category~D study (Wunderlich et al., Pareto fronts at comparable compute) is \meanWunderlichSC, and the sampling gap is negative in all \numStudiesDual\ dual-baseline studies (\Cref{sec:multitier}). A positive coordination return needs a baseline with at least as much of the declared resource ($k \ge k_{\text{tok}}$, \Cref{eq:ktok}); ruling one out needs only one with no more, such as nested initial-round voting (\texttt{PACE-Minimum}). With shared evidence communication adds no information (\Cref{prop:redundancy}), and in the controlled experiment on MMLU-Pro ($n=\numControlledMmluItems$, debate \meanControlledMmluDebateGap, MoA \meanControlledMmluMoaGap) debate did not exceed the sampling frontier; with split evidence ($n=\numControlledHpItems$) split-view debate exceeds split-view voting by \meanControlledHpDebateGap{} ($p_{\mathrm{Holm}}=\numControlledHpHolmP$) yet trails placebo debate and four-choice chance. Split evidence, an external verifier that filters changes, and models that make different errors are candidate settings for positive coordination returns, not effects established by this corpus; parallel tooling and sandboxed execution are further candidates this synthesis did not test.

\paragraph{Reproducibility.}\phantomsection\label{sec:openscience} Code, extraction tables, the audit ledger, and evaluation utilities are available at \url{https://github.com/mahmoudsajjadi/pace}; extraction verification procedures and the audit ledger are detailed in \Cref{app:moved_os_full}. The synthesis was not preregistered. \textbf{Broader impact.} Reporting baselines and resources can reduce wasted compute and overstated claims, and it can also be used to dismiss a system on an incomplete resource log, so the checklist asks for stated missing fields and not for imputed ones. The controlled experiments use small open-weight models and release their logs. No real-world deployment was evaluated; the experiments use offline benchmark and synthetic tasks.

\label{endmain}

\bibliography{references}

\appendix
\counterwithin{table}{section}
\counterwithin{figure}{section}
\section{Mathematical Formulations and Robustness Checks}
\label{sec:appendix_math}
\begin{table}[!htbp]
\centering
\scriptsize
\setlength{\tabcolsep}{4pt}
\renewcommand{\arraystretch}{1.14}
\caption{Notation and symbol reference separating corpus indexing, protocol dimensions, policy classes, frontier estimators, and communication diagnostics.}
\label{tab:notation_reference}
\begin{tabularx}{\textwidth}{@{} >{\raggedright\arraybackslash}p{0.20\textwidth} >{\raggedright\arraybackslash}X >{\raggedright\arraybackslash}p{0.26\textwidth} @{}}
\toprule
\textbf{Symbol} & \textbf{Meaning} & \textbf{Scope and unit} \\
\midrule
$S,\; m_s,\; M$ & Primary studies ($S$), contrasts in study $s$ ($m_s$), and total contrasts in a slice ($M = \sum_{s=1}^S m_s$) & Synthesis and meta-analysis indexing \\
$K,\; T,\; N$ & Agents or independent sample trajectories ($K$), communication rounds ($T$), and benchmark items ($N$) & Protocol execution and item sample count \\
$\mathcal{J},\; \mathbf{R},\; \mathbf{B}_{\mathcal{J}}$ & Predeclared matched resource subset ($\mathcal{J} \subseteq \{1,\dots,6\}$), resource vector with six dimensions $\langle R_{\mathrm{reason}}, R_{\mathrm{token}}, R_{\mathrm{cost}}, R_{\mathrm{lat}}, R_{\mathrm{sample}}, R_{\mathrm{sys}} \rangle$, and ceiling on $\mathcal{J}$ & Directional feasibility and budget accounting \\
$\Pi_0 / \Pi_{\mathrm{SA}}$ & Noncommunicating reference class $\Pi_0(\mathcal{M}_{\mathrm{mod}}, \mathcal{T}_{\mathrm{tool}}) \triangleq \Pi_{\mathrm{SA}}$: no call reads another thread, with fixed aggregation without a model & Reference class indexed by models $\mathcal{M}_{\mathrm{mod}}$ and tools $\mathcal{T}_{\mathrm{tool}}$ \\
$\Pi_{\mathrm{sample}},\; \Pi_{\mathrm{think}}$ & Independent parallel draws with fixed voting ($\Pi_{\mathrm{sample}}$) and sequential continuation on a single thread ($\Pi_{\mathrm{think}}$) & Subfamilies inside $\Pi_{\mathrm{SA}}$ \\
$\Pi_{\mathrm{agg}},\; \Pi_{\mathrm{agent}}$ & Learned model synthesis over $k$ independent draws ($\Pi_{\mathrm{agg}}$) and general protocols allowing reads across threads ($\Pi_{\mathrm{agent}}$) & Communicating architectures ($\Pi_{\mathrm{agg}} \subset \Pi_{\mathrm{agent}}$) \\
$f_{\mathcal{J}}(\mathbf{B}_{\mathcal{J}})$ & Projected single-agent frontier: supremum accuracy in $\Pi_{\mathrm{SA}}$ under $\mathbb{E}[\mathbf{R}(\pi)_{\mathcal{J}}] \preceq \mathbf{B}_{\mathcal{J}}$ & Satisfies $f_{\mathcal{J}}(\mathbf{B}_{\mathcal{J}}) \ge f(\mathbf{B})$ \\
$\hat{f}_{\mathrm{SC}}^-(C),\; \hat{f}_{\mathrm{SC}}^+(C)$ & Directional self-consistency frontier estimators at token cost $C$: two-fold cross-fitted lower bound ($\hat{f}_{\mathrm{SC}}^-$) and monotone envelope upper bound ($\hat{f}_{\mathrm{SC}}^+$) & Lower bound for ruling gains out ($\hat{f}_{\mathrm{SC}}^-$); upper bound for testing positive gains ($\hat{f}_{\mathrm{SC}}^+$) \\
$k_{\mathrm{tok}},\; k_{\mathrm{tok}}^{\mathrm{full}}$ & Independent single-agent draws equating total tokens of a multi-agent run: $KT + (T-1)K(K-1)r$ (previous round peers) and $KT + \tfrac{1}{2}K^2T(T-1)r$ (full history) & Pool sizing ($r = \ell/(c_p+\ell)$) \\
$K_{\mathrm{eff}}$ & Effective ensemble size $K / [1 + (K-1)\rho]$ matching mean variance under exchangeable error correlation $\rho$ & Aggregation variance scaling \\
$\rho_{\mathrm{within}},\; \rho_{\mathrm{shared}},\; \rho_{\mathrm{study}}$ & Within item replicate correlation ($\rho_{\mathrm{within}}$), shared baseline or benchmark block correlation ($\rho_{\mathrm{shared}}$), and study cluster correlation ($\rho_{\mathrm{study}}$) & Dependence across agents, blocks, and study clusters \\
$p^*,\; d,\; \tau_{\mathrm{interaction}}$ & Equilibrium threshold $\tau_{i\to c}/(\tau_{i\to c}+\tau_{c\to i})$, full private disclosure rate ($d$), and $2 \times 2$ interaction contrast ($\tau_{\mathrm{interaction}}$) & Transition, disclosure, and factorial diagnostics \\
\bottomrule
\end{tabularx}
\end{table}

\begin{table}[!htbp]
\centering
\scriptsize
\setlength{\tabcolsep}{4pt}
\renewcommand{\arraystretch}{1.14}
\caption{PACE terminology dictionary separating accuracy contrasts, resource feasibility rules, ensemble variance scaling, and transcript diagnostics.}
\label{tab:pace_glossary}
\begin{tabularx}{\textwidth}{@{} >{\raggedright\arraybackslash}p{0.23\textwidth} >{\raggedright\arraybackslash}p{0.27\textwidth} >{\raggedright\arraybackslash}X @{}}
\toprule
\textbf{Term} & \textbf{Formal expression} & \textbf{Operational interpretation} \\
\midrule
Raw accuracy difference & $\Delta_A = p_{\mathrm{MAS}} - p_{\mathrm{greedy}}$ & Difference against a single greedy pass (Category~A); mixes communication with sample multiplicity and expanded compute. \\
Sampling gap & $\Delta_{\mathrm{gap}} = (p_{\mathrm{MAS}} - p_{\mathrm{sampled}}) - (p_{\mathrm{MAS}} - p_{\mathrm{greedy}}) = p_{\mathrm{greedy}} - p_{\mathrm{sampled}}$ & Shift when replacing greedy decoding with a sampled baseline; $p_{\mathrm{MAS}}$ cancels algebraically, leaving a measure of baseline strength. \\
Frontier gap & $\Delta_{\mathrm{frontier}} = p_{\mathrm{MAS}} - \hat{f}_{\mathrm{SC}}(C_{\mathrm{MAS}})$ & Multi-agent accuracy minus the noncommunicating self-consistency frontier evaluated at the multi-agent system's measured mean token expenditure. \\
Resource feasibility & $\mathbf{R}_{\mathrm{base},\mathcal{J}} \preceq \mathbf{R}_{\mathrm{MAS},\mathcal{J}}$ (rule out) vs.\ $\succeq$ (credit gain) & Directional partial order on subset $\mathcal{J}$: a baseline using no more suffices to rule a positive gain out, whereas crediting a gain requires at least as much declared resource. \\
Effective ensemble size & $K_{\mathrm{eff}} = K / [1 + (K-1)\rho]$ & Number of uncorrelated draws producing the same variance of the mean as $K$ exchangeable draws with pairwise correlation $\rho$. \\
Communication diagnostic & $\tau_{i\to c},\; \tau_{c\to i},\; p^*,\; \Delta_{\mathrm{arch}},\; d$ & Transcript checks comparing flip rates against a control with no communication, semantic content against a placebo ($\Delta_{\mathrm{arch}}$), and full disclosure ($d$). \\
\bottomrule
\end{tabularx}
\end{table}

\begin{table}[!htbp]
\centering
\scriptsize
\setlength{\tabcolsep}{3.5pt}
\renewcommand{\arraystretch}{1.12}
\caption{Theorem-to-evidence map connecting formal PACE results, assumptions, mathematical conclusions, and empirical roles.}
\label{tab:theorem_evidence_map}
\begin{tabularx}{\textwidth}{@{} >{\raggedright\arraybackslash}p{0.11\textwidth} >{\raggedright\arraybackslash}p{0.17\textwidth} >{\raggedright\arraybackslash}p{0.21\textwidth} >{\raggedright\arraybackslash}p{0.24\textwidth} >{\raggedright\arraybackslash}X @{}}
\toprule
\textbf{Class} & \textbf{Result} & \textbf{Main Assumption} & \textbf{Main Conclusion} & \textbf{Empirical Relevance} \\
\midrule
Definition & Pareto dominance and $f_{\mathcal{J}}(\mathbf{B}_{\mathcal{J}})$ (\Cref{eq:projected_frontier}, \Cref{sec:appendix_pareto}) & Vector $\mathbf{R}$ with six axes and predeclared subset $\mathcal{J}$ & Dominance requires at least one strict coordinate gain; partial matching yields $f_{\mathcal{J}} \ge f$ & Classifies baseline rungs and prevents unmeasured axes from implying full parity \\
Proposition & Sharp partial identification bound (\Cref{prop:manski}) & Nonempty feasible set $\mathcal{F}_{\mathcal{J}}$ ($\mathbf{R}_{\mathrm{base},\mathcal{J}} \preceq \mathbf{R}_{\mathrm{comm},\mathcal{J}}$) & $\Delta_{\mathrm{frontier}} \le p_{\mathrm{comm}} - \max_{\mathcal{F}_{\mathcal{J}}} p(\pi_0)$ with sharp interval $[p_{\mathrm{comm}}-1, \overline{\Delta}_{\mathcal{J}}]$ & Adjudicates 516 corpus configurations ($356$ not ahead of a feasible baseline) \\
Diagnostic Identity & Telescope and sampling gap (\Cref{eq:decomposition,eq:sampling_gap_cancel}) & Paired greedy, sampled, and multi-agent runs on one task & $p_{\mathrm{MAS}} - p_G$ splits into sampling, extra pool, and frontier terms; $p_{\mathrm{MAS}}$ cancels in $\Delta_{\mathrm{gap}}$ & Shows 53 apparent Category~A wins turn nonpositive against Category~B \\
Proposition & No information gain under redundancy (\Cref{prop:redundancy}) & Closed world ($R_{\mathrm{sys}}=0$), Bayes readout, $\operatorname{I}(Y;X_{1:K}\mid X_{\mathrm{shared}})=0$ & Transcript carries zero conditional mutual information given $X_{\mathrm{shared}}$ & Shows why shared prompt MMLU-Pro debate gives no information advantage \\
Theorem & Compute and correlation hurdle (\Cref{thm:hurdle}) & Normal margin vote approximation; measured $k_{\mathrm{tok}}$ and $\rho_T$ & Debate beats frontier iff $z(p_T) > z(p_1)\Lambda(K,T,r,\rho_T)$ & MMLU-Pro ($K=5,T=3$) requires $\Lambda = 2.12$--$3.67\times$ single draw margin \\
Proposition & Conformity-to-truth threshold (\Cref{thm:conformity}) & Constant round transition rates $\tau_{i\to c},\tau_{c\to i}$ ($0 < s \le 1$) & Monotone drift toward $p^* = \kappa/(1+\kappa)$; decline when $p_1 > p^*$ & Stylized mechanism for degradation without oracle feedback \\
Corollary & Verifier thinned drift (Route~II, \Cref{sec:appendix_proofs}) & Verifier accepts true edits ($1-\epsilon_{\mathrm{FN}}$) and false edits ($\epsilon_{\mathrm{FP}}$) & Drift stays positive when $\epsilon_{\mathrm{FP}} < \kappa \frac{1-p_1}{p_1}(1-\epsilon_{\mathrm{FN}})$ & Requires granting the same $R_{\mathrm{sys}}$ verifier to the baseline \\
Diagnostic Identity & Plug-in threshold and resampling null (\Cref{sec:appendix_proofs}) & Round counts $n_{\mathrm{w}}, n_{\mathrm{c}} \ge 1$ and flips $F_{ic}, F_{ci}$ & $\hat{p}^* > p_1 \iff F_{ic} > F_{ci}$; independent redraws give $\hat{p}^*_{\mathrm{null}} = \bar{p}$ & Requires \texttt{-{}-null} control before attributing flips to persuasion \\
Proposition & Coalescent collapse under imitation (\Cref{prop:coalescent}) & Independent round 1 draws; uniform peer copying rate $\gamma \in (0,1)$ & $\rho_{\mathrm{within}}(t) = 1-(1-c)^{t-1}$, $K_{\mathrm{eff}}(t) \to 1$, majority approaches $p_1$ & Models accuracy loss from consensus collapse below $K$ sample voting \\
Proposition & Exponential disclosure bottleneck (\Cref{prop:disclosure}) & $S$ required private items disclosed independently at rate $q((1-\theta)\ell)$ & Full disclosure $d = q^S$ falls in private reasoning share $\theta$ and $S$ & Calibrates $d = 0.060$ on Hidden Profile and $0.000$ under $94\%$ private thinking \\
Proposition & Vote amplification, variance inflation, and scaffolding (\Cref{prop:vote_amp,prop:keff,prop:scaffold}) & Independent or exchangeable draws; full history prompt echo & Plurality beats single draw when $p_1 > 0.5$; $\operatorname{Var}(\bar Z)$ scales by $1+(K-1)\rho$; $B_{\mathrm{eff}}$ shrinks in $T$ & Quantifies Category~A inflation, correlation loss ($K_{\mathrm{eff}}$), and crowding \\
Proposition & Shared baseline variance deflation (\Cref{prop:deff}) & $m_b$ contrasts share baseline $b$ with variance $v_b$ and independent $a_j$ & Diagonal covariance understates mean variance by $\mathrm{DEff} = 1 + (m_b-1)\rho_{\mathrm{base}}$ & Motivates CHE covariance blocks $\mathbf{V}_k(\rho)$ and 72 baseline cells \\
\bottomrule
\end{tabularx}
\end{table}

\begin{table}[t]
\centering
\footnotesize
\caption{Exact-arithmetic regression checks of the closed-form results of Section~3 (an algebra and implementation check, not a proof). Settings are parameter combinations evaluated with rational arithmetic.}
\label{tab:math_regression}
\begin{tabular}{@{}lrr@{}}
\toprule
Check & Settings & Violations \\
\midrule
Independent vote amplification and Hoeffding bound & 441 & 0 \\
Variance-equivalent ensemble size & 287 & 0 \\
Conformity recurrence to $p^{*}$ & 900 & 0 \\
Token-equated pool size, both protocols & 48 & 0 \\
Shared-baseline covariance identity & 203 & 0 \\
\bottomrule
\end{tabular}
\end{table}

\subsection{Pareto Dominance Formalism and the Resource Vector}
\label{sec:appendix_pareto}

Let a system configuration be evaluated on a benchmark task. Performance is evaluated by empirical accuracy $p \in [0, 1]$ and a resource vector $\mathbf{R}$ across six dimensions:
\begin{equation}
\mathbf{R} = \big\langle R_{\text{reason}},\, R_{\text{token}},\, R_{\text{cost}},\, R_{\text{lat}},\, R_{\text{sample}},\, R_{\text{sys}} \big\rangle \in \mathbb{R}_{\ge 0}^6,
\end{equation}
where $R_{\text{reason}}$ is the internal reasoning token budget, $R_{\text{token}}$ is total token consumption (including system instructions and echoed histories), $R_{\text{cost}}$ is monetary cost in USD, $R_{\text{lat}}$ is execution latency, $R_{\text{sample}}$ is independent generation draws, and $R_{\text{sys}}$ is external tool or environment invocations.

\begin{definition}[Pareto Dominance]
A system $A = (p_A, \mathbf{R}_A)$ Pareto dominates system $B = (p_B, \mathbf{R}_B)$, denoted $A \succ_{\text{Pareto}} B$, if and only if:
\begin{equation}
\Big[\big(p_A \ge p_B\big) \;\land\; \big(R_{A,k} \le R_{B,k}\ \ \forall k \in \{1,\dots,6\}\big)\Big] \;\land\; \Big[\big(p_A > p_B\big) \;\lor\; \big(\exists k \in \{1,\dots,6\}: R_{A, k} < R_{B, k}\big)\Big].
\end{equation}
The second bracket requires at least one strict inequality among the seven coordinates, so equality in accuracy and in every resource dimension simultaneously is not dominance.
\end{definition}

\begin{definition}[Pareto Incomparability]
Systems $A$ and $B$ are incomparable in the Pareto sense, denoted $A \parallel_{\text{Pareto}} B$, if there exist dimensions $i, j \in \{0, 1, \dots, 6\}$ (treating $p$ as dimension 0 where larger is better) such that $A$ is superior along dimension $i$ while $B$ is superior along dimension $j$.
\end{definition}

\paragraph{Budget decomposition.} The additive decomposition of the apparent gain into a coordination effect and a compute scaling effect is stated in \Cref{sec:decomposition} of the main text.

\subsection{Proofs for Section 3 and Shared-Baseline Variance Deflation}
\label{sec:appendix_proofs}

\begin{proof}[Proof of \Cref{prop:redundancy}]
Let each agent $i$ send, in round $t$, a message $M_{i,t}=g_{i,t}(X_{\mathrm{shared}},X_i,\mathcal{H}^{(t-1)},U_{i,t})$, where $\mathcal{H}^{(t-1)}$ collects all earlier messages and $U_{i,t}$ is private randomness independent of $(Y,X_{\mathrm{shared}},X_1,\dots,X_K)$. By induction on $t$, $\mathcal{H}^{(T)}$ is a function of $(X_{\mathrm{shared}},X_1,\dots,X_K,U)$ with $U=(U_{i,t})$. Conditional mutual information is monotone under taking functions of the second argument, so
\[
\begin{aligned}\operatorname{I}(Y;\mathcal{H}^{(T)}\mid X_{\mathrm{shared}})\le\operatorname{I}(Y;X_1,\dots,X_K,U\mid X_{\mathrm{shared}}) \\ &=\operatorname{I}(Y;X_1,\dots,X_K\mid X_{\mathrm{shared}}) \\ &\quad+\operatorname{I}(Y;U\mid X_{\mathrm{shared}},X_1,\dots,X_K).
\end{aligned}\]
The first term is zero by the redundancy assumption and the second is zero because $U$ is independent of $Y$ given the other variables. Hence $Y$ and $\mathcal{H}^{(T)}$ are conditionally independent given $X_{\mathrm{shared}}$, so $\Pr(Y\mid X_{\mathrm{shared}},\mathcal{H}^{(T)})=\Pr(Y\mid X_{\mathrm{shared}})$ and the Bayes-optimal readout from $(X_{\mathrm{shared}},\mathcal{H}^{(T)})$ is the Bayes-optimal readout from $X_{\mathrm{shared}}$, with the same accuracy.
\end{proof}

\begin{proof}[Proof of \Cref{prop:vote_amp}]
Write $S_K\sim\mathrm{Bin}(K,p)$ with $p=p_1$ and $q=1-p$. For $p>\tfrac12$, Hoeffding's inequality gives $\Pr[S_K\le K/2]=\Pr[S_K-Kp\le-K(p-\tfrac12)]\le\exp(-2K(p-\tfrac12)^2)$, which is the stated lower bound on $p_{\mathrm{SC}}(K)=\Pr[S_K>K/2]$. For the sign, compare $K$ and $K+2$ draws, $K$ odd, and let $X,Y$ be the two extra draws. The majority changes from wrong to right only if $S_K=(K-1)/2$ and $X=Y=1$, and from right to wrong only if $S_K=(K+1)/2$ and $X=Y=0$. With $c=\binom{K}{(K-1)/2}=\binom{K}{(K+1)/2}$,
\[
p_{\mathrm{SC}}(K+2)-p_{\mathrm{SC}}(K)=c\,p^{(K+3)/2}q^{(K+1)/2}-c\,p^{(K+1)/2}q^{(K+3)/2}=c\,(pq)^{(K+1)/2}(p-q),
\]
which is positive if and only if $p>\tfrac12$. Since $p_{\mathrm{SC}}(1)=p$, induction over odd $K\ge3$ gives $p_{\mathrm{SC}}(K)>p$ for $p>\tfrac12$. At the boundary points $p\in\{0,\tfrac12,1\}$, symmetry or degeneracy gives $\Delta_{\mathrm{vote}}(K,p)=0$, and $p_{\mathrm{SC}}(K;p)=1-p_{\mathrm{SC}}(K;1-p)$ gives $\Delta_{\mathrm{vote}}(K,p)<0$ on $(0,\tfrac12)$. For multiclass items ($|\mathcal{Y}|>2$) with empirical vote counts $N_y$, any class $y\neq y^*$ beats $y^*$ only if $N_{y^*}-N_y\le 0$; each single draw changes $N_{y^*}-N_y$ by at most $\{-1,0,+1\}$ (bounded in $[-1,1]$ with range 2) and has expectation $p_{y^*}-p_y\ge d$, so Hoeffding's inequality gives $\Pr[N_{y^*}-N_y\le 0]\le\exp(-Kd^2/2)$, and a union bound over the $|\mathcal{Y}|-1$ incorrect classes yields $1-(|\mathcal{Y}|-1)\exp(-Kd^2/2)$.
\end{proof}

\begin{proof}[Proof of \Cref{prop:keff}]
Let $Z_1,\dots,Z_K$ have variance $\sigma^2$ and pairwise correlation $\rho\in[-\tfrac{1}{K-1},1]$. Then $\operatorname{Var}(\sum_kZ_k)=K\sigma^2+K(K-1)\rho\sigma^2$, so $\operatorname{Var}(\bar Z)=\frac{\sigma^2}{K}[1+(K-1)\rho]$. Setting this equal to $\sigma^2/K_{\mathrm{eff}}$ gives $K_{\mathrm{eff}}=K/(1+(K-1)\rho)$, which is at most $K$ for $\rho\ge0$, with equality if and only if $\rho=0$ (for $K\ge2$), and decreases in $\rho$.
\end{proof}

\begin{proof}[Proof of \Cref{prop:scaffold}]
In round $t\in\{1,\dots,T\}$, each of the $K$ agents has a prompt that contains the fixed scaffold $c_0$ and the $K(t-1)$ messages of earlier rounds, each of mean length $\bar\ell$. The scaffold tokens over all agents and rounds are $\sum_{t=1}^{T}K\big(c_0+K(t-1)\bar\ell\big)=KTc_0+K^2\bar\ell\,\frac{T(T-1)}{2}$. Subtracting from $B$, dividing by $KT$, and clamping at zero gives $B_{\mathrm{eff}}(K,T)=\max\!\big(0, B/(KT)-c_0-\tfrac12K(T-1)\bar\ell\big)$. Since $\mu$ is nondecreasing and $B_{\mathrm{eff}}(K,T)\le\max(0, B/(KT)-c_0)$, $\mu(B_{\mathrm{eff}}(K,T))\le\mu\big(\max(0,B/(KT)-c_0)\big)$. When each round $t\ge 2$ passes only the previous round's $(K-1)$ peer messages of normalized share $r$, total call tokens equal $KT+(T-1)K(K-1)r$ single-agent draws ($6+6r$ for $K=3, T=2$ versus $15+40r$ for $K=5, T=3$).
\end{proof}

\begin{proof}[Proof of \Cref{thm:conformity}]
The expected indicator of correctness after one round is $p_t(1-\tau_{c\to i})+(1-p_t)\tau_{i\to c}$, so $p_{t+1}-p_t=(1-p_t)\tau_{i\to c}-p_t\tau_{c\to i}=\tau_{i\to c}-s\,p_t$ with $s=\tau_{i\to c}+\tau_{c\to i}\in(0,1]$. Write $p^{*}=\tau_{i\to c}/s$; then $\Delta_{\mathrm{round}}(p)=s\,(p^{*}-p)$, so $\Delta_{\mathrm{round}}(p_1)<0$ if and only if $p_1>p^{*}$, and $p^{*}=\kappa/(1+\kappa)$ after dividing numerator and denominator by $\tau_{c\to i}$ (if $\tau_{c\to i}=0$ then $p^{*}=1$). The recursion $p_{t+1}-p^{*}=(1-s)(p_t-p^{*})$ gives $p_t-p^{*}=(1-s)^{t-1}(p_1-p^{*})$. Because $s\in(0,1]$, $0\le 1-s<1$, so $|p_t-p^{*}|$ decreases monotonically and $p_t$ stays between $p_1$ and $p^{*}$: if $p_1>p^{*}$ then $p^{*}\le p_{t+1}\le p_t\le p_1$, and symmetrically if $p_1<p^{*}$ then $p_1\le p_t\le p_{t+1}\le p^{*}$.

\end{proof}

\begin{corollary}[Asymmetric verification]\label{cor:verifier}
Under the assumptions of \Cref{thm:conformity}, let a verifier accept a change to a correct answer with probability $1-\epsilon_{\mathrm{FN}}$ and a change to a wrong answer with probability $\epsilon_{\mathrm{FP}}$. The transition rates become $\tau'_{i\to c}=\tau_{i\to c}(1-\epsilon_{\mathrm{FN}})$ and $\tau'_{c\to i}=\tau_{c\to i}\epsilon_{\mathrm{FP}}$, the one-round change $\Delta'_{\mathrm{round}}(p_1)$ is positive if and only if $\epsilon_{\mathrm{FP}}<\kappa\frac{1-p_1}{p_1}(1-\epsilon_{\mathrm{FN}})$, and $p^{*}\to 1$ as $\epsilon_{\mathrm{FP}}\to 0$.
\end{corollary}
\begin{proof}
Thinning the two transition probabilities gives the new rates. Then $\Delta'_{\mathrm{round}}(p_1)>0$ if and only if $(1-p_1)\tau_{i\to c}(1-\epsilon_{\mathrm{FN}})>p_1\tau_{c\to i}\epsilon_{\mathrm{FP}}$, that is $\epsilon_{\mathrm{FP}}<\kappa\frac{1-p_1}{p_1}(1-\epsilon_{\mathrm{FN}})$ with $\kappa=\tau_{i\to c}/\tau_{c\to i}$; the threshold $p^{*}=\tau'_{i\to c}/(\tau'_{i\to c}+\tau'_{c\to i})$ tends to $1$ as $\epsilon_{\mathrm{FP}}\to 0$.
\end{proof}

For route (I), Fano's inequality for a finite label set gives $P_e\ge(\operatorname{H}(Y\mid X_{\mathrm{shared}},X_i)-1)/\log_2|\mathcal{Y}|$ for any estimator of $Y$ from $(X_{\mathrm{shared}},X_i)$, with entropy in bits; an agent whose own view leaves more than one bit of conditional entropy cannot be correct with probability one from that view, this is an estimation bound for one agent's evidence, and an information gain from distributed evidence requires the joint condition $\operatorname{I}(Y;X_1,\dots,X_K\mid X_{\mathrm{shared}})>0$ (a pairwise term can vanish when evidence is synergistic). Route (III) follows from \Cref{prop:keff}.

\begin{proof}[Proof of \Cref{prop:manski}]
Every $\pi_0\in\mathcal{F}_{\mathcal{J}}$ satisfies the resource constraint that defines the supremum $f_{\mathcal{J}}$, so $f_{\mathcal{J}}\ge p(\pi_0)$ for each of them and $f_{\mathcal{J}}\ge\max_{\mathcal{F}_{\mathcal{J}}}p(\pi_0)$; subtracting from $p_{\mathrm{comm}}$ gives $\Delta_{\mathrm{frontier}}\le\overline{\Delta}_{\mathcal{J}}$. Since accuracies lie in $[0,1]$, $f_{\mathcal{J}}\le 1$ and $\Delta_{\mathrm{frontier}}\ge p_{\mathrm{comm}}-1$. The observed baselines restrict the frontier only through these two inequalities: for any $f\in[\max p(\pi_0),1]$ there is a completion of the unobserved part of $\Pi_{\mathrm{SA}}$ (an unobserved feasible policy of accuracy $f$, or none when $f=\max p(\pi_0)$) that is consistent with all data. Hence the interval is the identified set and its upper end cannot be lowered without further assumptions. If $\mathcal{F}_{\mathcal{J}}=\emptyset$ only $f_{\mathcal{J}}\ge 0$ is available. In the adjudication of \Cref{tab:adjudication}, $\mathcal{F}_{\mathcal{J}}$ is never empty, because the greedy single pass and the nested first-round vote of every configuration belong to it. The \numConfigFeasible{} configurations with $\overline{\Delta}_{\mathcal{J}}\le 0$ are those in which a feasible baseline matches or beats the system, so a positive coordination return is ruled out. The \numConfigAhead{} configurations are ahead of a baseline that holds at least as much of the declared resource on the declared axis, so a gain is claimable against that baseline. The remaining \numConfigUndet{} have $\overline{\Delta}_{\mathcal{J}}>0$ but no baseline that holds at least as much of the resource, so the gain is neither ruled out nor claimable and is undetermined. \end{proof}

\begin{proof}[Proof of \Cref{thm:hurdle}]
Let $M$ exchangeable $\{0,1\}$ answers have mean $p$ and pairwise correlation $\rho$. The vote share has mean $p$ and variance $p(1-p)[1+(M-1)\rho]/M=p(1-p)/M_{\mathrm{eff}}$ (\Cref{prop:keff}), and the vote is correct when the share exceeds $\tfrac12$, so in the normal approximation its accuracy is $\Phi\big(\sqrt{M_{\mathrm{eff}}}\,z(p)\big)$, increasing in $\sqrt{M_{\mathrm{eff}}}\,z(p)$. For the frontier $M=k_{\mathrm{tok}}$, $\rho=0$, $p=p_1$; for the debate $M=K$, $\rho=\rho_T$, $p=p_T$. The debate wins if and only if $\sqrt{K_{\mathrm{eff}}}\,z(p_T)>\sqrt{k_{\mathrm{tok}}}\,z(p_1)$, that is $z(p_T)>z(p_1)\sqrt{k_{\mathrm{tok}}/K_{\mathrm{eff}}}$, and $k_{\mathrm{tok}}/K_{\mathrm{eff}}=\big(T+(T-1)(K-1)r\big)\big(1+(K-1)\rho_T\big)=\Lambda^{2}$. For $K=5$, $T=3$ and $k_{\mathrm{tok}}=\mmluKtokExact$, $\Lambda^{2}=(k_{\mathrm{tok}}/5)(1+4\rho_T)$. \end{proof}

\begin{proof}[Proof of \Cref{prop:coalescent}]
Write $x_i^{t}\in\{0,1\}$ and let $a_t=\mathbb{E}[x_i^{t}x_j^{t}]$ for $i\neq j$, the same for all pairs by exchangeability. (i) $\mathbb{E}[x_i^{t+1}\mid x^{t}]=(1-\gamma)x_i^{t}+\gamma\,(S_t-x_i^{t})/(K-1)$, whose sum over $i$ is $S_t$; so $S_t$ is a martingale and $\mathbb{E}[x_i^{t}]=p_1$. (ii) Condition on which of $i,j$ copy and whom they copy. Both keep (probability $(1-\gamma)^2$): $a_t$. One copies (probability $2\gamma(1-\gamma)$): the product is $x_i^{t}x_{u}^{t}$ with $u$ uniform over the $K-1$ others of $j$, which is $x_i^{t}$ with probability $1/(K-1)$ (mean $p_1$) and a distinct pair otherwise (mean $a_t$). Both copy ($\gamma^2$): the two sources coincide with probability $(K-2)/(K-1)^{2}$ (mean $p_1$) and are otherwise a distinct pair (mean $a_t$). Collecting terms, $a_{t+1}=(1-c)a_t+c\,p_1$ with $c=\tfrac{2\gamma(1-\gamma)}{K-1}+\tfrac{(K-2)\gamma^{2}}{(K-1)^{2}}$. With $a_t=p_1^{2}+\rho_t\,p_1(1-p_1)$ this is $\rho_{t+1}-\rho_t=c(1-\rho_t)$, and $\rho_1=0$ gives $\rho_t=1-(1-c)^{t-1}$; then $K_{\mathrm{eff}}=K/(1+(K-1)\rho_t)$. (iii) Unanimity is absorbing and reachable from every state with positive probability (all dissenters copy a holder of the other answer while the rest keep theirs), so it is reached almost surely; $S_t/K$ is a bounded martingale, so $\Pr[\text{all correct}]=\mathbb{E}[S_\infty]/K=p_1$. At unanimity the vote equals the common answer. The direction matches the fall from $0.747$ (round-1 vote) to $0.715$ (debate) in \citet{choi2025debate}; the model is not fitted to those data. \end{proof}

\begin{proof}[Proof of \Cref{prop:disclosure}]
Independence of the $S$ disclosures gives $d=q^{S}$. Because $q$ is nondecreasing in $\ell_{\mathrm{vis}}$ and $\ell_{\mathrm{vis}}$ is decreasing in $\theta$, $d$ is nonincreasing in $\theta$; $q\in[0,1]$ makes $q^{S}$ nonincreasing in $S$ and $q^{S}\to 0$ as $q\to 0$. The accuracy identity is the law of total probability over the event that all pieces are disclosed. \end{proof}

\paragraph{Item-heterogeneity resampling null, influence models, and within-item $K_{\mathrm{eff}}$.}
For any empirical transcript with round-$t$ wrong and right agent-item counts $n_{\mathrm{w}}\ge 1$ and $n_{\mathrm{c}}\ge 1$ and flip counts $F_{ic}$ (wrong to right) and $F_{ci}$ (right to wrong) with $F_{ic}+F_{ci}\ge 1$ (with no flips, or with all agents correct or all wrong, $\hat{p}^*$ is undefined), the plug-in threshold $\hat{p}^*$ satisfies the algebraic identity
\[
\hat{p}^* - p_1 \;=\; \frac{n_{\mathrm{c}} \cdot n_{\mathrm{w}} \cdot (F_{ic} - F_{ci})}{(F_{ic}n_{\mathrm{c}} + F_{ci}n_{\mathrm{w}})(n_{\mathrm{c}} + n_{\mathrm{w}})},
\]
so $\hat{p}^*$ exceeds $p_1$ if and only if accuracy increased ($F_{ic} > F_{ci}$), preventing $\hat{p}^*$ from serving as an independent test on the same transitions. Next, let items $i=1,\dots,n$ have item-specific single-draw accuracy $p_i$ with dataset mean $\bar{p}=\frac{1}{n}\sum_{i=1}^n p_i$ and mean within-item variance $V=\frac{1}{n}\sum_{i=1}^n p_i(1-p_i)$. Under a no-communication null where each round's answer is an independent redraw with probability $p_i$, expected wrong-to-right and right-to-wrong transitions both equal $\sum_{i=1}^n K p_i(1-p_i)=nKV$, while expected initial wrong and right counts are $nK(1-\bar{p})$ and $nK\bar{p}$. Dividing gives $\tau_{i\to c}^{\mathrm{null}}=V/(1-\bar{p})$ and $\tau_{c\to i}^{\mathrm{null}}=V/\bar{p}$, so $\hat{p}^*_{\mathrm{null}}=\tau_{i\to c}^{\mathrm{null}}/(\tau_{i\to c}^{\mathrm{null}}+\tau_{c\to i}^{\mathrm{null}})=\bar{p}$ without any peer influence. Transition probabilities under this null vary by item, violating constant rates unless benchmarked against a control log. The companion utility \texttt{pace dynamics -{}-null} therefore reports adjacent-round transition rates and $\Delta\rho_{\mathrm{debate}}$ as excess over a no-communication control log alongside empirical transition tables keyed both by correct-peer count ($0$ to $K-1$ across debate, placebo, and item-agent paired difference) and by the share of peers holding the agent's own answer (separating truth tracking from majority following) to distinguish the three influence models in \Cref{tab:influence_models}.

\begin{table}[t]
\centering
\footnotesize
\caption{Three peer-influence models on excess transition rates above a no-communication control, where $x \in [0, 1]$ denotes the share of peers correct and $p$ is the share correct.}
\label{tab:influence_models}
\begin{tabularx}{\textwidth}{@{}l l >{\raggedright\arraybackslash}X l@{}}
\toprule
Model & Excess switch ($i\to c$, $c\to i$) & Drift in share correct $p$ (mean-field, $x=p$) & Signature \\
\midrule
Constant & $a,\; b$ regardless of peers & $a(1-p) - bp$; stable point $a/(a+b)$ & Flat in correct peers \\
Linear & $a x,\; b(1-x)$ & $(a-b)p(1-p)$; zero drift (martingale) when $a=b$ & Linear in correct peers \\
Convex & $a x^2,\; b(1-x)^2$ & $p(1-p)[ap - b(1-p)]$; unstable point $b/(a+b)$ & Convex; majority amplified \\
\bottomrule
\end{tabularx}
\end{table}

To separate shared item difficulty from communication-induced correlation in $K_{\mathrm{eff}}$, run $Q \ge 2$ replicate debates per item and let $Y_a\in\{0,1\}$ be the correctness of agent $a$. With the joint-correctness rate of two different agents in one replicate, $C_{\mathrm{same}}=\mathbb{E}[Y_a Y_b]$, and of two agents in different replicates of the same item, $C_{\mathrm{diff}}=\mathbb{E}[Y_a Y'_b]$, define
\[
\rho_{\mathrm{total}} = \frac{C_{\mathrm{same}} - \bar{p}^2}{\bar{p}(1-\bar{p})}, \qquad
\rho_{\mathrm{item}} = \frac{C_{\mathrm{diff}} - \bar{p}^2}{\bar{p}(1-\bar{p})}, \qquad
\rho_{\mathrm{within}} = \frac{C_{\mathrm{same}} - C_{\mathrm{diff}}}{\bar{p} - C_{\mathrm{diff}}},
\]
where $\bar{p}-C_{\mathrm{diff}}=\mathbb{E}[p_i(1-p_i)]$ is the mean within-item variance; $\rho_{\mathrm{within}}$ is defined when this quantity is positive and is reported as undefined otherwise. At round~1 the draws are independent, so $\rho_{\mathrm{within}} \approx 0$ provides a built-in check before $\rho_{\mathrm{within}}$ is substituted into $K_{\mathrm{eff}} = K/(1 + (K-1)\rho_{\mathrm{within}})$. On hidden-profile tasks (\texttt{pace hidden-profile}), pooling efficiency $(\mathrm{debate}-\mathrm{vote})/(\mathrm{full}-\mathrm{vote})$ is reported with two diagnostics, the rate $d$ at which all private criteria are disclosed in the transcript and the accuracies $a_1$ and $a_0$ with and without full disclosure, which satisfy $\mathrm{debate}=d\,a_1+(1-d)\,a_0$; structured evidence-partition hypotheses are tested as $(\mathrm{debate}-\mathrm{vote})_{\mathrm{split}} - (\mathrm{debate}-\mathrm{vote})_{\mathrm{shared}}$ and state-dependent conformity as debate minus placebo final counts within shared round-1 count-correct strata (canceling regression to the mean).

\begin{proposition}[Shared-baseline variance deflation]\label{prop:deff}
Let $m_b$ contrasts $d_j=a_j-b$, $j=1,\dots,m_b$, share one baseline estimate $b$ with variance $v_b$, where the system estimates $a_j$ are independent of each other and of $b$ and each has variance $v_a$. Then $\operatorname{Corr}(d_j,d_k)=\rho_{\mathrm{base}}\triangleq v_b/(v_a+v_b)$ for $j\ne k$, and the variance of the equally weighted mean $\bar d$ is
\[
\operatorname{Var}(\bar d)=\frac{v_a+v_b}{m_b}\,\mathrm{DEff}(m_b,\rho_{\mathrm{base}}),\qquad \mathrm{DEff}(m_b,\rho_{\mathrm{base}})=1+(m_b-1)\rho_{\mathrm{base}}.
\]
A diagonal sampling covariance, which treats the $d_j$ as independent, reports $(v_a+v_b)/m_b$ and therefore understates the variance by the factor $\mathrm{DEff}>1$ whenever $m_b\ge2$ and $v_b>0$.
\end{proposition}
\begin{proof}
$\operatorname{Var}(d_j)=v_a+v_b$ and, for $j\ne k$, $\operatorname{Cov}(d_j,d_k)=\operatorname{Cov}(a_j,a_k)-\operatorname{Cov}(a_j,b)-\operatorname{Cov}(a_k,b)+\operatorname{Var}(b)$. Independence of $a_j$, $a_k$ and $b$ sets the first three terms to zero, so $\operatorname{Cov}(d_j,d_k)=\operatorname{Var}(b)=v_b$; contrasts evaluated on the same items keep $\operatorname{Cov}(a_j,a_k)>0$, which raises the correlation above $v_b/(v_a+v_b)$. Hence $\operatorname{Corr}(d_j,d_k)=v_b/(v_a+v_b)$ and $\operatorname{Var}(\bar d)=m_b^{-2}\big[m_b(v_a+v_b)+m_b(m_b-1)v_b\big]=\frac{v_a+v_b}{m_b}\big[1+(m_b-1)\rho_{\mathrm{base}}\big]$. The factor exceeds one if and only if $(m_b-1)\rho_{\mathrm{base}}>0$.
\end{proof}
\Cref{prop:deff} is the reason the correlated and hierarchical effects (CHE) block covariance $\mathbf{V}_k(\rho)$ of \Cref{tab:appA_bc_sensitivity} places a within-block correlation among contrasts that share a baseline. In empirical benchmarks, contrasts evaluated on the same item set also share item-level covariance in $a_j$ and $a_k$, which further increases $\operatorname{Corr}(d_j,d_k)$ beyond $v_b/(v_a+v_b)$. The values $\rho\in\{0.0,0.5,0.8\}$ in that table are a sensitivity grid, not estimates: the source papers do not report $v_b$ and $v_a$ separately.

\subsection{Sensitivity Analysis of Category B/C REML Estimates}
\label{app:sensitivity}
\label{sec:appendix_sensitivity}

The Level~1 sampling variance $v_{jk}$ for the log odds ratio contrast $y_{jk} = \operatorname{logit}(\tilde{p}_{\text{MAS}}) - \operatorname{logit}(\tilde{p}_{\text{SAS}})$ is computed using the bivariate delta method:
\begin{equation}
v_{jk} = \frac{1}{N_{jk}} \left[ \frac{1}{\tilde{p}_{\text{MAS}}(1 - \tilde{p}_{\text{MAS}})} + \frac{1}{\tilde{p}_{\text{SAS}}(1 - \tilde{p}_{\text{SAS}})} - \frac{2\varphi}{\sqrt{\tilde{p}_{\text{MAS}}(1 - \tilde{p}_{\text{MAS}})\tilde{p}_{\text{SAS}}(1 - \tilde{p}_{\text{SAS}})}} \right],
\end{equation}
where $\tilde{p} = \frac{p N + 0.5}{N + 1}$ applies the Haldane--Anscombe continuity correction and $\varphi \in [0.00, 0.90]$ is the assumed correlation across paired item responses. On the accuracy scale the corresponding standard error, used for the noise bands of \Cref{sec:multitier}, is
\begin{equation}
\mathrm{SE}(\Delta)=\sqrt{\frac{p_{\text{MAS}}(1-p_{\text{MAS}})+p_{\text{base}}(1-p_{\text{base}})-2\varphi\sqrt{p_{\text{MAS}}(1-p_{\text{MAS}})\,p_{\text{base}}(1-p_{\text{base}})}}{N}}.
\label{eq:noise_band}
\end{equation}
In addition, Panel~B of \Cref{tab:appA_bc_sensitivity} replaces the diagonal sampling covariance with a correlated and hierarchical effects (CHE) block covariance matrix $V_k(\rho)$ assuming pairwise sampling error correlation $\rho \in \{0.0, 0.5, 0.8\}$ within shared-baseline blocks (149 blocks sharing study, benchmark, model, and baseline row), within shared-benchmark blocks (37 study--benchmark cells), or across an entire study (8 uniform study clusters). Panel~C repeats the fit while omitting one study at a time across all eight studies in Category~B/C, and Panel~D adds variance-bound, cluster-merging, sample-size, and raw-scale checks.

\begin{table}[t]
\centering
\footnotesize
\caption{Sensitivity checks for the combined Category B/C model (B9, log odds scale except the raw-$\Delta$ and Cohen's $h$ rows of Panel D). Panel A varies the assumed item-level correlation $\varphi$. Panel B fits a correlated-effects block covariance matrix $V_k(\rho)$ for shared baselines (149 blocks), shared test benchmarks (37 blocks), or uniform whole-study correlation across disjoint benchmarks (8 clusters). Panel C omits each of the eight studies in turn, including Jwalapuram et al.\ (upper-bound CoT-SC baseline) and Zhang et al.\ (largest contributor). Panel D checks Fr\'echet--Hoeffding feasible-$\varphi$ bounds, merged Jwalapuram--Ke study clustering ($K=7$), exclusion of Zhang et al.\ AGIEval placeholder-$N$ rows ($M=454$), exclusion of extreme low-performing tails and winsorized log odds, the raw accuracy difference ($\Delta$) scale, Cohen's $h$ without continuity correction, and clustering by normalized framework family ($K=19$).}
\label{tab:appA_bc_sensitivity}
\resizebox{\textwidth}{!}{
\begin{tabular}{lcrrccr}
\toprule
Specification & Studies/contrasts & $\hat\beta$ & SE & Wald 95\% CI & CR2 95\% CI & CR2 $\nu$ \\
\midrule
\multicolumn{7}{l}{\textit{Panel A: Item-correlation sweep for B9 ($K=8$, $M=474$)}} \\
$\varphi=0.00$ & 8/474 & $-0.194$ & 0.055 & [$-0.301$, $-0.087$] & [$-0.329$, $-0.059$] & 6.2 \\
$\varphi=0.25$ & 8/474 & $-0.195$ & 0.054 & [$-0.300$, $-0.090$] & [$-0.326$, $-0.063$] & 6.3 \\
$\varphi=0.50$ & 8/474 & $-0.195$ & 0.052 & [$-0.298$, $-0.092$] & [$-0.323$, $-0.067$] & 6.4 \\
$\varphi=0.75$ & 8/474 & $-0.195$ & 0.051 & [$-0.295$, $-0.095$] & [$-0.320$, $-0.071$] & 6.4 \\
$\varphi=0.90$ & 8/474 & $-0.195$ & 0.050 & [$-0.293$, $-0.097$] & [$-0.317$, $-0.073$] & 6.5 \\
\midrule
\multicolumn{7}{l}{\textit{Panel B: Correlated sampling covariance (CHE, $\varphi=0.50$, $K=8$, $M=474$)}} \\
Shared baseline (149 blocks, $\rho=0.0$) & 8/474 & $-0.195$ & 0.052 & [$-0.298$, $-0.092$] & [$-0.323$, $-0.067$] & 6.4 \\
Shared baseline (149 blocks, $\rho=0.5$) & 8/474 & $-0.191$ & 0.055 & [$-0.297$, $-0.084$] & [$-0.325$, $-0.057$] & 6.3 \\
Shared baseline (149 blocks, $\rho=0.8$) & 8/474 & $-0.187$ & 0.056 & [$-0.296$, $-0.078$] & [$-0.324$, $-0.050$] & 6.3 \\
Shared benchmark (37 blocks, $\rho=0.0$) & 8/474 & $-0.195$ & 0.052 & [$-0.298$, $-0.092$] & [$-0.323$, $-0.067$] & 6.4 \\
Shared benchmark (37 blocks, $\rho=0.5$) & 8/474 & $-0.197$ & 0.059 & [$-0.313$, $-0.080$] & [$-0.344$, $-0.049$] & 6.1 \\
Shared benchmark (37 blocks, $\rho=0.8$) & 8/474 & $-0.195$ & 0.062 & [$-0.316$, $-0.074$] & [$-0.350$, $-0.040$] & 6.0 \\
Whole-study block (8 blocks, $\rho=0.0$) & 8/474 & $-0.195$ & 0.052 & [$-0.298$, $-0.092$] & [$-0.323$, $-0.067$] & 6.4 \\
Whole-study block (8 blocks, $\rho=0.5$) & 8/474 & $-0.182$ & 0.077 & [$-0.332$, $-0.032$] & [$-0.371$, $+0.007$] & 6.1 \\
Whole-study block (8 blocks, $\rho=0.8$) & 8/474 & $-0.165$ & 0.092 & [$-0.345$, $+0.015$] & [$-0.389$, $+0.059$] & 6.2 \\
\midrule
\multicolumn{7}{l}{\textit{Panel C: Omitting one study at a time from B9 ($\varphi=0.50$, $K=7$)}} \\
Without Choi & 7/460 & $-0.194$ & 0.057 & [$-0.305$, $-0.082$] & [$-0.337$, $-0.050$] & 5.6 \\
Without Jwalapuram & 7/367 & $-0.190$ & 0.061 & [$-0.310$, $-0.070$] & [$-0.347$, $-0.034$] & 5.5 \\
Without Ke & 7/445 & $-0.184$ & 0.057 & [$-0.297$, $-0.072$] & [$-0.330$, $-0.038$] & 5.5 \\
Without Kim & 7/456 & $-0.195$ & 0.056 & [$-0.305$, $-0.085$] & [$-0.336$, $-0.054$] & 5.7 \\
Without Smit & 7/446 & $-0.211$ & 0.057 & [$-0.322$, $-0.099$] & [$-0.357$, $-0.065$] & 5.4 \\
Without Tran & 7/378 & $-0.233$ & 0.053 & [$-0.336$, $-0.130$] & [$-0.370$, $-0.095$] & 5.1 \\
Without Yang & 7/432 & $-0.205$ & 0.059 & [$-0.321$, $-0.090$] & [$-0.356$, $-0.055$] & 5.4 \\
Without Zhang & 7/334 & $-0.145$ & 0.039 & [$-0.221$, $-0.070$] & [$-0.246$, $-0.045$] & 5.1 \\
\midrule
\multicolumn{7}{l}{\textit{Panel D: Variance-bound, cluster-merging, sample-size, and raw-scale checks ($\varphi=0.50$)}} \\
Fr\'echet--Hoeffding feasible-$\varphi$ clamp & 8/474 & $-0.194$ & 0.052 & [$-0.296$, $-0.092$] & [$-0.322$, $-0.066$] & 6.4 \\
Merged Jwalapuram + Ke cluster & 7/474 & $-0.187$ & 0.058 & [$-0.299$, $-0.074$] & [$-0.333$, $-0.040$] & 5.5 \\
Excluding Zhang et al.\ AGIEval ($N=100$ placeholder) & 8/454 & $-0.203$ & 0.058 & [$-0.317$, $-0.090$] & [$-0.345$, $-0.062$] & 6.5 \\
Excluding AgentVerse$\times$Llama-8B \& AFlow $0.000$ tails & 8/466 & $-0.177$ & 0.042 & [$-0.260$, $-0.095$] & [$-0.281$, $-0.074$] & 6.2 \\
Winsorized log odds ($2.5\%$--$97.5\%$) & 8/474 & $-0.193$ & 0.046 & [$-0.284$, $-0.103$] & [$-0.306$, $-0.081$] & 6.5 \\
Raw accuracy difference ($\Delta$) scale & 8/474 & $-0.037$ & 0.010 & [$-0.057$, $-0.018$] & [$-0.061$, $-0.013$] & 6.3 \\
Cohen's $h$ scale (no continuity correction, $\varphi = 0.50$) & 8/474 & $-0.086$ & 0.022 & [$-0.130$, $-0.042$] & [$-0.141$, $-0.031$] & 6.3 \\
Clustered by framework family ($K = 19$ families) & 19/474 & $-0.241$ & 0.075 & [$-0.387$, $-0.095$] & [$-0.397$, $-0.085$] & 17.2 \\
\bottomrule
\end{tabular}}
\end{table}

\begin{figure}[t]
\centering
\includegraphics[width=\textwidth]{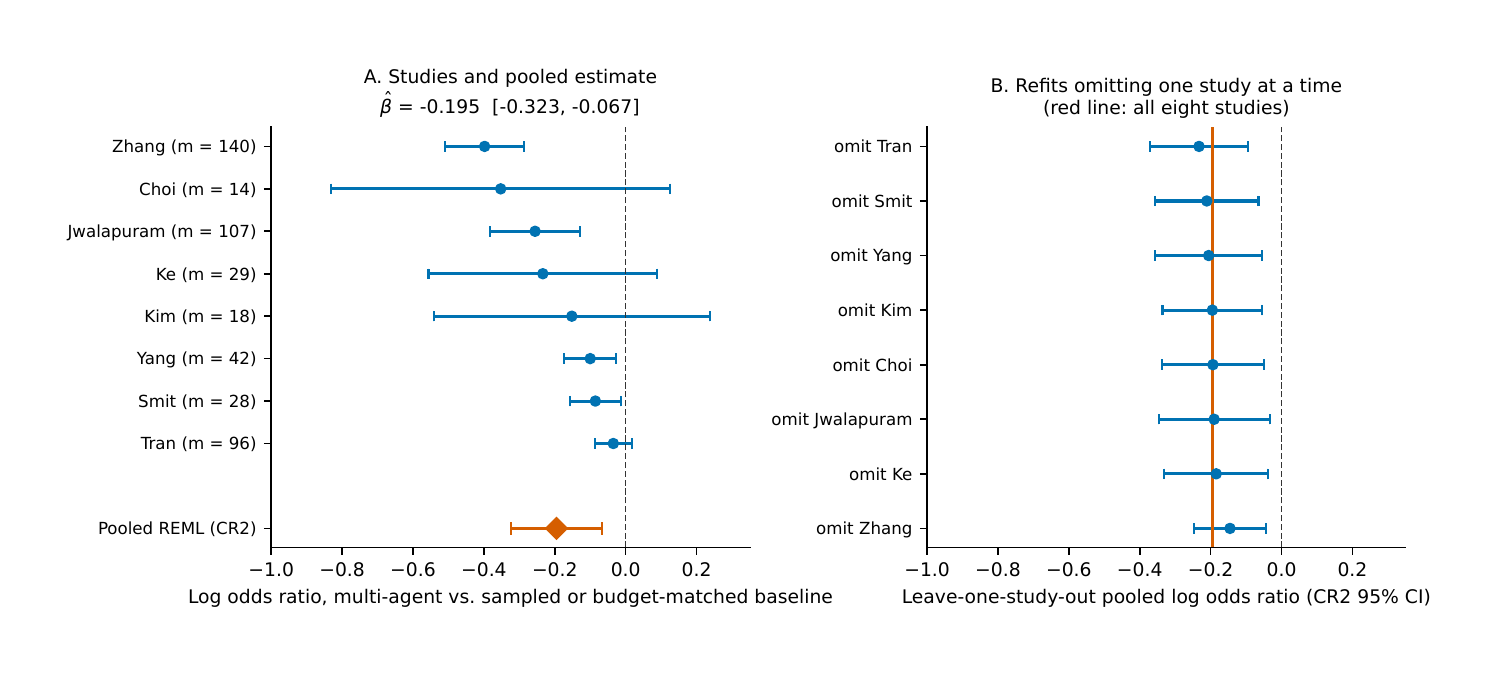}
\caption{Category B/C log odds ratio estimates across eight studies ($M=474$ contrasts, $\varphi=0.50$). Panel A: unweighted study mean log odds ratios with 95\% $t$ intervals over $m$ contrasts per study and the pooled REML estimate with its CR2 95\% confidence interval. Panel B: pooled REML estimates and CR2 95\% confidence intervals when omitting one study at a time. All eight Category~B/C study means are negative; the Category~D comparison of Wunderlich et al.\ ($+0.020$) is not part of this estimate.}
\label{fig:reml_forest}
\end{figure}

Specifically, $\hat\beta$ varies from \remlBNinePhiMinBeta\ to \remlBNinePhiMaxBeta\ across $\varphi \in [0.00, 0.90]$ with CR2 intervals below zero, stays negative across all eight checks omitting one study at a time (\remlBNineLooMinBeta\ to \remlBNineLooMaxBeta; upper CR2 bound at most \remlBNineLooWorstCrHi), and stays below zero under CHE block covariance up to $\rho=0.80$ within shared baselines (\remlCheBaseEightCrLo\ to \remlCheBaseEightCrHi) and shared benchmarks (upper bound \remlCheBenchEightCrHi). Merging Jwalapuram et al.\ and Ke et al.\ into one cluster for shared authorship ($K=7$, $M=474$; Panel~D) yields $\hat\beta = \remlBNineMergedJKBeta$ (CR2 $95\%$ CI [\remlBNineMergedJKCrLo, \remlBNineMergedJKCrHi], $\nu = \remlBNineMergedJKNu$, $p = \remlBNineMergedJKP$; the smallest two-sided $p$-value a cluster sign-flip test can return with seven clusters is $2/2^7 = 0.016$, attained here because all seven cluster means are negative, so the cluster-robust test conveys little beyond the common sign), alongside consistent negative estimates under Fr\'echet--Hoeffding feasible-$\varphi$ bounds, removal of AGIEval placeholder-$N$ rows, the raw accuracy difference scale, Cohen's $h$ without continuity correction ($\hat\beta = \remlBNineCohenHBeta$, CR2 [\remlBNineCohenHCrLo, \remlBNineCohenHCrHi]), and clustering by normalized framework family ($K=19$ families, $\hat\beta = \remlBNineFwClustBeta$, CR2 [\remlBNineFwClustCrLo, \remlBNineFwClustCrHi]). \Cref{fig:reml_forest} displays the Category~B/C estimate by study (Panel~A, descriptive) and the pooled estimate after omitting each study in turn (Panel~B).

As reported in \Cref{tab:appA_bc_sensitivity}, the combined Category~B/C estimate ($\hat\beta = \remlBNineBeta$, CR2 95\% CI [\remlBNineCrLo, \remlBNineCrHi]) exhibits the following behavior across sensitivity checks:
\begin{enumerate}
    \item Assumed item correlation ($\varphi \in [0.00, 0.90]$): Point estimates stay between \remlBNinePhiMinBeta\ and \remlBNinePhiMaxBeta\ across the full grid, with CR2 95\% confidence intervals bounded below zero ($\nu$ between \remlBNinePhiNuMin\ and \remlBNinePhiNuMax; worst bounds [\remlBNinePhiWorstCrLo, \remlBNinePhiWorstCrHi]). Because contrast-level heterogeneity ($\hat\sigma^2_{(2)} = 0.230$) is far larger than median within-cell sampling variance ($0.012$), insensitivity across $\varphi$, Fr\'echet--Hoeffding clamping (the exact bound $\varphi_{\max} = \exp(-|\ln\mathrm{OR}|/2)$ binds in $19/474$ [$4.0\%$], $70/474$ [$14.8\%$], and $238/474$ [$50.2\%$] of contrasts at $\varphi=0.50, 0.75, 0.90$), and AGIEval's placeholder-$N$ follows from variance components by construction.
    \item Correlated sampling covariance ($\rho \in \{0.0, 0.5, 0.8\}$): Under block covariance on shared baselines (149 blocks) or shared benchmarks (37 blocks), CR2 confidence intervals stay below zero through $\rho = 0.80$ (shared baseline: [\remlCheBaseEightCrLo, \remlCheBaseEightCrHi]; shared benchmark upper bound: \remlCheBenchEightCrHi). Under an adversarial whole-study correlation block assuming uniform covariance across disjoint benchmarks, upper CR2 limits reach \remlCheStudyHalfCrHi\ at $\rho=0.50$ and \remlCheStudyEightCrHi\ at $\rho=0.80$.
    \item Omitting one study at a time: Dropping each of the eight studies in turn keeps point estimates between \remlBNineLooMinBeta\ and \remlBNineLooMaxBeta\ with all eight CR2 upper confidence limits below zero (at most \remlBNineLooWorstCrHi), including when omitting Zhang et al.\ (B9$-$Z: \remlBNineNoZBeta, CR2 [\remlBNineNoZCrLo, \remlBNineNoZCrHi]) or Jwalapuram et al., whose upper-bound CoT-SC row is a maximum over entries (B9$-$J: \remlBNineNoJBeta, CR2 [\remlBNineNoJCrLo, \remlBNineNoJCrHi]).
    \item Variance-bound, cluster-merging, sample-size, alternative-scale, and family-clustering checks (Panel~D): Applying Fr\'echet--Hoeffding feasible-$\varphi$ bounds gives $\hat\beta = \remlBNineFrechetBeta$ (CR2 [\remlBNineFrechetCrLo, \remlBNineFrechetCrHi]); merging Jwalapuram et al.\ and Ke et al.\ into one cluster for shared authorship ($K=7$, $M=474$) gives $\hat\beta = \remlBNineMergedJKBeta$ (CR2 [\remlBNineMergedJKCrLo, \remlBNineMergedJKCrHi], $\nu = \remlBNineMergedJKNu$, $p = \remlBNineMergedJKP$, beside a cluster sign-flip $p = 2/2^7 = 0.016$); dropping the 20 AGIEval placeholder $N$ rows in Zhang et al.\ ($M=454$) yields $\hat\beta = \remlBNineNoAgiBeta$ (CR2 [\remlBNineNoAgiCrLo, \remlBNineNoAgiCrHi]); fitting on raw accuracy differences ($\Delta$) yields $\hat\beta = \remlBNineRawDeltaBeta$ (CR2 [\remlBNineRawDeltaCrLo, \remlBNineRawDeltaCrHi]); fitting on Cohen's $h$ without continuity correction yields $\hat\beta = \remlBNineCohenHBeta$ (CR2 [\remlBNineCohenHCrLo, \remlBNineCohenHCrHi]); and clustering across studies by normalized framework family ($K=19$ families: \texttt{ADAS}, \texttt{AFlow}, \texttt{AgentVerse}, \texttt{Centralized}, \texttt{ChatEval}, \texttt{Claude-Sonnet-4.5-orch}, \texttt{DyLAN}, \texttt{Ensemble Refinement}, \texttt{EoT}, \texttt{Expert MAS}, \texttt{GPT-5-orch}, \texttt{MAD/Deb/SoM}, \texttt{MAS-GPT}, \texttt{MAS-Orchestra}, \texttt{MAS-Zero}, \texttt{MaAS}, \texttt{Multi-Persona}, \texttt{Seq}, \texttt{ToolOrchestra}, merging Society of Mind with Du-style debate) yields $\hat\beta = \remlBNineFwClustBeta$ (CR2 [\remlBNineFwClustCrLo, \remlBNineFwClustCrHi]); the close agreement with $\mathrm{B}_{\mathrm{only}}$ is a coincidence, since the two fits use different contrasts and clusters. For the maxima in Jwalapuram et al., $\operatorname{Var}(\text{CoT-SC}_5)=\operatorname{Var}(\text{CoT})/5$ gives an expected inflation of $1.267\,s_{\mathrm{CoT}}$ for the CoT row against $0.567\,s_{\mathrm{CoT}}$ for the CoT-SC row (six entries), which attenuates that study's sampling gap toward zero by $0.700\,s_{\mathrm{CoT}}$ (\Cref{app:scope_notes}, item (xi)).
\end{enumerate}

\section{Benchmark Contrast Ledger and Disaggregation across Baselines}
\label{sec:appendix_ledger}

\subsection{The Complete Benchmark Contrast Ledger}
\label{sec:appendix_ledger_full}

\Cref{tab:appB_ledger} provides the disaggregated ledger for all 516 unique multi-agent configurations across the nine studies (from 531 raw extracted table rows after excluding 14 majority-voting rows in Choi et al.\ and 1 self-consistency baseline row in Wunderlich et al.), by study and benchmark.

\FloatBarrier
{\footnotesize
\setlength{\tabcolsep}{4pt}
\setlength{\LTpre}{4pt}\setlength{\LTpost}{4pt}
\begin{longtable}{llrrrrr}
\caption{Contrast ledger for all 516 unique multi-agent configurations across the nine studies (from 531 raw extracted table rows after excluding 14 majority-voting rows in Choi et al.\ and 1 self-consistency baseline row in Wunderlich et al.), by study and benchmark, against each study's reference baseline (Choi et al.: single-run baseline, yielding $\bar\Delta=+0.045$, versus $-0.031$ against majority voting in \Cref{tab:baseline_tiers}; Wunderlich et al.: Debate ($0.700$) and MoA ($0.714$) against 10-sample self-consistency at comparable compute read off Pareto fronts ($0.687$, versus CoT $0.643$), yielding $m=2$, $\bar p_{\mathrm{MAS}}=0.707$, $\bar\Delta=+0.020$; Smit et al.\ use the median over configurations). Benchmark labels match source papers: Choi et al.'s medical split is MMLU professional medicine, \texttt{BrowseComp-Plus} (Jwalapuram et al.) and \texttt{BrowseComp+} (Ke et al.) refer to the same benchmark, and Ke et al.'s \texttt{GPQA} ($N=198$) corresponds to GPQA-Diamond. Differences $\bar\Delta$ are computed before rounding. $m$ is the number of paired multi-agent and single-agent contrasts; $\bar p_{\mathrm{SAS}}$ and $\bar p_{\mathrm{MAS}}$ are unweighted mean accuracies.}
\label{tab:appB_ledger}\\
\toprule
Study & Benchmark & $m$ & $\bar p_{\mathrm{SAS}}$ & $\bar p_{\mathrm{MAS}}$ & $\bar\Delta$ & Share $\Delta>0$ \\
\midrule
\endfirsthead
\toprule
Study & Benchmark & $m$ & $\bar p_{\mathrm{SAS}}$ & $\bar p_{\mathrm{MAS}}$ & $\bar\Delta$ & Share $\Delta>0$ \\
\midrule
\endhead
Tran \& Kiela & FRAMES & 48 & 0.450 & 0.454 & $+0.004$ & 0.31 \\
 & MuSiQue (4-hop) & 48 & 0.341 & 0.323 & $-0.017$ & 0.35 \\
\midrule
Kim et al. & SWE-bench Verified & 9 & 0.522 & 0.506 & $-0.017$ & 0.33 \\
 & Terminal-Bench & 9 & 0.344 & 0.278 & $-0.067$ & 0.22 \\
\midrule
Jwalapuram et al. & GPQA-Diamond & 23 & 0.718 & 0.728 & $+0.010$ & 0.52 \\
 & HLE-Maths & 23 & 0.168 & 0.194 & $+0.025$ & 0.65 \\
 & SWE-bench Lite & 17 & 0.310 & 0.246 & $-0.064$ & 0.12 \\
 & BrowseComp-Plus & 23 & 0.737 & 0.682 & $-0.055$ & 0.17 \\
 & SMFR & 21 & 0.292 & 0.329 & $+0.038$ & 0.57 \\
\midrule
Choi et al. & Arithmetics & 2 & 0.773 & 0.790 & $+0.017$ & 0.50 \\
 & GSM8K & 2 & 0.805 & 0.840 & $+0.035$ & 1.00 \\
 & MMLU (prof. med.) & 2 & 0.765 & 0.796 & $+0.031$ & 1.00 \\
 & MMLU (formal logic) & 2 & 0.435 & 0.540 & $+0.105$ & 1.00 \\
 & HellaSwag & 2 & 0.707 & 0.740 & $+0.033$ & 1.00 \\
 & CommonsenseQA & 2 & 0.746 & 0.792 & $+0.046$ & 1.00 \\
 & HH-RLHF & 2 & 0.461 & 0.510 & $+0.049$ & 1.00 \\
\midrule
Wunderlich et al. & MMLU-Pro & 2 & 0.687 & 0.707 & $+0.020$ & 1.00 \\
\midrule
Ke et al. & AIME24 & 7 & 0.500 & 0.491 & $-0.009$ & 0.57 \\
 & AIME25 & 7 & 0.450 & 0.418 & $-0.032$ & 0.43 \\
 & BrowseComp+ & 4 & 0.011 & 0.033 & $+0.022$ & 0.50 \\
 & GPQA & 7 & 0.605 & 0.494 & $-0.112$ & 0.43 \\
 & HotpotQA & 4 & 0.336 & 0.376 & $+0.040$ & 0.75 \\
\midrule
Yang et al. & AIME & 21 & 0.145 & 0.127 & $-0.018$ & 0.29 \\
 & MATH500 & 21 & 0.770 & 0.747 & $-0.023$ & 0.05 \\
\midrule
Zhang et al. & AGIEval & 20 & 0.559 & 0.563 & $+0.004$ & 0.50 \\
 & ARC-Challenge & 20 & 0.906 & 0.837 & $-0.070$ & 0.05 \\
 & CommonsenseQA & 20 & 0.790 & 0.702 & $-0.088$ & 0.05 \\
 & GSM8K & 20 & 0.879 & 0.753 & $-0.126$ & 0.25 \\
 & HumanEval & 20 & 0.608 & 0.561 & $-0.048$ & 0.40 \\
 & MATH & 20 & 0.451 & 0.414 & $-0.037$ & 0.55 \\
 & MBPP & 20 & 0.528 & 0.447 & $-0.082$ & 0.15 \\
 & MMLU & 20 & 0.707 & 0.657 & $-0.050$ & 0.25 \\
 & MMLU-Pro & 20 & 0.512 & 0.485 & $-0.027$ & 0.25 \\
\midrule
Smit et al. & CIAR & 4 & 0.500 & 0.463 & $-0.037$ & 0.00 \\
 & Chess & 4 & 0.180 & 0.255 & $+0.075$ & 1.00 \\
 & CosmosQA & 4 & 0.430 & 0.410 & $-0.020$ & 0.00 \\
 & GPQA & 4 & 0.280 & 0.263 & $-0.018$ & 0.25 \\
 & MMLU & 4 & 0.720 & 0.705 & $-0.015$ & 0.25 \\
 & MedQA & 4 & 0.590 & 0.595 & $+0.005$ & 0.75 \\
 & PubMedQA & 4 & 0.700 & 0.713 & $+0.013$ & 0.75 \\
\midrule
\multicolumn{2}{l}{All contrasts} & 516 & 0.522 & 0.495 & $-0.026$ & 0.35 \\
\bottomrule
\end{longtable}
}

Across the 237 unique multi-agent configurations from the initial five studies, multi-agent systems score above their reference baseline in 41\% of contrasts ($\bar\Delta = -0.006$); across all 516 unique multi-agent configurations in the nine-study corpus, multi-agent systems score ahead in 35\% of contrasts ($\bar\Delta = -0.026$). On software engineering benchmarks (SWE-bench Verified, SWE-bench Lite, Terminal-Bench), multi-agent systems exhibit negative differences ($\bar\Delta$ from $-0.067$ to $-0.017$).

\subsection{Disaggregation across Benchmarks and Baselines}
\label{sec:appendix_dual_baseline}

\Cref{tab:appB_dual} presents the audit across two baselines for the five benchmarks evaluated by Jwalapuram et al.\ (107 contrasts).

\begin{table}[t]
\centering
\small
\caption{Breakdown for the 107 contrasts of Jwalapuram et al.\ against the CoT row and the CoT-SC row of their Table 4 (the source describes CoT-SC as a 5-sample majority vote and states that the rows ``report the best observed score across MAS system entries''; per-entry baseline scores are not published, so the rows are an upper-bound single-agent baseline). $\bar\Delta_{\mathrm{CoT}}$ and $\bar\Delta_{\mathrm{SC}}$ are our calculation: multi-agent accuracy minus each row. The last two columns give the median ratio of the multi-agent system's inference cost to each baseline's cost.}
\label{tab:appB_dual}
\resizebox{\textwidth}{!}{
\begin{tabular}{lrrrrrrrr}
\toprule
Benchmark & $m$ & $\bar p_{\mathrm{CoT}}$ & $\bar p_{\mathrm{CoT\text{-}SC}}$ & $\bar p_{\mathrm{MAS}}$ & $\bar\Delta_{\mathrm{CoT}}$ & $\bar\Delta_{\mathrm{SC}}$ & Cost ratio vs CoT & Cost ratio vs SC \\
\midrule
GPQA-Diamond & 23 & 0.718 & 0.738 & 0.728 & $+0.010$ & $-0.009$ & 7.8 & 1.1 \\
HLE-Maths & 23 & 0.168 & 0.194 & 0.194 & $+0.025$ & $0.000$ & 8.5 & 1.6 \\
SWE-bench Lite & 17 & 0.310 & 0.357 & 0.246 & $-0.064$ & $-0.111$ & 7.6 & 0.9 \\
BrowseComp-Plus & 23 & 0.737 & 0.759 & 0.682 & $-0.055$ & $-0.077$ & 10.9 & 1.9 \\
SMFR & 21 & 0.292 & 0.351 & 0.329 & $+0.038$ & $-0.021$ & 6.7 & 1.7 \\
\midrule
All & 107 & 0.455 & 0.489 & 0.448 & $-0.007$ & $-0.041$ & 8.1 & 1.6 \\
\bottomrule
\end{tabular}
}
\end{table}

Benchmarking against greedy CoT yields a mean difference of $\bar\Delta_{\text{CoT}} = -0.007$ despite multi-agent systems incurring an $8.1\times$ median inference cost. Against the CoT-SC row the mean difference is $\bar\Delta_{\text{SC}} = -0.041$. The baseline caveat for the CoT-SC row is given in \Cref{sec:meta}.

\subsection{Breakdown by Multi-Agent Framework}
\label{sec:appendix_frameworks}

\Cref{tab:appB_systems} disaggregates the contrasts across two baselines by multi-agent framework architecture.

\begin{table}[t]
\centering
\small
\caption{Results by multi-agent framework (Jwalapuram et al., 107 contrasts) against the CoT and CoT-SC rows of their Table 4. Share columns give the fraction of contrasts in which the framework scores above the respective row.}
\label{tab:appB_systems}
\begin{tabular}{lrrrrr}
\toprule
Framework & $m$ & $\bar\Delta_{\mathrm{CoT}}$ & $\bar\Delta_{\mathrm{SC}}$ & Share $>$ CoT & Share $>$ CoT-SC \\
\midrule
DyLAN & 18 & $+0.018$ & $-0.014$ & 0.44 & 0.39 \\
MAS-Zero & 18 & $-0.025$ & $-0.057$ & 0.33 & 0.28 \\
AFlow & 18 & $-0.031$ & $-0.063$ & 0.33 & 0.06 \\
MaAS & 18 & $+0.008$ & $-0.025$ & 0.56 & 0.28 \\
ADAS & 18 & $-0.059$ & $-0.091$ & 0.28 & 0.17 \\
MAS-Orchestra & 14 & $+0.012$ & $-0.024$ & 0.50 & 0.21 \\
Expert MAS & 3 & $+0.226$ & $+0.167$ & 1.00 & 1.00 \\
\bottomrule
\end{tabular}
\end{table}

Automated orchestration frameworks show mixed performance against greedy CoT (DyLAN wins 44\% and MAS-Orchestra wins 50\% of contrasts, while AFlow, MAS-Zero, and ADAS win 28\%--33\%). Win rates against the CoT-SC row appear in the table. Expert MAS wins all three of its contrasts against both rows on SMFR.

To connect the audit findings across our corpus back to the literature introducing named methods, \Cref{tab:original_vs_audit} catalogs the named external multi-agent frameworks evaluated in the nine-study corpus (excluding generic architectures without a distinct named system in Tran and Kiela and Kim et al., human-designed Expert MAS, and Ke et al.'s orchestrator ablations), alongside their introducing papers, corpus audit outcomes by baseline category, and author-independence flags. Following \citet{zhang2025stop} and \citet{smit2024understanding}, Society of Mind is cited to \citet{du2023improving} (the ``society of minds'' formulation in their title and setup, rather than \citealp{zhuge2023mindstorms}), and Multi-Persona is cited to \citet{liang2023encouraging}. Note that MAS-Zero and MAS-Orchestra share authors with Jwalapuram et al.\ and Ke et al., so their evaluations in those two studies are not independent audits.

\begin{table}[t]
\centering
\scriptsize
\setlength{\tabcolsep}{3.0pt}
\caption{Named external multi-agent frameworks evaluated in the nine-study corpus, with the paper each auditing study cites for the method and the corpus mean accuracy differences ($\bar\Delta$) against single-pass (Cat.~A) and sampled (Cat.~B) baselines. Originals and audits differ in models, benchmarks, and prompts; original-paper results tables were not re-extracted here.}
\label{tab:original_vs_audit}
\begin{tabularx}{\textwidth}{@{}>{\raggedright\arraybackslash}Xllrrrrl@{}}
\toprule
Framework & Cited source & Auditing studies & $m_{\text{A}}$ & $\bar\Delta_{\text{A}}$ & $m_{\text{B}}$ & $\bar\Delta_{\text{B}}$ & Indep. \\
\midrule
Society of Mind & \citet{du2023improving} & Zhang, Smit & 43 & $-0.012$ & 35 & $-0.030$ & Yes \\
Decentralized debate\textsuperscript{a} & \citet{du2023improving} & Choi, Yang & 14 & $+0.045$ & 56 & $-0.023$ & Yes \\
Multi-Persona & \citet{liang2023encouraging} & Zhang, Smit & 43 & $-0.101$ & 35 & $-0.112$ & Yes \\
ChatEval & \citet{wangeval2023chateval} & Zhang, Smit & 43 & $-0.009$ & 35 & $-0.016$ & Yes \\
AgentVerse & \citet{chen2023agentverse} & Zhang & 36 & $-0.098$ & 28 & $-0.125$ & Yes \\
Exchange-of-Thought & \citet{yin2023exchange} & Zhang & 36 & $-0.048$ & 28 & $-0.057$ & Yes \\
Ensemble Refinement\textsuperscript{\dag} & \citet{singhal2023towards} & Smit & 7 & $+0.004$ & 7 & $-0.017$ & Yes \\
DyLAN & \citet{liu2023dynamic} & Jwalapuram & 18 & $+0.018$ & 18 & $-0.014$ & Yes \\
ADAS & \citet{hu2024automated} & Jwalapuram & 18 & $-0.059$ & 18 & $-0.091$ & Yes \\
AFlow & \citet{zhang2024aflow} & Jwalapuram, Ke & 21 & $-0.014$ & 21 & $-0.050$ & Yes \\
MaAS & \citet{zhang2025maas} & Jwalapuram, Ke & 21 & $-0.013$ & 21 & $-0.049$ & Yes \\
MAS-Zero & \citet{ke2025maszero} & Jwalapuram & 18 & $-0.025$ & 18 & $-0.057$ & No\textsuperscript{\ddag} \\
MAS-Orchestra, introducing paper & \citet{ke2026masorchestra} & Ke & 5 & $+0.125$ & 5 & $+0.089$ & Introducing \\
MAS-Orchestra, later study & \citet{ke2026masorchestra} & Jwalapuram & 14 & $+0.012$ & 14 & $-0.024$ & No\textsuperscript{\ddag} \\
MAS-GPT & \citet{ye2025masgpt} & Ke & 3 & $+0.034$ & 3 & $-0.022$ & Yes \\
ToolOrchestra & \citet{su2025toolorchestra} & Ke & 5 & $-0.174$ & 5 & $-0.210$ & Yes \\
\bottomrule
\end{tabularx}
\par\smallskip\raggedright\scriptsize
\textsuperscript{a}In Tran and Kiela, the generic \texttt{Deb} arm ($m_{\text{C}}=48$) averages $\bar\Delta_{\text{C}}=+0.002$ ($-0.014$ above 100 thinking tokens).
\textsuperscript{\dag}Smit et al.\ group Ensemble Refinement with ensemble methods; it is listed because its second stage is a model call reading peer outputs (\Cref{sec:decomposition}).
\textsuperscript{\ddag}Auditing study shares authors with the introducing paper. Not listed: Expert MAS ($m=3$ per category), Ke et al.'s orchestrator ablations ($m=10$), and the generic architectures in Tran and Kiela and Kim et al.
\end{table}

\paragraph{Quantitative synthesis across the named external frameworks.}
\Cref{tab:supp_original_vs_audit_quant} expands \Cref{tab:original_vs_audit} with contrast ranges, win percentages, paired binomial noise-band splits ($-/\pm/+$ at $\pm 1.96\,\mathrm{SE}$, $\varphi = 0.50$), paired sampling shifts ($\bar\Delta_{\mathrm{gap}}$), and benchmark overlap with each introducing paper. Across the 15 rows that are not Ke et al.'s introducing row, 13 of which are independent of the introducing authors (excluding only Ke et al.'s introducing rows for \texttt{MAS-Orchestra}, where $\bar\Delta_{\mathrm{B}} = +0.089$ across $m_{\mathrm{B}}=5$), the mean Category~B accuracy difference is negative (from $-0.014$ for \texttt{DyLAN} and $-0.016$ for \texttt{ChatEval} to $-0.125$ for \texttt{AgentVerse} and $-0.210$ for \texttt{ToolOrchestra}), and every paired sampling shift $\bar\Delta_{\mathrm{gap}}$ is negative.

\begin{landscape}
\begin{table}[t]
\centering
\scriptsize
\setlength{\tabcolsep}{2.6pt}
\caption{Quantitative audit ledger of the named external multi-agent frameworks evaluated across the nine-study corpus, expanding \Cref{tab:original_vs_audit}. Columns report the introducing paper, auditing studies ($K$) and benchmarks ($B$), Category~A contrast count ($m_{\mathrm{A}}$), mean ($\bar\Delta_{\mathrm{A}}$), min--max range, and win percentage ($\Delta > 0$), Category~B/C contrast count ($m_{\mathrm{BC}}$), mean ($\bar\Delta_{\mathrm{BC}}$), min--max range, win percentage, paired binomial noise-band counts ($-/\pm/+$ at $\pm 1.96\,\mathrm{SE}$, $\varphi = 0.50$), paired sampling shift ($\bar\Delta_{\mathrm{gap}} = \bar\Delta_{\mathrm{B}} - \bar\Delta_{\mathrm{A}}$), benchmark overlap with the original paper, and author independence. Following \citet{du2023improving}, Society of Minds (Zhang, Smit) and decentralized debate (Choi, Yang) appear on separate rows matching \Cref{tab:original_vs_audit}; MAS-Orchestra is split between its introducing paper (Ke et al.) and the later study (Jwalapuram et al.).}
\label{tab:supp_original_vs_audit_quant}
\resizebox{\textwidth}{!}{
\begin{tabular}{llrrrrrrrrrrll}
\toprule
Framework & Original paper & Studies ($K$, $B$) & $m_{\mathrm{A}}$ & $\bar\Delta_{\mathrm{A}}$ [min, max] & Win\textsubscript{A} & $m_{\mathrm{BC}}$ & $\bar\Delta_{\mathrm{BC}}$ [min, max] & Win\textsubscript{BC} & Noise ($-/\pm/+$) & $\bar\Delta_{\mathrm{gap}}$ ($m_{\mathrm{p}}$) & Orig.\ bench.\ overlap & Independent? \\
\midrule
Society of Minds & \citet{du2023improving} & Smit, Zhang (2, 15) & 43 & $-0.012$ [$-0.16$, $+0.11$] & 40\% & 35 & $-0.030$ [$-0.16$, $+0.05$] & 26\% & 15 / 15 / 5 & $-0.017$ (35) & GSM8K, MMLU, Chess & Yes \\
Decentralized debate\textsuperscript{a} & \citet{du2023improving} & Choi, Yang (2, 9) & 14 & $+0.045$ [$-0.05$, $+0.14$] & 93\% & 56 & $-0.023$ [$-0.23$, $+0.03$] & 23\% & 9 / 47 / 0 & $-0.076$ (14) & GSM8K, MMLU, MATH & Yes \\
Multi-Persona & \citet{liang2023encouraging} & Smit, Zhang (2, 15) & 43 & $-0.101$ [$-0.35$, $+0.11$] & 7\% & 35 & $-0.112$ [$-0.39$, $+0.08$] & 9\% & 24 / 10 / 1 & $-0.017$ (35) & GSM8K, TriviaQA/CSQA & Yes \\
ChatEval & \citet{wangeval2023chateval} & Smit, Zhang (2, 15) & 43 & $-0.009$ [$-0.13$, $+0.22$] & 37\% & 35 & $-0.016$ [$-0.10$, $+0.15$] & 26\% & 19 / 12 / 4 & $-0.017$ (35) & Transfer (eval $\to$ QA/code) & Yes \\
AgentVerse & \citet{chen2023agentverse} & Zhang (1, 9) & 36 & $-0.098$ [$-0.75$, $+0.08$] & 33\% & 28 & $-0.125$ [$-0.74$, $+0.09$] & 21\% & 17 / 8 / 3 & $-0.016$ (28) & HumanEval, GSM8K, MMLU & Yes \\
Exchange-of-Thought & \citet{yin2023exchange} & Zhang (1, 9) & 36 & $-0.048$ [$-0.20$, $+0.07$] & 25\% & 28 & $-0.057$ [$-0.18$, $+0.02$] & 14\% & 19 / 8 / 1 & $-0.016$ (28) & GSM8K, MATH, CSQA & Yes \\
Ensemble Refinement\textsuperscript{\dag} & \citet{singhal2023towards} & Smit (1, 7) & 7 & $+0.004$ [$-0.04$, $+0.07$] & 57\% & 7 & $-0.017$ [$-0.06$, $+0.04$] & 14\% & 1 / 5 / 1 & $-0.021$ (7) & MedQA, PubMedQA, MMLU & Yes \\
DyLAN & \citet{liu2023dynamic} & Jwalapuram (1, 5) & 18 & $+0.018$ [$-0.13$, $+0.14$] & 44\% & 18 & $-0.014$ [$-0.17$, $+0.07$] & 39\% & 3 / 13 / 2 & $-0.032$ (18) & Transfer (MMLU/HumanEval $\to$ HLE/SWE) & Yes \\
ADAS & \citet{hu2024automated} & Jwalapuram (1, 5) & 18 & $-0.059$ [$-0.28$, $+0.07$] & 28\% & 18 & $-0.091$ [$-0.37$, $+0.02$] & 17\% & 10 / 8 / 0 & $-0.032$ (18) & GPQA-Diamond, Math/MGSM & Yes \\
AFlow & \citet{zhang2024aflow} & Jwalapuram, Ke (2, 8) & 21 & $-0.014$ [$-0.29$, $+0.12$] & 43\% & 21 & $-0.050$ [$-0.27$, $+0.05$] & 19\% & 8 / 13 / 0 & $-0.036$ (21) & HotpotQA, Math/Coding & Yes \\
MaAS & \citet{zhang2025maas} & Jwalapuram, Ke (2, 8) & 21 & $-0.013$ [$-0.20$, $+0.10$] & 48\% & 21 & $-0.049$ [$-0.25$, $+0.08$] & 24\% & 4 / 16 / 1 & $-0.036$ (21) & Math, Coding, QA & Yes \\
MAS-Zero & \citet{ke2025maszero} & Jwalapuram (1, 5) & 18 & $-0.025$ [$-0.15$, $+0.09$] & 33\% & 18 & $-0.057$ [$-0.23$, $+0.04$] & 28\% & 10 / 8 / 0 & $-0.032$ (18) & GPQA, SWE-Bench Lite, AIME & No\textsuperscript{\ddag} \\
MAS-Orchestra (introducing) & \citet{ke2026masorchestra} & Ke (1, 5) & 5 & $+0.125$ [$+0.05$, $+0.16$] & 100\% & 5 & $+0.089$ [$+0.02$, $+0.14$] & 100\% & 0 / 3 / 2 & $-0.036$ (5) & AIME24/25, GPQA, BrowseComp+ & Introducing \\
MAS-Orchestra (later study) & \citet{ke2026masorchestra} & Jwalapuram (1, 5) & 14 & $+0.012$ [$-0.05$, $+0.14$] & 50\% & 14 & $-0.024$ [$-0.15$, $+0.06$] & 21\% & 2 / 11 / 1 & $-0.036$ (14) & GPQA-Diamond, HLE, SWE, BrowseComp & No\textsuperscript{\ddag} \\
MAS-GPT & \citet{ye2025masgpt} & Ke (1, 3) & 3 & $+0.034$ [$-0.02$, $+0.09$] & 67\% & 3 & $-0.022$ [$-0.08$, $+0.01$] & 67\% & 0 / 3 / 0 & $-0.055$ (3) & MATH/AIME, GPQA & Yes \\
ToolOrchestra & \citet{su2025toolorchestra} & Ke (1, 5) & 5 & $-0.174$ [$-0.34$, $+0.04$] & 40\% & 5 & $-0.210$ [$-0.40$, $+0.02$] & 40\% & 3 / 2 / 0 & $-0.036$ (5) & BrowseComp+/QA, Math & Yes \\
\bottomrule
\multicolumn{13}{l}{\rule{0pt}{2.4ex}\textsuperscript{a}In Tran and Kiela, the generic \texttt{Deb} arm ($m_{\mathrm{C}}=48$) averages $\bar\Delta_{\mathrm{C}}$ = $+0.002$ ($-0.014$ above 100 thinking tokens; noise split 10/31/7).}\\ 
\multicolumn{13}{l}{\textsuperscript{\dag}Smit et al.\ group Ensemble Refinement with ensemble methods; it is listed because its second stage is a model call reading peer outputs (\Cref{sec:decomposition}).}\\ 
\multicolumn{13}{l}{\textsuperscript{\ddag}Auditing study shares authors with the introducing paper. Not listed: \texttt{Expert MAS} ($m=3$ per category), Ke et al.'s orchestrator ablations ($m=10$), and generic architectures in Tran and Kiela and Kim et al.}
\end{tabular}}
\end{table}

\end{landscape}

\providecommand{\allegVerified}{}\renewcommand{\allegVerified}{1}
\providecommand{\allegListed}{}\renewcommand{\allegListed}{15}
\providecommand{\allegUnverified}{}\renewcommand{\allegUnverified}{14}
\providecommand{\allegIntroAuditB}{}\renewcommand{\allegIntroAuditB}{+0.149}
\providecommand{\allegAllegianceTerm}{}\renewcommand{\allegAllegianceTerm}{+0.113}
\providecommand{\allegRungTerm}{}\renewcommand{\allegRungTerm}{+0.036}

\begin{figure}[t]
\centering
\includegraphics[width=0.5\textwidth]{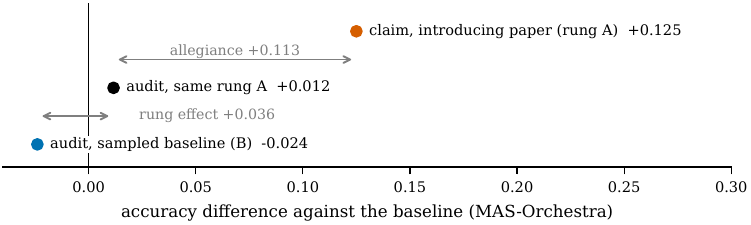}
\caption{Allegiance gap for the decomposable framework: the claim in the introducing paper (red) and the audit against the same rung (black), accuracy difference against the baseline. Frameworks whose introducing-paper claim is not available locally are not plotted.}
\label{fig:allegiance}
\end{figure}

\paragraph{Allegiance-gap decomposition.} The gap between an introducing paper's claim and the audit against a sampled baseline splits into an allegiance or implementation term and a rung effect, $\Delta_{\mathrm{intro}}-\Delta_{\mathrm{audit},B}=[\Delta_{\mathrm{intro}}-\Delta_{\mathrm{audit},r}]+[\Delta_{\mathrm{audit},r}-\Delta_{\mathrm{audit},B}]$, where $r$ is the rung of the baseline in the introducing paper. \Cref{tab:allegiance} lists \allegListed{} named frameworks; the introducing-paper claim is available locally for \allegVerified{} (MAS-Orchestra, whose introducing study is in the corpus), and the other \allegUnverified{} are marked UNVERIFIED in \texttt{evidence/introducing\_papers.csv} and are not estimated. For MAS-Orchestra the claimed difference against a single pass is $\allegIntroAuditB{}$ in total against the later audit's sampled baseline, of which $\allegAllegianceTerm{}$ is the allegiance term and $\allegRungTerm{}$ the rung effect (\Cref{fig:allegiance}); the two studies differ in benchmarks and models, so the allegiance term also contains task differences. With one decomposable framework the clustered REML pooled estimate is not estimable, so no pooled allegiance effect is claimed (\texttt{review/ALLEGIANCE.md}).
\begin{table}[t]
\centering
\footnotesize
\caption{Allegiance-gap decomposition (accuracy-difference scale). 15 named frameworks were listed; only those whose introducing-paper claim is in the corpus can be decomposed, and the rest are UNVERIFIED and not estimated. The introducing study of MAS-Orchestra (Ke et al.) and its later audit (Jwalapuram et al.) differ in benchmarks and models, so the allegiance term mixes implementation and task differences. Pooled estimate: not estimable: 1 verified framework (from 1 cluster) of 15; 14 UNVERIFIED.}
\label{tab:allegiance}
\resizebox{\textwidth}{!}{
\begin{tabular}{@{}lccrrrrr@{}}
\toprule
Framework & Rung $r$ & Studies & $\Delta_{\mathrm{intro}}$ & $\Delta_{\mathrm{audit},r}$ & $\Delta_{\mathrm{audit},B}$ & Allegiance & Rung effect \\
\midrule
MAS-Orchestra & A & Ke, Jwalapuram & +0.125 & +0.012 & -0.024 & +0.113 & +0.036 \\
Remaining 14 frameworks & --- & --- & UNVERIFIED & UNVERIFIED & UNVERIFIED & --- & --- \\
\bottomrule
\end{tabular}}
\end{table}

\subsection{Homogeneous versus Heterogeneous Debate in Zhang et al.}
\label{sec:appendix_hetermad}

\citet{zhang2025stop} also report a heterogeneous-model debate method, Heter-MAD, in which each agent call is routed to GPT-4o-mini or Llama3.1-70b with probability 0.5 (their Table 4, nine benchmarks, arXiv:2502.08788v3). Their comparison baseline is greedy CoT averaged over the two models (CoT-Average); the table has no self-consistency row, so no contrast of heterogeneous models against self-consistency can be formed from it. These results are not part of the 474 Category~B/C contrasts or of any pooled estimate in the main text; \Cref{tab:appB_hetermad} reports the Average column of their Table 4 and the differences computed from it.

\begin{table}[t]
\centering
\small
\caption{Average accuracy over nine benchmarks (percent) in Table 4 of \citet{zhang2025stop}: each debate framework with GPT-4o-mini only, Llama3.1-70b only, and the heterogeneous Heter-MAD variant. The last three columns are computed here from the printed averages, in percentage points. CoT-Average is 72.5.}
\label{tab:appB_hetermad}
\begin{tabular}{lrrrrrr}
\toprule
Framework & GPT & Llama & Heter & \makecell[r]{Heter $-$\\mean(GPT, Llama)} & \makecell[r]{Heter $-$\\CoT-Average} & \makecell[r]{Heter $-$\\best single} \\
\midrule
SoM & 74.3 & 69.9 & 76.7 & +4.60 & +4.2 & +2.4 \\
EoT & 73.4 & 66.5 & 75.7 & +5.75 & +3.2 & +2.3 \\
ChatEval & 74.8 & 69.6 & 75.1 & +2.90 & +2.6 & +0.3 \\
AgentVerse & 75.7 & 68.4 & 74.0 & +1.95 & +1.5 & $-$1.7 \\
\bottomrule
\end{tabular}
\end{table}

Heter-MAD is above the mean of its two single-model variants and above CoT-Average for every framework on the Average column. Against the better of the two single-model variants the sign differs by framework: Heter-SoM, Heter-EoT and Heter-ChatEval are above it, whereas Heter-AgentVerse is below GPT-only AgentVerse, so the gain over the stronger single model is not uniform. The comparison baseline is greedy CoT, not a sampled or token-matched baseline, so the contrast belongs to Category~A in the taxonomy of \Cref{sec:meta}; the table reports no compute. Per-benchmark cells, including AGIEval, are in the source table. We read the table from the arXiv PDF (v3) and computed the differences above; we did not rerun any of their experiments.

\subsection{Moderator Synthesis by Topology, Task Domain, and Model Scale}
\label{sec:appendix_moderators}

To test whether the pooled Category~B/C finding masks positive sub-populations across architectural topologies, benchmark domains, or model capability groups, \Cref{tab:supp_topology_archetypes,tab:supp_domain_moderators,tab:supp_model_scale_moderators} disaggregate all 358 Category~A and 474 Category~B/C contrasts along three moderator axes using raw accuracy differences ($\bar\Delta$), Haldane--Anscombe log odds ratios ($\overline{\ln\mathrm{OR}}_{\mathrm{BC}}$), paired binomial noise bands ($\pm 1.96\,\mathrm{SE}$ at $\varphi = 0.50$), and paired sampling gaps ($\bar\Delta_{\mathrm{gap}}$). Across all five topology archetypes (\Cref{tab:supp_topology_archetypes}), all four benchmark domains (\Cref{tab:supp_domain_moderators}), and both native test-time reasoning and standard instruction-tuned model families (\Cref{tab:supp_model_scale_moderators}), the Category~B/C mean accuracy difference, median difference, mean log odds ratio, and paired sampling shift remain negative.

\begin{landscape}\begin{table}[t]
\centering
\scriptsize
\setlength{\tabcolsep}{3.5pt}
\caption{Disaggregation of extracted contrasts by multi-agent topology archetype across all 358 Category~A and 474 Category~B/C contrasts in \texttt{baseline\_tier\_contrasts.csv}. $K$: contributing studies ($K_{\mathrm{A}}/K_{\mathrm{BC}}$). $\bar\Delta_{\mathrm{A}}$ and $\bar\Delta_{\mathrm{BC}}$: unweighted mean accuracy differences (multi-agent minus single-agent). $\overline{\ln\mathrm{OR}}_{\mathrm{BC}}$: unweighted mean Haldane--Anscombe log odds ratio across Category~B/C contrasts. Not ahead: percentage of Category~B/C contrasts with $\Delta \le 0$. Noise-band split ($-/\pm/+$) and Within~$1.96\,\mathrm{SE}$: counts and percentage of Category~B/C contrasts falling below $-1.96\,\mathrm{SE}$, within $\pm 1.96\,\mathrm{SE}$, and above $+1.96\,\mathrm{SE}$ under the paired binomial standard error at $\varphi = 0.50$. $\bar\Delta_{\mathrm{gap}}$: mean paired shift ($\Delta_{\mathrm{B}} - \Delta_{\mathrm{A}}$) across $m_{\mathrm{pair}}$ dual-baseline pairs. Footnotes report Tran and Kiela's three supplementary feedforward arms ($m=144$, \Cref{tab:tran_all_arms}), Wunderlich et al.'s Category~D Pareto comparisons ($m=2$, \Cref{tab:study_overview}), and the effect of excluding human-designed \texttt{Expert MAS}.}
\label{tab:supp_topology_archetypes}
\resizebox{\textwidth}{!}{
\begin{tabular}{lrrrrrrrrrrr}
\toprule
Topology archetype & $K$ ($K_{\mathrm{A}}/K_{\mathrm{BC}}$) & $m_{\mathrm{A}}$ & $\bar\Delta_{\mathrm{A}}$ & $m_{\mathrm{BC}}$ & $\bar\Delta_{\mathrm{BC}}$ & $\overline{\ln\mathrm{OR}}_{\mathrm{BC}}$ & Not ahead & Noise ($-/\pm/+$) & Within $1.96\,\mathrm{SE}$ & $m_{\mathrm{pair}}$ & $\bar\Delta_{\mathrm{gap}}$ \\
\midrule
Symmetric Peer Debate & 5 (3/5) & 179 & $-0.035$ & 237 & $-0.035$ & $-0.203$ & 75.5\% & 96 / 123 / 18 & 51.9\% & 147 & $-0.023$ \\
Feedforward Layered / Ensemble Aggregation\textsuperscript{a} & 1 (1/1) & 7 & $+0.004$ & 7 & $-0.017$ & $-0.065$ & 85.7\% & 1 / 5 / 1 & 71.4\% & 7 & $-0.021$ \\
Centralized Orchestrator \& Role Graph\textsuperscript{b} & 4 (3/4) & 91 & $-0.034$ & 101 & $-0.058$ & $-0.264$ & 68.3\% & 28 / 62 / 11 & 61.4\% & 83 & $-0.029$ \\
Automated Graph / Prompt Search & 2 (2/2) & 81 & $-0.024$ & 81 & $-0.059$ & $-0.356$ & 76.5\% & 32 / 48 / 1 & 59.3\% & 81 & $-0.035$ \\
Single-Context / Iterative Self-Refinement & 1 (0/1) & 0 & --- & 48 & $-0.015$ & $-0.071$ & 75.0\% & 14 / 30 / 4 & 62.5\% & 0 & --- \\
\midrule
All primary contrasts (Cat.~A \& B/C) & 8 (5/8) & 358 & $-0.032$ & 474 & $-0.042$ & $-0.227$ & 74.3\% & 171 / 268 / 35 & 56.5\% & 318 & $-0.028$ \\
\bottomrule
\multicolumn{12}{l}{\rule{0pt}{2.4ex}\textit{Archetype membership:} \textbf{Symmetric Peer Debate}: \texttt{MAD} (Choi, Yang 2$\times$8/4$\times$4/8$\times$2), \texttt{Deb} (Tran), \texttt{Multi-Persona}/\texttt{MP} (Smit, Zhang), \texttt{ChatEval} (Smit, Zhang),}\\ 
\multicolumn{12}{l}{\texttt{Society of Mind}/\texttt{SoM} (Smit, Zhang), \texttt{EoT} (Zhang). \textbf{Feedforward / Ensemble}: \texttt{Ensemble Refinement} (Smit). \textbf{Centralized Orchestrator}: \texttt{AgentVerse} (Zhang),}\\ 
\multicolumn{12}{l}{\texttt{DyLAN} (Jwalapuram), \texttt{MAS-Orchestra} (Jwalapuram, Ke), \texttt{Centralized} (Kim), \texttt{Claude-Sonnet-4.5}/\texttt{GPT-5 as orchestrator} (Ke), \texttt{ToolOrchestra} (Ke), \texttt{Expert MAS} (Jwalapuram).}\\ 
\multicolumn{12}{l}{\textbf{Automated Search}: \texttt{ADAS}, \texttt{MAS-Zero} (Jwalapuram), \texttt{AFlow}, \texttt{MaAS} (Jwalapuram, Ke), \texttt{MAS-GPT} (Ke). \textbf{Single-Context / Iterative}: \texttt{Seq} (Tran).}\\ 
\multicolumn{12}{l}{\textsuperscript{a}Adding Tran and Kiela's three supplementary feedforward arms (\texttt{Subtask-parallel}, \texttt{Parallel roles}, \texttt{Ensemble}; $m=144$ Cat.~C contrasts in \Cref{tab:tran_all_arms}) yields $m_{\mathrm{BC}}=151$, $\bar\Delta_{\mathrm{BC}}=-0.024$.}\\ 
\multicolumn{12}{l}{In Category~D (Wunderlich et al., $N=1{,}000$), \texttt{Debate} scores $+0.013$ and \texttt{Mixture-of-Agents} (\texttt{MoA}) scores $+0.027$ against 10-sample self-consistency.}\\ 
\multicolumn{12}{l}{\textsuperscript{b}Excluding human-designed \texttt{Expert MAS} ($m=3$ on SMFR, $\bar\Delta_{\mathrm{A}}=+0.226, \bar\Delta_{\mathrm{B}}=+0.167$) from Centralized Orchestrator yields $m_{\mathrm{A}}=88, \bar\Delta_{\mathrm{A}}=-0.043$ and $m_{\mathrm{BC}}=98, \bar\Delta_{\mathrm{BC}}=-0.065$.}
\end{tabular}}
\end{table}
\end{landscape}
\begin{landscape}\begin{table}[t]
\centering
\scriptsize
\setlength{\tabcolsep}{3.5pt}
\caption{Disaggregation of extracted contrasts by task domain and verification affordance across all 358 Category~A and 474 Category~B/C contrasts in \texttt{baseline\_tier\_contrasts.csv}. Columns report study counts $K$ ($K_{\mathrm{A}}/K_{\mathrm{BC}}$), Category~A contrasts ($m_{\mathrm{A}}$) and mean difference ($\bar\Delta_{\mathrm{A}}$), Category~B/C contrasts ($m_{\mathrm{BC}}$), mean ($\bar\Delta_{\mathrm{BC}}$), median ($\mathrm{Med}\,\Delta_{\mathrm{BC}}$), mean log odds ratio ($\overline{\ln\mathrm{OR}}_{\mathrm{BC}}$), share with $\Delta \le 0$ (Not ahead), paired binomial noise-band counts and shares at $\pm 1.96\,\mathrm{SE}$ ($\varphi = 0.50$), and mean paired sampling gap ($\bar\Delta_{\mathrm{gap}}$).}
\label{tab:supp_domain_moderators}
\resizebox{\textwidth}{!}{
\begin{tabular}{lrrrrrrrrrrrr}
\toprule
Task domain & $K$ ($K_{\mathrm{A}}/K_{\mathrm{BC}}$) & $m_{\mathrm{A}}$ & $\bar\Delta_{\mathrm{A}}$ & $m_{\mathrm{BC}}$ & $\bar\Delta_{\mathrm{BC}}$ & $\mathrm{Med}\,\Delta_{\mathrm{BC}}$ & $\overline{\ln\mathrm{OR}}_{\mathrm{BC}}$ & Not ahead & $<-1.96\,\mathrm{SE}$ & Within $\pm 1.96\,\mathrm{SE}$ & $>+1.96\,\mathrm{SE}$ & $\bar\Delta_{\mathrm{gap}}$ ($m_{\mathrm{pair}}$) \\
\midrule
Symbolic \& Competition Mathematics & 5 (4/5) & 81 & $-0.035$ & 123 & $-0.055$ & $-0.025$ & $-0.329$ & 75.6\% & 43 (35.0\%) & 73 (59.3\%) & 7 (5.7\%) & $-0.037$ (81) \\
Multihop \& Graduate Knowledge QA\textsuperscript{a} & 6 (5/6) & 172 & $-0.026$ & 268 & $-0.031$ & $-0.024$ & $-0.159$ & 70.9\% & 96 (35.8\%) & 151 (56.3\%) & 21 (7.8\%) & $-0.019$ (172) \\
Executable Software Eng.\ \& Terminal\textsuperscript{b} & 3 (2/2) & 57 & $-0.065$ & 35 & $-0.075$ & $-0.068$ & $-0.455$ & 82.9\% & 12 (34.3\%) & 22 (62.9\%) & 1 (2.9\%) & $-0.047$ (17) \\
Search, Financial \& Scientific Agents\textsuperscript{c} & 2 (2/2) & 48 & $-0.008$ & 48 & $-0.044$ & $-0.035$ & $-0.174$ & 83.3\% & 20 (41.7\%) & 22 (45.8\%) & 6 (12.5\%) & $-0.036$ (48) \\
\midrule
All primary contrasts (Cat.~A \& B/C) & 8 (5/8) & 358 & $-0.032$ & 474 & $-0.042$ & $-0.027$ & $-0.227$ & 74.3\% & 171 (36.1\%) & 268 (56.5\%) & 35 (7.4\%) & $-0.028$ (318) \\
\bottomrule
\multicolumn{13}{l}{\rule{0pt}{2.4ex}\textit{Benchmark partition:} \textbf{Mathematics}: \texttt{GSM8K}, \texttt{MATH}, \texttt{MATH500}, \texttt{AIME}, \texttt{AIME24}, \texttt{AIME25}, \texttt{HLE-Maths}, \texttt{Arithmetics}. \textbf{Knowledge QA}: \texttt{MMLU}, \texttt{MMLU-Pro}, \texttt{MMLU\_FormLog},}\\ 
\multicolumn{13}{l}{\texttt{MMLU\_ProMed}, \texttt{GPQA}, \texttt{GPQA-Diamond}, \texttt{FRAMES}, \texttt{MuSiQue\_4hop}, \texttt{AGIEval}, \texttt{MedQA}, \texttt{PubMedQA}, \texttt{ARC-Challenge}, \texttt{CommonsenseQA}, \texttt{CosmosQA}, \texttt{HellaSwag}, \texttt{HotpotQA}, \texttt{CIAR}, \texttt{Chess}, \texttt{HH-RLHF}.}\\ 
\multicolumn{13}{l}{\textbf{Software Eng.\ \& Terminal}: \texttt{HumanEval}, \texttt{MBPP}, \texttt{SWE-Bench Lite}, \texttt{SWE-bench Verified}, \texttt{Terminal-Bench}. \textbf{Search \& Scientific Agents}: \texttt{BrowseComp+}, \texttt{BrowseComp-Plus}, \texttt{SMFR}.}\\ 
\multicolumn{13}{l}{\textsuperscript{a}Omitting Tran and Kiela's 16 truncated 100-token rows on \texttt{FRAMES} and \texttt{MuSiQue\_4hop} yields $m_{\mathrm{BC}}=252$, $\bar\Delta_{\mathrm{BC}}=-0.037$, and 13 instead of 21 contrasts above $+1.96\,\mathrm{SE}$.}\\ 
\multicolumn{13}{l}{\textsuperscript{b}Zhang et al.\ report \texttt{HumanEval} ($m=20$) and \texttt{MBPP} ($m=20$) only against greedy CoT (Cat.~A); Category~B/C software contrasts comprise \texttt{SWE-Bench Lite} ($m=17$), \texttt{SWE-bench Verified} ($m=9$), and \texttt{Terminal-Bench} ($m=9$).}\\ 
\multicolumn{13}{l}{\textsuperscript{c}Of the 6 contrasts above $+1.96\,\mathrm{SE}$ in Search \& Scientific Agents, 3 are human-designed \texttt{Expert MAS} on \texttt{SMFR}; automated MAS frameworks on \texttt{BrowseComp-Plus}/\texttt{SMFR} have 3 of 45 contrasts above $+1.96\,\mathrm{SE}$.}
\end{tabular}}
\end{table}
\end{landscape}
\begin{landscape}\begin{table}[t]
\centering
\scriptsize
\setlength{\tabcolsep}{3.5pt}
\caption{Backbone model capability and native test-time reasoning moderator analysis across all 358 Category~A and 474 Category~B/C contrasts in \texttt{baseline\_tier\_contrasts.csv}. Panel~A partitions contrasts into native test-time reasoning models (configured with internal thinking budgets or reasoning chains) versus standard instruction-tuned models, also showing native reasoning models at non-truncated budgets ($\ge 500$ thinking tokens in Tran and Kiela). Panel~B partitions contrasts by parameter scale and deployment group.}
\label{tab:supp_model_scale_moderators}
\resizebox{\textwidth}{!}{
\begin{tabular}{lrrrrrrrrrrrr}
\toprule
Backbone model group & $K$ ($K_{\mathrm{A}}/K_{\mathrm{BC}}$) & $m_{\mathrm{A}}$ & $\bar\Delta_{\mathrm{A}}$ & $m_{\mathrm{BC}}$ & $\bar\Delta_{\mathrm{BC}}$ & $\mathrm{Med}\,\Delta_{\mathrm{BC}}$ & $\overline{\ln\mathrm{OR}}_{\mathrm{BC}}$ & Not ahead & $<-1.96\,\mathrm{SE}$ & Within $\pm 1.96\,\mathrm{SE}$ & $>+1.96\,\mathrm{SE}$ & $\bar\Delta_{\mathrm{gap}}$ ($m_{\mathrm{pair}}$) \\
\midrule
\multicolumn{13}{l}{\textit{Panel A: Native Test-Time Reasoning vs.\ Standard Instruction-Tuned Backbones}} \\
Native Test-Time Reasoning Models (all budgets) & 4 (2/4) & 105 & $-0.006$ & 213 & $-0.028$ & $-0.020$ & $-0.139$ & 64.8\% & 57 (26.8\%) & 136 (63.8\%) & 20 (9.4\%) & $-0.041$ (105) \\
\quad Native Reasoning ($\ge 500$ thinking tokens)\textsuperscript{a} & 4 (2/4) & 105 & $-0.006$ & 197 & $-0.037$ & $-0.024$ & $-0.177$ & 70.1\% & 57 (28.9\%) & 128 (65.0\%) & 12 (6.1\%) & $-0.041$ (105) \\
Standard Instruction-Tuned Models & 6 (4/6) & 253 & $-0.042$ & 261 & $-0.053$ & $-0.031$ & $-0.298$ & 82.0\% & 114 (43.7\%) & 132 (50.6\%) & 15 (5.7\%) & $-0.021$ (213) \\
\midrule
\multicolumn{13}{l}{\textit{Panel B: Model Parameter Scale \& Deployment Group}} \\
Small Open-Weight ($\le 14\text{B}$) & 3 (2/3) & 59 & $-0.084$ & 73 & $-0.090$ & $-0.039$ & $-0.468$ & 78.1\% & 38 (52.1\%) & 34 (46.6\%) & 1 (1.4\%) & $-0.036$ (49) \\
Mid/Large Open-Weight ($15\text{B}$--$120\text{B}$)\textsuperscript{b} & 5 (3/5) & 99 & $-0.012$ & 143 & $-0.032$ & $-0.023$ & $-0.139$ & 66.4\% & 43 (30.1\%) & 86 (60.1\%) & 14 (9.8\%) & $-0.028$ (89) \\
Frontier \& Proprietary API\textsuperscript{c} & 6 (3/6) & 200 & $-0.026$ & 258 & $-0.034$ & $-0.027$ & $-0.207$ & 77.5\% & 90 (34.9\%) & 148 (57.4\%) & 20 (7.8\%) & $-0.025$ (180) \\
\midrule
All primary contrasts (Cat.~A \& B/C) & 8 (5/8) & 358 & $-0.032$ & 474 & $-0.042$ & $-0.027$ & $-0.227$ & 74.3\% & 171 (36.1\%) & 268 (56.5\%) & 35 (7.4\%) & $-0.028$ (318) \\
\bottomrule
\multicolumn{13}{l}{\rule{0pt}{2.4ex}\textit{Model classification:} \textbf{Native Reasoning}: \texttt{DeepSeek-R1-Distill-Llama-70B}, \texttt{Qwen3-30B-A3B}, \texttt{Gemini-2.5-Flash/Pro} (Tran, Jwalapuram, Kim), \texttt{gemini-3-flash} (Kim), \texttt{GPT-5}/\texttt{gpt-5-mini}/\texttt{nano}}\\ 
\multicolumn{13}{l}{(Jwalapuram, Kim), \texttt{GPT-OSS-120B} (Jwalapuram), and \texttt{Qwen-7B orchestrator; gpt-oss-120b sub-agents} (Ke). \textbf{Standard Instruction}: \texttt{GPT-4o}, \texttt{GPT-4o-mini}, \texttt{claude-3-5-haiku}, \texttt{claude-sonnet-4/4-5},}\\ 
\multicolumn{13}{l}{\texttt{gemini-2.0-flash}, \texttt{Llama3.1-8B/70B-Instruct}, \texttt{Qwen2.5-1.5B/3B/7B/14B/32B}, and Smit et al.\ entries. \textsuperscript{a}Excludes Tran and Kiela's 16 contrasts at the truncated budget of 100 thinking tokens ($\bar\Delta=+0.077$).}\\ 
\multicolumn{13}{l}{\textsuperscript{b}Includes \texttt{Qwen2.5-32B}, \texttt{Qwen3-30B-A3B}, \texttt{DeepSeek-R1-Distill-Llama-70B}, \texttt{Llama3.1:70b-instruct}, \texttt{GPT-OSS-120B}, and Ke et al.'s \texttt{gpt-oss-120b} sub-agent configurations.}\\ 
\multicolumn{13}{l}{\textsuperscript{c}Includes all proprietary OpenAI, Anthropic, and Google Gemini API models across Jwalapuram, Kim, Tran, Yang, Zhang, and Smit.}
\end{tabular}}
\end{table}
\end{landscape}

\paragraph{Six-slice adjudication breakdown.}
\Cref{tab:adjudication_summary} expands \Cref{tab:adjudication} across six empirical adjudication slices: the full Category~B/C contrast pool, the directionally feasible nonpositive subset, shared-ceiling and matched-generation contrasts, cost-observed Category~B contrasts, deduplicated dual-baseline cells, and the complete nine-study configuration corpus.

\begin{table}[t]
\centering
\scriptsize
\setlength{\tabcolsep}{3.2pt}
\caption{Summary of empirical adjudication slices across the nine-study corpus (\Cref{sec:multitier}). Each row states the analysis unit, denominator ($M$ or cells), number of studies ($K$), resource relation on $\mathbf{R}$, unweighted mean difference ($\bar\Delta$), share not ahead ($\Delta \le 0$) and ahead ($\Delta > 0$), paired binomial noise-band split ($-/\pm/+$ at $\pm 1.96\,\mathrm{SE}$, $\varphi = 0.50$), and scope qualification.}
\label{tab:adjudication_summary}
\begin{tabular}{@{}>{\raggedright\arraybackslash}p{0.15\textwidth}>{\raggedright\arraybackslash}p{0.065\textwidth}rr>{\raggedright\arraybackslash}p{0.15\textwidth}r>{\raggedright\arraybackslash}p{0.105\textwidth}>{\raggedright\arraybackslash}p{0.075\textwidth}>{\raggedright\arraybackslash}p{0.19\textwidth}@{}}
\toprule
Adjudication slice & Unit & $K$ & Denom. & Resource relation on $\mathbf{R}$ & $\bar\Delta$ & Not ahead / Ahead & Noise ($-/\pm/+$) & Scope \& qualification \\
\midrule
1. Full Category B/C pool & Contrasts & 8 & 474 & Sampled (B) or budget-matched (C) & $-0.042$ & 352 (74\%) / 122 (26\%) & 171 / 268 / 35 & Pooled Category~B/C contrast set \\
2. Directionally feasible ($\Delta \le 0$) & Contrasts & 5 & 474 & $\mathbf{R}_{\mathrm{base},\mathcal{J}} \preceq \mathbf{R}_{\mathrm{MAS},\mathcal{J}}$ \& $\Delta \le 0$ & --- & 171 (36\%) verified & 56 / 115 / 0\textsuperscript{a} & 120 / 367 (33\%) without Jwalapuram \\
3. Shared-ceiling / matched-gen. & Contrasts & 3 & 156 & Equal ceiling (Tran, Kim) or 16-gen.\ (Yang) & $-0.015$ & 112 (72\%) / 44 (28\%) & 29 / 115 / 12 & Above 100 tok.\ ($M=140$): 112 (80\%) not ahead, 28 (20\%) ahead ($\bar\Delta$ = $-0.025$) \\
4. Cost-observed Category B & Contrasts & 1 & 105 & Per-contrast USD cost vs.\ CoT-SC(5) & $-0.041$ & 79 (75\%) / 26 (25\%) & 34 / 65 / 6 & 51 Pareto-dominated, 28 not ahead at lower cost; 7 ahead at lower cost, 19 at higher cost \\
5. Deduplicated dual-baseline cells & Cells & 5 & 72 & Unique (study, benchmark, model) pairs & $-0.034$ & 64 (89\%) $\Delta_{\mathrm{gap}} < 0$ & --- & Mean weighted equally by study: $-0.036$ \\
6. Full 9-study configuration corpus & Configs & 9 & 516 & Best available baseline per configuration & --- & 356 (69\%) $\le$ feasible/Cat.~A & --- & \numConfigAhead{} (\pctConfigAhead\%) ahead at equal ceiling or lower cost; \numConfigUndet{} (\pctConfigUndet\%) undetermined (incl. 2 Wunderlich Pareto configurations) \\
\bottomrule
\multicolumn{9}{p{0.97\textwidth}}{\rule{0pt}{2.4ex}\textsuperscript{a}Among the 171 directionally feasible nonpositive contrasts (51 Jwalapuram, 64 Tran, 35 Yang, 13 Kim, 8 Choi), 163 have $\Delta < 0$, 8 are exact ties ($\Delta = 0$), and 56 fall below $-1.96\,\mathrm{SE}$.}
\end{tabular}
\end{table}

\section{Candidate Study Screening and Untested Positive-Return Hypotheses}
\label{app:candidates}

\subsection{Screening of Candidate Studies}
Seven candidate papers identified by the authors during corpus expansion (six additional papers and one earlier exclusion) were screened against written eligibility criteria. A paper was eligible for extraction if it reported accuracy for at least one multi-agent and one single-agent system on the same benchmark items in a table or text, with an identifiable single-agent baseline and objective scoring. Unmatched resource budgets were not an exclusion criterion, and the direction of results was not considered. Titles and identifiers were checked against each paper's arXiv page. Decisions appear in \Cref{tab:candidates}. This screening is not a systematic literature search.

\begin{table}[t]
\centering
\small
\caption{Screening of candidate studies. Status: Included (eligible and extracted in the nine-study synthesis); Excluded (relevant but not suitable for direct tabular comparison).}
\label{tab:candidates}
\begin{tabular}{>{\raggedright\arraybackslash}p{0.28\textwidth}c>{\raggedright\arraybackslash}p{0.54\textwidth}}
\toprule
Paper (arXiv id) & Status & Reason \\
\midrule
Ke et al., MAS-Orchestra (2601.14652) & Included & Table with accuracy for standalone agents and orchestration systems; test sizes stated. \\
Yang et al., Revisiting Multi-Agent Debate as Test-Time Scaling (2505.22960) & Included & Accuracy for self-consistency, refinement, and debate at equal maximum generations; AIME size derived as 60 (30 problems in each of two years); MATH500 stated as 500. \\
Zhang et al., Stop Overvaluing Multi-Agent Debate (2502.08788) & Included & Single-agent, CoT, self-consistency, and five debate frameworks on nine benchmarks; dataset sizes stated in the appendix; evaluated counts and AGIEval count not stated ($N=100$ placeholder for AGIEval). \\
Smit et al., Should we be going MAD? (2311.17371) & Included & Best and median scores over hyperparameter configurations; dataset sizes stated for seven benchmarks, evaluated counts not always confirmed. \\
Chu et al., MAS-Bench (2610.04672) & Excluded & Reports only multi-agent settings with communication costs; no single-agent condition. \\
Parmar et al., Multi-Agent Debate for Data Cleaning (2606.02866) & Excluded & F1-type metrics on error detection, mainly one backbone; not item accuracy. \\
Wang et al., Reasoning in Token Economies (2406.06461) & Excluded & Compares debate with self-consistency under token budgets, but only in plots. \\
\bottomrule
\end{tabular}
\end{table}

\subsection{Proposed Mechanisms for Positive Returns (Untested Hypotheses)}
\label{app:hypotheses}
Several studies and mechanisms report isolated conditions in which multi-agent configurations exceed single-agent baselines. Because these claims either come from single benchmarks, exclude noncommunicating baselines with matched tooling, or lie outside matched resource vector $\mathbf{R}$ evaluations in our extracted tables, \Cref{tab:positive_hypotheses} catalogs them explicitly as untested hypotheses for future controlled experiments rather than established empirical laws.

\begin{table}[t]
\centering
\footnotesize
\caption{Reported settings and proposed mechanisms for positive multi-agent coordination returns, listed as untested hypotheses rather than validated generalizations across benchmark distributions.}
\label{tab:positive_hypotheses}
\begin{tabular}{>{\raggedright\arraybackslash}p{0.20\textwidth}>{\raggedright\arraybackslash}p{0.28\textwidth}p{0.44\textwidth}}
\toprule
Source & Reported Pattern & Untested Coordination Hypothesis \\
\midrule
\citet{wunderlich2026mixture} & Debate ($70.0\%$) and MoA ($71.4\%$) exceed 10-sample SC ($68.7\%$) at comparable compute read off Pareto fronts on MMLU-Pro ($N=1{,}000$; Llama 3.1 70B). The MoA proposers are copies of one model (single-proposer setting), not different models. & Dense multiple-choice knowledge verification and aggregation by a synthesizing layer may yield positive gains over sampling at comparable compute; the source reports no cross-model comparison in this setting. \\
ColMAD \citep{colmad2026collaborative} & Collaborative debate outperforms standard competitive debate toward downstream judge accuracy at matched tokens. & Converting inter-agent debate from zero-sum competition into supportive critique aimed at a downstream judge may mitigate oversight errors under token budgets. \\
\citet{zhu2026demystifying} & Confidence-weighted aggregation and diverse initializations exceed majority voting on six QA tasks. & Preserving panelist diversity and weighting responses by calibrated confidence may avoid consensus collapse observed in homogeneous debate. \\
\citet{zhang2025stop} & Homogeneous debate methods show isolated positive cells (the AGIEval mean over four models and five methods is $\bar\Delta=+0.004$) despite $-0.058$ overall vs.\ CoT. A separate heterogeneous-model experiment in the source (Heter-MAD, its Table 4) reports gains on AGIEval that our extraction does not include; \Cref{sec:appendix_hetermad} reports the Average column of that table. & Model diversity may supply complementary priors on selected tasks, though net aggregate differences across nine benchmarks remain negative ($-0.073$ vs.\ SC) for the homogeneous methods we extracted. \\
Expert MAS \citep{jwalapuram2026illusion} & Reaches $96.51\%$ at \$555 on SMFR ($N=588$) vs.\ $48.5\%$ CoT, $57.0\%$ CoT-SC, and $20.2\%$--$63.0\%$ automated MAS. & Domain-specific external tooling and verifiers ($R_{\text{sys}}$) appear to drive accuracy gains when compared against unequipped single-agent baselines. \\
\bottomrule
\end{tabular}
\end{table}

\section{mae-eval Evaluation Utilities and Runnable Example}
\label{sec:appendix_mae_eval}

To support reproducible resource accounting, the companion archive provides the \texttt{mae-eval} prototype package (\texttt{code/mae\_eval/}) implementing the six-dimension vector $\mathbf{R}$ and the PACE Reviewer Checklist:
\begin{itemize}
    \item Resource Tracking: \texttt{ResourceTracker} instruments model generation calls to record thinking tokens ($R_{\text{reason}}$), total tokens ($R_{\text{token}}$), monetary cost ($R_{\text{cost}}$), execution latency ($R_{\text{lat}}$), sample count ($R_{\text{sample}}$), and external tool invocations ($R_{\text{sys}}$).
    \item Parity Assertion: \texttt{EvaluationAuditor} compares multi-agent JSONL trace logs against single-agent baselines, flagging unmatched resource dimensions and checking declared parity tolerances.
\end{itemize}
Unit tests covering token accounting, concurrency latency, CLI trace parsing, and parity assertions are maintained in \texttt{tests/test\_mae\_eval.py}. Running the bundled trace file \texttt{code/examples/sample\_trace.jsonl} exercises the auditor on one single-agent call versus three concurrent workers plus a judge call, using illustrative placeholder API pricing of \$1.00/\$4.00 per million input/output tokens. The trace is a synthetic illustrative example written by hand to exercise the auditor; it does not come from a model run. A ratio is printed as n/a when both systems report 0 for a dimension (here \texttt{r\_sys}), because 0/0 is undefined. The command prints:
\begin{lstlisting}[language=bash,basicstyle=\ttfamily\scriptsize]
$ PYTHONPATH=code python -m mae_eval audit --log code/examples/sample_trace.jsonl \
    --price-in 1.0 --price-out 4.0
SAS: r_reason=900.0, r_token=2000.0, r_cost=0.0056, r_lat=10.0, r_sample=1, r_sys=0, scaffold=0.0
MAS: r_reason=2300.0, r_token=11800.0, r_cost=0.0232, r_lat=14.0, r_sample=4, r_sys=0, scaffold=20.3
MAS/SAS ratios: r_reason=2.56, r_token=5.9, r_cost=4.14, r_lat=1.4, r_sample=4, r_sys=n/a
FLAG r_reason [Unmatched (Measured)]: reasoning-token budgets differ: thinking compute is not matched
FLAG r_token [Unmatched (Measured)]: total tokens differ: scaffold, history echoes, or extra calls consume more budget
FLAG r_cost [Unmatched (Measured)]: monetary cost differs: comparison is not cost-matched
FLAG r_lat [Unmatched (Measured)]: wall-clock latency differs (serialization or concurrency)
FLAG r_sample [Unmatched (Measured)]: sample counts differ: gains may reflect ensembling (more draws), not coordination
Parity on r_token within 0.05: VIOLATED
\end{lstlisting}
Note that legacy \texttt{mae\_eval} reports total model calls in \texttt{r\_sample} (\texttt{4} for three parallel workers plus one judge), whereas \texttt{pace} records independent generation trajectories ($R_{\text{sample}}=3$) separately from total model calls ($N_{\text{calls}}=4$).

\subsection{The \texttt{pace} command line}
\label{sec:appendix_cme_cli}
The companion package installs two command-line entry points. \texttt{mae-eval} audits a JSONL trace of one multi-agent and one single-agent system (the listing above). \texttt{pace} adds the baseline rung (\texttt{pace classify}), the audit card (\texttt{pace audit}), the self-consistency frontier and frontier gap (\texttt{pace frontier}, \texttt{pace report}), debate dynamics (\texttt{pace dynamics}), and the hidden-profile generator (\texttt{pace hidden-profile}); the sections above use \texttt{pace} for these. The three listings below are the unedited output of the commands on the files in the repository, run from its root with \texttt{PYTHONPATH=code}. The debate trace \texttt{code/examples/sample\_debate\_dynamics.jsonl} is synthetic: 40 items, 5 agents and 3 rounds, generated by a seeded simulation in which an incorrect agent becomes correct with probability 0.30 and a correct agent becomes incorrect with probability 0.10 per round, with every record marked \texttt{"synthetic": true}. It does not come from a model run. Because the command compares round 1 with round 3, its transition rates accumulate over two rounds and exceed the per-round values of the simulation. The files \texttt{demo\_pool.jsonl} and \texttt{demo\_run.jsonl} are also synthetic. For open-ended outputs that no exact-match rule can vote over, \texttt{pace aggregate-open} reads JSONL records with a \texttt{candidates} list and returns the medoid candidate (the one most similar to the others) under a fixed non-model similarity chosen with \texttt{--metric}: \texttt{code\_structural}, \texttt{ngram\_cosine}, \texttt{token\_jaccard} or \texttt{sequence\_ratio}; it is the noncommunicating aggregator against which a multi-round open-ended system is compared in \Cref{sec:decomposition}.

\paragraph{Hidden-profile generator.} The command writes 500 items and checks, from the generated scores, that the decoy leads every individual view without ties and the target leads the pooled evidence without ties (\Cref{sec:redundancy}). The check output is printed on the standard error stream.
\begin{lstlisting}[language={},basicstyle=\ttfamily\scriptsize,breaklines=true]
$ PYTHONPATH=code python -m pace hidden-profile --n 500 --seed 42 --out hp.jsonl
generated 500 items (agents 5, criteria 10, split 6, margin 3, seed 42, format text); decoy leads a view in violation: 0, target not leading pooled: 0, uncovered split criteria: 0
\end{lstlisting}

\paragraph{Debate dynamics.} Flip counts between the first and the last round, the two transition rates of \Cref{thm:conformity} and the implied threshold, the correlation of correctness among agents of an item per round (including shared item difficulty), its change from the first to the last round (the correlation added by debate), the effective team size of \Cref{prop:keff}, and tokens per call, with 95\% percentile intervals from a bootstrap over items. Answers also change between rounds without any peer message, so the two transition rates are read against a no-communication control log of the same task (the placebo run S7 or a repeated independent run); the tool reports the excess of each adjacent-round rate and of the added correlation over that control, and rates without such a control describe the run but do not attribute the change to peer content.
\begin{lstlisting}[language={},basicstyle=\ttfamily\scriptsize,breaklines=true]
$ PYTHONPATH=code python -m pace dynamics code/examples/sample_debate_dynamics.jsonl --n-boot 200
Debate dynamics: 40 items, rounds [1, 2, 3] (first 1 to last 3), 200 bootstrap resamples over items
flips (item-agent pairs): wrong->right 27, right->wrong 25, right->right 116, wrong->wrong 32
tau_ic (wrong -> right):  0.458 [0.321, 0.614]
tau_ci (right -> wrong):  0.177 [0.117, 0.237]
p_star_hat = tau_ic / (tau_ic + tau_ci): 0.721 [0.622, 0.818]
p1 (first round accuracy): 0.705 [0.650, 0.765]    p_last: 0.715 [0.645, 0.785]
delta_rho_debate = rho_total(round 3) - rho_total(round 1): 0.029 [-0.101, 0.144]
round 1: rho_total 0.007 [-0.078, 0.095], K_eff 4.86 [3.62, 5.00]
round 2: rho_total 0.044 [-0.062, 0.129], K_eff 4.25 [3.30, 5.00]
round 3: rho_total 0.036 [-0.056, 0.131], K_eff 4.37 [3.28, 5.00]
mean tokens per call by round: r1 180, r2 300, r3 420
rho_total is the correlation of correctness across agents of an item and includes shared item difficulty; delta_rho_debate is its change from the first to the last round, in which shared difficulty largely cancels, so it estimates the correlation added by communication. The tau rates assume the constant-rate model.
\end{lstlisting}

\paragraph{Frontier gap and rollout-card compute block.} The report command writes \texttt{card.json}, \texttt{card.yaml} and \texttt{card.tex}: rows for the annotations stream of a rollout card \citep{masters2026rollout} in the namespace \texttt{pace.compute}, and a reporting-rule entry with its drops manifest. The block follows Appendix E of that paper; it is a compute block and not a complete card, and it has not been run through the validator of the authors of that paper. A dimension that the log does not report is printed as n/a and stored as null.
\begin{lstlisting}[language={},basicstyle=\ttfamily\scriptsize,breaklines=true]
$ PYTHONPATH=code python -m pace report --pool code/pace/examples/demo_pool.jsonl --logs code/pace/examples/demo_run.jsonl --system debate --n-boot 200 --export-rollout-card card
items 40 | pool k=8 | system accuracy 0.525 at 1029 tokens/item | frontier 0.587 (inside) | gap -0.062 [-0.214, +0.111]
MAS/SAS ratios: r_reason=n/a, r_token=3.28, r_cost=n/a, r_lat=1.78, r_sample=5, r_sys=n/a
wrote card.json
wrote card.yaml
wrote card.tex
\end{lstlisting}

\clearpage
\section{Supplementary Sensitivity and Threshold Analyses}
\label{app:exploratory}

\begin{figure}[!htbp]
\centering
\includegraphics[width=0.92\textwidth]{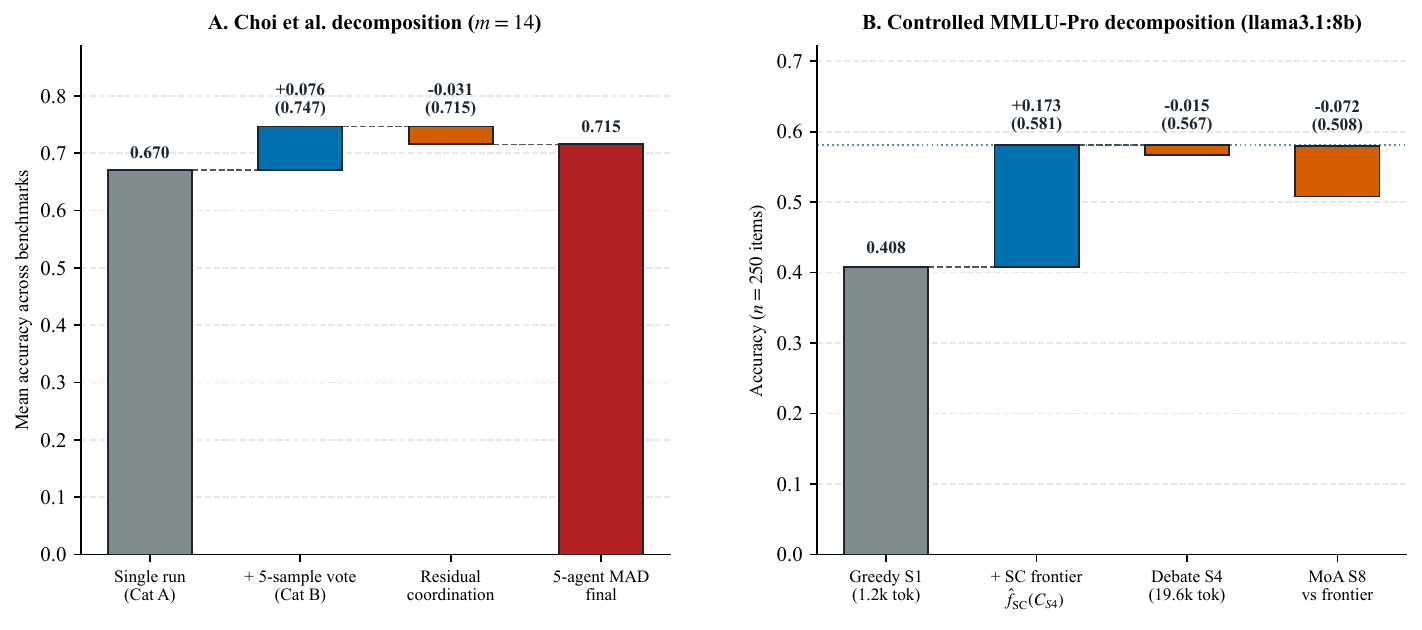}
\caption{Accounting decomposition on \citet{choi2025debate} and controlled MMLU-Pro. Panel~A (\citet{choi2025debate}, $m=14$ model and benchmark configurations): unweighted mean accuracy for the single-run baseline ($0.670$), majority voting over five samples ($0.747$, difference $+0.076$ before rounding), and five-agent decentralized debate ($T=2$ rounds, $0.715$; difference $-0.031$ relative to majority voting and $+0.045$ relative to the single-run baseline). Panel~B (controlled MMLU-Pro, $n=250$ items, \texttt{llama3.1:8b}): greedy decoding S1 ($\mmluGreedyAcc$ at $\mmluGreedyTok$ tokens), isotonic self-consistency frontier $\hat{f}_{\mathrm{SC}}(C_{\mathrm{S4}})$ ($\mmluDebateFrontier$), debate S4 ($\mmluDebateAcc$ at $\mmluDebateTok$ tokens, gap \meanControlledMmluDebateGap), and mixture-of-agents S8 ($\mmluMoaAcc$ at $\mmluMoaTok$ tokens, gap \meanControlledMmluMoaGap{} relative to its matched frontier).}
\label{fig:waterfall}
\end{figure}

\begin{figure}[!htbp]
\centering
\includegraphics[width=\textwidth]{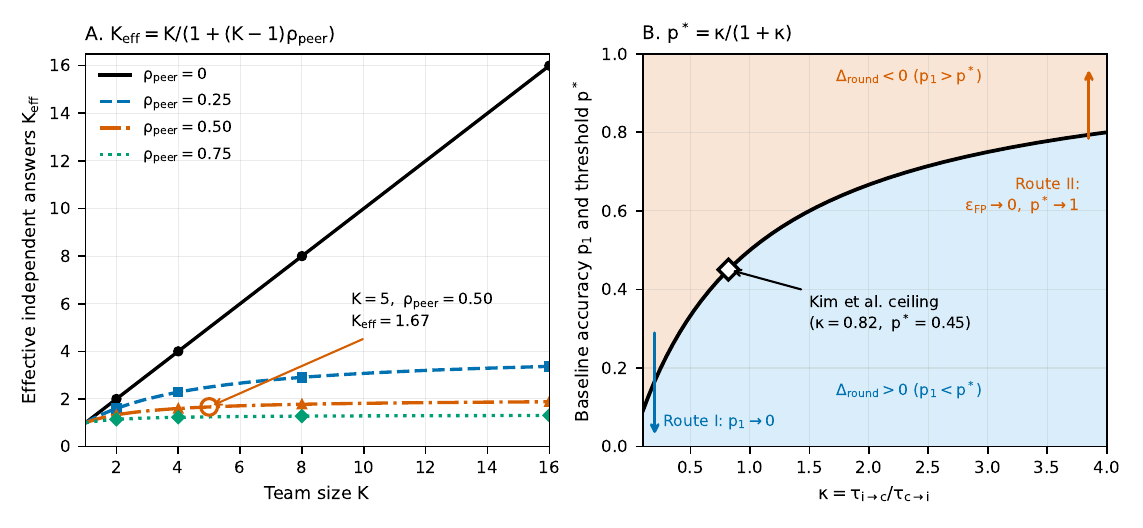}
\caption{Theoretical curves from \Cref{prop:keff} and \Cref{thm:conformity}. Panel A: variance-equivalent effective ensemble size $K_{\mathrm{eff}}=K/(1+(K-1)\rho_{\mathrm{peer}})$ under exchangeable error correlation for team size $K$ from 1 to 16 and peer correlation $\rho_{\mathrm{peer}}\in\{0,0.25,0.50,0.75\}$; the circle marks $K=5$, $\rho_{\mathrm{peer}}=0.50$, where $K_{\mathrm{eff}}=1.67$. Panel B: threshold $p^{*}=\kappa/(1+\kappa)$ for $\kappa=\tau_{i\to c}/\tau_{c\to i}$ from 0.1 to 4.0; a baseline accuracy $p_1$ above the curve gives $\Delta_{\mathrm{round}}<0$ and below it $\Delta_{\mathrm{round}}>0$. The diamond marks $\kappa=0.82$, $p^{*}=0.45$, a value fitted to the threshold reported by \citet{kim2026scaling} and not measured. Arrows mark Route~I ($p_1\to 0$) and Route~II ($\epsilon_{\mathrm{FP}}\to 0$, $p^{*}\to 1$).}
\label{fig:theory_phase_diagram}
\end{figure}

\begin{figure}[!htbp]
\centering
\includegraphics[width=0.92\textwidth]{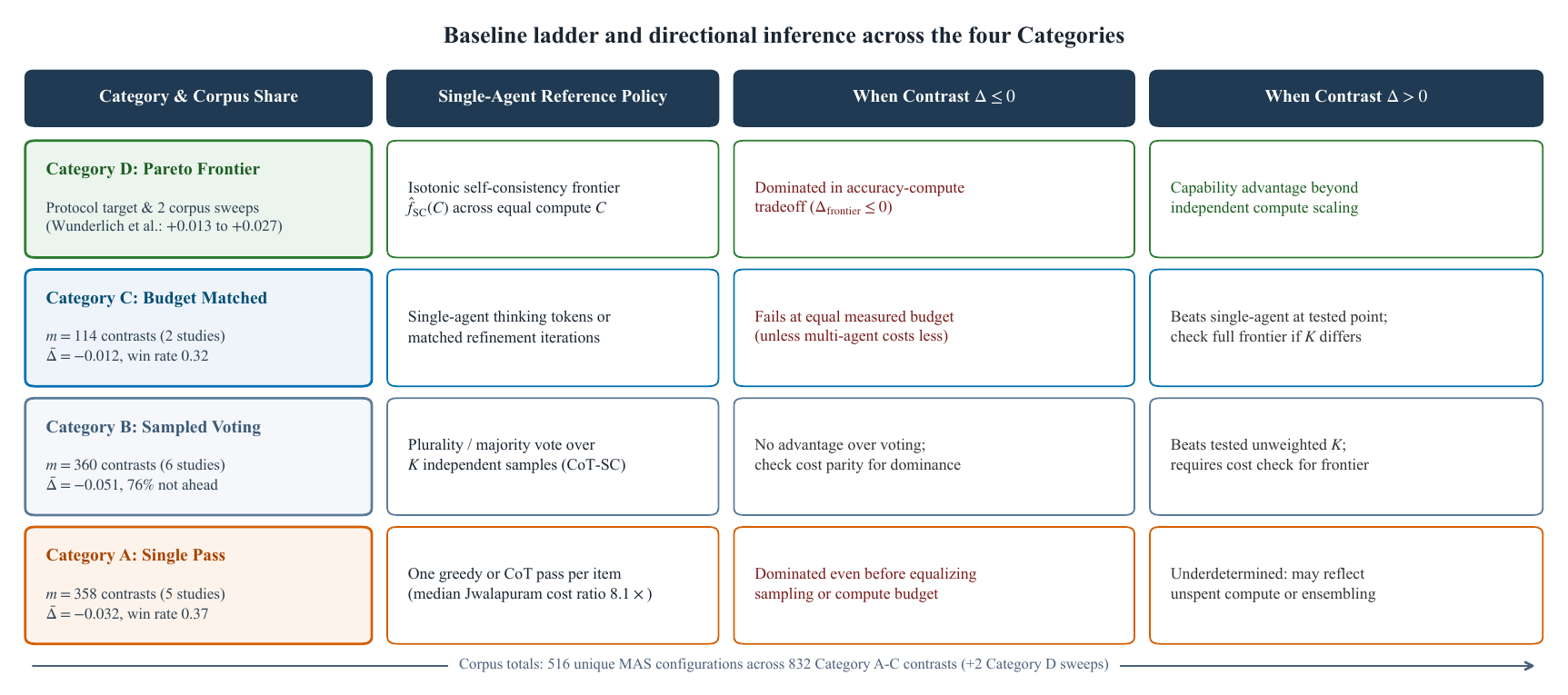}
\caption{The Baseline Ladder: single-agent baseline categories and resource conditions.
Categories~A and~B plot baselines without verified total test-time compute parity (Category~A: greedy single pass at $k=1$; Category~B: uncommunicating sampled baseline of $k$ draws). Categories~C and~D plot controlled compute comparisons on declared dimensions of the resource vector $\mathbf{R}$ (Category~C: matched resource budget or shared resource ceiling across arms; Category~D: Pareto frontier sweeping token expenditure or dollar cost across operating levels).}
\label{fig:baseline_ladder}
\end{figure}

The figures across the paper and appendices (\Cref{fig:baseline_ladder,fig:waterfall,fig:theory_phase_diagram,fig:resource_matrix,fig:evidence_flow,fig:empirical_ladder,fig:pareto_frontier,fig:reml_forest,fig:disaggregated_scaling,fig:controlled_frontier,fig:coordination_taxonomy,fig:decision_flowchart}) present the baseline ladder, accounting decomposition, theory phase diagram, resource matrix, evidence flow, empirical study ladder, SMFR frontier, Category B/C REML estimates, scaling curves, controlled confirmatory frontier, failure taxonomy, and deployment flow. This appendix reports supplementary sensitivity tables, relocated reference tables and checklists, extended scope notes, and capability threshold diagnostics.

\subsection{Tran and Kiela: all five multi-agent arms}
The main analysis uses the Debate and Sequential arms of Tran and Kiela. \Cref{tab:tran_all_arms} adds the subtask-parallel, parallel-roles, and ensemble arms against the same single-agent baseline at equal budgets of thinking tokens. Pooling all five arms gives a mean difference of $-0.017$ over 240 contrasts, so arm selection does not change the conclusion.
\begin{table}[t]
\centering
\footnotesize
\caption{Sensitivity to arm selection in Tran and Kiela: mean accuracy difference of each multi-agent arm from the single agent at the same budget of thinking tokens. The main analysis uses Debate and Sequential ($m=96$); the last row pools all five arms ($m=240$).}
\label{tab:tran_all_arms}
\begin{tabular}{lrrrr}
\toprule
Arm & $m$ & Mean $\Delta$ & Median $\Delta$ & MAS ahead \\
\midrule
Debate & 48 & $+0.002$ & $-0.012$ & 0.42 \\
Sequential & 48 & $-0.015$ & $-0.025$ & 0.25 \\
Subtask-parallel & 48 & $-0.027$ & $-0.024$ & 0.12 \\
Parallel roles & 48 & $-0.004$ & $-0.004$ & 0.40 \\
Ensemble & 48 & $-0.042$ & $-0.033$ & 0.19 \\
\midrule
All five arms & 240 & $-0.017$ & & \\
\bottomrule
\end{tabular}
\end{table}

\FloatBarrier

\subsection{Threshold analysis: what was claimed}
Evaluating 260 configurations across six benchmarks (the arXiv abstract gives 180 configurations in versions 1 and 2 and 260 in version 3; we did not check the published supplement) in their \textit{Nature Machine Intelligence} study, \citet{kim2026scaling} report capability saturation: on tasks where single-agent accuracy already exceeds approximately $45\%$, adding communicating agents yields negative returns across all four studied multi-agent topologies (Centralized, Decentralized, Hybrid, Independent). They also report that coordination efficacy depends on task structure, ranging from gains on decomposable financial reasoning to losses on sequential software planning. The $45\%$ figure represents a ceiling on gains (agents help less, or degrade performance, once the baseline model is competent), rather than a floor below which coordination fails.

\subsection{Why a threshold claim is hard to establish}
Three methodological caveats apply to any hypothesis of a capability threshold. First, accuracy is bounded in $[0, 1]$, so absolute gains compress near the ceiling; apparent nonlinear breaks in raw accuracy differences can arise from measurement scale alone \citep{schaeffer2024are}. Testing requires metrics invariant to scale such as the log odds ratio. Second, baseline accuracy is confounded with task domain and model family: across existing studies, contrasts with high or low baseline accuracy originate from distinct benchmarks. Third, isolating a capability ceiling at high accuracy requires a baseline using no more compute when establishing failure and at least as much when attributing gains; otherwise, apparent gains at low baseline accuracy may reflect additional inference compute rather than communicative architecture.

\subsection{What the extracted contrasts show}
We first split the multi-agent contrasts across the initial studies (excluding Pareto sweeps in Wunderlich et al.\ and majority-voting baseline rows in Choi et al., yielding $m=235$ configurations) at a single-agent accuracy of $0.45$ as a descriptive diagnostic (\Cref{tab:threshold_split}), and then fit an exploratory segmented regression with a knot at the same value (\Cref{tab:threshold_regression}).

\begin{table}[t]
\centering
\small
\caption{Mean accuracy difference $\bar\Delta$ (MAS $-$ SAS) for multi-agent contrasts with baseline accuracy below versus at or above $0.45$, by study ($m=235$ configurations excluding majority-voting baseline rows in Choi et al.\ and Wunderlich et al.\ Pareto sweeps). Descriptive only.}
\label{tab:threshold_split}
\begin{tabular}{lrrrr}
\toprule
Study (condition) & $m_{<.45}$ & $\bar\Delta_{<.45}$ & $m_{\ge .45}$ & $\bar\Delta_{\ge .45}$ \\
\midrule
Tran (matched thinking tokens) & 70 & $-0.004$ & 26 & $-0.014$ \\
Jwalapuram (unmatched cost) & 54 & $-0.003$ & 53 & $-0.011$ \\
Kim (matched iterations) & 9 & $+0.056$ & 9 & $-0.139$ \\
Choi (matched on agent count; MAD only) & 2 & $+0.112$ & 12 & $+0.034$ \\
\bottomrule
\end{tabular}
\end{table}

The empirical pattern in \Cref{tab:threshold_split} indicates that any capability threshold depends on domain and resource condition:
\begin{itemize}
    \item In the study matched on thinking tokens (Tran and Kiela) and the cost audit (Jwalapuram et al.), accuracy differences stay near zero or negative on both sides of $0.45$ ($-0.004$ vs.\ $-0.014$ in Tran; $-0.003$ vs.\ $-0.011$ in Jwalapuram). These studies show no structural break dependent on baseline accuracy.
    \item In Kim et al.'s tabulated contrasts (SWE-bench Verified and Terminal-Bench), the pattern matches their reported ceiling: gains below $0.45$ ($+0.056$) and losses above it ($-0.139$). However, this table rests on $n=20$ instances per cell across nine model configurations, offering qualitative alignment rather than high statistical power.
    \item In Choi et al.'s decentralized debate configurations ($m=14$, excluding majority-voting rows treated as a sampled baseline), differences remain positive against the single-run baseline on both sides of $0.45$ ($+0.112$ across two cells below $0.45$, $+0.034$ across twelve cells at or above $0.45$), reflecting unconstrained ensembling over five draws relative to a single run.
\end{itemize}

\begin{table}[t]
\centering
\small
\caption{Exploratory segmented regression of the log odds ratio on baseline single-agent accuracy ($m=235$ multi-agent configurations across four studies), with a knot at $p_{\mathrm{SAS}}=0.45$ (logit scale). Cells show slope (HC3 standard error). The pooled model has study fixed effects, so slopes are identified within studies. HC3 errors ignore clustering within studies and shared baselines, so they are optimistic. Majority-voting rows in Choi et al.\ (baseline arm) and Wunderlich et al.\ (Pareto sweep) are excluded. Each slope includes a mechanical term of $-1$ because the baseline appears in both the regressor and the log odds ratio; the difference between slopes does not. The analysis was run after the data were extracted.}
\label{tab:threshold_regression}
\resizebox{\textwidth}{!}{
\begin{tabular}{lrccc}
\toprule
Sample & $m$ & Slope below knot & Slope above knot & Difference (above $-$ below) \\
\midrule
All four studies, study fixed effects & 235 & $-0.06$ (0.09) & $-0.03$ (0.09) & $+0.03$ (0.17) \\
Excluding Tran at 100 thinking tokens & 219 & $-0.05$ (0.09) & $-0.03$ (0.09) & $+0.03$ (0.17) \\
Tran and Kiela only & 96 & $-0.08$ (0.05) & $+0.05$ (0.07) & $+0.13$ (0.11) \\
Jwalapuram et al. only & 107 & $+0.05$ (0.09) & $-0.08$ (0.12) & $-0.13$ (0.20) \\
Kim et al. only & 18 & $-1.04$ (0.36) & $+0.40$ (0.40) & $+1.44$ (0.73) \\
Choi et al. only & 14 & \multicolumn{3}{c}{fewer than 3 contrasts on one side of the knot} \\
\bottomrule
\end{tabular}
}
\end{table}

The regression models the log odds ratio as piecewise linear in the logit of baseline accuracy, with a knot at $0.45$ and, in the pooled model, a separate intercept for each study. Pooling all four studies ($m=235$) gives a slope difference of $+0.03$ (HC3 standard error $0.17$) between segments above and below the knot, showing no evidence of a break. Within studies, the difference is $+0.13$ (0.11) in Tran and Kiela, $+1.44$ (0.73) in Kim et al., and $-0.13$ (0.20) in Jwalapuram et al.; Choi et al.\ has only $2$ contrasts below $0.45$ and $12$ at or above $0.45$, leaving fewer than $3$ contrasts below the knot so no within-study Choi break is estimated. A capability ceiling removing gains above $0.45$ would predict a negative change in slope above the knot; only Jwalapuram et al.\ exhibits that sign, with an interval spanning zero. Because baseline accuracy enters both the regressor and the log odds ratio, every slope contains a mechanical term of $-1$ that cancels in the difference. Standard errors ignore clustering within studies and shared baselines, and the analysis was conducted after extraction.

\subsection{Differences in Task Structure and Compute Allocation}
The divergence between Kim et al.\ and Tran and Kiela reflects task structure and compute allocation. Kim et al.\ evaluate sequential tool-using tasks (code editing and terminal interaction) matched on iterations, attributing the ceiling to coordination overhead and error propagation when single agents are already capable. In contrast, Tran and Kiela evaluate multihop question answering matched on thinking tokens; there, single-agent reasoning matches or outperforms debate across nonvacuous budgets ($\ge 500$ tokens) regardless of baseline accuracy.

\paragraph{Scaling across budgets in Tran and Kiela.}
At 100 thinking tokens, multi-agent debate holds an unweighted advantage ($\bar\Delta = +0.077$). However, the source authors document that 100 thinking tokens yield truncated reasoning outputs for both single-agent and multi-agent systems. At 500 tokens, the difference shifts to $-0.012$; from 1,000 to 10,000 tokens, single agents lead ($-0.023$ to $-0.035$). Disaggregated scaling curves (\Cref{fig:disaggregated_scaling}) show this pattern across both FRAMES and MuSiQue 4-hop deduction.

\subsection{Summary of the Threshold Evidence}
Three conclusions follow: (i) extracted data do not support a universal capability threshold when reasoning tokens are matched; (ii) saturation in Kim et al.\ depends on task structure (present in sequential tool execution, absent in multihop deduction); and (iii) testing a general scaling threshold requires scale-invariant metrics such as log odds ratios, resource accounting across $\mathbf{R}$, and large item counts across continuous baseline accuracy variation.

\FloatBarrier

\subsection{PACE Constructs and Estimands Summary}
\label{app:pace_reference}
\Cref{tab:pace_master_reference} consolidates the policy classes, frontier definitions, feasibility criteria, allocation identities, transition parameters, and baseline rungs introduced in \Cref{sec:theory}.

\begin{table}[t]
\centering
\caption{Summary of PACE constructs, estimands, and diagnostic rules. The equations describe different components of multi-agent evaluation and should not be interpreted as interchangeable measures of coordination.}
\label{tab:pace_master_reference}
\scriptsize
\setlength{\tabcolsep}{3.0pt}
\renewcommand{\arraystretch}{1.12}
\begin{tabularx}{\textwidth}{@{} >{\raggedright\arraybackslash}p{1.55cm} >{\raggedright\arraybackslash}p{2.25cm} >{\raggedright\arraybackslash}X >{\raggedright\arraybackslash}X @{}}
\toprule
\textbf{Layer} & \textbf{Construct} & \textbf{Definition} & \textbf{Interpretation and diagnostic rule} \\
\midrule
Policy space & Policy classes & $\Pi_0\triangleq\Pi_{\mathrm{SA}}$, which contains $\Pi_{\mathrm{sample}}$ and $\Pi_{\mathrm{think}}$ as subsets; $\Pi_{\mathrm{agent}}$: cross-thread information flow & Classify policies by information flow, not architecture name. A multi-call workflow is not necessarily communicating. \\
Frontier & Noncommunicating frontier & $f(\mathbf{B}; t, \mathcal{M}_{\mathrm{mod}}, \mathcal{T}_{\mathrm{tool}}) = \sup_{\pi \in \Pi_0:\, \mathbb{E}[\mathbf{R}(\pi)] \preceq \mathbf{B}} \mathbb{E}[\operatorname{Accuracy}(\pi; t)]$ & Best performance available to the declared noncommunicating policy class under the stated ceiling. An observed baseline lower-bounds the frontier. \\
Feasibility & Directional resource feasibility & $\mathbf{R}_{\mathrm{base},\mathcal{J}} \preceq \mathbf{R}_{\mathrm{MAS},\mathcal{J}}$ & A baseline using no more of the declared resources rules out a positive frontier gap when it matches or exceeds the MAS. The reverse direction provides a demanding comparator but does not alone identify $f$. \\
Information & Contextual redundancy & $\operatorname{I}(Y; X_{1:K} \mid X_{\mathrm{shared}}) = 0 \implies \operatorname{I}(Y; \mathcal{H}^{(T)} \mid X_{\mathrm{shared}}) = 0$ & The transcript adds no target information conditional on shared context. A joint readout of $(X_{\mathrm{shared}}, \mathcal{H}^{(T)})$ has the same Bayes accuracy as $X_{\mathrm{shared}}$ alone. \\
Allocation & Effective per-call budget & $B_{\mathrm{eff}}(K,T) = \max\!\big(0, \frac{B - C_{\mathrm{tot}}(K,T)}{KT}\big)$ & Message history and scaffolding consume resources that could otherwise fund reasoning tokens. Exact overhead depends on the communication protocol. \\
Baseline strength & Sampling gap & $\Delta_{\mathrm{gap}} = (p_{\mathrm{MAS}} - p_{\mathrm{sampled}}) - (p_{\mathrm{MAS}} - p_{G}) = p_{G} - p_{\mathrm{sampled}}$ & Measures the change in reference baseline strength. Because $p_{\mathrm{MAS}}$ cancels, this is a baseline-strength check rather than an independent MAS effect. \\
Aggregation & Effective sample size & $K_{\mathrm{eff}} = K / [1 + (K-1)\rho_{\mathrm{peer}}]$ & For exchangeable answers, positive pairwise correlation reduces the effective size of an average; a variance-equivalence identity, not a universal voting accuracy formula. \\
Compute match & Token-equated pool size & $k_{\mathrm{tok}} = KT + (T-1)K(K-1)r$ & Protocol-specific rule for sizing an independent single-agent pool under stated message-length assumptions. Actual token counts should be measured directly. \\
Transitions & Conformity threshold & $p^* = \tau_{i\to c}/(\tau_{i\to c} + \tau_{c\to i}) = \kappa/(1+\kappa)$ & Under constant-rate transitions, accuracy drifts downward when $p_1 > p^*$. A model diagnostic, not a universal empirical threshold. \\
Verification & Asymmetric verifier & $\epsilon_{\mathrm{FP}} < \kappa\,\frac{1-p_1}{p_1}(1-\epsilon_{\mathrm{FN}})$ & Gives positive round drift under the verifier model. Report false-positive and false-negative rates and grant the same verifier affordance to the baseline. \\
Evaluation & Baseline ladder & $\mathrm{A}$: greedy, $\mathrm{B}$: sampled, $\mathrm{C}$: named resource, $\mathrm{D}$: frontier & Report the baseline category, controlled dimensions $\mathcal{J}$, resource-status vector, test size, uncertainty, task structure, and search expenditure. \\
\bottomrule
\end{tabularx}
\end{table}

\subsection{Adjudication Summary, Moderator Synthesis, and Auxiliary Corpus Figures}
\label{app:adjudication_notes}
Across topology archetypes (\Cref{tab:supp_topology_archetypes}), Category~B/C means range from $-0.015$ for single-context iterative refinement ($m=48$) and $-0.017$ for two-stage feedforward ensemble refinement ($m=7$; $-0.024$ over $m=151$ including Tran and Kiela's supplementary feedforward arms) to $-0.035$ for symmetric peer debate ($m=237$), $-0.058$ for centralized orchestrators ($m=101$; $-0.065$ excluding Expert MAS), and $-0.059$ for automated graph and prompt search ($m=81$, where only 1 contrast exceeds $+1.96\,\mathrm{SE}$ and $\bar\Delta_{\mathrm{gap}}=-0.035$). Across task domains (\Cref{tab:supp_domain_moderators}), means remain negative in multihop and graduate knowledge QA ($-0.031$, $m=268$), search, financial, and scientific agent tasks ($-0.044$, $m=48$), symbolic and competition mathematics ($-0.055$, $m=123$), and executable software engineering and terminal tasks ($-0.075$, $m=35$). Across backbone groups (\Cref{tab:supp_model_scale_moderators}), standard instruction-tuned models average $-0.053$ ($82.0\%$ not ahead, $\bar\Delta_{\mathrm{gap}}=-0.021$), native test-time reasoning models average $-0.028$ ($-0.037$ omitting 16 truncated 100-token rows, $\bar\Delta_{\mathrm{gap}}=-0.041$), and parameter groups remain negative across small open-weight ($\le 14\text{B}$, $-0.090$), mid/large open-weight ($15\text{B}$--$120\text{B}$, $-0.032$), and frontier proprietary APIs ($-0.034$).

\begin{table}[t]
\centering
\small
\caption{Adjudication and denominator summary. Panel~A partitions the \numContrastsTotal\ unique configurations by what reported resources license; Panel~B gives counts among the \numContrastsCatBC\ Category~B/C contrasts; Panel~C summarizes dual-baseline pairing. Signs reflect reported point estimates with ties counted as not ahead, and noise-band counts use \Cref{eq:noise_band} at $\varphi=0.50$. Full six-slice breakdowns appear in \Cref{tab:adjudication_summary}.}
\label{tab:adjudication}
\begin{tabular}{@{}p{0.68\textwidth}rr@{}}
\toprule
Outcome & Count & Share \\
\midrule
\multicolumn{3}{@{}l}{\textit{Panel A: unique multi-agent configurations ($M=\numContrastsTotal$, 9 studies)}} \\
A baseline with no more of a known resource, or under the same ceiling, scored at least as high & \numConfigFeasible & \pctConfigFeasible\% \\
System ahead under an equal ceiling or lower observed cost & \numConfigAhead & \pctConfigAhead\% \\
Undetermined: ahead only of baselines that used less compute or lacked per-arm cost & \numConfigUndet & \pctConfigUndet\% \\
\midrule
\multicolumn{3}{@{}l}{\textit{Panel B: Category~B/C contrasts ($M=\numContrastsCatBC$, 8 studies)}} \\
Not ahead of the baseline & 352 & \pctNotAheadBC\% \\
Not ahead of a baseline with no more of the controlled or reported resource & 171 & 36\% \\
\quad of which below $-1.96\,\mathrm{SE}$ & \numFeasibleNoiseBelow & \\
Baseline dominates on accuracy and observed cost & 51 & \pctParetoDominatedBC\% \\
Ahead at a shared ceiling or matched generation budget (of 156) & 44 & \pctSharedCeilAhead\% \\
\quad of which above $+1.96\,\mathrm{SE}$ & \numSharedCeilNoiseAbove & \\
\midrule
\multicolumn{3}{@{}l}{\textit{Panel C: paired dual-baseline checks (\numStudiesDual\ studies)}} \\
Paired Category~B vs.\ Category~A contrasts with negative shift ($\Delta_{\text{gap}} < 0$) & 279 / 318 & 88\% \\
Deduplicated (study, benchmark, model) baseline cells with $\Delta_{\text{gap}} < 0$ & 64 / 72 & 89\% \\
\bottomrule
\end{tabular}
\end{table}

\begin{figure}[t]
\centering
\includegraphics[width=0.92\textwidth]{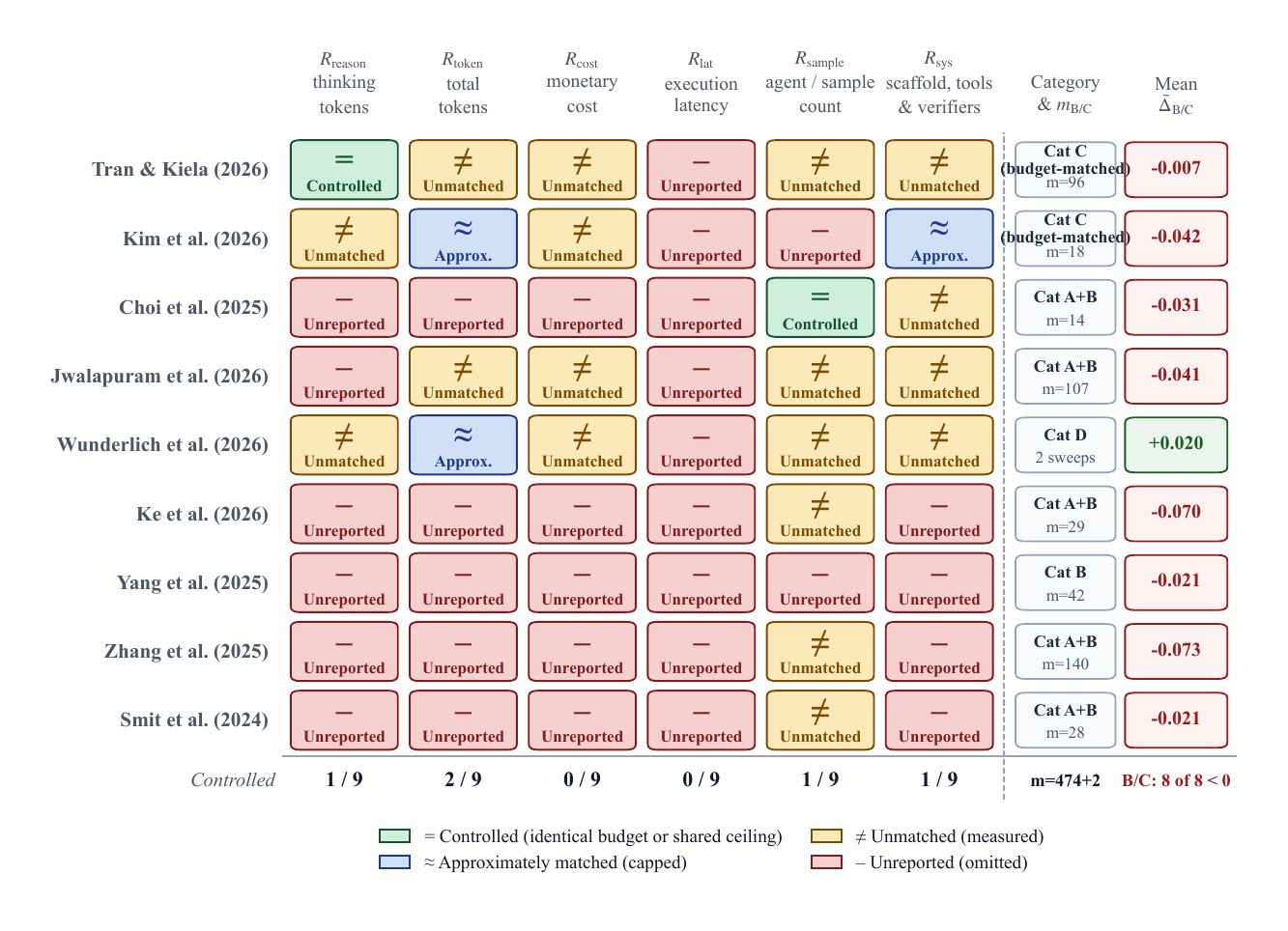}
\caption{Resource accounting matrix across all nine corpus studies on the six dimensions of $\mathbf{R}$, showing verification status (controlled, approximately matched, unmatched, unreported; the iteration cap of Kim et al.\ and the generation cap of Yang et al.\ bound the number of calls $N_{\mathrm{calls}}$, which is logged separately, so their $R_{\mathrm{sample}}$ is unreported), baseline category, contrast counts ($m_{\mathrm{B/C}}$), and mean accuracy differences ($\bar{\Delta}_{\mathrm{B/C}}$).}
\label{fig:resource_matrix}
\end{figure}

\begin{figure}[t]
\centering
\includegraphics[width=0.82\textwidth]{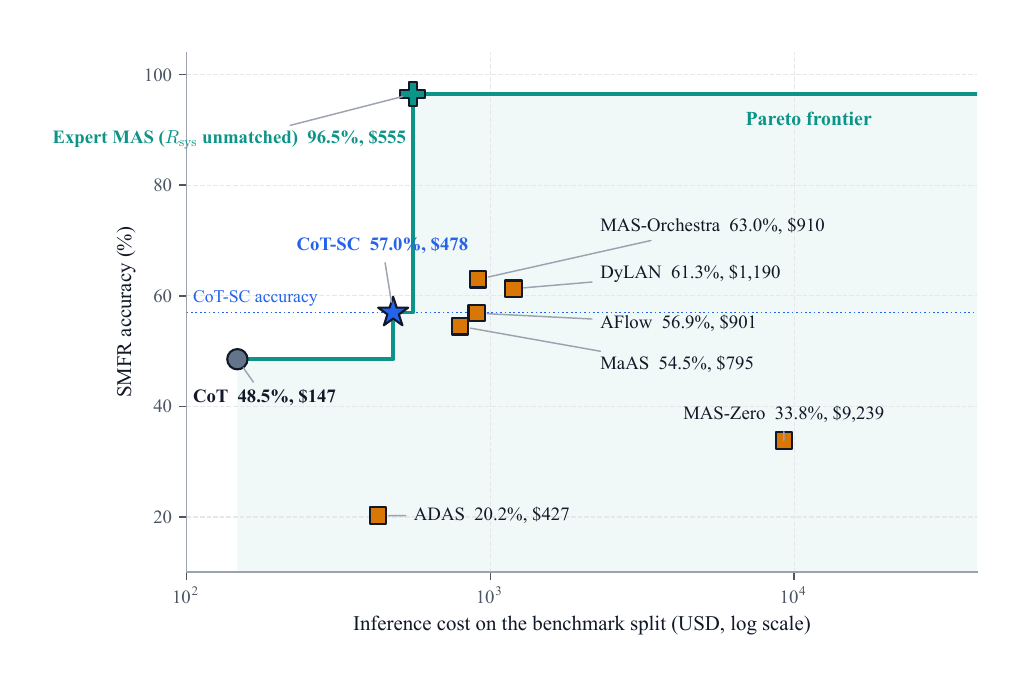}
\caption{Accuracy versus inference cost on the SMFR benchmark (GPT-5, $N=588$). Plotted values are accuracy and API dollar costs reported in Table~4 of \citet{jwalapuram2026illusion} across automated multi-agent systems, CoT, CoT-SC, and Expert MAS.}
\label{fig:pareto_frontier}
\end{figure}

\begin{figure}[t]
\centering
\includegraphics[width=0.92\textwidth]{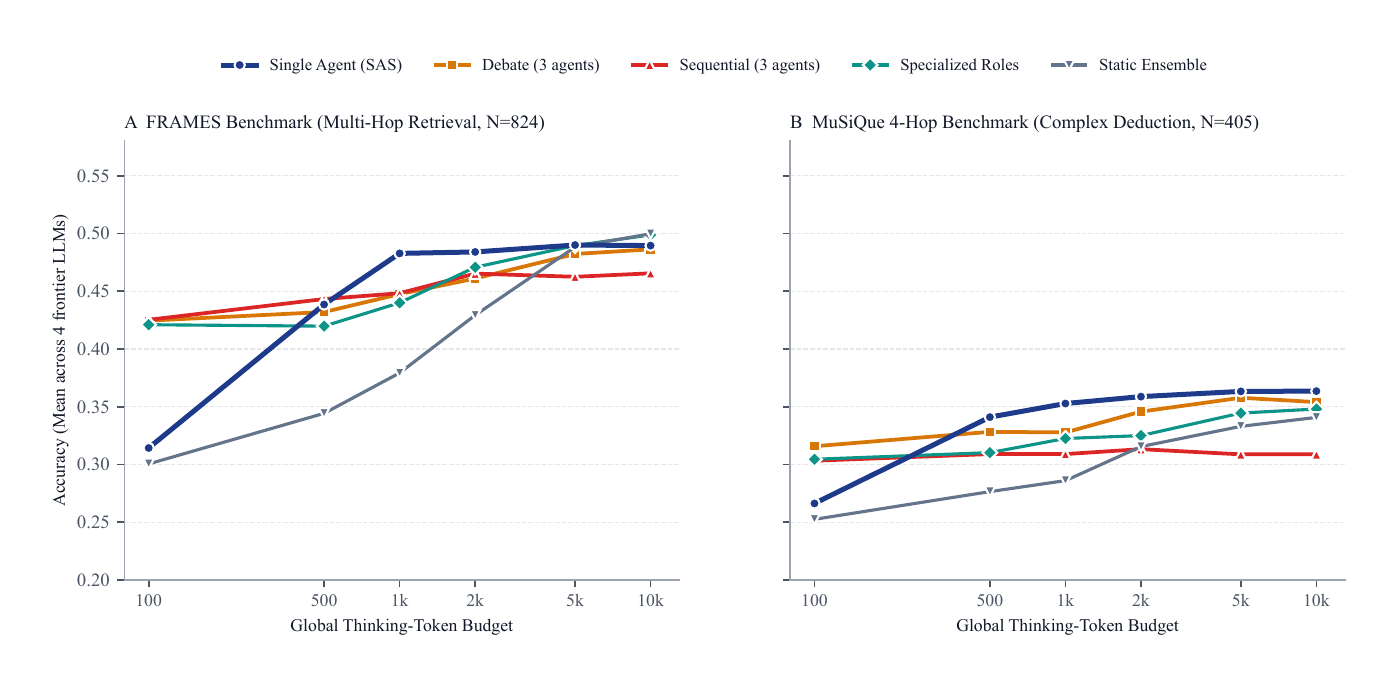}
\caption{Mean accuracy across four models on FRAMES ($N=824$) and MuSiQue 4-hop ($N=405$) plotted against global budget of thinking tokens (100--10,000 tokens, log scale) for single-agent scaling, Debate, Sequential, Roles, and Ensemble in \citet{tran2026thinking}.}
\label{fig:disaggregated_scaling}
\end{figure}

\begin{table}[t]
\centering
\scriptsize
\setlength{\tabcolsep}{4pt}
\caption{Operational entry points for four AI research communities derived from \texttt{baseline\_tier\_contrasts.csv} and the audited source corpus, listing extracted Category~B/C contrast counts ($m_{\mathrm{BC}}$), unweighted mean differences ($\bar\Delta_{\mathrm{BC}}$), and the required PACE check.}
\label{tab:pace_subfield_hooks}
\resizebox{\textwidth}{!}{
\begin{tabular}{p{0.22\textwidth}p{0.29\textwidth}p{0.41\textwidth}}
\toprule
Research community & Corpus slice and empirical summary & Primary PACE check and command \\
\midrule
Native reasoning models & $m_{\mathrm{BC}}=197$ contrasts at untruncated thinking budgets ($\ge 500$ thinking tokens): $\bar\Delta_{\mathrm{BC}}=-0.037$, $70.1\%$ not ahead ($\bar\Delta_{\mathrm{gap}}=-0.041$ across $105$ paired contrasts). & Compare against single-agent CoT-SC with $k \ge k_{\mathrm{tok}}$ (\Cref{eq:ktok}) at untruncated thinking budgets; separate thinking and visible output tokens in $\mathbf{R}$ (\texttt{pace report}). \\
Automated workflow and graph search & $m_{\mathrm{BC}}=81$ contrasts across two studies: $\bar\Delta_{\mathrm{BC}}=-0.059$, $\overline{\ln\mathrm{OR}}_{\mathrm{BC}}=-0.356$, $76.5\%$ not ahead, $1/81$ ($1.2\%$) above $+1.96\,\mathrm{SE}$, $\bar\Delta_{\mathrm{gap}}=-0.035$. & Log offline validation/search tokens and test inference tokens per query separately inside $R_{\mathrm{token}}$ and $R_{\mathrm{cost}}$; compare against single-agent tuning at equal expenditure (\texttt{pace audit}). \\
Software engineering and terminal agents & $m_{\mathrm{BC}}=35$ contrasts (\texttt{SWE-Bench Lite}, \texttt{SWE-bench Verified}, \texttt{Terminal-Bench}): $\bar\Delta_{\mathrm{BC}}=-0.075$, median $-0.068$, $82.9\%$ not ahead, $\bar\Delta_{\mathrm{gap}}=-0.047$ ($17$ pairs). & Supply identical filesystem, shell, test runner, and state reset permissions ($R_{\mathrm{sys}}$) to both arms; report $R_{\mathrm{lat}}$, iteration cap, and tool calls per item (\texttt{pace classify}). \\
Collaborative and heterogeneous ensembles & \texttt{Mixture-of-Agents} leads CoT-SC with 10 samples by $+0.027$ at comparable compute read off Pareto fronts using Llama-3.1-70B copies; \texttt{Heter-MAD} in \citet{zhang2025stop} lacks sampled mixed comparators (\Cref{sec:appendix_hetermad}). & Benchmark heterogeneous agent debate against an uncommunicating vote over the exact same constituent model pool and against the strongest single constituent model at equal $R_{\mathrm{cost}}$ (\texttt{pace dynamics}). \\
\bottomrule
\end{tabular}}
\end{table}

\paragraph{Operational summary across the four research communities.}
Across these four communities (\Cref{tab:pace_subfield_hooks}), PACE translates empirical failure channels into concrete experimental protocols. \emph{Native test-time reasoning evaluations} should verify that individual chains operate above starvation limits ($\ge 500$ thinking tokens) and anchor multi-agent claims against $k \ge k_{\mathrm{tok}}$ self-consistency sweeps via \texttt{pace report}. \emph{Automated workflow and graph search pipelines} should separate offline topology search cost from test-time execution in $R_{\mathrm{cost}}$ via \texttt{pace audit}. \emph{Software engineering and terminal agents} should hold execution environment and tool affordances ($R_{\mathrm{sys}}$) identical across multi-agent and single-agent runs while logging $R_{\mathrm{lat}}$ via \texttt{pace classify}. \emph{Heterogeneous model ensembles} should test Route~III against noncommunicating votes over identical model pools and inspect error-correction versus conformity rates via \texttt{pace dynamics}.

\subsection{Extended Scope and Anticipated Objections}
\label{app:scope_notes}
\paragraph{Limitations of the corpus (continued from the limitations of the corpus synthesis in \Cref{app:moved_s5_limits}).}
(vii)~\emph{Implementation fidelity.} Each framework is evaluated as configured by the auditing study; automated-design frameworks (ADAS, AFlow, MaAS, MAS-Zero) may not have been searched again for each benchmark, and Smit et al.\ report medians over hyperparameter configurations. In Zhang et al., AgentVerse with Llama-3.1-8B scores between $0.054$ and $0.246$ on six benchmarks on which the chain-of-thought baseline of the same model scores between $0.401$ and $0.864$; gaps of this size may reflect a failure of the pipeline with a small model and not a property of debate. Omitting the seven Category~B contrasts of that pair changes the Category~B/C contrast mean from \meanCatBC\ to $-0.036$, and omitting its nine Category~A contrasts changes the Category~A mean from \meanCatA\ to $-0.024$. (viii)~\emph{Task coverage.} Software-engineering and web-browsing benchmarks (SWE-bench Lite, SWE-bench Verified, Terminal-Bench, BrowseComp-Plus) supply 62 of the \numContrastsCatBC\ Category~B/C contrasts, 18 of which rest on 20 items each. The corpus therefore says little about long-horizon tasks in which agents hold separate context or tool state. (ix)~\emph{Model mix and cost.} No pooled contrast compares a mixed-model panel with a vote over the same models or with the best single model at equal cost, so Route~III is untested here. (x)~\emph{Choice of axis.} No study controls latency (\Cref{fig:resource_matrix}), so the corpus cannot show whether parallel agents reach a given accuracy sooner than sequential sampling. PACE reports the gap on every measured axis and does not treat tokens as the only budget. (xi)~\emph{Baseline maxima.} The CoT and CoT-SC rows of Jwalapuram et al.\ are maxima over entries. If $n$ entries differ by run-to-run noise with standard deviation $s$, the expected maximum exceeds the mean by about $1.27s$ for $n=6$ and $1.35s$ for $n=7$; with the run-level standard deviations reported in that table, the inflation can be of the same order as the mean difference against CoT-SC ($-0.041$). Because $\operatorname{Var}(\text{CoT-SC}_5)=\operatorname{Var}(\text{CoT})/5$, the expected maximum over six entries inflates the CoT row by about $1.267\,s_{\mathrm{CoT}}$ and the CoT-SC row by only $0.567\,s_{\mathrm{CoT}}$, which attenuates that study's sampling gap ($-0.033$) toward zero by $0.700\,s_{\mathrm{CoT}}$, so the reported gap is conservative. Estimates without that study are in \Cref{tab:appA_bc_sensitivity}. (xii)~\emph{Item counts.} The noise bands use dataset sizes where evaluated counts are not stated (Zhang et al., Tran and Kiela); capping every assumed cell size at $N \le 500$ moves 25 negative and eight positive contrasts inside the band (146 below $-1.96\,\mathrm{SE}$, 301 within $\pm 1.96\,\mathrm{SE}$, and 27 above $+1.96\,\mathrm{SE}$ vs.\ 171 / 268 / 35), so the count of contrasts below $-1.96\,\mathrm{SE}$ falls from 171 to 146 and the count above $+1.96\,\mathrm{SE}$ from 35 to 27 if evaluated subsets were smaller; the sign of the pooled mean does not change. (xiii)~\emph{Search and tuning budgets.} Automated topology and prompt search studies often tune graph structure, router parameters, or prompt templates on development sets before evaluating test accuracy, while reporting only inference cost on the test split. Within PACE, offline search tokens and dollar charges are logged separately from test expenditure per query inside $R_{\mathrm{token}}$ and $R_{\mathrm{cost}}$, and equivalent tuning compute is allotted to single-agent prompt search or sampling schedules before attributing test differences to runtime multi-agent coordination.

\paragraph{Additional context outside the extracted corpus.} Two related long-context and cross-model collaboration frameworks outside our nine-study extraction are cited for context \citep{zhang2024chainofagents,chen2024reconcile}.

\subsection{Practitioner Route Tests, Decision Flow, and Reporting Templates}
\label{app:checklist_templates}

\begin{table}[t]
\centering
\small
\caption{Three route tests. Each row gives the condition under which communication can add to a noncommunicating policy, the control that should accompany the claim, and the quantity to report. To our knowledge none of the pooled contrasts includes one of these controls.}
\label{tab:route_tests}
\begin{tabularx}{\textwidth}{@{}l >{\raggedright\arraybackslash}X >{\raggedright\arraybackslash}X >{\raggedright\arraybackslash}X@{}}
\toprule
Route & Condition to state & Control to run & Quantity to report \\
\midrule
I. Split evidence & Some agent holds evidence that others lack & Vote over the same split views; the same items with shared evidence; one agent with all evidence & Pooling efficiency and the split-minus-shared difference \\
II. External verifier & A verifier with a low false-accept rate is available & Sampling filtered by the same verifier at equal tokens and equal $R_{\mathrm{sys}}$ & Frontier gap against verifier-filtered sampling \\
III. Model diversity & Within-item correlation is lower across models than within one model & Vote over the same models; the best single model with self-consistency at equal cost & Debate minus mixed-model vote; $K_{\mathrm{eff}}$ from $\rho_{\mathrm{within}}$ \\
\bottomrule
\end{tabularx}
\end{table}

\clearpage
\begin{tcolorbox}[colback=white,colframe=black!60,boxrule=0.6pt,left=4pt,right=4pt,top=3pt,bottom=3pt]
\textbf{PACE decision rules.}
(1)~Before crediting communication, run self-consistency with $k \ge k_{\text{tok}}$ samples (\Cref{eq:ktok}) and report the frontier gap with its interval.
(2)~Credit coordination only if the gap is positive on the declared axis against a baseline that received at least as much of that resource; conclude that there is no gain if a baseline that received no more matches or exceeds the system.
(3)~If the claim rests on split evidence, a verifier, or model diversity, run the control of \Cref{tab:route_tests}.
(4)~Give any verifier or tool to both arms and report $R_{\text{sys}}$.
\end{tcolorbox}

\begin{figure}[t]
\centering
\includegraphics[width=0.92\textwidth]{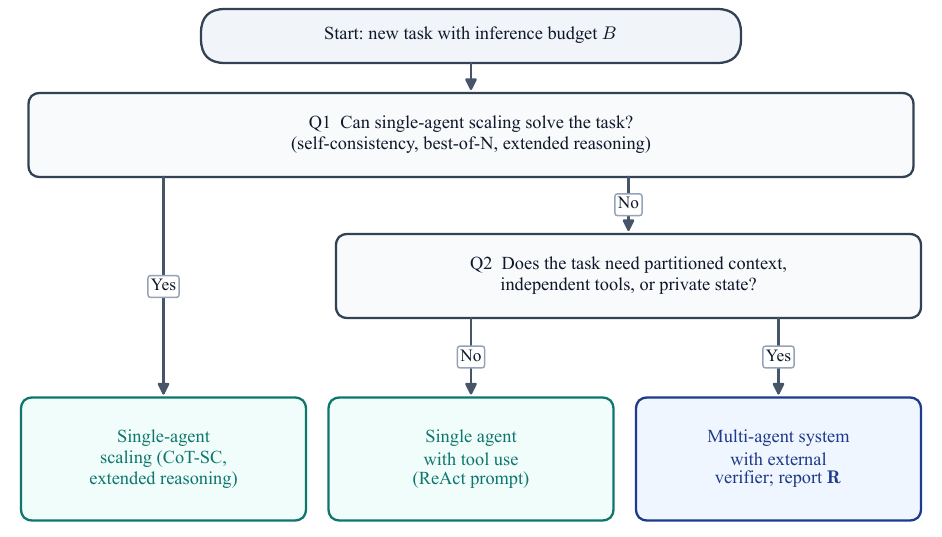}
\caption{Decision flow across single-agent scaling and multi-agent architectures conditioned on feasibility checks and resource vector $\mathbf{R}$.}
\label{fig:decision_flowchart}
\end{figure}

\begin{center}
\fbox{\begin{minipage}{0.94\textwidth}
\small
\textbf{Suggested reporting text.}
\par\smallskip
\textit{For authors (experimental setup).} ``Following the Parity-Aware Compute Evaluation (PACE) protocol, we report the resource vector $\mathbf{R}$ with six dimensions for every system. We compare our multi-agent system with both a single-pass baseline (Category~A, Greedy Single-Pass) and a sampled single-agent baseline whose total tokens are at least those of our system (Category~B or C; CoT-SC with $k$ samples satisfying $k \ge k_{\text{tok}}$), and we report the gap to the self-consistency frontier at our system's measured tokens with a paired bootstrap interval.''
\par\smallskip
\textit{For reviewers (baseline parity request).} ``The submission compares only against a single greedy pass (Category~A). Under the PACE reporting checklist, please add a Category~B (sampled baseline, e.g.\ CoT-SC with $k$ samples) or Category~C (feasible or budget-matched baseline) comparison, and report the resource vector $\mathbf{R}$ for both. Without feasible or compute-matched controls, the reported advantage cannot be disentangled from sample multiplicity or inference expansion.''
\par\smallskip
\textit{Reviewer short form.} (a)~Which rung is the strongest single-agent baseline? (b)~For each reported resource, did that baseline receive less, about the same, or more than the multi-agent system? (c)~Does the stated conclusion follow from (a) and (b) under the feasibility rule?
\end{minipage}}
\end{center}

\begin{center}
\fbox{\begin{minipage}{0.94\textwidth}
\small
\textbf{Minimum Viable PACE Audit: Three Steps for Limited-Compute Evaluations.}
\par\smallskip
\textbf{Step 1: Record the execution.} For the multi-agent run and its noncommunicating reference, record all six resource dimensions: $R_{\mathrm{reason}}$, $R_{\mathrm{token}}$, $R_{\mathrm{cost}}$, $R_{\mathrm{lat}}$, $R_{\mathrm{sample}}$, and $R_{\mathrm{sys}}$. Record independent answer trajectories ($R_{\mathrm{sample}}$) separately from total model-generation calls ($N_{\mathrm{calls}}$), judge calls, and tool calls. Mark every unreported dimension as unreported rather than assigning zero.
\par\smallskip
\textbf{Step 2: Construct a nested sampled baseline.} When initial independent answers are retained, aggregate the round-1 answers before inter-agent communication using the same voting or aggregation rule used by the evaluation (as in \citealp{choi2025debate}). This nested baseline requires no additional generation calls and provides a Category~B sampled reference with no more independent trajectories than the subsequent debate run, though it is not a token-, cost-, or latency-matched Category~C baseline.
\par\smallskip
\textbf{Step 3: Report the directional comparison.} Compare the final multi-agent result with the nested baseline, report paired item outcomes or an uncertainty interval, and publish the PACE resource-status vector. When compute permits, add one single-agent condition under a declared thinking, generation, or iteration ceiling.
\end{minipage}}
\end{center}

\begin{table}[t]
\centering
\small
\caption{Three PACE reporting levels by evaluation budget and study scope. \texttt{PACE-Minimum} provides a reporting floor for limited-compute evaluations without claiming full frontier parity.}
\label{tab:reporting_levels}
\begin{tabularx}{\textwidth}{@{}l l >{\raggedright\arraybackslash}X@{}}
\toprule
Reporting level & Intended users & Required reporting elements \\
\midrule
\texttt{PACE-Minimum} & Limited-compute evaluations & One greedy reference (Category~A), one nested round-1 or sampled reference (Category~B) or named-resource feasible reference where available, all six resource statuses in $\mathbf{R}$, independent trajectories versus total model calls ($N_{\mathrm{calls}}$), benchmark item count ($N$), run/seed details, and paired accuracy difference. \\
\texttt{PACE-Recommended} & Method and benchmark papers & Multiple sample counts or thinking budgets ($k \ge k_{\text{tok}}$), paired item outcomes with bootstrap intervals, variance across runs, resource ratios ($R_{\text{token}}, R_{\text{cost}}, R_{\text{lat}}$), task decomposability note, and equal tool/verifier access ($R_{\text{sys}}$). \\
\texttt{PACE-Extended} & Full scaling and audit studies & Empirical Pareto frontier across budgets, cost and latency curves, within-item correlation ($\rho_{\mathrm{within}}$ and $K_{\mathrm{eff}}$), cumulative architecture-search expenditure, and machine-readable item traces. \\
\bottomrule
\end{tabularx}
\end{table}

\paragraph{Connection to ensemble covariance.}
The effective ensemble size $K_{\mathrm{eff}} = K/[1+(K-1)\rho]$ in \Cref{sec:theory} parallels the classical variance-covariance decomposition for averaged predictors \citep{krogh1994neural,ueda1996generalization}. For $K$ exchangeable estimators with common bias $b$, marginal variance $\sigma^2$, and pairwise error correlation $\rho$, the mean-squared error of the ensemble average $\bar f = \frac{1}{K}\sum_{a=1}^K f_a$ satisfies
\[
\operatorname{MSE}(\bar f) = b^2 + \frac{\sigma^2}{K}\big[1 + (K-1)\rho\big] = b^2 + \frac{\sigma^2}{K_{\mathrm{eff}}}.
\]
Whereas this identity is exact for linear averaging under squared loss, discrete majority voting over zero-one correctness indicators is nonlinear, so $K_{\mathrm{eff}}$ serves in our framework as a first-order summary of correlated binary indicators rather than an exact binomial sample-size equivalence.

\paragraph{Separating architecture benefit from communication overhead (protocol diagnostic).}
Let $p_{\mathrm{base}}$, $p_{\mathrm{comm}}$, and $p_{\mathrm{placebo}}$ denote accuracy of the noncommunicating baseline matched on resource budget, the communicating multi-agent system, and a placebo topology that keeps the turn schedule and tool permissions and replaces peer content by messages from another item, matched in character length but not in tokens (the placebo design of \Cref{sec:controlled}). Defining $\Delta_{\mathrm{arch}} = p_{\mathrm{comm}} - p_{\mathrm{placebo}}$ and $\Delta_{\mathrm{overhead}} = p_{\mathrm{base}} - p_{\mathrm{placebo}}$ separates semantic communication effects from raw context expansion ($p_{\mathrm{comm}} - p_{\mathrm{base}} = \Delta_{\mathrm{arch}} - \Delta_{\mathrm{overhead}}$), requiring both a positive net accuracy advantage ($p_{\mathrm{comm}} > p_{\mathrm{base}}$, equivalently $\Delta_{\mathrm{arch}} > \Delta_{\mathrm{overhead}}$) and $\Delta_{\mathrm{arch}} > 0$.

\paragraph{Comparing verification with aggregation (untested protocol objective).}
Rather than reporting error correction rates alone, compare the net verifier transition drift $\Delta p_{\mathrm{ver}} = (1 - p_1)\tau_{i\to c}^{\mathrm{ver}} - p_1\tau_{c\to i}^{\mathrm{ver}}$ against the net aggregation drift $\Delta p_{\mathrm{agg}} = (1 - p_1)\tau_{i\to c}^{\mathrm{agg}} - p_1\tau_{c\to i}^{\mathrm{agg}}$, testing whether $\Delta p_{\mathrm{ver}} > \Delta p_{\mathrm{agg}}$ while tracking both error correction ($\tau_{i\to c}$) and corruption ($\tau_{c\to i}$) transitions (\Cref{thm:conformity}).

\paragraph{Cost efficiency as a separate design objective (untested protocol objective).}
Treat monetary cost reduction $\Delta C = C_{\mathrm{MAS}} - C_{\mathrm{baseline}} < 0$ as a separate deployment objective rather than a substitute for accuracy parity: any claim of lower dollar expenditure must co-report accuracy, latency ($R_{\mathrm{lat}}$), and every remaining component of $\mathbf{R}$.

\paragraph{Additional untested condition: parallel tools and separated state (untested protocol objective).}
Beyond Routes~I--III, parallel tool invocation inside separated execution sandboxes represents a candidate structural condition for multi-agent workflows when serial execution would incur long wait times or workspace contamination. Evaluating this candidate requires a noncommunicating baseline granted equivalent tool permissions, state reset policies, and resource ceilings.

\paragraph{Itemwise resource matching ($k_i$).}
When stopping rules or trace lengths vary across items, grow the sampled single-agent pool individually for each item $i$ until cumulative single-agent tokens or dollar expenditure match the realized multi-agent budget $B_i^{\mathrm{MAS}}$ on that item, distinguishing itemwise dominance from parity only in expectation.

\paragraph{Adaptive stopping.}
Whenever early exit, dynamic routing, or consensus thresholding controls generation count, record the exact stopping criterion, declared resource ceiling, item distribution of realized expenditure, and termination reason alongside $R_{\mathrm{token}}$ and $R_{\mathrm{sample}}$.

\paragraph{Cached token accounting.}
In addition to raw prompt tokens, completion tokens, and monetary spend ($R_{\mathrm{token}}$, $R_{\mathrm{cost}}$), pipelines using prompt caching may report weighted billing tokens $C_{\mathrm{tok}}^{\mathrm{adj}}(\boldsymbol{\alpha}) = C_{\mathrm{in,uncached}} + \alpha_{\mathrm{cache}}C_{\mathrm{in,cached}} + \alpha_{\mathrm{out}}C_{\mathrm{out}}$ alongside the raw counts, stating the chosen weight vector $\boldsymbol{\alpha}$ explicitly rather than substituting billed weights for raw context consumption.

\paragraph{Hybrid policy classification.}
Classify hybrid pipelines directly from the directed call graph of each execution trace: assign policy class $\Pi_{\mathrm{agent}}$ whenever any model call reads an output generated on another reasoning thread, and report counts of independent trajectories, continuation edges along a single thread, edges across distinct threads, model aggregation calls, judge calls, and external tool invocations.

\section{Controlled comparison: protocol and sensitivity}
\label{app:controlled}

\paragraph{Plan and hashes.} The plan, with its addendum, was committed and hashed before any outcome was analyzed; it was not registered on a public registry. SHA-256 of \texttt{docs/PREREGISTRATION.md}: {\footnotesize\texttt{\preregShaMain}}; of \texttt{docs/PREREGISTRATION\_ADDENDUM\_1.md}: {\footnotesize\texttt{\preregShaAddOne}}.

\paragraph{Strategies and calls per item.} \Cref{tab:controlled_strategies} lists the strategies and the number of model calls per item as implemented. Round 1 of S4, S5, S7 and S8 reuses the first five samples of the pool of S2 on MMLU-Pro (and the five split-view draws of S2v for S4, S4v, S5 and S7 on Hidden Profile), and the tokens of those calls are charged to the strategy that uses them. In S7 the borrowed text comes from other items and is not charged.
\begin{table}[t]
\centering
\small
\caption{Strategies of the controlled comparison and calls per item as implemented (five agents, three rounds, pool of 16 samples).}
\label{tab:controlled_strategies}
\begin{tabular}{lp{0.66\textwidth}r}
\toprule
Label & Description & Calls per item \\
\midrule
S1 & Greedy, one call & 1 \\
S2 & 16 independent samples, temperature 0.7, plurality vote (ties share credit) & 16 \\
S3 & One call with more thinking (server think flag, or an extended-reasoning prompt) & 1 \\
S4 & Vanilla debate, five copies, three rounds & 15 (10 beyond the pool) \\
S5 & Debate with confidence shown to peers and confidence-weighted vote & 15 (10) \\
S7 & Placebo debate: each peer message replaced by a sample from a different item, truncated or repeated to match the character length of the message it replaces, not token-matched & 15 (10) \\
S8 & Mixture-of-agents, two layers (five aggregators, then one) & 11 (6) \\
\bottomrule
\end{tabular}
\end{table}

\paragraph{Positive control.} Each item has four candidates scored on ten equally weighted criteria. Four criteria are shared and favor a decoy; six are split so that each of five agents sees two of them and the pooled total favors the target. An agent that reads one view correctly answers the decoy, so independent voting over views has no route to the target. The items are synthetic, generated with seed 42, and the structural properties were checked by script on 500 items (dataset card in the supplementary material).

\paragraph{Completed cell breakdown and live exploratory runs.} \Cref{tab:confirmatory_cells} reports all 17 completed cell-strategy rows in the confirmatory batch. For \texttt{llama3.1:8b} on Hidden Profile ($n=200$), rows in \Cref{tab:confirmatory_cells} are referenced against the full-evidence 16-sample pool ($0.905$), which highlights the cost of partial information relative to unsplit single-agent access, whereas H2 in \Cref{tab:confirmatory_primary} evaluates split-view debate against the split-view voting frontier ($\hpFrontier$). Across an exploratory live validation suite (\texttt{outputs/experiments/live\_validation/}), \texttt{llama3.1:8b} likewise trails the self-consistency frontier on GSM8K ($n=35$: debate $0.771$ at $6{,}038$ tokens vs.\ pool $0.843$ at $2{,}387$ tokens; frontier gap $-0.071$ $[-0.164, -0.008]$; placebo S7 and shuffled-peer S13 both match S4 at $0.800$ on the paired 15-item slice) and on MATH-500 levels 4--5 ($n=25$: debate $0.173$ vs.\ pool $0.286$; frontier gap $-0.094$ $[-0.230, -0.000]$), while adaptive gating reduces average calls without improving GSM8K accuracy ($0.800$ gated vs.\ $0.800$ ungated, $n=15$). The GSM8K and MATH-500 levels 4--5 cells are not rows of \Cref{tab:confirmatory_cells}, which lists only the confirmatory batch; they are read from the live-validation logs, with $n=35$ and $n=25$ items.
\begin{table}[t]
\centering
\scriptsize
\setlength{\tabcolsep}{1.6pt}
\caption{All completed cells: accuracy, mean tokens per item (prompt plus completion), the self-consistency frontier at those tokens, and the gap to it with its 95\% paired bootstrap interval. Cells outside the confirmatory family are labeled. S2v is independent voting on the five split views, S4v is debate in which each agent sees only its own view and the peer messages from round 2 on, and S4 on hidden profile reuses the five split-view draws in round 1 before giving every agent the full evidence in rounds 2 and 3.}
\label{tab:confirmatory_cells}
\begin{tabular}{llrlrrrrl}
\toprule
Model & Task & $n$ & Strategy & Accuracy & Tokens & Frontier & Gap & 95\% interval \\
\midrule
deepseek-r1:14b (exploratory) & hidden profile & 50 & greedy & 1.000 & 2{,}338 & 0.000 & \ensuremath{+1.000} & [\ensuremath{+1.000}, \ensuremath{+1.000}] \\
deepseek-r1:14b & hidden profile & 50 & independent vote on split views & 0.000 & 16{,}808 & 0.000 & \ensuremath{+0.000} & [\ensuremath{+0.000}, \ensuremath{+0.000}] \\
deepseek-r1:14b & hidden profile & 50 & split-view debate & 0.000 & 39{,}532 & 0.000 & \ensuremath{+0.000} & [\ensuremath{+0.000}, \ensuremath{+0.000}] \\
llama3.1:8b & hidden profile & 200 & greedy & 0.925 & 777 & 0.864 & \ensuremath{+0.061} & [\ensuremath{+0.033}, \ensuremath{+0.090}] \\
llama3.1:8b & hidden profile & 200 & 16-sample pool & 0.905 & 15{,}018 & 0.905 & \ensuremath{-0.000} & [\ensuremath{-0.005}, \ensuremath{+0.000}] \\
llama3.1:8b & hidden profile & 200 & independent vote on split views & 0.077 & 3{,}965 & 0.891 & \ensuremath{-0.814} & [\ensuremath{-0.858}, \ensuremath{-0.766}] \\
llama3.1:8b & hidden profile & 200 & debate & 0.815 & 25{,}668 & 0.905 & \ensuremath{-0.090} & [\ensuremath{-0.150}, \ensuremath{-0.035}] \\
llama3.1:8b & hidden profile & 200 & split-view debate & 0.182 & 23{,}604 & 0.905 & \ensuremath{-0.723} & [\ensuremath{-0.785}, \ensuremath{-0.662}] \\
llama3.1:8b & hidden profile & 200 & confidence-weighted debate & 0.185 & 23{,}527 & 0.905 & \ensuremath{-0.720} & [\ensuremath{-0.785}, \ensuremath{-0.653}] \\
llama3.1:8b & hidden profile & 200 & placebo debate & 0.248 & 34{,}819 & 0.905 & \ensuremath{-0.657} & [\ensuremath{-0.723}, \ensuremath{-0.595}] \\
llama3.1:8b & mmlu pro & 250 & greedy & 0.408 & 1{,}172 & 0.464 & \ensuremath{-0.056} & [\ensuremath{-0.101}, \ensuremath{-0.011}] \\
llama3.1:8b & mmlu pro & 250 & 16-sample pool & 0.581 & 13{,}982 & 0.581 & \ensuremath{+0.000} & [\ensuremath{-0.005}, \ensuremath{+0.000}] \\
llama3.1:8b & mmlu pro & 250 & more thinking & 0.464 & 1{,}120 & 0.463 & \ensuremath{+0.001} & [\ensuremath{-0.043}, \ensuremath{+0.045}] \\
llama3.1:8b & mmlu pro & 250 & debate & 0.567 & 19{,}571 & 0.581 & \ensuremath{-0.015} & [\ensuremath{-0.065}, \ensuremath{+0.035}] \\
llama3.1:8b & mmlu pro & 250 & confidence-weighted debate & 0.516 & 19{,}761 & 0.581 & \ensuremath{-0.065} & [\ensuremath{-0.112}, \ensuremath{-0.019}] \\
llama3.1:8b & mmlu pro & 250 & placebo debate & 0.550 & 19{,}144 & 0.581 & \ensuremath{-0.031} & [\ensuremath{-0.072}, \ensuremath{+0.008}] \\
llama3.1:8b & mmlu pro & 250 & mixture-of-agents & 0.508 & 13{,}409 & 0.580 & \ensuremath{-0.072} & [\ensuremath{-0.118}, \ensuremath{-0.028}] \\
\bottomrule
\end{tabular}
\end{table}

\paragraph{Sensitivity analyses.} \Cref{tab:confirmatory_sensitivity} repeats the gaps of the primary MMLU-Pro cell with a frontier built from completion tokens only and with a pool of 8 samples. The planned analysis that excludes the pilot items is not needed for MMLU-Pro, because the item sampling was corrected after the pilot and the items differ; and the planned analysis with a scorer that accepts only an answer line is not run.
\begin{table}[t]
\centering
\small
\caption{Sensitivity of the frontier gap on the primary MMLU-Pro cell to the token definition and the pool size (95\% paired bootstrap intervals).}
\label{tab:confirmatory_sensitivity}
\begin{tabular}{llrl}
\toprule
Frontier definition & Strategy & Gap & 95\% interval \\
\midrule
prompt plus completion tokens, pool of 16 (primary) & more thinking & \ensuremath{+0.001} & [\ensuremath{-0.043}, \ensuremath{+0.045}] \\
prompt plus completion tokens, pool of 16 (primary) & debate & \ensuremath{-0.015} & [\ensuremath{-0.065}, \ensuremath{+0.035}] \\
prompt plus completion tokens, pool of 16 (primary) & mixture-of-agents & \ensuremath{-0.072} & [\ensuremath{-0.118}, \ensuremath{-0.028}] \\
completion tokens only & more thinking & \ensuremath{-0.001} & [\ensuremath{-0.044}, \ensuremath{+0.045}] \\
completion tokens only & debate & \ensuremath{+0.002} & [\ensuremath{-0.043}, \ensuremath{+0.047}] \\
completion tokens only & mixture-of-agents & \ensuremath{-0.048} & [\ensuremath{-0.087}, \ensuremath{-0.008}] \\
pool of 8 & more thinking & \ensuremath{+0.007} & [\ensuremath{-0.039}, \ensuremath{+0.054}] \\
pool of 8 & debate & \ensuremath{+0.014} & [\ensuremath{-0.033}, \ensuremath{+0.061}] \\
pool of 8 & mixture-of-agents & \ensuremath{-0.045} & [\ensuremath{-0.093}, \ensuremath{+0.002}] \\
\bottomrule
\end{tabular}
\end{table}

\IfFileExists{tables/k24_sensitivity_text.tex}{\paragraph{Post hoc sensitivity (k=24).} After the planned analysis (fixed and hashed before any outcome was analyzed, but not publicly registered; see \Cref{sec:controlled}), eight further samples per item (indices 17--24) were generated for each of the $n=250$ items of the primary MMLU-Pro cell, and the self-consistency frontier was rebuilt from 24 samples; this analysis is post hoc and the planned $k=16$ results above remain the primary ones. At the token cost of debate (S4), the frontier is $0.593$ and the gap is $-0.026$ ($95\%$ interval [$-0.079$, $+0.026$]); for mixture-of-agents (S8) the gap is $-0.072$ ([$-0.116$, $-0.027$]). The debate gap moves from \confOneGap{} to $-0.026$ and its interval still contains zero, and the mixture-of-agents gap is unchanged (script \texttt{analysis/run\_k24\_sensitivity.py}; report \texttt{review/K24\_SENSITIVITY.md}).
}{}

\subsection{Certification and calibration checks}
\label{app:certification}
Values in this subsection are computed from the confirmatory logs (seed 42, 5{,}000 bootstrap replicates; supplementary certification script). The first, second and fourth items are empirical; the third is a Monte Carlo calibration, not a model run.

\paragraph{Simultaneous frontier bands.} Each bootstrap replicate refits the isotonic frontier $\hat f_{\mathrm{SC}}(C)$ (stratified by category on MMLU-Pro, paired item resampling on Hidden Profile), and percentile bands $[L_f(C),U_f(C)]$ are calibrated to a joint coverage of at least $95\%$ over the whole curve. A gain is certified only if $L(p_{\mathrm{MAS}})-U_f(C)>0$, and parity or a deficit if $U(p_{\mathrm{MAS}})-L_f(C)\le 0$. Marginal bands do not cancel the item correlation shared by the two accuracies, so non-overlap is more conservative than the paired gap interval. Under this criterion all ten contrasts of that file are indeterminate, including MMLU-Pro S4 ($-0.0145$, $[-0.0629,+0.0343]$) and S8 ($-0.0719$, $[-0.1186,-0.0273]$, Holm-adjusted $p=0.0224$ over ten contrasts) and split-view debate against the five-view pool ($+0.0733$, $[+0.0254,+0.1247]$, Holm-adjusted $p=0.0224$; \hpThirtyProvisional{}). The Holm family here (ten contrasts) differs from that of \Cref{tab:confirmatory_primary}; the two values of $0.0224$ are not a duplicated cell, since MMLU-Pro S8 is rank 7 ($7\times 0.0032$) and the five-view contrast rank 8 ($8\times 0.0028$) in exact Holm arithmetic, where $p$-values in this ten-contrast exploratory audit are two-sided ($H_a:\text{gap}\neq 0$, explaining why negative S8 yields $p_{\mathrm{Holm}}=0.0224$), whereas \Cref{tab:confirmatory_primary} reports one-sided positive-gain $p$-values ($H_a:\text{gap}>0$).

\paragraph{Evidence-by-communication interaction.} On the 200 paired Hidden Profile items, $\tau_{\mathrm{interaction}}=(\mu_{\mathrm{split},1}-\mu_{\mathrm{split},0})-(\mu_{\mathrm{shared},1}-\mu_{\mathrm{shared},0})=+0.1950$ ($[+0.1200,+0.2725]$, $p=0.0004$), from $\tau_{\mathrm{shared}}=-0.0900$ ($[-0.1500,-0.0325]$; \texttt{S2} with 16 samples to \texttt{S4}) and $\tau_{\mathrm{split}}=+0.1050$ ($[+0.0500,+0.1600]$; \texttt{S2v} to \texttt{S4v}); with five samples as the shared baseline it is $+0.1900$ ($[+0.1125,+0.2650]$). \texttt{S4} reads split views in round 1 and full evidence in rounds 2 and 3, whereas \texttt{S4v} and \texttt{S7} keep split views throughout, so the interaction mixes the evidence factor with that protocol difference. The semantic contrast of split-view debate against placebo debate is $-0.0654$ ($[-0.1492,+0.0125]$, $p=0.1072$). These contrasts are exploratory.

\paragraph{Search-adjusted budget.} If a multi-agent topology is chosen on development items before it is compared with a baseline, the total budget is $B_{\mathrm{total}}=B_{\mathrm{search}}+B_{\mathrm{inference}}$ and the development gain overstates the test gain by $O=\Delta_{\mathrm{dev}}-\Delta_{\mathrm{test}}$. A simulation over 500 disjoint splits ($N_{\mathrm{dev}}=100$, $N_{\mathrm{test}}=250$; sharp null, baseline accuracy $0.581$, item correlation $0.65$, $19{,}571$ and $13{,}982$ tokens per item for the system and baseline) gives, for $K=1$, $5$ and $20$ candidate topologies, mean optimism $+0.0026$, $+0.0386$ and $+0.0688$, naive false certification rates of $0.024$, $0.062$ and $0.246$, nested-evaluation rates of $0.036$, $0.016$ and $0.020$, and $B_{\mathrm{total}}/B_{\mathrm{inference}}$ of $1.40$, $2.33$ and $5.83$.

\paragraph{Directional frontier estimators.} The preregistered isotonic frontier $\hat f_{\mathrm{SC}}$ is the primary estimator. Because a fitted monotone curve can pool a genuine drop of the voting curve, \Cref{tab:directional_frontier} adds the monotone upper envelope $\hat f^{+}_{\mathrm{SC}}(C)=\max_{k:\bar C_k\le C}\hat a(k)$, used for positive claims, and the two-fold item cross-fit $\hat f^{-}_{\mathrm{SC}}$ (200 splits), used for nonpositive claims (2{,}000 paired bootstrap replicates, seed 0; supplementary audit script). The five-view curve of Hidden Profile falls from $0.1409$ at $k=1$ (the envelope is $0.1477$ at $k=2$ because empty outputs are excluded from the plurality) to $0.0775$ at $k=5$, so the isotonic value $0.1092$ understates the best single-view accuracy; the H2 gap shrinks from $+0.0733$ to $+0.0348$ and its interval then includes zero.
\begin{table}[t]
\centering
\footnotesize
\caption{Frontier gap (system accuracy minus frontier) under three estimators; 95\% paired bootstrap interval in brackets. Intervals here use 2{,}000 paired resamples (seed 0), whereas the primary intervals of \Cref{tab:confirmatory_primary} use the 10{,}000 paired resamples of the confirmatory analysis and the simultaneous bands of \Cref{app:certification} use 5{,}000 (seed 42), so the last digit of an interval can differ between tables.}
\label{tab:directional_frontier}
\scriptsize\setlength{\tabcolsep}{3pt}
\begin{tabular}{@{}lrrrr@{}}
\toprule
Slice & System & Isotonic & Envelope $\hat f^{+}$ & Cross-fit $\hat f^{-}$ \\
\midrule
MMLU-Pro S4 ($k=16$) & 0.5668 & $-0.0145$ [$-0.0667,+0.0338$] & $-0.0145$ [$-0.0667,+0.0323$] & $-0.0117$ [$-0.0651,+0.0356$] \\
MMLU-Pro S8 ($k=16$) & 0.5080 & $-0.0719$ [$-0.1153,-0.0293$] & $-0.0719$ [$-0.1153,-0.0306$] & $-0.0698$ [$-0.1134,-0.0283$] \\
Hidden Profile S4v, five views (H2) & 0.1825 & $+0.0733$ [$+0.0254,+0.1203$] & $+0.0348$ [$-0.0128,+0.0815$] & $+0.0348$ [$-0.0128,+0.0816$] \\
Hidden Profile S4, $k=16$ pool & 0.8150 & $-0.0904$ [$-0.1475,-0.0354$] & $-0.0907$ [$-0.1482,-0.0363$] & $-0.0874$ [$-0.1463,-0.0331$] \\
\bottomrule
\end{tabular}
\end{table}

\paragraph{Item sampling.} The MMLU-Pro items are drawn from 4,000 rows spread over 13 of its 14 categories.

\paragraph{Placebo token breakdown.} Mean tokens per item (prompt, completion): MMLU-Pro S4 $13{,}250$ and $6{,}321$, S7 $12{,}000$ and $7{,}144$; Hidden Profile S4v $16{,}467$ and $7{,}136$, S7 $22{,}876$ and $11{,}943$. S7 over S4 is $0.906$ (prompt), $1.130$ (completion) and $0.978$ (total) on MMLU-Pro, and S7 over S4v is $1.389$, $1.674$ and $1.475$ on Hidden Profile.

\paragraph{Specification curve and fail-safe $K$.} Over 288 specifications of the B9 model (three Tran and Kiela arms, with and without the 100-token cell, two Smit budgets, three Kim arms, Jwalapuram and AGIEval in or out, two clusterings) the pooled log odds ratio has median $-0.1935$ and range $-0.2186$ to $-0.1681$; all 288 are negative, 286 have a CR2 upper bound below zero, and the least negative upper bound is $+0.0010$. Adding $K$ synthetic studies, each with $36$ contrasts (the median cluster size), effect $+0.125$ and the median contrast variance, brings the CR2 upper bound to zero at $K^{*}=2$ added studies and the pooled estimate to zero at $K_0=12$.

\paragraph{Truncation and unparsable answers.} \Cref{tab:r8_truncation} lists, for each strategy of the controlled comparison, the share of calls that hit the token cap and the share with no parsable answer (existing logs, \texttt{analysis/run\_r8\_truncation.py}). Both shares are highest for the single-call strategies S1 and S3 on MMLU-Pro (about $13$ to $16\%$) and below $5\%$ for most debate strategies; such calls count as incorrect.
\begin{table}[t]
\centering
\footnotesize
\caption{Calls per strategy that reached the token cap or returned no parsable answer (\texttt{llama3.1:8b}; calls reused from the first round count for each strategy that used them).}
\label{tab:r8_truncation}
\begin{tabular}{@{}llrrrr@{}}
\toprule
Task & Strategy & Items & Calls & Hit cap & Unparsable \\
\midrule
MMLU-Pro & S1 & 250 & 250 & 39 (15.6\%) & 40 (16.0\%) \\
MMLU-Pro & S2 & 250 & 4{,}000 & 243 (6.1\%) & 283 (7.1\%) \\
MMLU-Pro & S3 & 250 & 250 & 32 (12.8\%) & 37 (14.8\%) \\
MMLU-Pro & S4 & 250 & 3{,}750 & 105 (2.8\%) & 154 (4.1\%) \\
MMLU-Pro & S5 & 250 & 3{,}750 & 106 (2.8\%) & 157 (4.2\%) \\
MMLU-Pro & S7 & 250 & 3{,}750 & 151 (4.0\%) & 189 (5.0\%) \\
MMLU-Pro & S8 & 250 & 2{,}750 & 103 (3.7\%) & 131 (4.8\%) \\
Hidden Profile & S1 & 200 & 200 & 0 (0.0\%) & 0 (0.0\%) \\
Hidden Profile & S2 & 200 & 3{,}200 & 0 (0.0\%) & 6 (0.2\%) \\
Hidden Profile & S2v & 200 & 1{,}000 & 6 (0.6\%) & 46 (4.6\%) \\
Hidden Profile & S4 & 200 & 3{,}000 & 14 (0.5\%) & 111 (3.7\%) \\
Hidden Profile & S4v & 200 & 3{,}000 & 14 (0.5\%) & 69 (2.3\%) \\
Hidden Profile & S5 & 200 & 3{,}000 & 9 (0.3\%) & 56 (1.9\%) \\
Hidden Profile & S7 & 200 & 3{,}000 & 55 (1.8\%) & 126 (4.2\%) \\
\bottomrule
\end{tabular}
\end{table}

\paragraph{Reasoning-model cell.} For \texttt{deepseek-r1:14b} the completed $N=50$ slices have $800$ calls per cell ($500$ debate, $250$ pool or first-round and $50$ greedy calls). On MMLU-Pro \confDsCapMmlu{} of \confDsCallsMmlu{} calls reached the token cap and \confDsParseMmlu{} of \confDsParseOfMmlu{} were unparsable (the $50$ items are stratified by category; matching the calls against an unstratified draw of $50$ items finds only $560$ of the $800$ calls and $47$ unparsable), and thinking characters were \confReasonShareMmlu{} of thinking plus peer-message characters (the one definition used throughout, over debate and pool calls); on Hidden Profile $1$ of $800$ reached the cap and the share was \confReasonShareHp{}. Measured against the visible output of the debate calls alone the shares are $0.950$ and $0.993$.

\paragraph{HP30 comparator (post hoc).} HP30 covers the 200 Hidden Profile items with six trajectories of five split-view calls each, 30 calls per item (the five round-1 split-view calls reused plus 25 new calls), and a plurality vote over the 30 answers; it uses a mean of $23{,}002.4$ tokens per item against \hpDebateTok{} for split-view debate, so it is a near-budget comparator, not an exact token match.

\paragraph{Calibration and early stopping (exploratory).} On MMLU-Pro the round-3 vote shares of vanilla debate (S4) are poorly calibrated: their expected calibration error is $0.271$ against $0.079$ for a vote over five independent draws (difference $+0.192$ [$+0.133$, $+0.229$]), and $16.4\%$ of items end unanimous and wrong against $2.4\%$ for the five-draw vote. Stopping debate after round 1 and voting the round-1 answers would, counterfactually on the logged calls, save $78.4\%$ of its tokens at an accuracy change of $-0.008$ [$-0.051$, $+0.034$].

\paragraph{Exploratory cells on Routes II and III.} Two further cells are not part of the hashed plan and are exploratory (\texttt{scripts/run\_revision\_cells.py}, cells E2 and E3; summaries and paired item bootstrap intervals with $B=2000$ from \texttt{analysis/run\_e2\_e3\_exploratory\_summary.py}). The Route~III cell asks whether errors decorrelate across models on MMLU-Pro, with three independent samples per model and item. On the 250 items answered by \texttt{llama3.1:8b} and \texttt{qwen2.5:7b}, neither of which timed out, the cross-model intraclass correlation of correctness is below the within-model one ($\rho_{\mathrm{het}}-\rho_{\mathrm{hom}}=-0.175$ [$-0.248$, $-0.101$]), yet a mixed-model majority is no more accurate than a single-model one ($+0.003$ [$-0.020$, $+0.025$]). A third model, \texttt{deepseek-r1:14b}, was stopped at $n=80$ and timed out on 91 of its 240 samples, which the runner scored as incorrect, so no three-model estimate is reported. The Route~II cell uses $n=80$ synthetic Python tasks whose three complementary constraints are split across three \texttt{llama3.1:8b} agents, with the unit tests applied as a verifier to every final candidate. Split-view debate exceeds placebo debate by $+0.087$ [$-0.013$, $+0.188$] in the share of items with a passing candidate, at $8{,}502$ against $8{,}113$ tokens per item, and its per-candidate pass rate is below that of a full-evidence single pass by $-0.092$ [$-0.200$, $+0.021$].

\begin{figure}[t]
\centering
\includegraphics[width=0.5\textwidth]{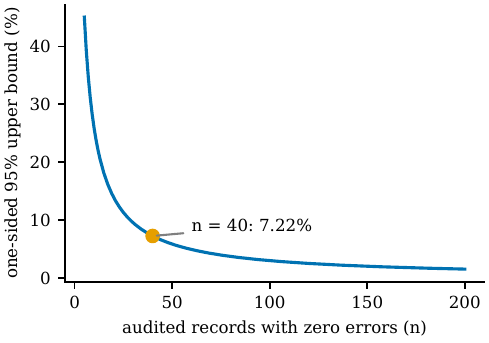}
\caption{Exact one-sided $95\%$ upper bound on the per-record error rate after zero errors in $n$ audited records, $1-0.05^{1/n}$; at $n=40$ it is \auditBoundPct\%. Because the 40-record sample was stratified by study rather than drawn at random, the bound is an illustrative reference rather than a corpus-wide error bound.}
\label{fig:audit_bound}
\end{figure}

\begin{figure}[t]
\centering
\includegraphics[width=0.92\textwidth]{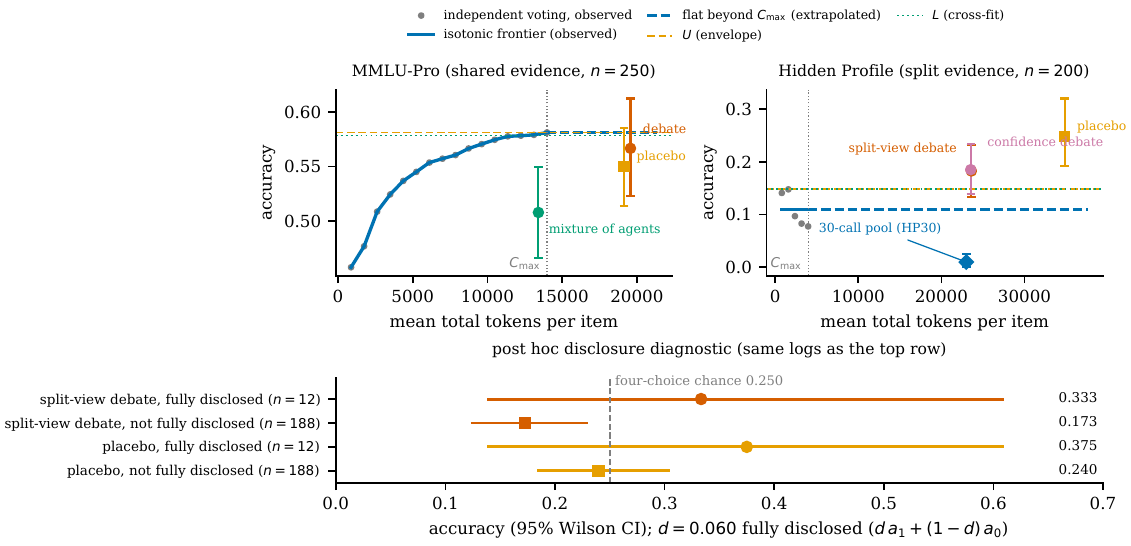}
\caption{Frontier estimators and disclosure diagnostic. Top: as in \Cref{fig:controlled_frontier}, with the envelope estimator $U$ (\frontUmmlu, \frontUhp) and the cross-fitted frontier estimator $L$ (\frontLmmlu, \frontLhp). Bottom: post hoc diagnostic; every agent mentions all its private criteria in only \hpDiscD{} of items ($a_1$, \hpDiscNOne{} items; $a_0$, $188$), so accuracy $=d\,a_1+(1-d)\,a_0$; points are accuracies with 95\% Wilson intervals.}
\label{fig:disclosure_diag}
\end{figure}

\begin{figure}[t]
\centering
\includegraphics[width=0.98\textwidth]{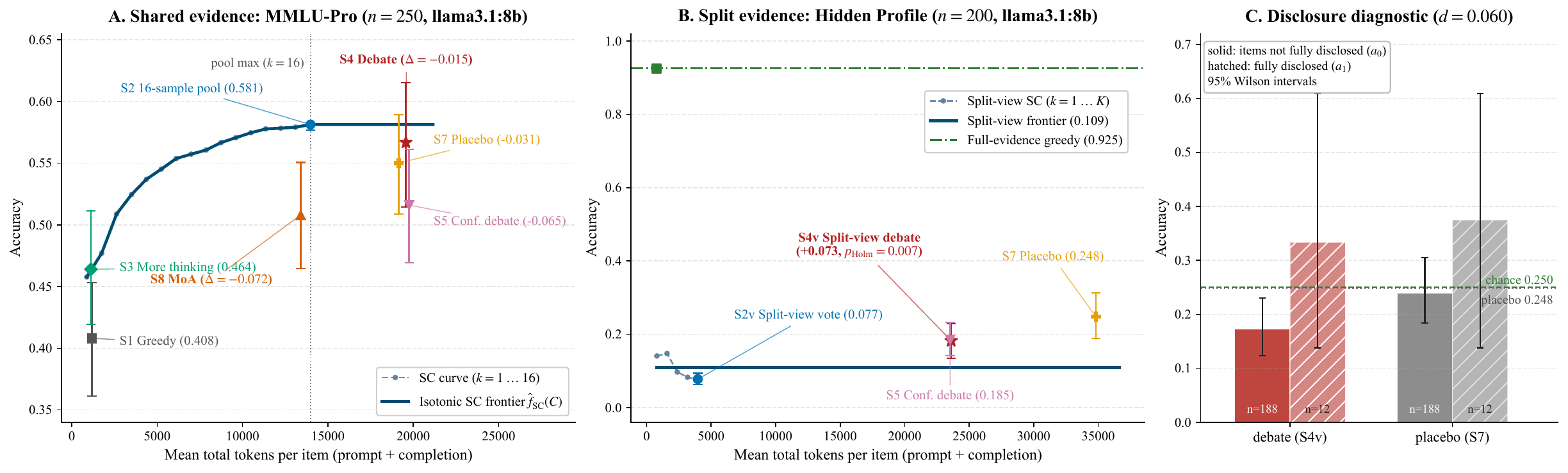}
\caption{Controlled experiment self-consistency frontiers and multi-agent strategies on \texttt{llama3.1:8b}. Panel A plots MMLU-Pro accuracy ($n=250$) against mean total tokens (prompt plus completion) for $k=1\dots 16$ self-consistency samples, the isotonic frontier $\hat{f}_{\mathrm{SC}}(C)$, and strategies S1--S8 with 95\% paired bootstrap intervals. Panel B plots Hidden Profile split-view self-consistency ($n=200$), the split-view frontier ($\hpFrontier$), split-view strategies (S2v, S4v, S5, S7), and the full-evidence greedy accuracy ($0.925$). Panel C is a post hoc disclosure diagnostic computed from the same logs: \hpDiscRate{} of the private criteria are mentioned in the agents' round-1 messages, and in only \hpDiscD{} of items (\hpDiscNOne{} items) does every agent mention all of its private criteria; accuracy on those items ($a_1=\hpDiscAOne$ for debate, $\hpDiscPlAOne$ for placebo) and on the others ($a_0=\hpDiscAZero$, $\hpDiscPlAZero$), so that accuracy $=d\,a_1+(1-d)\,a_0$.}
\label{fig:controlled_frontier_full}
\end{figure}

\begin{figure}[t]
\centering
\includegraphics[width=0.92\textwidth]{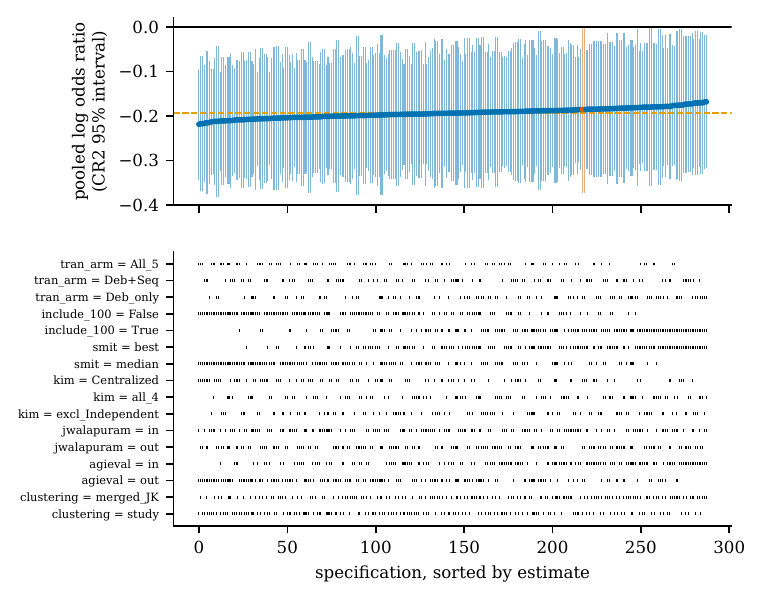}
\caption{Specification curve of the B9 model over \specN{} specifications (the choices listed in the left margin of the lower panel). Points are pooled log odds ratios sorted by estimate, with CR2 95\% intervals; the dashed line is the median. All \specNegative{} estimates are negative and \specCrNegative{} have an upper bound below zero.}
\label{fig:spec_curve}
\end{figure}

\begin{figure}[t]
\centering
\includegraphics[width=0.48\textwidth]{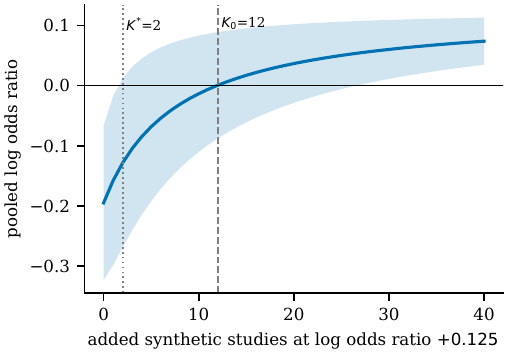}\hfill
\includegraphics[width=0.48\textwidth]{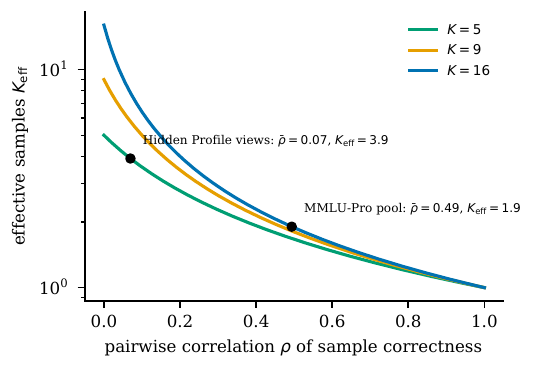}
\caption{Left: fail-safe $K$. Adding synthetic studies at log odds ratio $+0.125$ brings the CR2 upper bound to zero at $K^{*}=\failsafeKstar$ and the pooled estimate to zero at $K_0=\failsafeKzero$. Right: effective number of independent samples $K_{\mathrm{eff}}=K/(1+(K-1)\rho)$ against the pairwise correlation $\rho$ of sample correctness (\Cref{prop:keff}); points mark the measured mean pairwise correlation of the sampled pools ($\bar\rho=\rhoMmlu$ for the 16 MMLU-Pro samples, $\rhoHp$ for the five Hidden Profile views).}
\label{fig:failsafe_rho}
\end{figure}

\paragraph{Wild bootstrap with few clusters.} Category~B/C has $G=8$ study clusters, so \Cref{tab:wild_bootstrap} adds the exact Rademacher wild cluster bootstrap of \citet{joshi2022cluster}, enumerating all $2^G$ sign patterns for the CR2 statistic ($\varphi=0.50$, log odds ratio with Haldane--Anscombe correction). Enumeration returns $p$-values no smaller than $2/2^G$; Webb six-point weights give $p=0.0016$ for the primary model, and the unrestricted percentile-$t$ interval is $[-0.3315,-0.0586]$.
\begin{table}[t]
\centering
\footnotesize
\caption{Cluster-robust inference for the pooled log odds ratio: CR2 interval (Satterthwaite degrees of freedom $\nu$) and exact wild cluster bootstrap interval with $p$-value (sign patterns with at least as extreme a statistic; two for each model).}
\label{tab:wild_bootstrap}
\begin{tabular}{@{}lrrrr@{}}
\toprule
Model & $G$ / $M$ & $\hat\beta$ & CR2 95\% CI ($\nu$) & Exact wild bootstrap 95\% CI ($p$) \\
\midrule
B9 (primary) & 8 / 474 & $-0.1950$ & $[-0.3233,-0.0668]$ (6.36) & $[-0.3370,-0.0729]$ (0.0078) \\
B9 without Zhang et al. & 7 / 334 & $-0.1454$ & $[-0.2457,-0.0451]$ (5.09) & $[-0.2415,-0.0527]$ (0.0156) \\
B9, Jwalapuram and Ke merged & 7 / 474 & $-0.1867$ & $[-0.3331,-0.0403]$ (5.47) & $[-0.3577,-0.0708]$ (0.0156) \\
\bottomrule
\end{tabular}
\end{table}

\section{Extended Theory}\label{app:moved_material}

\subsection{Three channels and one inferential layer}\label{app:moved_s3_channels}

\paragraph{Three channels and one inferential layer.}
A multi-agent system can differ from a noncommunicating system through three channels. The first is evidence: agents may receive different or complementary information, which decides whether communication can supply what an individual agent lacks. The second is budget: communication and orchestration consume part of the inference budget and so change the computation available per call. The third is aggregation: agents combine and update correlated answers, so the number of agents need not equal the effective number of independent answers. PACE adds a comparison layer on top of these channels. Policy classes, directional resource feasibility, and the noncommunicating frontier determine which performance differences can be attributed relative to a specified baseline and which remain unidentified.

\FloatBarrier

\subsection{Context Crowding and the Data Processing Bound}\label{app:moved_s3_crowd}

\label{sec:dpi_context}
\citet{tran2026thinking} invoke an information-theoretic bound rooted in the Data Processing Inequality: given a fixed budget of thinking tokens and uncorrupted context retention, passing intermediate reasoning traces through an inter-agent communication channel cannot increase mutual information with the target solution relative to allocating all tokens directly to internal single-agent thinking. In empirical evaluation, this bound operates under two specific conditions: (i) the total reasoning-token budget is held constant, and (ii) the single agent's reasoning capacity does not degrade over longer contexts. When single-agent reasoning suffers context fragmentation, or when tasks require partitioned contexts, parallel external tools, or sandboxed execution, multi-agent topologies can circumvent this constraint. Rather than relying on asymptotic information bounds, we model the operational trade-off directly through scaffolding overhead and context crowding.

\subsubsection{Token Dilution and Scaffolding Overhead}
\label{sec:scaffolding}
Context echoing reduces the token budget available for reasoning. When $K$ agents communicate across $T$ rounds under a global ceiling of $B$ tokens, conversational scaffolding consumes a growing share of the available budget (\Cref{eq:beff}). System prompts, role headers, and message histories accumulate across rounds as dialogue histories are echoed:
\begin{equation}
  B_{\text{eff}} \;=\; \max\!\left(0,\; \frac{B - \sum_{k=1}^{K} C^{(k)}_{\text{scaffold}}}{K\cdot T}\right).
  \label{eq:beff}
\end{equation}
A single agent has $K = T = 1$ and a single scaffold cost $C_0$, yielding $B^{\text{SAS}}_{\text{eff}} = \max(0, B - C_0)$. As $K$ and $T$ increase, effective reasoning tokens per agent call diminish. When question answering is matched on thinking tokens, single agents use this larger per-call budget to match or exceed multi-agent debate across nonvacuous token budgets \citep{tran2026thinking}. At small token allocations (e.g., 100 tokens), both single-agent and multi-agent systems produce truncated reasoning chains, rendering early accuracy differences uninformative. Throughout this paper, $K$ denotes the number of agents or parallel samples, $T$ the number of interaction rounds, $N$ the number of benchmark evaluation items, $M$ the total number of multi-agent configurations in a corpus (with $m$ indexing contrast counts within a study or slice), and $\mathbf{B}$ the resource ceiling vector. In \Cref{sec:reml_main} and \Cref{app:sensitivity} only, $K$ and $M$ follow meta-analytic convention and denote the numbers of clusters and contrasts in a fitted model.

\FloatBarrier

\subsection{Directional feasibility: interpretation and partial identification}\label{app:moved_s3_ident}

\paragraph{Interpretation of directional feasibility.}
Directional resource feasibility establishes one-sided conclusions relative to the declared resource subset $\mathcal{J}$ and distinguishes two estimands: the observed pairwise difference $\Delta_{\text{baseline}} \triangleq p_A - p_{\text{baseline}}$ and the frontier gap $\Delta_{\text{frontier}} \triangleq p_A - f(\mathbf{B}_A)$. If a noncommunicating baseline uses no more of each resource in $\mathcal{J}$ than the multi-agent system ($\mathbf{R}_{\text{base},\mathcal{J}} \preceq \mathbf{R}_{A,\mathcal{J}}$), then it is feasible under the system's resource ceiling when shared model and tool conditions hold, so $p_{\text{baseline}} \le f(\mathbf{B}_A)$ and $\Delta_{\text{frontier}} \le \Delta_{\text{baseline}}$; a nonpositive $\Delta_{\text{baseline}}$ therefore rules out a positive coordination gap over the frontier. In the opposite direction, a baseline using at least as much of the declared resources ($\mathbf{R}_{\text{base},\mathcal{J}} \succeq \mathbf{R}_{A,\mathcal{J}}$) provides a demanding comparator for claiming a gain over that baseline, but one observed score lower-bounds rather than upper-bounds $f(\mathbf{B}_A)$, so a positive frontier claim still requires estimating or upper-bounding the best feasible noncommunicating policy. When only $\mathcal{J}$ is controlled, the comparison conditions on the projected frontier of \Cref{eq:projected_frontier} rather than the full six-dimensional ceiling.

\paragraph{Partial identification under incomplete resource reporting.}
When only a subset $\mathcal{J}$ of $\mathbf{R}$ is observed or controlled, the comparison does not induce a total ordering over the full resource space: PACE reports the resource relation on $\mathcal{J}$ while leaving unreported dimensions unidentified. This connects directional feasibility to partial-identification analysis \citep{manski2003partial}, in which observed constraints license one-sided bounds without identifying a unique point estimate across unmeasured dimensions.

\FloatBarrier

\subsection{Accounting decomposition on Choi et al. and controlled MMLU-Pro}\label{app:moved_s3_waterfall}

\FloatBarrier

\subsection{Connection to the comparison of information channels}\label{app:moved_s3_blackwell}

\paragraph{Connection to comparison of information channels.}
When the agents' private evidence is conditionally redundant given $X_{\mathrm{shared}}$, the transcript $\mathcal{H}^{(T)}$ is a randomized transformation of information already present in $X_{\mathrm{shared}}$. Viewed through \citet{blackwell1953equivalent}, the transcript is a garbled signal conditional on $X_{\mathrm{shared}}$, so adding $\mathcal{H}^{(T)}$ alongside $X_{\mathrm{shared}}$ cannot raise the Bayes-optimal expected decision value when $X_{\mathrm{shared}}$ remains observable. As with \Cref{prop:redundancy}, this equivalence applies to statistical information rather than the computational efficiency of a bounded inference procedure under fixed token limits.

\FloatBarrier

\subsection{Two testable predictions of the framework}\label{app:moved_s3_predictions}

The framework makes two testable predictions. With a single shared prompt, as in MMLU-Pro, \Cref{eq:redundancy_bound} can be tested by comparing debate with the self-consistency frontier at measured tokens; a violation would show that communication uses tokens more efficiently than sampling. When evidence is partitioned so that every single view favors a wrong answer and only the pooled evidence favors the right one, independent voting cannot recover the answer, and the premium of communication measures how well agents pool the partition. This is the hidden-profile design of \citet{stasser1985pooling}, and a synthetic task built on it can be used to check whether an evaluation detects such a premium. The task generator is released with the code (\texttt{pace hidden-profile}; see \Cref{sec:appendix_mae_eval}).\ifcsname r@sec:poscontrol\endcsname{} Its protocol and results are in \Cref{sec:poscontrol}.\fi\ifcsname r@sec:controlled\endcsname{} The controlled comparison that tests both cases is in \Cref{sec:controlled}.\fi

\FloatBarrier

\subsection{Sizing the sample pool: worked examples}\label{app:moved_s3_pool}

For $K=3$, $T=2$ under Eq.~\eqref{eq:ktok} this is $6+6r$, between 6 and 12 ($6+9r$, below 15, if an agent's own message is echoed), which a 16-sample pool covers. For $K=5$, $T=3$ it is $15+40r$, which exceeds a 16-sample pool once $r>0.025$, so $r$ should be measured from logged tokens, counted by a tokenizer because server-side counts can omit cached prompt tokens. By construction the token-equated baseline costs as many tokens as one more run of the system. Eq.~\eqref{eq:scaffold_total} ($C_{\mathrm{tot}}$ and $B_{\mathrm{eff}}$) and Eq.~\eqref{eq:ktok} ($k_{\text{tok}}$) apply the same token-accounting framework to two distinct message-passing protocols: $C_{\mathrm{tot}}(K,T)$ and $B_{\mathrm{eff}}(K,T)$ record cumulative full-history expenditure and remaining reasoning allowance under ceiling $B$, while $k_{\text{tok}}$ maps previous-round message volume to an independent single-agent sampling pool, making $k_{\text{tok}}$ protocol- and tokenizer-dependent rather than a universal function of $K$ and $T$ alone.

\FloatBarrier

\subsection{Influence regimes, plug-in threshold identity, and null-corrected audits}\label{app:moved_s3_conformity}

The constant-rate Markov chain in \Cref{thm:conformity} is a stylized marginal reference among three influence regimes: (a) constant rates independent of peers, (b) linear peer-share rates proportional to opposing peers (which yield zero drift when strengths match, matching the martingale model of \citet{choi2025debate}), and (c) convex majority-amplification rates. When transition rates vary across rounds or when peer flips are correlated within an item, round-$T$ majority-vote accuracy depends on the joint distribution of agent answers rather than marginal $p_t$ alone. On any transcript with $n_{\mathrm{w}}\ge 1$ wrong and $n_{\mathrm{c}}\ge 1$ correct agent-item pairs in round $t$ and flip counts $F_{ic}$ and $F_{ci}$ with $F_{ic}+F_{ci}\ge 1$, the algebraic identity $\hat{p}^* - p_1 = \frac{n_{\mathrm{c}}n_{\mathrm{w}}(F_{ic} - F_{ci})}{(F_{ic}n_{\mathrm{c}} + F_{ci}n_{\mathrm{w}})(n_{\mathrm{c}} + n_{\mathrm{w}})}$ shows that $\hat{p}^*$ exceeds $p_1$ if and only if accuracy rises ($F_{ic} > F_{ci}$), so $\hat{p}^*$ restates the net change and cannot serve as an independent test on the same transitions; with no flips, or with all agents correct or all wrong, $\hat{p}^*$ is undefined. Under a no-communication null where each agent draws a fresh answer with item accuracy $p_i$ (mean $\bar{p}=\frac{1}{n}\sum_i p_i$, mean within-item variance $V=\frac{1}{n}\sum_i p_i(1-p_i)>0$), expected wrong-to-right and right-to-wrong counts over $n$ items both equal $nKV$, so the plug-in transition rates converge to $V/(1-\bar{p})$ and $V/\bar{p}$ and return $\hat p^*=\bar{p}$ without any peer influence. Empirical transcript audits should therefore report transition rates and peer correlations as excess over a no-communication control alongside empirical transition tables keyed both by correct-peer count and by peer share holding the agent's own answer (\texttt{pace dynamics -{}-null}; see \Cref{sec:appendix_mae_eval}).

\FloatBarrier

\subsection{Three further formal results: hurdle, imitation and disclosure}\label{app:moved_s3_formal}

\begin{theorem}[Compute--correlation hurdle]\label{thm:hurdle}
Let a $K$-agent, $T$-round debate have final-round per-agent accuracy $p_T$ and pairwise answer correlation $\rho_T$, so that its vote has $K_{\mathrm{eff}}=K/(1+(K-1)\rho_T)$ (\Cref{prop:keff}; \Cref{app:moved_s3_prop2}), and let the token-equated frontier be a vote over $k_{\mathrm{tok}}=K\,[T+(T-1)(K-1)r]$ independent draws of accuracy $p_1$ (\Cref{eq:ktok}). In the standardized-margin (normal) approximation to a majority vote, with $z(p)=(p-\tfrac12)/\sqrt{p(1-p)}$, the debate's vote beats the frontier if and only if
\begin{equation}
z(p_T)\;>\;z(p_1)\,\Lambda(K,T,r,\rho_T),\qquad \Lambda=\sqrt{\big(T+(T-1)(K-1)r\big)\big(1+(K-1)\rho_T\big)}.
\label{eq:hurdle}
\end{equation}
\end{theorem}
On MMLU-Pro (\texttt{llama3.1:8b}, $K=5$, $T=3$, measured $k_{\mathrm{tok}}=\mmluKtokExact$) $\Lambda=\lamHurdleZero$, $\lamHurdleQuarter$ and $\lamHurdleHalf$ at $\rho_T=0,0.25,0.50$ (normal approximation; \Cref{sec:appendix_proofs}). These values use the binary normal approximation, which does not describe ten-option MMLU-Pro with single-sample accuracy below one half; read them as how $\Lambda$ grows with $\rho_T$, not as a prediction for this task.

\begin{proposition}[Coalescent collapse under peer imitation]\label{prop:coalescent}
Let $K\ge 2$ agents hold $\{0,1\}$ answers to one item that are independent with mean $p_1$ at round~1; in each later round every agent independently, with probability $\gamma\in(0,1)$, copies the answer of a uniformly chosen other agent and otherwise keeps its own. Then $\mathbb{E}[p_t]=p_1$ for all $t$, the pairwise correlation is $\rho_{\mathrm{within}}(t)=1-(1-c)^{t-1}$ with $c=\tfrac{2\gamma(1-\gamma)}{K-1}+\tfrac{(K-2)\gamma^{2}}{(K-1)^{2}}$, so $K_{\mathrm{eff}}(t)=K/\big[K-(K-1)(1-c)^{t-1}\big]$, and unanimity is reached almost surely and is correct with probability $p_1$, so for odd $K$ the majority vote moves from $p_{\mathrm{SC}}(K;p_1)$ to $p_1$ as $t\to\infty$.
\end{proposition}

\begin{diagnostic}[Disclosure bottleneck, one-parameter calibration]\label{prop:disclosure}
Suppose the pooled answer needs $S$ pieces of private evidence, each put into the visible peer messages independently with probability $q((1-\theta)\ell)$, nondecreasing in the visible length left after a share $\theta$ of the generated length $\ell$ goes to private reasoning. Then every piece is disclosed with probability $d=q^{S}$ (nonincreasing in $\theta$ and $S$, and tending to $0$ as $q\to 0$), and $\operatorname{Acc}=d\,a_1+(1-d)\,a_0$ for the accuracies $a_1$, $a_0$ on fully and not fully disclosed items.
\end{diagnostic}

\FloatBarrier

\subsection{Disclosure model against the measured d}\label{app:moved_s3_disclosure_check}

With ten private agent-criterion pairs per item and the measured per-criterion rate $q=\hpDiscRate$, independent disclosure gives $d=q^{10}=\hpDiscQPowTen$, against $\hpDiscD$ observed (bottom row of \Cref{fig:disclosure_diag}); the excess shows positive dependence of disclosures within an item. The same $q$ raised to the power six gives $\hpDiscQPowSix$, which equals the observed $d$ only as a one-parameter coincidence and is not evidence for the model. In \texttt{deepseek-r1:14b}, where thinking characters are \confReasonShareHp{} of thinking plus peer-message characters (\Cref{app:controlled}), the model implies $d\to 0$, consistent with the observed accuracy of $0.000$.

\FloatBarrier

\subsection{Two formal results: scope}\label{app:moved_s3_twoformal}

\paragraph{Two formal results.} The next two propositions formalize the confounds that the baseline ladder is meant to remove, building on Condorcet's jury theorem \citep{condorcet1785}, its extension to correlated votes \citep{ladha1992condorcet}, Kish's design effect \citep{kish1965survey}, and a two-state Markov chain, and state the vote-amplification, variance-inflation, scaffolding and conformity-threshold results that the ladder needs in the notation of this paper. Both propositions describe stylized models under the assumptions listed in them; proofs are in \Cref{sec:appendix_proofs}. Throughout, $p_1$ is the accuracy of a single random draw at the sampling temperature. Greedy decoding is not a random draw at $p_1$, so the accuracy of a Category~A baseline can lie above or below $p_1$, and \Cref{prop:vote_amp} compares a vote with a single sampled draw.

\FloatBarrier

\subsection{Proposition: majority-vote amplification, variance inflation, and scaffolding dilution}\label{app:moved_s3_prop2}

\begin{proposition}[Independent vote amplification]\label{prop:vote_amp}
Let an agent answer a binary item correctly with probability $p_1$ on each independent draw, and let $K\ge 3$ be odd. The plurality-vote accuracy $p_{\mathrm{SC}}(K)=\Pr[\mathrm{Bin}(K,p_1)>K/2]$ satisfies $p_{\mathrm{SC}}(K)\ge 1-\exp\!\big(-2K(p_1-\tfrac12)^2\big)$ for $p_1>\tfrac12$, and
\begin{equation}
\Delta_{\mathrm{vote}}(K,p_1)\;\triangleq\;p_{\mathrm{SC}}(K)-p_1\;\;\begin{cases}>0 & p_1\in(\tfrac12,1),\\ =0 & p_1\in\{0,\tfrac12,1\},\\ <0 & p_1\in(0,\tfrac12).\end{cases}
\label{eq:cat_a_bias}
\end{equation}
On multiclass items ($|\mathcal{Y}|>2$), if the correct class $y^*$ has margin $p_{y^*}\ge \max_{y\neq y^*} p_y + d$ ($d>0$), plurality voting over $K$ independent draws succeeds with probability at least $1-(|\mathcal{Y}|-1)\exp(-Kd^2/2)$.
\end{proposition}
\begin{proposition}[Variance-equivalent ensemble size]\label{prop:keff}
If $K$ answers have common mean, variance $\sigma^2$ and pairwise correlation $\rho_{\mathrm{peer}}\in[-\tfrac{1}{K-1},1]$, then $\operatorname{Var}(\bar Z)=\frac{\sigma^2}{K}\big[1+(K-1)\rho_{\mathrm{peer}}\big]$, so the variance-equivalent effective ensemble size (the number of independent answers with the same variance of the mean) is $K_{\mathrm{eff}}=K/\big(1+(K-1)\rho_{\mathrm{peer}}\big)\le K$ for $\rho_{\mathrm{peer}}\ge 0$ under exchangeable error correlation, with equality only if $\rho_{\mathrm{peer}}=0$.
\end{proposition}
\begin{proposition}[Scaffolding dilution]\label{prop:scaffold}
In a protocol in which every call carries a fixed scaffold of $c_0$ tokens and, in round $t$, each of $K$ agents reads all $K$ messages (mean length $\bar\ell$) of each earlier round, the scaffold tokens over $T$ rounds are
\begin{equation}
C_{\mathrm{tot}}(K,T)\;=\;KTc_0+\tfrac12K^2T(T-1)\bar\ell,
\label{eq:scaffold_total}
\end{equation}
and under a total budget $B$ the thinking allowance per call is $B_{\mathrm{eff}}(K,T)=\max\!\big(0,\big(B-C_{\mathrm{tot}}(K,T)\big)/(KT)\big)=\max\!\big(0,B/(KT)-c_0-\tfrac12K(T-1)\bar\ell\big)$, which is nonincreasing in $K$ and $T$. For any nondecreasing accuracy-versus-budget curve $\mu$, per-call accuracy is at most $\mu\big(\max(0,B/(KT)-c_0)\big)$.
\end{proposition}

\Cref{prop:vote_amp} says that a vote over $K$ independent draws is more accurate than one sampled draw when single-draw accuracy exceeds one half on the item and less accurate when it is below one half; by Jensen's inequality, a benchmark average $\frac{1}{n}\sum_i p_{\mathrm{SC}}(K; p_i)$ can differ from $p_{\mathrm{SC}}(K; \bar{p})$ depending on item heterogeneity. The multiclass bound counts ties as errors, and for even $K$ with uniform tie-breaking $p_{\mathrm{SC}}(K)=p_{\mathrm{SC}}(K-1)$. Because greedy decoding is not a sampled draw, \Cref{prop:vote_amp} does not by itself sign the difference between a vote and a greedy pass. \Cref{prop:keff} holds for any aggregation of correlated answers, spanning the upper endpoint $\rho_{\mathrm{peer}}=1$ ($K_{\mathrm{eff}}=1$, zero diversity) down to the antithetic boundary $\rho_{\mathrm{peer}}=-1/(K-1)$ (where $\operatorname{Var}(\bar Z)=0$); because cross-item difficulty heterogeneity alone induces nonzero between-agent correlation $\rho_{\text{total}}$, communication effects must be isolated via round-to-round excess correlation or within-item replicate correlation $\rho_{\text{within}} = (C_{\text{same}} - C_{\text{diff}})/(\bar{p} - C_{\text{diff}})$ from $Q \ge 2$ replicate debates per item (checked by $\rho_{\text{within}} \approx 0$ at round 1; \Cref{sec:appendix_proofs}). \Cref{prop:scaffold} is an accounting identity for the full-history protocol.

\FloatBarrier

\subsection{Routes to a positive coordination return}\label{app:moved_s3_routes}

\paragraph{Routes to a positive coordination return.} Within the model, communication raises accuracy exactly when $\kappa>p_1/(1-p_1)$. We identify three routes by which this can happen; we do not claim that they are exhaustive. \emph{(I) Information partitioning.} If agents hold different evidence, the transition rates are not constant: an agent that cannot answer from its own view can become correct only by receiving information. For a finite label set, Fano's inequality bounds below the error probability of any estimator of $Y$ from $(X_{\mathrm{shared}},X_i)$ by $\big(\operatorname{H}(Y\mid X_{\mathrm{shared}},X_i)-1\big)/\log_2|\mathcal{Y}|$, positive when the conditional entropy exceeds one bit; this is an estimation bound for one agent's evidence and does not by itself show that distributed information is needed, which is why $\operatorname{I}(Y;X_1,\dots,X_K\mid X_{\mathrm{shared}})>0$ is necessary for an information gain (jointly, since synergistic evidence can have vanishing pairwise terms). Benchmarks with distributed information, such as Silo-Bench \citep{zhang2026silobench}, probe this route directly: their reported failure to integrate exchanged information is the case in which pooled evidence is available but is not used, and the shared-notebook result of \citet{ou2025analyzing} is an instance in which information sharing helped. \emph{(II) Asymmetric verification.} A verifier $V$ that accepts a proposed change with probability $1-\epsilon_{\mathrm{FN}}$ if the new answer is correct and $\epsilon_{\mathrm{FP}}$ if it is wrong replaces $(\tau_{i\to c},\tau_{c\to i})$ by $\big(\tau_{i\to c}(1-\epsilon_{\mathrm{FN}}),\,\tau_{c\to i}\,\epsilon_{\mathrm{FP}}\big)$, so the gain is positive when $\epsilon_{\mathrm{FP}}<\kappa\,\frac{1-p_1}{p_1}(1-\epsilon_{\mathrm{FN}})$; as $\epsilon_{\mathrm{FP}}\to 0$, $p^{*}\to 1$ (\Cref{cor:verifier}). The Blackwell interpretation applies when verifier output augments the original candidate and the decision rule can retain or ignore the original signal; when a verifier replaces the candidate with a noisy verdict, its false-positive and false-negative rates determine whether the resulting decision rule improves accuracy. Verification of this kind is a tool or environment call and is counted in $R_{\mathrm{sys}}$. \emph{(III) Cross-model error decorrelation.} By \Cref{prop:keff}, $K_{\mathrm{eff}}$ decreases in $\rho_{\mathrm{peer}}$, so a heterogeneous panel with $\rho_{\mathrm{het}}<\rho_{\mathrm{hom}}$ aggregates more independent evidence at the same $K$; this acts on the aggregation side and does not by itself change $\kappa$. Model heterogeneity is a candidate source of functional diversity rather than a substitute for measuring it \citep{hong2004groups}: different model identities do not guarantee complementary representations or independent errors, so Route~III requires reporting pairwise error overlap, within-item correctness correlation, and performance relative to a noncommunicating mixed-model ensemble.

\FloatBarrier

\subsection{Comparing multi-agent teams against greedy single-agent baselines}\label{app:moved_s3_compare}

Comparing multi-agent teams against greedy single-agent baselines mixes two phenomena: variance reduction from drawing multiple stochastic trajectories, and coordination through communication. Under shared token or iteration ceilings (Category~C), directionally feasible comparisons establish whether the single agent matches or exceeds the multi-agent system under the same ceiling. Under matching on agent count (Category~B), the comparison reflects unconstrained sample multiplicity unless the baseline is nested inside the multi-agent run, as with round-1 majority voting over five samples in \citet{choi2025debate}. \Cref{tab:pace_master_reference} in \Cref{app:pace_reference} summarizes all PACE policy classes, estimands, and diagnostic rules in one reference table.

\FloatBarrier

\subsection{Hypothetical worked example of the protocol}\label{app:moved_s4_worked}

\def\paceShow{table}\def\paceShow{both}

\FloatBarrier

\section{Extended Corpus Analysis}

\subsection{Apparent wins against a sampled baseline}\label{app:moved_fig_hero}

\begin{figure}[!htbp]
\centering
\includegraphics[width=0.5\textwidth]{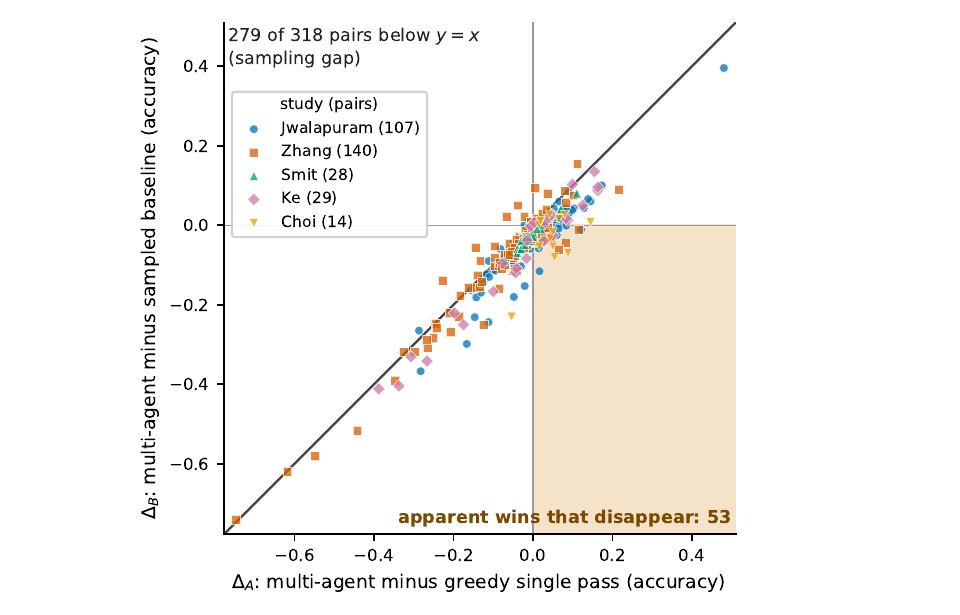}
\caption{\numHeroApparentWins{} apparent wins disappear against a sampled baseline. Each point is one multi-agent configuration of the five studies that report both baselines (\numContrastsDual{} pairs, \numCellsDual{} unique baseline cells, \numNegCellsDual{} negative): $x=\Delta_A$ is its accuracy difference against the greedy single pass (Category~A), $y=\Delta_B$ against the sampled baseline (Category~B). The shaded quadrant ($\Delta_A>0$, $\Delta_B\le 0$) holds those sign flips, and \numHeroBelow{} of \numContrastsDual{} points lie below the diagonal $y=x$, the sampling gap. Categories: A greedy single pass, B sampled, C resource-matched, D Pareto frontier (\Cref{fig:baseline_ladder}).}
\label{fig:hero_scatter}
\end{figure}

\FloatBarrier

\subsection{The baseline ladder}\label{app:moved_fig1_ladder}

\FloatBarrier

\subsection{Related work: controlled audits of multi-agent superiority}\label{app:moved_rw_audits}

Controlled audits challenge unconditional multi-agent superiority. When budgets are matched on thinking tokens, a single agent matches or outperforms debate across reasoning benchmarks \citep{tran2026thinking}. Evaluating 260 configurations that use tools across six benchmarks in \textit{Nature Machine Intelligence}, \citet{kim2026scaling} report that multi-agent gains diminish as baseline model competence rises, turning negative beyond a $\sim 45\%$ threshold. \citet{ferreira2026beyond} evaluate 23 models across eleven vendor families with 5,500 runs with generation-budget matching, finding that debate often ties or trails majority voting once generation budgets are matched. \citet{bertalanic2026cost} observe that multi-agent consensus consumes 2.1--3.4$\times$ more tokens than self-correction without improving accuracy, while nominal agent scaling exhibits coordination degradation analogous to the Ringelmann effect \citep{bertalanic2026ringelmann}. In shared-context token budgeting, \citet{nguyen2026think} observe that self-consistency matches or outperforms deployed debate in 93\% of window-matched settings, and \citet{ann2026interaction} identify an interaction tax under matched budgets on verifier-scored tasks.

\FloatBarrier

\subsection{Related work: matching definitions, cost-controlled agent evaluation, closest prior work, statistics, and task structure}\label{app:moved_rw_more}

Matching definitions diverge across papers: some audits equate internal scratchpad tokens while ignoring prompt inflation; others match iteration counts or monitor monetary API spend \citep{jwalapuram2026illusion,wunderlich2026mixture}. At comparable compute read off Pareto fronts on MMLU-Pro, \citet{wunderlich2026mixture} observe that debate and mixture-of-agents outperform self-consistency by $+1.3$ and $+2.7$ percentage points, reporting positive accuracy differences on dense multiple-choice items. A system matched on token count can remain unmatched on execution latency or financial cost. Recent benchmarking efforts such as the Holistic Agent Leaderboard \citep{kapoor2026holistic} emphasize standardized multidimensional evaluation. The resource vector $\mathbf{R}$ with six dimensions formalizes these reporting distinctions (\Cref{sec:cme}).

\citet{kapoor2024ai} argue that agent evaluations must be cost-controlled and report, on a code-generation benchmark, that simple baselines can match complex agent architectures at lower cost. \citet{wang2024tokeneconomies} likewise compare multi-agent debate with self-consistency under token budgets. The resource vector $\mathbf{R}$ with six dimensions formalizes these cost-aware principles into a standardized accounting framework that we apply across primary studies.

Closest prior empirical work compares debate-style systems with single-agent and self-consistency baselines on sets of benchmarks. \citet{smit2024understanding}, \citet{zhang2025stop}, and \citet{yang2025revisiting} differ in what they hold fixed: Yang et al.\ give self-consistency, refinement and debate the same maximum of 16 samples; Smit et al.\ report the best and the median score over hyperparameter configurations; the tables of Zhang et al.\ do not report matched tokens or cost. In our extraction of their tables, debate-based systems are ahead of the sampled baseline in a minority of contrasts (19\% for Zhang et al., 17\% for Yang et al., 21\% for Smit et al.; \Cref{sec:meta}). To our knowledge none of these papers places its results beside studies with other matching conditions (thinking tokens, iterations, cost, agent count) in one resource-accounting framework, or compares the same multi-agent systems against greedy and sampled baselines across studies; that placement is what we add.

Statistical comparisons of classifiers \citep{demsar2006statistical} and error bar practices for LLM evaluations \citep{miller2024adding} address variance within a benchmark, but not synthesis across papers. Dependent effect sizes within a study are handled by multilevel models \citep{konstantopoulos2011fixed,cheung2014modeling} with robust variance estimation \citep{hedges2010robust,tipton2015small}. Raw accuracy differences are bounded and ceiling compressed \citep{schaeffer2024are}, which motivates our log odds scale.

Task decomposability governs multi-agent returns. On MASBENCH, orchestrator architectures achieve gains on parallel tasks, but fail on ordered sequential deduction where a single chain of thought scores higher \citep{ke2026masorchestra}. Routing topology also affects token efficiency under partial observability \citep{chu2026masbench}. Analyzing 1,600 annotated traces across seven frameworks, \citet{cemri2025why} catalog 14 recurring system-design, inter-agent misalignment, and verification failure modes, while on HiddenBench \citet{li2026hiddenbench} observe across 15 models that multi-agent groups reach $30.1\%$ accuracy with distributed information versus $80.7\%$ when a single agent holds complete information. \citet{zhu2026demystifying} report that while unguided homogeneous debate trails simple majority voting in most settings, confidence-weighted aggregation and diverse initializations outperform both vanilla debate and majority voting across six reasoning QA benchmarks. Furthermore, framing debate as a collaborative non-zero-sum game with supportive critique oriented toward a downstream judge improves error detection and oversight accuracy at matched token budgets \citep{colmad2026collaborative}. While recent surveys review agent benchmarks \citep{nageshwaran2026agenteval}, prior work does not synthesize effect sizes across distinct matching conditions and baseline sampling strategies. We analyze paired contrasts extracted directly from primary empirical tables across the literature.

\FloatBarrier

\subsection{Related work: single-agent baselines, aggregation, and reproducibility standards}\label{app:moved_rw_extra}

\paragraph{Single-agent baselines and aggregation.} \citet{wang2024rethinking} find that a single agent with strong prompts almost matches the best discussion approach on a wide range of reasoning tasks, and that multi-agent discussion is better only when the prompt contains no demonstration; this is an early baseline-strength result for discussion. \citet{li2025rethinking} show for Mixture-of-Agents that mixing weaker heterogeneous models degrades synthesis relative to Self-MoA, which aggregates multiple sampled outputs from one strong model, and \citet{xu2026rethinking} demonstrate across seven benchmarks that executing homogeneous workflows inside one shared single-agent context matches or exceeds multi-agent accuracy while reducing inference cost through KV-cache reuse. \citet{yang2026agentauditor} note that most multi-agent frameworks aggregate by majority vote, which is brittle when agents share correlated biases, and propose an evidence-structured adjudicator; correlated answers are the case that \Cref{prop:keff} describes through the effective number of independent answers. \citet{ou2025analyzing} study information sharing in a multi-agent travel-planning system and report that a shared notebook reduces hallucination-related errors and that an orchestrator reduces errors within focused sub-areas, against a single-agent baseline with a 7.5\% pass rate. \citet{zhang2026silobench} introduce Silo-Bench, a benchmark of 30 algorithmic tasks in which information is distributed across agents, and report a communication--reasoning gap: agents exchange information but fail to integrate it into correct answers.

\paragraph{Reproducibility standards and noise floors.} \citet{rabanser2026reliability} propose twelve metrics that decompose agent reliability into consistency, robustness, predictability, and safety, and report that recent capability gains have yielded only small reliability improvements; run-to-run consistency is the same property that a noise floor measures. \citet{pattnayak2026reproevalcard} audit 55 LLM pipeline papers published from 2022 to 2025 for evaluation artifacts and report that randomness controls are missing in 75\% and intermediate execution traces in 61\% of them; their coding concerns artifact availability, whereas the audit here concerns baseline strength. \citet{masters2026rollout} propose rollout cards, which preserve the rollout record behind a reported score together with the reporting rules and drops manifests used to compute it, and document reporting-rule differences in cost and token accounting. The PACE resource vector can be written as a compute block in a card's annotation namespace and reporting-rule registry (\texttt{pace report -{}-export-rollout-card}; an example is in \Cref{sec:appendix_mae_eval}). \citet{kaliyev2026coordination} propose a paired noise-floor protocol for coordination gains. Using Claude Haiku 4.5 on $\tau^2$-bench retail, the clean configuration-equivalent contrast gives $+10$ and $0$ percentage points across two seeds of $n=100$ (pooled $+5$ points, Wilson CI $-2$ to $+12$). The envelope of the observed paired gaps spans $-3$ to $+18$ points, and the $+18$-point contrast did not reproduce at the second seed. A noise floor of this kind and a baseline-strength audit address different threats: the first asks how large a difference must be to exceed run-to-run variation, and the second asks whether the baseline received comparable compute.

\FloatBarrier

\subsection{The nine studies and the extraction procedure}\label{app:moved_s5_nine}

We extracted results directly from the primary tables of nine studies. The five initial studies are \citet{tran2026thinking} (Table~1), \citet{choi2025debate} (Table~1), \citet{kim2026scaling} (Table~16 of the arXiv version), \citet{jwalapuram2026illusion} (Table~4 and Appendix Table~6), and \citet{wunderlich2026mixture} (accuracies stated in Section~4). Tables were parsed from source PDFs with \texttt{pdftotext} and inspected against rendered pages. We synthesize \numStudiesTotal\ primary empirical studies comprising \numContrastsTotal\ unique multi-agent configurations across \numRawRows\ raw extracted table rows (\Cref{tab:study_overview}). From the five initial studies (252 raw extracted rows), excluding 14 majority-voting baseline rows in Choi et al.\ and 1 self-consistency baseline row in Wunderlich et al.\ yields 237 unique multi-agent configurations spanning thinking tokens, iteration limits, monetary cost, agent counts, and Pareto compute sweeps. Seven additional candidate papers identified for eligibility screening were then screened against written eligibility criteria requiring paired tabular multi-agent and single-agent accuracy on identical benchmark splits with objective scoring (\Cref{app:candidates}): three were excluded (including \citet{wang2024tokeneconomies}, which reports token-budget curves only in plots), and four met all criteria (\citet{ke2026masorchestra}, 29 configurations; \citet{yang2025revisiting}, 42; \citet{zhang2025stop}, 180; \citet{smit2024understanding}, 28; totaling 279 configurations). This screening is not a systematic literature search, and this paper makes no claim of completeness. Sample sizes come from source appendices where stated; the MuSiQue 4-hop test size is the official dev split ($N=405$), FRAMES ($N=824$) uses the published benchmark size, and SMFR ($N=588$) comes from Appendix Table~6 of Jwalapuram et al.

\FloatBarrier

\subsection{Bounded corpus selection and screening}\label{app:moved_s5_bounded}

\paragraph{Bounded corpus selection and screening.} The corpus is a bounded synthesis, not a full systematic literature review. Five initial studies were selected to span compute dimensions. The second stage screened seven candidates identified by the authors (six additional papers plus one earlier exclusion); four were included and three excluded, yielding nine studies. Eligibility required: (i) accuracy reported for at least one multi-agent and one single-agent system; (ii) evaluation on the same benchmark items; (iii) results in a table or text; and (iv) an identifiable single-agent baseline with objective scoring. Unmatched resource budgets were retained because PACE classifies resource relations rather than requiring parity for inclusion; excluding them would remove comparisons needed to examine baseline strength across matching conditions. Candidates were screened against these criteria; titles and identifiers were checked against arXiv pages. No calendar-date cutoff or preregistered corpus freeze is reported; the boundary is the screen completed in stage two, not exhaustive coverage.

\FloatBarrier

\subsection{Evidence flow and audit}\label{app:moved_s5_flow}

\begin{figure}[!htbp]
\centering
\includegraphics[width=0.92\textwidth]{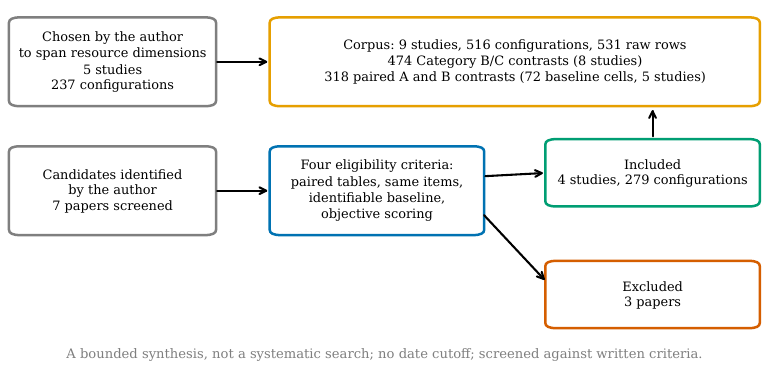}
\caption{Evidence flow and configuration counts across baseline categories. Initial selection (five studies; 237 unique multi-agent configurations; 252 raw rows) and formal screening of seven candidate papers against four eligibility criteria (four included contributing 279 configurations; three excluded) yield nine included studies, 516 unique multi-agent configurations, and 531 raw extracted rows (15 standalone baseline rows). Extracted contrasts total 358 in Category~A (five studies), 360 in Category~B (six studies), 114 in Category~C (two studies), and 2 in Category~D (one study, Pareto front), spanning 474 combined Category~B/C contrasts (eight studies) and 318 paired Category~A and~B contrasts across 72 unique baseline cells (five studies).}
\label{fig:evidence_flow}
\end{figure}

\paragraph{Evidence flow and audit.} As summarized in \Cref{fig:evidence_flow}, the five initial studies were chosen without a systematic search protocol to span different compute dimensions, after which seven candidate papers were screened against written eligibility criteria (\Cref{app:candidates}), yielding four included studies and three exclusions. Combined, the \numStudiesTotal\ studies provide \numContrastsTotal\ unique multi-agent configurations across \numRawRows\ raw table rows; \Cref{tab:baseline_tiers} gives the split by baseline category. Tran and Kiela matched thinking tokens, Kim et al.\ matched iterations, and Yang et al.\ matched generations (16 samples); the other added studies compare multi-agent graphs against self-consistency without verified budget equality. Values were extracted from primary source tables, candidate papers were screened against written eligibility criteria, programmatic PDF concordance checks verified 336 of 336 table values cell-by-cell for Tran \& Kiela alongside page-level value checks across studies, and a stratified 40-record ledger (\texttt{review/EXTRACTION\_AUDIT\_LEDGER.md}, four to five records per study across all nine studies) was checked against the rendered primary tables with zero numerical discrepancies. The nine-study corpus is therefore a criterion-based synthesis of identified evaluations (12 papers considered across two stages: five initial studies selected to span resource dimensions plus seven candidates screened in \Cref{app:candidates}, of which four were included and three excluded), not a probability sample of all multi-agent research. Across the corpus, 531 raw extracted table rows minus 15 baseline-only rows (14 majority-voting rows in Choi et al.\ and 1 self-consistency row in Wunderlich et al.) yield 516 unique multi-agent configurations; pairing these configurations with reported single-agent references forms 834 baseline-specific comparison records (358 Category~A, 360 Category~B, 114 Category~C, and 2 Category~D Pareto comparisons, which overlap when a study reports multiple baselines), 474 combined Category~B/C comparison records, and 318 linked Category~A and Category~B pairs across five dual-baseline studies that reduce to 72 unique $(\mathrm{study},\mathrm{benchmark},\mathrm{model})$ baseline cells.

\FloatBarrier

\subsection{Pairing rules and study overview}\label{app:moved_s5_pairing}

\paragraph{Pairing rules.}
Tran and Kiela report five multi-agent architectures; the main analysis uses Sequential (their primary baseline) and Debate. Pooling all five (subtask-parallel, parallel roles, and ensemble in addition) gives 240 contrasts with a mean difference of $-0.017$ against $-0.007$ for the two arms (\Cref{tab:tran_all_arms}); only Debate has a positive mean ($+0.002$). For Tran and Kiela we pair SAS with the Sequential and Debate architectures at the same budget of thinking tokens across four models, two benchmarks, and six budgets ($m=96$). For \citet{choi2025debate}, majority voting over five samples ($0.747$) is treated as a Category~B baseline rather than a multi-agent system: the $m=14$ decentralized debate configurations ($T=2$, five agents, mean $0.715$) are paired against single-run decoding (Category~A, mean $0.670$, $\bar\Delta=+0.045$) and against majority voting over five samples (Category~B, $\bar\Delta=-0.031$). For Kim et al.\ we pair Single with Centralized across nine models on SWE-bench Verified and Terminal-Bench ($m=18$, $n=20$ instances per cell); Centralized was fixed before computing summaries, and the other three architectures appear in the released extraction table. For Jwalapuram et al.\ we pair chain-of-thought with each of six automatic MAS frameworks and Expert MAS across five benchmarks (GPQA-Diamond, HLE-Maths, SWE-Bench Lite, BrowseComp-Plus, and SMFR) and up to four models ($m=107$). For \citet{wunderlich2026mixture}, self-consistency with 10 samples ($0.687$) is treated as the comparable-compute Category~B/D baseline rather than a multi-agent arm, pairing debate ($0.700$, $+0.013$) and mixture-of-agents ($0.714$, $+0.027$) against it ($m=2$, $\bar p_{\text{MAS}}=0.707$, $\bar\Delta=+0.020$; vs.\ greedy CoT $0.643$). The four added studies contribute 279 configurations against their primary baselines (Ke et al.\ 29, Yang et al.\ 42, Zhang et al.\ 180, Smit et al.\ 28). Total unique multi-agent configurations: $M=\numContrastsTotal$ (from $\numRawRows$ raw extracted rows).

\begin{table}[t]
\centering
\scriptsize
\setlength{\tabcolsep}{2.8pt}
\caption{The nine studies. $m$: contrasts against each study's reference baseline ($K=9$ corpus, 516 unique multi-agent configurations across 531 raw extracted rows after excluding 14 majority-voting rows in Choi et al.\ and 1 self-consistency baseline row in Wunderlich et al.); $\bar\Delta$: mean accuracy difference, multi-agent minus single-agent. Category: baseline classification used in \Cref{tab:baseline_tiers} and \Cref{fig:baseline_ladder}. The first five studies are the initial five-study corpus; the last four were added by screening (\Cref{app:candidates}). Wunderlich et al.\ contributes Pareto comparisons against CoT ($+0.057$ for debate and $+0.071$ for MoA; unweighted mean $+0.064$, up to $20\times$ compute) and two contrasts against self-consistency at comparable compute read off Pareto fronts ($+0.013$ debate, $+0.027$ MoA; mean $\bar\Delta = +0.020$, $\bar p_{\mathrm{MAS}}=0.707$ vs.\ $\bar p_{\mathrm{SC}}=0.687$); self-consistency is treated as a baseline rather than a multi-agent arm.}
\label{tab:study_overview}
\begin{tabular*}{\textwidth}{@{\extracolsep{\fill}}lllrll@{}}
\toprule
Study & Matching condition & Baseline(s) & $m$ & \multicolumn{1}{r}{$\bar\Delta$} & Category \\
\midrule
Tran and Kiela & Thinking tokens & SAS, equal budget & 96 & $-0.007$ & C (Budget-matched) \\
Kim et al. & Iterations & Single, budget-matched & 18 & $-0.042$ & C (Budget-matched) \\
Jwalapuram et al. & Cost audited & CoT; CoT-SC(5) & 107 & $-0.007$ & A (Greedy), B (Sampled) \\
Choi et al. & Agent count & Single run; majority vote & 14 & $+0.045$ / $-0.031$ & A (Greedy), B (Sampled) \\
Wunderlich et al. & Comparable (Pareto) & CoT; SC (Pareto) & 2 & $+0.020$ & D (Pareto) \\
\midrule
Ke et al. & Unreported & CoT agent; SC agent & 29 & $-0.028$ & A (Greedy), B (Sampled) \\
Yang et al. & Approx. (16 samples) & SC (16) & 42 & $-0.021$ & B (Sampled) \\
Zhang et al. & Unmatched / unreported & CoT; SC & 180 & $-0.058$ & A (Greedy), B (Sampled) \\
Smit et al. & Unreported & Single agent; SC (median) & 28 & $+0.000$ & A (Greedy), B (Sampled) \\
\bottomrule
\end{tabular*}
\end{table}

\FloatBarrier

\subsection{Descriptive results and sensitivity across two baselines}\label{app:moved_s5_two_baselines}

\Cref{tab:appB_ledger} gives unweighted mean accuracy differences ($\bar\Delta$) by study and benchmark; in the multilevel analyses we also use the log odds ratio, which is less sensitive to ceiling compression than the raw difference \citep{schaeffer2024are}. No standard errors or $p$-values are attached to descriptive summaries in this subsection because contrasts within a study share baselines.

\paragraph{Budget-matched and cost-audited evaluations (Categories C and A).}
When thinking tokens are matched, $\bar\Delta=-0.007$ over 96 contrasts, with one third favoring MAS. These contrasts compare the single-agent arm with the Debate and Sequential arms of Table~1 in the source (4 models, 2 benchmarks, 6 budgets, 2 arms). The Subtask, Roles, and Ensemble arms and the longer-thinking single-agent variant are not included, so the result describes two multi-agent designs rather than all five. The pattern depends on budget: at 100 thinking tokens MAS averages $+0.077$, where the source authors document truncated reasoning outputs for both systems; from 500 tokens upward the mean difference is negative (about $-0.012$ at 500 tokens and $-0.023$ to $-0.035$ from 1,000 to 10,000 tokens). Under matched iterations, $\bar\Delta=-0.042$ over 18 contrasts ($-0.017$ on SWE-bench Verified, $-0.067$ on Terminal-Bench), though each contrast rests on 20 instances.\footnote{On SWE-bench Verified, although the unweighted accuracy difference is $-0.017$, the unweighted mean log odds ratio computed on the same nine contrasts is $+0.056$. This divergence arises from the nonlinearity of the logit near the floor of the accuracy scale, where a small absolute change corresponds to a large change in log odds: on \texttt{gpt-5-nano}, MAS rises from $5\%$ to $35\%$ ($+30\%$), yielding $\ln\text{OR} = +1.98$ that offsets smaller negative differences across eight other models.}

\paragraph{Comparison against greedy baseline versus self-consistency in Jwalapuram et al.}
In the evaluation of Jwalapuram et al.\ ($m=107$) where dollar costs were audited, multi-agent systems incur median dollar costs $8.1\times$ higher than greedy chain-of-thought (mean $27.1\times$, ranging up to $225\times$). Against greedy CoT, the unweighted accuracy difference averages $\bar\Delta = -0.007$, showing small gains on GPQA-Diamond ($+0.010$) and HLE-Maths ($+0.025$). The same paper also reports a self-consistency row with five samples (CoT-SC). Against that row the unweighted difference is $\bar\Delta=-0.041$ (multi-agent systems ahead in $25\%$ of contrasts versus $42\%$ against greedy CoT; \Cref{tab:appB_dual}). The source caption states that the single-agent CoT and CoT-SC rows ``report the best observed score across MAS system entries''; because per-entry baseline scores are not published, differences are computed against these reported rows, representing an upper-bound single-agent baseline. Because the rows are maxima over entries, differences against them are biased against the multi-agent systems.

\paragraph{Tooling versus prompting on SMFR.}
\Cref{fig:pareto_frontier} plots accuracy against cost for GPT-5 on SMFR ($N=588$). Automated multi-agent frameworks score between $20.2\%$ (ADAS) and $63.0\%$ (MAS-Orchestra) at costs from \$427 to \$9,239, compared with $48.5\%$ for the CoT row and $57.0\%$ for the CoT-SC row. Tooling built by humans (Expert MAS) reaches $96.51\%$ at \$555, suggesting that gains on this benchmark were associated with domain-specific tooling rather than conversational prompting alone; this observation reflects one benchmark and one specialized system.

\paragraph{Sampled comparisons and voting in Choi et al.\ (Category B).}
In \citet{choi2025debate}, majority voting over five samples without communication attains an unweighted gain of $+0.076$ over single-run baseline across 14 model and benchmark cells ($0.747$ vs.\ $0.670$). Decentralized debate with two rounds among five agents attains $+0.045$ against single-run baseline ($0.715$). When debate is compared directly against majority voting (holding agent count equal at five, where voting aggregates round-1 answers prior to communication), debate trails by $\bar\Delta = -0.031$ across the 14 cells: debate is ahead in 6 cells, tied in 1, and behind in 7, with a cell-level median difference of $-0.008$ (the mean is pulled downward by \texttt{Qwen2.5-7B-Instruct} on Arithmetics dropping from $0.990$ voting to $0.760$ debate; omitting that cell yields mean $-0.016$). Across all nine debate configurations (Decentralized, Sparse, and Centralized MAD at $T \in \{2,3,5\}$) averaged over the 14 cells in Table~1 of the source, debate trails voting. This pattern is consistent with the interpretation that positive agent count differences reflect variance reduction from repeated sampling rather than communicative coordination.

\paragraph{Pareto frontiers (Category D).}
The stated configurations in \citet{wunderlich2026mixture} on MMLU-Pro with Llama-3.1-70B ($N=1{,}000$) exceed greedy CoT ($64.3\%$) at up to about $20\times$ compute, reaching $68.7\%$ for self-consistency with 10 samples, $70.0\%$ for debate, and $71.4\%$ for mixture-of-agents. At comparable compute read off Pareto fronts, debate and mixture-of-agents exceed self-consistency by $+1.3$ and $+2.7$ percentage points ($\bar p_{\text{MAS}}=70.7\%$, $\bar\Delta=+0.020$ across the two multi-agent configurations). Among extracted studies in our corpus, Wunderlich et al.\ is the sole positive exception, at comparable compute read off Pareto fronts. Recent investigations of debate mechanisms (\citealp{zhu2026demystifying} on confidence calibration and \citealp{colmad2026collaborative} on collaborative debate designed for a downstream judge, whereas \citealp{ferreira2026beyond} observe that debate often ties or trails majority voting at matched generation budgets) suggest candidate design hypotheses when cognitive diversity or oversight structure is altered.

\FloatBarrier

\subsection{Headline by baseline category}\label{app:moved_s5_headline}

\paragraph{Headline by baseline category.} A single pooled mean mixes baselines of different strength, so \Cref{tab:baseline_tiers} and \Cref{fig:empirical_ladder} disaggregate outcomes by category. Category~A compares multi-agent systems against a single-pass baseline (\numContrastsCatA\ contrasts, five studies): the contrast mean is \meanCatA\ (\meanCatAStudy\ when weighted equally by study), with multi-agent systems ahead in \winRateCatA\ of contrasts (\meanCatANoZhang\ contrast mean and \meanCatANoZhangStudy\ study mean without Zhang et al., \winRateCatANoZhang\ ahead). Category~B uses sampled baselines (\numContrastsCatB\ contrasts, six studies): \meanCatB\ contrast mean (\meanCatBStudy\ study mean), ahead in \winRateCatB\ (\meanCatBNoZhang\ without Zhang et al.). Category~C uses budget-matched baselines (\numContrastsCatC\ contrasts, two studies): \meanCatC\ contrast mean (\meanCatCStudy\ study mean). Combined Category~B/C encompasses \numContrastsCatBC\ contrasts across eight studies (\meanCatBC\ contrast mean overall, \meanCatBCStudy\ study mean; \meanCatBCNoZhang\ contrast mean without Zhang et al.). The multi-agent system's accuracy equals or falls below its baseline's in \numNotAheadBC\ of these \numContrastsCatBC\ contrasts (\pctNotAheadBC\%; the share ahead is \winRateCatBC), and in \numNotAheadB\ of the \numContrastsCatB\ contrasts against sampled baselines (\pctNotAheadB\%).

\FloatBarrier

\subsection{Dominance, directional feasibility, noise bands, adjudication, and study means}\label{app:moved_s5_sens}

\paragraph{Dominance on accuracy and observed cost.} By the Pareto-dominance definition of \Cref{sec:appendix_pareto}, a baseline Pareto dominates a multi-agent system only if its accuracy is at least as high and it uses no more of every resource in $\mathbf{R}$; we checked the resource condition contrast by contrast against the cost recorded in the contrast data. Cost is observed per contrast only in Jwalapuram et al.: out of 80 not-ahead contrasts against CoT-SC(5) (since 27 of 107 are ahead), \numParetoCostObserved\ have observed cost and 1 lacks a cost entry; a baseline uses no more cost, with at least one strict inequality, in \numParetoDominated\ (Pareto dominance on accuracy and cost) and costs more than the multi-agent system in \numParetoCheaperMAS. In the other \numParetoUnobserved\ contrasts in which the multi-agent system is not ahead, the data record no per-arm cost, so dominance on accuracy and cost cannot be assessed. The share with verified dominance on accuracy and cost is therefore \numParetoDominated\ of \numContrastsCatBC\ contrasts (\pctParetoDominatedBC\%), and \pctNotAheadBC\% is an upper bound on the share a baseline dominates.

\paragraph{Directional feasibility.} Under directional resource feasibility on the controlled or reported dimension ($\mathbf{R}_{\text{base},\mathcal{J}} \preceq \mathbf{R}_{\text{MAS},\mathcal{J}}$; \Cref{sec:decomposition}), a baseline using no more resource matches or exceeds the multi-agent system in $36\%$ of comparable contrasts ($171$ of \numContrastsCatBC: 51 in Jwalapuram et al., 64 in Tran and Kiela, 35 in Yang et al., 13 in Kim et al., and 8 in Choi et al.\ where round-1 voting is nested inside the debate run). Three caveats qualify this total: Tran and Kiela and Kim et al.\ share a budget or iteration ceiling rather than equal measured per-item consumption (so under the same ceiling the single agent did at least as well), omitting Jwalapuram et al.\ (whose CoT-SC row is a maximum over entries) leaves $120$ of $367$ contrasts ($33\%$), and ``not ahead'' includes ties and finite-sample noise on small benchmarks ($N=20$ in Kim et al.).

\paragraph{Noise bands.} Because raw sign counts pool exact ties and small-sample noise (such as the $N=20$ coding items in \citealt{kim2026scaling}), applying the paired binomial noise band of \Cref{eq:noise_band} at $\varphi = 0.50$ separates differences beyond the band from those within it: across all \numContrastsCatBC\ Category~B/C contrasts, \numNoiseBelowBC\ fall below $-1.96\,\mathrm{SE}$ (a different set from the $171$ feasible contrasts above; the equal counts are a coincidence), \numNoiseWithinBC\ lie within $\pm 1.96\,\mathrm{SE}$, and \numNoiseAboveBC\ exceed $+1.96\,\mathrm{SE}$ (at $\pm 1.0\,\mathrm{SE}$, the split is 257 below, 159 within, and 58 above). Within the $171$ feasible not-ahead contrasts, 163 are negative ($\Delta < 0$), 8 are exact ties ($\Delta = 0$), 110 fall below $-1.0\,\mathrm{SE}$, and \numFeasibleNoiseBelow\ fall below $-1.96\,\mathrm{SE}$; within the \numSharedCeil\ contrasts that share a ceiling or generation budget in \citet{tran2026thinking}, \citet{kim2026scaling}, and \citet{yang2025revisiting}, 29 fall below $-1.96\,\mathrm{SE}$, 115 lie within $\pm 1.96\,\mathrm{SE}$, and \numSharedCeilNoiseAbove\ exceed $+1.96\,\mathrm{SE}$. These 156 contrasts do not share one compute constraint. Of them, 96 come from Tran and Kiela, who matched the number of thinking tokens, 18 from Kim et al., who matched the number of iterations, and 42 from Yang et al., who matched the number of generations. Paired binomial noise bands are diagnostic classifications under stated item-count and $\varphi=0.50$ assumptions (with AGIEval using an $N=100$ placeholder and certain benchmarks using dataset sizes) rather than independent item-level hypothesis tests reconstructed from raw response logs.

\paragraph{Contrasts in which the system is ahead.} Conversely, at a shared ceiling or matched generation budget in Tran and Kiela, Kim et al., and Yang et al., the multi-agent system is ahead in \numSharedCeilAhead\ of \numSharedCeil\ contrasts (\pctSharedCeilAhead\%), or \numSharedCeilNoHundredAhead\ of \numSharedCeilNoHundred\ (\pctSharedCeilNoHundredAhead\%, with $112$ not ahead) when omitting Tran and Kiela's truncated 100-token budget (where all 16 contrasts favor MAS). Across Jwalapuram et al.'s 105 Category~B contrasts with observed cost ($2$ of 107 lack cost: 1 ahead, 1 not ahead), 79 are not ahead (51 with a baseline at equal or lower cost and 28 with a costlier baseline) and 26 are ahead (7 ahead at lower cost and 19 ahead at higher cost).

\paragraph{Configuration-level adjudication (\Cref{tab:adjudication}).} Aggregating to the \numContrastsTotal\ unique configurations across all nine studies (where a single greedy pass in Category~A never makes more model calls than the multi-agent system), a baseline using no more of a known resource or sharing the same ceiling scored at least as high in \numConfigFeasible\ configurations (\pctConfigFeasible\%: $171$ in Category~B/C plus 185 additional configurations with $\Delta_{\text{Cat A}} \le 0$ among the 224 nonpositive Category~A rows), the multi-agent system is ahead under an equal ceiling or lower observed cost in \numConfigAhead\ configurations (\pctConfigAhead\%: \numSharedCeilAhead\ in Tran/Kim/Yang [\numSharedCeilNoHundredAhead\ above 100 tokens] and 7 in Jwalapuram), and \numConfigUndet\ (\pctConfigUndet\%, including the 2 configurations in Wunderlich et al.\ where per-item compute is not matched) remain undetermined (ahead only of baselines that used less compute or lacked per-arm cost).

\paragraph{Two-field coding.} Separating baseline type from resource relation in a pair of fields moves Yang et al.'s 42 generation-matched contrasts and Choi et al.'s 14 agent-matched contrasts alongside Tran and Kiela and Kim et al.\ (170 contrasts at an equal ceiling or generation budget, mean $-0.016$, and 304 unmatched sampled contrasts, mean $-0.056$), without changing the combined 474-contrast Category~B/C pool. Within Category~C ($114$ contrasts, $-0.012$), $48$ contrasts from Tran and Kiela evaluate a single-context iterative refinement arm (\texttt{Sequential}), while the remaining $66$ contrasts cover Debate and Kim et al.

\paragraph{Study means and overlap.} All eight Category~B/C study means are negative, from $-0.007$ (Tran and Kiela) to $-0.073$ (Zhang et al.; \Cref{fig:empirical_ladder}), while Wunderlich et al.\ averages $+0.020$ across its two stated configurations against self-consistency (with fronts not digitized and compute not matched). The categories overlap because five studies compare identical multi-agent systems against both Category~A and Category~B baselines. Cost ratios are observed only in Jwalapuram et al.\ (median \medianCostRatioCoT$\times$ vs.\ greedy CoT; \medianCostRatioSC$\times$ vs.\ CoT-SC). \Cref{tab:adjudication} in \Cref{app:adjudication_notes} reports the full configuration-level adjudication, Category~B/C noise-band breakdown, and dual-baseline cell counts (see also \Cref{tab:adjudication_summary}).

\FloatBarrier

\subsection{Dependence on Zhang et al. and moderator synthesis}\label{app:moved_s5_zhang_mod}

\begin{figure}[!htbp]
\centering
\includegraphics[width=0.82\textwidth]{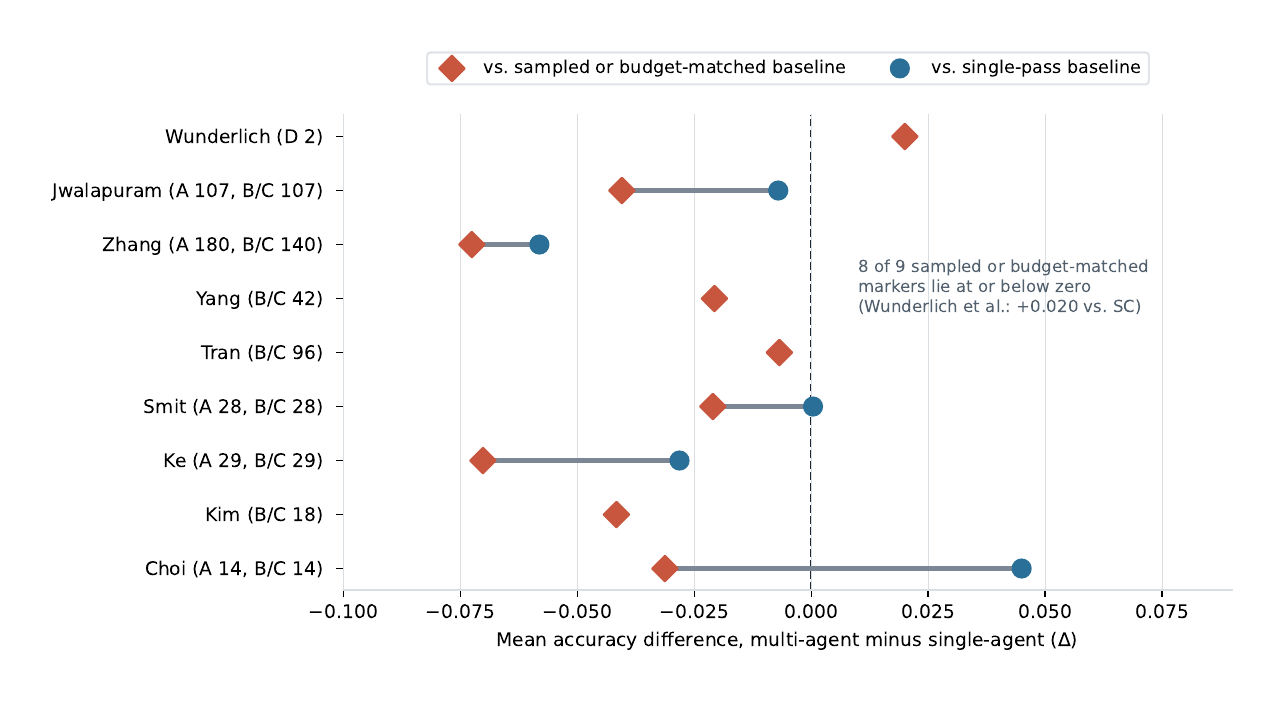}
\caption{Empirical Study Ladder and the Sampling Gap ($\Delta_{\text{gap}}$). All eight Category~B/C study means are negative, and the one Category~D study (Wunderlich et al., $+0.020$) is positive. Mean accuracy differences of multi-agent configurations against greedy single passes (circles, Category~A) and sampled, budget-matched, or Pareto baselines (diamonds, Categories~B/C/D) across nine studies ($m_{\mathrm{A}}=358$, $m_{\mathrm{BC}}=474$, $m_{\mathrm{D}}=2$). Connecting segments show the paired sampling gap ($\Delta_{\text{gap}}$) across 318 paired contrasts (72 unique baseline cells, five studies); Wunderlich et al.\ ($m=2$) plots the comparison against CoT-SC with 10 samples at comparable compute read off Pareto fronts ($+0.020$). Parenthetical numbers give plotted contrast counts per study in Category~A ($m_{\mathrm{A}}$) and Categories~B/C ($m_{\mathrm{BC}}$; Zhang et al.\ exclude HumanEval and MBPP from Category~B).}
\label{fig:empirical_ladder}
\end{figure}

\paragraph{Dependence on Zhang et al.} Zhang et al.\ supply 180 multi-agent configurations against greedy CoT (Category~A) but 140 against self-consistency (Category~B): the source reports no self-consistency baseline for the two code benchmarks, HumanEval and MBPP, so seven benchmarks, five methods and four models give 140 contrasts. In \Cref{fig:empirical_ladder} the label of a study gives its Category~A and Category~B/C contrast counts separately, and in \Cref{fig:reml_forest}, $m$ counts Category~B/C contrasts. Zhang et al.\ contribute \numContrastsZhang\ of the \numContrastsTotal\ unique multi-agent configurations and yield the lowest study mean ($-0.058$ vs.\ CoT, $-0.073$ vs.\ SC). For Category~A, only the mean weighted equally by study changes sign when omitting Zhang et al.\ (from \meanCatAStudy\ with Zhang et al.\ to \meanCatANoZhangStudy\ without), whereas the contrast-weighted Category~A mean remains negative in both cases (\meanCatA\ with Zhang et al.\ vs.\ \meanCatANoZhang\ without). Category~B/C remains negative across every weighting with or without Zhang et al.\ (contrast mean \meanCatBC\ vs.\ \meanCatBCNoZhang; study mean \meanCatBCStudy\ vs.\ \meanCatBCNoZhangStudy), as does the sampling gap in all five paired studies. Across the full nine-study corpus against each study's reference baseline (\Cref{tab:appB_ledger}), the 516 multi-agent configurations average $-0.026$. For named frameworks, the gap between an introducing paper's claim and a later audit is decomposed in \Cref{tab:allegiance}; it is available for \allegVerified{} of \allegListed{} frameworks. Item counts for the added studies come from each source paper's text (\texttt{evidence/extension\_long.csv}, column \texttt{n\_basis}); AGIEval (Zhang et al.) lacks a published item count ($N=100$ placeholder), and capping every cell size at 100 preserves every sign.

\paragraph{Moderator synthesis across topology, task domain, and model scale.} Disaggregating all 474 Category~B/C contrasts across moderators shows negative average accuracy differences across every topology archetype (range $-0.015$ to $-0.059$), every task domain (range $-0.031$ to $-0.075$), and every backbone model class or parameter tier (\Cref{app:adjudication_notes}, \Cref{tab:pace_subfield_hooks}, and \Cref{tab:supp_topology_archetypes,tab:supp_domain_moderators,tab:supp_model_scale_moderators} in \Cref{sec:appendix_moderators}).

\FloatBarrier

\subsection{Category table: contrast and study means by baseline category}\label{app:moved_s5_tiers_table}

\begin{table}[!htbp]
\centering
\footnotesize
\setlength{\tabcolsep}{3.5pt}
\caption{Headline synthesis by baseline category. Category A: single-agent baseline is one pass (Jwalapuram CoT, Zhang CoT, Smit single agent, Ke CoT agent, Choi single run). Category B: sampled baseline (Jwalapuram CoT-SC, Zhang SC, Smit SC, Ke SC agent, Yang SC with 16 samples, Choi majority voting). Category C: budget-matched baseline (Tran thinking tokens, Kim iterations). Zhang et al.\ supplies 180 of the 516 unique multi-agent configurations in the corpus (531 raw extracted rows including 14 majority-voting rows in Choi et al.\ and 1 self-consistency row in Wunderlich et al.); rows are shown with and without it. The sampling gap is the change in $\Delta$ when the same multi-agent arm is compared with the Category B instead of Category A baseline, available for the five studies that report both ($318$ paired contrasts spanning $72$ unique baseline cells; $64$ of $72$ cells negative, averaging $-0.036$ weighted equally by study, $-0.042$ without Zhang et al., and $-0.034$ weighted by cell). Cost ratios exist only for Jwalapuram et al.\ (median $8.1\times$ against CoT, $1.6\times$ against CoT-SC). Descriptive only.}
\label{tab:baseline_tiers}
\begin{tabular}{llrrrrr}
\toprule
Category & Set & Studies & Contrasts & Contrast mean & Study mean & MAS ahead \\
\midrule
Cat.~A (Single-pass) & with Zhang & 5 & 358 & $-0.032$ & $-0.010$ & 0.37 \\
Cat.~A (Single-pass) & without Zhang & 4 & 178 & $-0.005$ & $+0.003$ & 0.48 \\
Cat.~B (Sampled) & with Zhang & 6 & 360 & $-0.051$ & $-0.043$ & 0.24 \\
Cat.~B (Sampled) & without Zhang & 5 & 220 & $-0.038$ & $-0.037$ & 0.27 \\
Cat.~C (Budget-matched) & all & 2 & 114 & $-0.012$ & $-0.024$ & 0.32 \\
Cat.~B/C (Combined) & with Zhang & 8 & 474 & $-0.042$ & $-0.038$ & 0.26 \\
Cat.~B/C (Combined) & without Zhang & 7 & 334 & $-0.029$ & $-0.033$ & 0.29 \\
\midrule
\multicolumn{6}{l}{\textit{Sampling gap (paired contrasts; $\Delta$ vs Category B minus $\Delta$ vs Category A)}} & \textit{$\Delta_{\text{gap}} < 0$} \\
\quad Jwalapuram & & & 107 & $-0.007$ $\to$ $-0.041$ & $-0.033$ & 0.95 \\
\quad Zhang & & & 140 & $-0.056$ $\to$ $-0.073$ & $-0.016$ & 0.79 \\
\quad Smit & & & 28 & $+0.000$ $\to$ $-0.021$ & $-0.021$ & 1.00 \\
\quad Ke & & & 29 & $-0.028$ $\to$ $-0.070$ & $-0.042$ & 0.86 \\
\quad Choi & & & 14 & $+0.045$ $\to$ $-0.031$ & $-0.076$ & 1.00 \\
\quad Pooled, studies equal & & 5 & 318 & & $-0.038$ & \\
\quad Pooled without Zhang & & 4 & 178 & & $-0.043$ & \\
\quad Deduplicated baseline cells (5 studies) & & 5 & 72 & $-0.034$ & $-0.036$ & 0.89 \\
\bottomrule
\end{tabular}
\end{table}

\FloatBarrier

\subsection{Denominators and caveats}\label{app:moved_s5_denoms}

\paragraph{Denominators and caveats.} The \numRawRows{} raw extracted rows less \numRowsExcluded{} excluded rows (14 majority-voting rows in Choi et al.\ and 1 self-consistency baseline row in Wunderlich et al.) give \numContrastsTotal{} unique multi-agent configurations. They form \numContrastRowsAC{} contrasts against Categories~A to~C (plus two Category~D Pareto configurations); \numContrastsCatBC{} are Category~B/C contrasts, \numContrastsDual{} are paired against both a greedy and a sampled baseline, and those pairs reduce to \numCellsDual{} unique baseline cells. \Cref{tab:pace_verdict} applies one adjudication rule: a configuration is ruled out when a baseline with no more of a known resource scored at least as high (\numConfigFeasible{}), ahead when it beats a baseline at an equal ceiling or lower observed cost (\numConfigAhead{}), and undetermined otherwise (\numConfigUndet{}). Three caveats bear on the rows. The Jwalapuram et al.\ baselines are maximum rows. Tran and Kiela's 16 contrasts at the truncated budget of 100 thinking tokens (mean $+0.077$) are reported separately where a table says so. Zhang et al.\ supply 180 configurations against greedy chain of thought but 140 against self-consistency (there is no self-consistency baseline for HumanEval and MBPP), and AGIEval items use an $N=100$ placeholder in the noise bands.

\FloatBarrier

\subsection{Empirical study ladder and the sampling gap}\label{app:moved_s5_ladder_fig}

\FloatBarrier

\subsection{Disaggregated study-by-benchmark summaries and leave-one-out checks}\label{app:moved_s5_disagg}

Disaggregated study-by-benchmark summaries (\Cref{tab:appB_ledger}) and sensitivity analyses omitting one study at a time (\Cref{app:sensitivity}) appear in the appendices.

\FloatBarrier

\subsection{Exploratory multilevel model}\label{app:moved_s5_reml}

\label{sec:reml_main}
We fit exploratory REML models with three levels (contrasts nested in studies) on the log odds ratio, assuming item correlation $\varphi=0.50$ and using CR2 cluster-robust intervals grouped by study with Satterthwaite degrees of freedom \citep{cheung2014modeling,tipton2015small}. \Cref{tab:k9_reml} reports the exact category split matching \Cref{tab:baseline_tiers}: Category~A alone (A9; \remlANineK\ studies, \remlANineM\ contrasts), Category~B alone ($\mathrm{B}_{\mathrm{only}}$; \remlBOnlyK\ studies, \remlBOnlyM\ contrasts), combined Category~B/C (B9; \remlBNineK\ studies, \remlBNineM\ contrasts), Category~B/C without Zhang et al.\ (B9$-$Z; \remlBNineNoZK\ studies, \remlBNineNoZM\ contrasts), and the baseline-strength check (D; \remlGapK\ studies, \remlGapM\ paired contrasts modeling $\operatorname{logit}(p_{\text{A}}) - \operatorname{logit}(p_{\text{B}})$). The outcome in D does not depend on the multi-agent arm, so its \remlGapM\ rows repeat \remlGapCellsM\ distinct baseline cells and its interval understates uncertainty; the deduplicated cell-level REML fit ($\mathrm{D}_{\mathrm{cells}}$: \remlGapCellsK\ studies, \remlGapCellsM\ unique cells, $\hat\beta = \remlGapCellsBeta$, CR2 95\% CI [\remlGapCellsCrLo, \remlGapCellsCrHi], $\nu = \remlGapCellsNu$) is reported alongside it in \Cref{tab:k9_reml}, and the descriptive cell summary appears in \Cref{tab:baseline_tiers}. Category~C alone contains only two studies (Tran \& Kiela and Kim et al.) and is kept descriptive in \Cref{tab:baseline_tiers}.

Under comparable sampled or budget-matched baselines (B9), the pooled log odds estimate is $\hat\beta = \remlBNineBeta$ (SE \remlBNineSE; CR2 $95\%$ CI [\remlBNineCrLo, \remlBNineCrHi], $\nu=\remlBNineNu$), remaining negative when restricted to sampled baselines ($\mathrm{B}_{\mathrm{only}}$: $\hat\beta = \remlBOnlyBeta$, CR2 [\remlBOnlyCrLo, \remlBOnlyCrHi]), when omitting Zhang et al.\ (B9$-$Z: $\hat\beta = \remlBNineNoZBeta$, CR2 [\remlBNineNoZCrLo, \remlBNineNoZCrHi]), and when omitting Jwalapuram et al.\ whose CoT-SC row is an upper-bound maximum (B9$-$J: \remlBNineNoJK\ studies, \remlBNineNoJM\ contrasts, $\hat\beta = \remlBNineNoJBeta$, CR2 [\remlBNineNoJCrLo, \remlBNineNoJCrHi]). Because heterogeneity across contrasts is high ($I^2_{(2)}=\remlBNineITwo\%$, contrast-level variance $\hat\sigma^2_{(2)} = 0.230$ vs.\ median cell variance $0.012$), the B9 $95\%$ prediction interval includes positive values for a new study mean ([$-0.515$, $+0.125$], $K-2 = 6$ degrees of freedom, CR2 SE) as well as for a single new contrast ([\remlBNineContrastPiLo, \remlBNineContrastPiHi]), showing that individual new configurations can exceed zero around the negative pooled mean. That dominance of $\hat\sigma^2_{(2)}$ over sampling variance also makes the pooled REML estimate $\hat\beta$ nearly insensitive to assumed $\varphi$, Fr\'echet--Hoeffding clamping (which binds in $4.0\%$, $14.8\%$, and $50.2\%$ of contrasts at $\varphi=0.50, 0.75, 0.90$), and placeholder sample sizes on AGIEval.

In B9 sensitivity checks (\Cref{app:sensitivity}, \Cref{tab:appA_bc_sensitivity}, \Cref{fig:reml_forest}), $\hat\beta$ remains negative across $\varphi \in [0.00, 0.90]$ (\remlBNinePhiMinBeta\ to \remlBNinePhiMaxBeta), all eight fits that omit one study at a time (\remlBNineLooMinBeta\ to \remlBNineLooMaxBeta), CHE block covariance up to $\rho=0.80$ within shared baselines and shared benchmarks, merged Jwalapuram--Ke authorship clustering, Fr\'echet--Hoeffding bounds, raw accuracy difference, Cohen's $h$, and 19-framework clustering; only an adversarial uniform $\rho=0.50$--$0.80$ across disjoint benchmarks within a study pushes upper CR2 bounds past zero (\remlCheStudyHalfCrHi\ to \remlCheStudyEightCrHi).

Across five to eight studies, within-study heterogeneity remains high ($I^2_{(2)}=\remlBNineITwo\%$ in B9), $\varphi$ is assumed rather than observed, and baseline category coincides with individual papers; descriptive summaries remain the primary basis for inference. As an exploratory summary, B9 pools across two comparison designs that impose distinct baseline constraints (Category~B sample multiplicity without full resource verification versus Category~C declared resource ceilings) and should not be interpreted as estimating a single uniform intervention; $\mathrm{B}_{\mathrm{only}}$ is therefore reported separately alongside B9, while Category~C remains descriptive across two studies. Because both 95\% prediction intervals include positive values, the negative B9 mean describes the included Category~B/C contrasts and does not imply that every new configuration or domain will trail a noncommunicating baseline. Finally, because the effective number of study clusters in B9 is eight rather than 474, the multilevel REML and CR2 intervals are exploratory; clustering by 19 normalized framework families (a sensitivity taxonomy distinct from the five topology archetypes in \Cref{tab:supp_topology_archetypes}) depends on the family mapping and does not add independent studies.
\begin{table}[t]
\centering
\footnotesize
\caption{Exploratory REML model with three levels across baseline categories (log odds scale; $\varphi=0.50$ assumed; contrasts nested in studies). A9 pools Category A contrasts; $\mathrm{B}_{\mathrm{only}}$ pools Category B contrasts; B9 combines Categories B and C; B9$-$Z omits Zhang et al.\ as an influential-study check; D is a baseline-strength check on the five dual-baseline studies, modeling the log odds of the Category A baseline minus the Category B baseline across 318 paired contrasts spanning 72 unique baseline cells. Category C alone spans only two studies and is reported descriptively in \Cref{tab:baseline_tiers}. CR2 intervals cluster by study with Satterthwaite $\nu$; sensitivity to $\varphi$, omitting one study at a time, and block covariance is given in \Cref{app:sensitivity}. For B9 the CR2 $p$-value is $p=0.0094$ ($t=-3.67$, $\nu=6.4$). The item correlation $\varphi$ is assumed, not estimated, because item-level outcomes are not available for the primary studies; the B9 point estimate stays between $-0.194$ and $-0.195$ for assumed $\varphi$ from 0 to 0.90.}
\label{tab:k9_reml}
\resizebox{\textwidth}{!}{
\begin{tabular}{lcrrccr}
\toprule
Model & Studies/contrasts & $\hat\beta$ & SE & Wald 95\% CI & CR2 95\% CI & $I^2_{(2)}/I^2_{(3)}$ \\
\midrule
A9: Category A, single-pass & 5/358 & $-0.071$ & 0.087 & [$-0.241$, $+0.100$] & [$-0.318$, $+0.177$] ($\nu=3.8$) & 90 / 9 \\
$\mathrm{B}_{\mathrm{only}}$: Category B, sampled & 6/360 & $-0.235$ & 0.056 & [$-0.345$, $-0.125$] & [$-0.388$, $-0.082$] ($\nu=4.5$) & 95 / 4 \\
B9: Category B/C, matched & 8/474 & $-0.195$ & 0.052 & [$-0.298$, $-0.092$] & [$-0.323$, $-0.067$] ($\nu=6.4$) & 93 / 6 \\
B9$-$Z: Category B/C without Zhang et al. & 7/334 & $-0.145$ & 0.039 & [$-0.221$, $-0.070$] & [$-0.246$, $-0.045$] ($\nu=5.1$) & 86 / 4 \\
D: Baseline-strength check (sampling gap, 318 pairs) & 5/318 & $-0.169$ & 0.047 & [$-0.260$, $-0.077$] & [$-0.298$, $-0.039$] ($\nu=3.9$) & 72 / 21 \\
$\mathrm{D}_{\mathrm{cells}}$: Baseline-strength check (72 unique cells) & 5/72 & $-0.178$ & 0.058 & [$-0.292$, $-0.063$] & [$-0.347$, $-0.009$] ($\nu=3.7$) & 71 / 22 \\
\bottomrule
\end{tabular}}
\end{table}

\FloatBarrier

\subsection{Why the studies appear to disagree}\label{app:moved_s5_disagree}

The evidence is consistent with a baseline explanation of the apparent divergence: studies comparing multi-agent graphs against single greedy passes without budget parity show positive averages in isolated cases such as Choi et al.\ ($+0.045$, whereas four of five Category~A study means are zero or negative, $-0.032$ overall), while studies holding token or iteration budgets constant (Category~C) or comparing against sampled baselines (Category~B) yield nonpositive averages, and in Choi et al.\ decentralized debate trails majority voting over five samples on average ($\bar\Delta = -0.031$, median $-0.008$; ahead in 6 cells, tied in 1, behind in 7). Because matching condition coincides with study, paper-specific differences cannot be excluded.

\FloatBarrier

\subsection{Limitations of the corpus synthesis}\label{app:moved_s5_limits}

(i) The corpus spans nine studies, with study, matching condition, and baseline category partially confounded; one paper contributes 180 of 516 unique multi-agent configurations (531 raw extracted rows), and the five initial studies were selected without a search protocol, permitting selection bias. (ii) Descriptive means are unweighted; the exploratory REML models with three levels span five to eight studies (except two studies in descriptive Category~C), assume item correlation $\varphi$, and rely on assumed block correlations $\rho$ in CHE sensitivity models when shared baselines are modeled explicitly. (iii) Log odds use the Haldane--Anscombe continuity correction $\tilde{p} = (pN + 0.5)/(N + 1)$, which shifts near-boundary cells toward 0.5. (iv) Some contrasts rely on small test samples (such as $N=20$ in Kim et al.). (v) Primary multi-agent arms (Centralized, debate at $T=2$) were selected prior to pooling, constituting an analytical degree of freedom. (vi) Sample sizes for Tran and Kiela come from published benchmark definitions rather than the primary table text. Extended scope notes on implementation fidelity, task coverage, model mix, latency axes (\Cref{fig:resource_matrix}), baseline maxima, and item-count capping appear in \Cref{app:scope_notes}.

\FloatBarrier

\subsection{The \texorpdfstring{$\sim 45\%$}{~45\%} Capability Threshold: What the Extracted Data Support}\label{app:moved_s5_threshold}

\label{sec:threshold}
\citet{kim2026scaling} report that on tasks where single-agent accuracy already exceeds about $45\%$, adding communicating agents yields negative returns across all four topologies they study, and that returns depend on task structure. A threshold claim is hard to establish for three reasons: accuracy is bounded, so raw differences compress near the ceiling \citep{schaeffer2024are}; baseline accuracy is confounded with benchmark and model family; and a ceiling can only be attributed to coordination using directionally feasible baselines (a baseline using no more compute suffices to establish failure, whereas claiming gains requires at least as much compute).

\paragraph{A fitted illustration, not empirical evidence.} \Cref{thm:conformity} gives one way a ceiling of this kind could arise. If, without ground-truth feedback, an unverified step in a multi-step plan is easier to break than to repair, then $\tau_{c\to i}$ exceeds $\tau_{i\to c}$, $\kappa<1$, and communication lowers per-agent accuracy for every baseline above $p^{*}=\kappa/(1+\kappa)$. A value $\kappa\approx0.82$ gives $p^{*}\approx0.45$. That value is fitted to the reported threshold, one parameter to one number, so it shows the scale of $\kappa$ needed and is not evidence for the model: we have no transition rates for the systems of \citet{kim2026scaling}. The explanation predicts that adding an external verifier raises the threshold (route II in \Cref{sec:redundancy}), and, because equal wrong-to-right and right-to-wrong counts are what independent redraws produce (\Cref{sec:redundancy}), it would be refuted if, for baselines above $0.45$, the net change in accuracy under debate were no lower than under a no-communication control. The segmented regression in \Cref{app:exploratory} finds no break in the extracted contrasts, so we do not use the model to claim that a threshold exists.

Where thinking tokens are matched or bounded by a shared ceiling (Tran and Kiela), single agents lead from $500$ tokens upward across both FRAMES and MuSiQue 4-hop deduction regardless of baseline accuracy (\Cref{fig:disaggregated_scaling}), while the unexplained advantage at $100$ tokens rests on incomplete reasoning outputs. We therefore treat capability thresholds as task-dependent rather than universal. Tables, budget curves, and regression details are in \Cref{app:exploratory}.

The fitted threshold is a descriptive discontinuity in study-level means and not a causal threshold, because baseline accuracy was not randomized across studies.

\FloatBarrier

\section{Extended Controlled Experiment}

\subsection{Logged call quantities along a policy}\label{app:moved_s4_timeline}

\begin{figure}[t]
\centering
\includegraphics[width=0.92\textwidth]{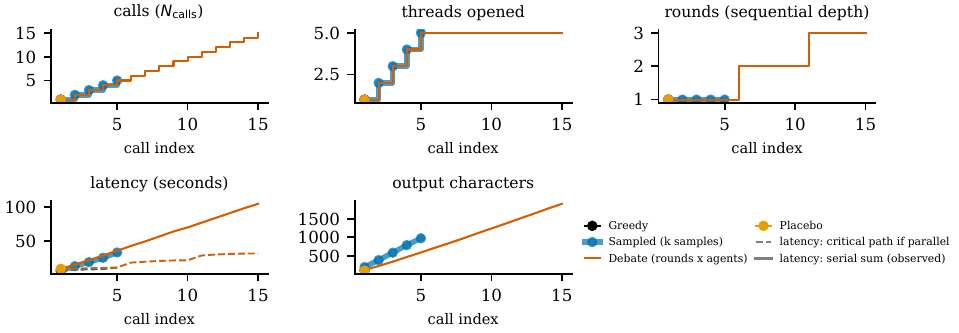}
\caption{Logged trace quantities along each policy's call sequence, mean over \traceItems{} traced items (\texttt{qwen2.5-coder:7b}, arithmetic). Greedy and placebo are single calls (points); the five samples are independent, whereas debate depth grows by round. Solid: observed; dashed latency: derived if the calls of a round run in parallel. Characters stand in for tokens, which the traces omit.}
\label{fig:resource_timeline}
\end{figure}

\FloatBarrier

\subsection{Controlled comparison: motivation}\label{app:moved_s6_intro}

The synthesis in \Cref{sec:meta} compares published systems whose budgets are matched in different ways, and few of its contrasts use a sampled baseline that was run for the purpose. This section reports a controlled comparison in which every strategy is run on the same items with the same model, and the multi-agent strategies reuse the first five samples of the sampling pool as their first round, so that debate and voting differ only by the messages. The design, hypotheses and analysis plan were written before the runs; \confRegStatus.

\FloatBarrier

\subsection{Protocol deviations and post hoc extensions}\label{app:moved_s6_deviations}

\begin{table}[!htbp]
\centering
\small
\caption{Protocol deviations and post hoc extensions.}
\label{tab:protocol_deviations}
\begin{tabular}{p{0.28\textwidth}p{0.62\textwidth}}
\toprule
Item & Disclosure \\
\midrule
Frontier tie and parsing rule & Empty parses are excluded from plurality counts; ties receive equal credit in S2 and the reconstructed frontier. \\
Pool extension & Eight additional independent MMLU-Pro samples ($k=17$--$24$) were collected post hoc with the original prompt, temperature, seed construction, and item list; the planned $k=16$ result remains primary. \\
Inference count & The $k=24$ extension is exploratory and does not alter any original call or result record for $k=1$--$16$ (samples are numbered from 1 here; the logs index them from 0). \\
Planned cells not run & The plan named replications of H4 on gpt-oss-20b and gpt-oss-120b (hosted free tier) and a full-history token-growth check of the scaffolding formula on 100 items; neither was run, so H4 rests on llama3.1:8b only, and the reasoning-model cells (deepseek-r1:14b) are 50-item exploratory slices. \\
\bottomrule
\end{tabular}
\end{table}

\FloatBarrier

\subsection{Exploratory live-suite debate-dynamics traces}\label{app:moved_s6_livesuite}

Exploratory $K=3$ live-suite debate-dynamics traces on \texttt{llama3.1:8b} illustrate transition asymmetry: on GSM8K ($n=35$), communication increases correctness correlation, whereas on MATH-500 levels 4--5 ($n=25$) corruption dominates correction.

\FloatBarrier

\subsection{Reasoning-model cell: disclosure and token caps}\label{app:moved_s6_deepseek}

In \texttt{deepseek-r1:14b}, thinking characters are \confReasonShareHp{} of thinking plus visible peer-message characters on Hidden Profile (\confReasonShareMmlu{} on the MMLU-Pro slice; debate and pool calls), which is consistent with too few disclosed criterion details in the visible peer messages for agents to pool the split evidence (accuracy $0.000$ on both split-view voting and split-view debate, \Cref{tab:confirmatory_cells}); \confDsCapHp{} of \confDsCallsHp{} hidden-profile calls reached the 8,192-token cap (\confDsParseHp{} of \confDsParseOfHp{} answers unparsable) and, on the MMLU-Pro slice, \confDsCapMmlu{} of \confDsCallsMmlu{} (\confDsParseMmlu{} of \confDsParseOfMmlu{} unparsable).

\FloatBarrier

\subsection{Replications and exploratory cells}\label{app:moved_s6_repl}

\label{sec:controlled_secondary}
Full strategy-level breakdowns across all 17 completed cell-strategy rows, together with exploratory live validation runs on GSM8K ($n=35$), MATH-500 levels 4--5 ($n=25$), and adaptive gating, appear in \Cref{tab:confirmatory_cells} and \Cref{app:controlled}. MATH-500 levels 4 to 5 is a secondary cell with restricted headroom (pilot greedy accuracy 0.36 to 0.38 for two of the models). The targeted run of a reasoning model (\texttt{deepseek-r1:14b}) with up to 8,192 output tokens covers 50 hidden-profile items and a category-stratified slice of 50 MMLU-Pro items (with 500 debate calls, 250 pool calls and 50 greedy calls logged for the 50 MMLU-Pro items) and is exploratory. Because the server does not report reasoning tokens, the reasoning share is computed on characters: on the MMLU-Pro slice the thinking characters of a debate item are \confReasonCharsMmlu{} on average and the peer-message characters shown to agents are \confCommCharsMmlu{}, a share of \confReasonShareMmlu{} thinking; on the hidden-profile slice the corresponding values are \confReasonCharsHp{}, \confCommCharsHp{} and \confReasonShareHp{} (with greedy full-evidence accuracy $1.000$ versus $0.000$ on both split-view voting and split-view debate). In \texttt{deepseek-r1:14b}, allocating $94\%$ of generated characters to private \texttt{<think>} blocks leaves too few disclosed criterion details in the visible peer messages for agents to pool the split evidence.

\FloatBarrier

\subsection{Positive control: the two cases}\label{app:moved_s6_framing}

\Cref{sec:redundancy} separates two cases. With a shared context, every agent holds the same evidence, $\operatorname{I}(Y;X_1,\dots,X_K\mid X_{\mathrm{shared}})=0$, and communication cannot add information. With partitioned evidence, $\operatorname{I}(Y;X_1,\dots,X_K\mid X_{\mathrm{shared}})>0$, and an information gain is possible, though not guaranteed. A positive control should show that the evaluation can detect the second case; a measurement that reports no coordination premium on shared-context benchmarks is informative only if it could have reported one.

\FloatBarrier

\subsection{Positive control: design}\label{app:moved_s6_hpdesign}

The control follows the hidden-profile design of \citet{stasser1985pooling}. Each item has four candidates scored from 1 to 5 on ten equally weighted criteria. Four criteria are shared and favor a decoy candidate; the other six favor the target and are divided among five agents, each of whom sees the four shared criteria and two of the six. In every single view the decoy leads by 6 points, so an agent that sees only its own view picks the decoy. Over the pooled evidence the target leads (38 against 32 points). The generator uses seed 42 and exports 500 items; a script checks, for every item, that the decoy leads in each view, that the target leads over the pooled evidence, and that all six split criteria appear (no violation in 500 items). Independent voting over the views is therefore expected to favor the decoy by construction, and the control tests whether debate, which lets agents exchange their views, recovers the target.

\FloatBarrier

\subsection{Hidden-profile positive control: table}\label{app:moved_s6_pctable}

\edef\hpNtest{\hpN}\edef\hpNna{\confNA}\ifx\hpNtest\hpNna
\begin{table}[!htbp]
\centering
\small
\caption{Hidden-profile positive control, llama3.1:8b: protocol and predictions. Each agent sees the four shared criteria and two of the six split criteria. The predictions follow from the construction and from \Cref{sec:redundancy}; they are not results. Status: protocol preview shown before completed cells populate the observed-values branch. The plan was committed and hashed before any outcome was analyzed and has not been registered on a public registry. The observed accuracies replace this table when every needed cell is complete.}
\label{tab:poscontrol}
\begin{tabular}{p{0.27\textwidth}p{0.30\textwidth}p{0.33\textwidth}}
\toprule
Strategy & Information per agent & Prediction when $\operatorname{I}(Y;X_{1:K}\mid X_{\mathrm{shared}})>0$ \\
\midrule
Independent voting over the five views (Category B) & shared criteria and two split criteria; no exchange & Each view favors the decoy, so the vote is expected to favor the decoy; accuracy near zero if agents add the visible scores correctly \\
Debate, five agents, three rounds & the same, plus the other agents' messages & Pooling is possible, so accuracy above independent voting; the size of the gap is not predicted \\
Frontier of independent voting at the debate's tokens (hypothesis H2) & as independent voting & Gap of debate over the frontier above zero \\
\bottomrule
\end{tabular}
\end{table}
\else
\begin{table}[!htbp]
\centering
\small
\caption{Hidden-profile positive control, llama3.1:8b. Accuracy over $n=\hpN$ items; tokens are prompt plus completion tokens per item. The gap is debate accuracy minus the isotonic self-consistency frontier of independent voting over the same views, evaluated at the debate's mean tokens, with a 95\% paired bootstrap interval over items. }
\label{tab:poscontrol}
\begin{tabular}{lrr}
\toprule
Strategy (same split views) & Accuracy & Mean tokens per item \\
\midrule
Independent voting, five views (Category B) & \hpVoteAcc & \hpVoteTok \\
Debate, five agents, three rounds & \hpDebateAcc & \hpDebateTok \\
\midrule
Frontier of independent voting at the debate's tokens & \hpFrontier & \hpDebateTok \\
Gap (debate minus frontier) [95\% interval] & \multicolumn{2}{r}{\hpGapEst\ [\hpGapLo, \hpGapHi]} \\
\bottomrule
\end{tabular}
\end{table}
\fi

\FloatBarrier

\subsection{What the stress test does and does not show}\label{app:moved_s6_controlnote}

The control checks the measurement, not the principle: it shows whether the frontier-gap protocol detects a premium when one is built into the task. A positive gap here, together with gaps near zero on shared-context tasks, would be consistent with \Cref{sec:redundancy}; it would not show that partitioned information is the only condition under which communication helps. The synthetic task is one construction with a fixed split, and results on it do not transfer to natural tasks.

\FloatBarrier

\section{Extended Standard and Reporting}

\subsection{Reporting requirements and measured resources}\label{app:moved_s4_reporting}

\paragraph{Reporting requirements.}
For each comparison, researchers should report (i) the verification state of each dimension in $\mathbf{R}$, (ii) the single-agent baseline's policy and its resource relation to the multi-agent system on each reported dimension (for example self-consistency with $k \ge k_{\text{tok}}$ samples), (iii) the number of evaluation items, and (iv) binary outcomes for each item or paired statistics so that paired variance can be computed. \Cref{sec:discussion} turns these into a reviewer checklist.

\paragraph{Measured resources and inferred resource requirements.} The resource vector distinguishes quantities logged by a source study from quantities inferred for protocol sizing: a dimension is classified as controlled only when a common ceiling is enforced, approximately matched when a stated bound is applied, unmatched when measured but allowed to differ, and unreported when omitted. The expression $k_{\mathrm{tok}}$ (\Cref{eq:ktok}) is a protocol-specific pool-sizing calculation under the message-length assumptions in \Cref{sec:theory} rather than a replacement for logged uncached and cached prompt tokens, output tokens, or provider billing records, so a Category~B sample-count match does not imply equality of total tokens, monetary cost, or execution latency.

\FloatBarrier

\subsection{Discussion: structural cases and decision rules}\label{app:moved_s7_lead}

Positive multi-agent returns in the corpus and source papers cluster in four structural cases rather than contradicting the pooled result: incomplete compute control on Pareto fronts (Wunderlich et al.), truncated individual reasoning chains (Tran and Kiela at 100 thinking tokens), domain-engineered workflows with external verification counted in $R_{\mathrm{sys}}$ (the Expert MAS of Jwalapuram et al.), and parallelizable task decomposition (Kim et al.). Three rules follow: measure the single-agent frontier before crediting a topology; choose agents by task structure, not accuracy targets; and give any verifier to both arms and report $R_{\mathrm{sys}}$. A semantic benefit of communication also requires $\Delta_{\mathrm{arch}}=p_{\mathrm{comm}}-p_{\mathrm{placebo}}>0$, which the controlled experiment did not show on Hidden Profile, where the placebo is content-mismatched and uses more tokens (\Cref{app:moved_s7_synth}, \Cref{app:moved_s7_decision}).

\FloatBarrier

\subsection{What the evidence supports: a four-case diagnostic synthesis}\label{app:moved_s7_synth}

The evidence supports an empirical conclusion conditional on baseline strength. Against a single greedy pass (Category~A), multi-agent systems show an unweighted contrast mean of \meanCatA\ across \numContrastsCatA\ contrasts (\winRateCatA\ of contrasts ahead) and \meanCatAStudy\ weighted equally by study; omitting Zhang et al., who supply \numContrastsZhang\ of these configurations, yields \meanCatANoZhang\ contrast-weighted (\winRateCatANoZhang\ ahead) and \meanCatANoZhangStudy\ weighted equally by study. Against sampled baselines (Category~B; \numContrastsCatB\ contrasts), multi-agent systems trail on average (\meanCatB\ contrast mean, \winRateCatB\ ahead; \meanCatBNoZhang\ without Zhang et al.). When thinking tokens or iterations are controlled (Category~C; \numContrastsCatC\ contrasts), the average is \meanCatC\ (\meanCatCStudy\ study mean), and with matched thinking tokens single agents match or beat debate from 500 tokens upward. Across combined sampled and budget-matched baselines (\numContrastsCatBC\ contrasts across eight studies), the contrast mean is \meanCatBC\ (\remlBNineBeta\ on log odds in REML). All eight Category~B/C study means are negative; the one Category~D study, and the sole positive exception among extracted studies, is \citet{wunderlich2026mixture}, where debate ($+0.013$) and mixture-of-agents ($+0.027$) lead self-consistency at comparable compute read off Pareto fronts (not an exact budget match) on MMLU-Pro with Llama-3.1-70B ($+0.020$ mean; the mixture-of-agents proposers are copies of one model). In the cost audit, the mean multi-agent difference is $-0.041$ against CoT-SC with five samples and $-0.007$ against greedy chain-of-thought (\Cref{sec:meta}).

Across the extracted corpus, positive Category~B/C exceptions cluster into four structural cases rather than contradicting the pooled Category~B/C result. First, in Case~1 (incomplete compute control on Pareto fronts), \citet{wunderlich2026mixture} report that debate ($+0.013$) and mixture-of-agents ($+0.027$) lead self-consistency on MMLU-Pro with Llama-3.1-70B at comparable compute read off Pareto fronts rather than under an exact budget match per item. Second, in Case~2 (truncated individual reasoning chains), \citet{tran2026thinking} observe that multi-agent configurations average $+0.077$ at 100 thinking tokens where individual reasoning chains are truncated, whereas single-agent scaling leads from 500 thinking tokens upward ($-0.012$ at 500 tokens and $-0.023$ to $-0.035$ across 1{,}000--10{,}000 tokens). Third, in Case~3 (workflow engineering and external verification affordances), the domain-engineered \texttt{Expert MAS} workflow on SMFR in \citet{jwalapuram2026illusion} reaches $96.51\%$ at \$555 against $57.0\%$ for CoT-SC with five samples and $20.2\%$--$63.0\%$ for automated multi-agent topologies, reflecting specialized workflow engineering and verification affordances ($R_{\text{sys}}$) rather than unverified peer conversation alone. Fourth, in Case~4 (parallelizable task decomposition), \citet{kim2026scaling} observe positive multi-agent returns on parallelizable financial retrieval tasks alongside negative returns on sequential software and planning tasks, indicating that task decomposability moderates whether multi-agent coordination helps or hurts.

\FloatBarrier

\subsection{Practitioner decision rules}\label{app:moved_s7_decision}

We translate the PACE protocol into three operational rules, detailed alongside the route controls (\Cref{tab:route_tests}) and deployment flow (\Cref{fig:decision_flowchart}) in \Cref{app:checklist_templates}.

\paragraph{Rule 1: Measure the single-agent frontier before crediting a multi-agent topology.}
Before attributing accuracy gains to multi-agent communication, compare against single-agent self-consistency with $k \ge k_{\text{tok}}$ samples (\Cref{eq:ktok}) and report the frontier gap with its uncertainty interval \citep{wang2022selfconsistency,snell2024scaling,brown2024large}. In deduction matched on thinking tokens, single agents match or outperform debate from 500 tokens upward; and under matching on agent count, majority voting without communication exceeds decentralized debate with two rounds ($0.747$ vs.\ $0.715$). Replacing peer messages by messages from another item, matched in character length but not in tokens (strategy S7 in \Cref{sec:controlled}; see also \citet{bertalanic2026cost}), separates semantic architecture effects $\Delta_{\mathrm{arch}} = p_{\mathrm{comm}} - p_{\mathrm{placebo}}$ from raw context expansion overhead $\Delta_{\mathrm{overhead}} = p_{\mathrm{base}} - p_{\mathrm{placebo}}$ (where $p_{\mathrm{base}}$ denotes the resource-matched noncommunicating baseline), requiring both a positive net accuracy advantage ($p_{\mathrm{comm}} > p_{\mathrm{base}}$, equivalently $\Delta_{\mathrm{arch}} > \Delta_{\mathrm{overhead}}$) and positive semantic architecture benefit ($\Delta_{\mathrm{arch}} > 0$). When systems target efficiency rather than raw accuracy alone, report the paired dollar cost differential $\Delta C = C_{\mathrm{MAS}} - C_{\mathrm{baseline}} < 0$ at matched accuracy separately from $\Delta_{\mathrm{acc}}$ (\Cref{app:checklist_templates}).

\paragraph{Rule 2: Deploy multiple agents based on task structure, not accuracy targets.}
Teams of agents are justified under specific structural constraints: split evidence across agents (\Cref{tab:route_tests}), input contexts exceeding single-model limits, independent parallel tools, or separated execution states \citep{kim2026scaling,ke2026masorchestra,chu2026masbench}, including the fourth untested candidate condition in which parallel tool execution and isolated workspace sandboxes avoid serial bottlenecks or context contamination (\Cref{app:checklist_templates}). When automated topology or prompt search discovers the graph offline, separate offline search expenditure from per-query inference cost inside $R_{\mathrm{token}}$ and $R_{\mathrm{cost}}$.

\begin{figure*}[!htbp]
\centering
\includegraphics[width=0.96\textwidth]{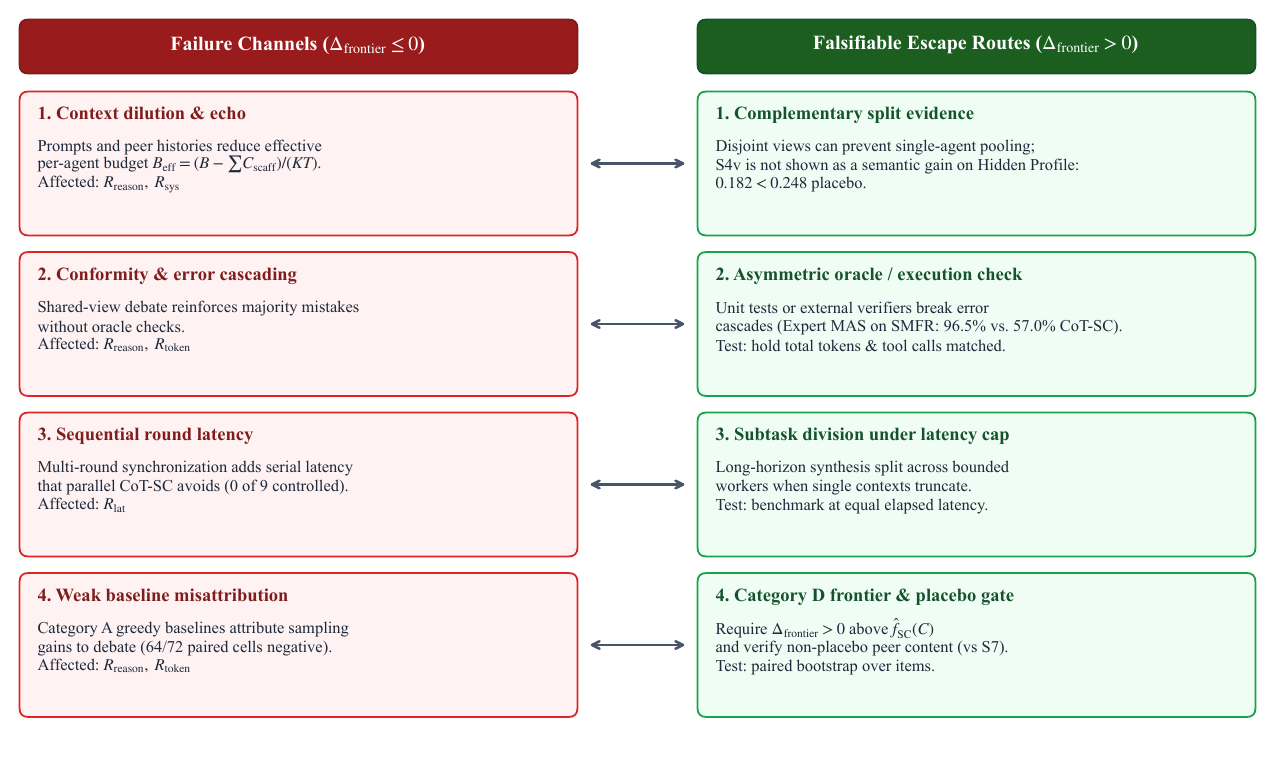}
\caption{Failure channels ($\Delta_{\mathrm{frontier}} \le 0$) and falsifiable escape routes ($\Delta_{\mathrm{frontier}} > 0$) across four mechanisms: (1)~context dilution and echo versus split evidence, (2)~error cascading without oracle checks versus asymmetric verifiers, (3)~sequential round latency ($R_{\mathrm{lat}}$ controlled in 0 of 9 studies) versus latency-bounded subtask division, and (4)~weak baseline misattribution versus Category~D frontier and placebo testing.}
\label{fig:coordination_taxonomy}
\end{figure*}

\paragraph{Rule 3: Provide any external verifier to both arms and report $R_{\text{sys}}$.}
Without compiler checks, unit tests, or database queries, multi-agent debate risks propagating initial errors \citep{huang2023cannot}; positive net drift requires the verification transition differential $\Delta p_{\mathrm{ver}}$ to outweigh conformity aggregation drift $\Delta p_{\mathrm{agg}}$, consistent with the transition threshold $\tau_{i\to c} > \frac{p_1}{1-p_1} \tau_{c\to i}$ (\Cref{thm:conformity}). Whenever external verifiers, tool affordances, or isolated execution environments are supplied to the multi-agent system, that same $R_{\text{sys}}$ affordance must be granted to the noncommunicating baseline so tool and environment state are not conflated with agent conversation. On SMFR, human tooling reaches $96.51\%$ versus $57.0\%$ for CoT-SC with five samples and $20.2\%$--$63.0\%$ for automated graphs. \Cref{fig:coordination_taxonomy} pairs the four failure channels against falsifiable escape routes (\Cref{fig:decision_flowchart} in \Cref{app:checklist_templates} gives the deployment decision tree).

\FloatBarrier

\subsection{Short OpenReview reviewer block}\label{app:moved_s7_orblock}

\begin{center}
\fbox{\begin{minipage}{0.94\textwidth}
\small
\textbf{Short OpenReview reviewer block.}
\par\smallskip
\texttt{1.\ Baseline rung:} State strongest single-agent Category~[A/B/C/D] (policy and $k$).\\
\texttt{2.\ Resource vector R:} Verify [$R_{\mathrm{reason}}, R_{\mathrm{token}}, R_{\mathrm{cost}}, R_{\mathrm{lat}}, R_{\mathrm{sample}}, R_{\mathrm{sys}}$] on both arms.\\
\texttt{3.\ Feasibility direction:} Positive claim requires baseline resource $\ge$ MAS; null requires $\le$.\\
\texttt{4.\ Frontier gap \& controls:} Require $\Delta_{\mathrm{frontier}}$ 95\% CI and Route I/II/III controls (\Cref{app:checklist_templates}).
\end{minipage}}
\end{center}

\FloatBarrier

\subsection{Evaluation Utilities and Prototype Implementation}\label{app:moved_s7_util}

To support resource reporting for both systems, we provide prototype utilities in the companion repository (\texttt{code/mae\_eval}), implementing resource tracking across $\mathbf{R}$ and automated parity checks via CLI (\Cref{sec:appendix_mae_eval}). The package installs two entry points: \texttt{mae-eval}, which audits a JSONL trace of a multi-agent and a single-agent system, and \texttt{pace}, which adds the baseline rung, the audit card, the frontier gap, debate dynamics and the hidden-profile generator (\Cref{sec:appendix_cme_cli}); the text uses \texttt{pace} unless it names the trace audit.

\subsubsection{Beyond Large Language Models}
\label{sec:beyond_llms}
The problem of separating coordination from sample multiplicity and compute is not specific to LLMs. Bagging reduces variance without communication \citep{breiman1996bagging}, cooperative multi-agent reinforcement learning \citep{lowe2017maddpg,foerster2018counterfactual} requires interaction steps and parameter accounting before crediting coordination, and in social choice independent voting gains accuracy with group size \citep{condorcet1785} while deliberation can induce cascades \citep{kerr2004group,surowiecki2004wisdom}. The \numContrastsTotal\ multi-agent configurations (\numRawRows\ raw extracted rows) here cover LLM reasoning and agent tasks only; \Cref{app:checklist_templates} notes the classical ensemble MSE covariance identity relating $K_{\mathrm{eff}}$ to $\rho$.

\FloatBarrier

\subsection{Open Science and audit trail}\label{app:moved_os_full}

\paragraph{Audit trail and data extraction.}
All \numContrastsTotal\ unique multi-agent configurations (across \numRawRows\ raw extracted table rows) of the nine studies were extracted from primary source tables using scripted \texttt{pdftotext} parsing and visual inspection of rendered tables. Candidate study screening (\Cref{app:candidates}) followed written eligibility criteria, and programmatic PDF concordance checks verified 336 of 336 cells on Tran and Kiela and page-level value concordance across the remaining studies. A stratified 40-record extraction audit (40 of 531 raw table rows, 7.5\%; four to five records per study across all nine studies; \texttt{review/EXTRACTION\_AUDIT\_LEDGER.md}) against rendered primary tables recorded zero numerical discrepancies: 16 records matched on a single PDF text line, 19 had at least one value on the cited page, and five with scrambled text layers (Jwalapuram Table~4, Kim Table~16, Yang Table~1) were verified against rendered tables. For reference, zero detected errors in 40 independent checks with perfect detection gives a one-sided 95\% binomial upper limit of $7.22\%$ ($1-0.05^{1/40}$); because the ledger is stratified across raw extraction records rather than a simple random sample of the Category~B/C set, this value is an illustrative reference rather than a corpus-wide error bound. Not every individual cell outside the automated concordance checks and stratified 40-record audit was re-keyed by hand. The authors defined the research question, specified inclusion criteria, directed the analyses, reviewed compiled drafts and verification logs, and take responsibility for all claims. Kim et al.\ values were additionally compared against the source-data file of the published version (see below). Item counts for the four added studies come from each paper's text, with the basis recorded per row (stated, dataset size stated, derived, or estimated).

\paragraph{Analysis protocol.}
The synthesis was not preregistered, and the analysis choices below were made while analyzing the extracted data, not fixed in advance. They are documented here for transparency: (i) the evidence base consists of nine studies (\numContrastsTotal\ unique multi-agent configurations from \numRawRows\ raw table rows), with five initial studies chosen to span different compute dimensions and four added studies selected from seven candidates screened against written criteria; (ii) descriptive summaries and sensitivity analyses are prioritized over inferential pooling because cluster-robust degrees of freedom are limited and matching condition is confounded with study; (iii) questions regarding capability thresholds are examined in light of task structure and baseline levels; (iv) studies are characterized along the resource vector $\mathbf{R}$; (v) source extraction boundaries are preserved explicitly: greedy CoT and CoT-SC baseline ceilings for \citet{jwalapuram2026illusion} follow maxima reported in the source, heterogeneous debate contrasts from \citet{zhang2025stop} are separated from homogeneous self-consistency pairings in \Cref{sec:appendix_hetermad}, and overlapping authorship flags between \texttt{MAS-Zero} and \texttt{MAS-Orchestra} are documented in \Cref{tab:original_vs_audit}.

\paragraph{Reference verification.}
The bibliography was audited against Crossref, OpenAlex, and arXiv records. The audit corrected author lists, titles, venues, and publication years, including for Kim et al., whose work appeared in \textit{Nature Machine Intelligence} in 2026. Our extracted values for Kim et al.\ (Table~16) come from the arXiv preprint (v3). Comparing resolution rates with the source-data file of the published version shows 80 of 90 values agree; the other 10 values, for \texttt{gemini-3-flash}, are absent from that file and appear in Table~S11 of the published supplement, matched by value but not by table position.

\paragraph{Data and code availability.}
The analysis protocol, reproduction code, extraction tables (including \path{contrasts_anchor5_core.csv} and \path{contrasts_dual_baseline.csv}), the stratified audit ledger, the PACE Reviewer Checklist, and the evaluation utilities are available at \url{https://github.com/mahmoudsajjadi/pace}.

\FloatBarrier

\end{document}